\documentclass[dvipsnames]{article} 
\usepackage{iclr2025_conference,times}

\usepackage[breaklinks]{hyperref}
\usepackage[utf8]{inputenc} 
\usepackage[T1]{fontenc}    
\usepackage{url}            
\usepackage{booktabs}       
\usepackage{amsfonts}       
\usepackage{nicefrac}       
\usepackage{microtype}      
\usepackage{xcolor}         
\usepackage{amsmath}
\usepackage{amssymb}
\usepackage{mathtools}
\usepackage{amsthm}
\usepackage[capitalize,noabbrev]{cleveref}
\usepackage{tabularx}
\usepackage{tabularray}
\usepackage{listings}
\usepackage[inline]{enumitem}
\usepackage{caption}
\usepackage{subcaption}
\usepackage{graphicx}
\usepackage{glossaries}
\usepackage{adjustbox}
\usepackage{pgfplots}
\usepackage{mleftright}
\usepackage{derivative}
\usepackage{pifont}
\usepackage{titlesec}
\usepackage{tablefootnote}
\usepackage{algorithm,algorithmicx,algpseudocode}
\usepackage{soul}
\usepackage{wrapfig}
\usepackage{bm}
\usepackage[bottom]{footmisc}
\usepackage{afterpage}
\usepackage{listing}
\usepackage{colortbl}
\usepackage{fancyvrb}
\usepackage[most]{tcolorbox}
\usepackage{multirow}
\usepackage{appendix}
\usepackage{cleveref}
\usepackage{tikz}
\usepackage{makecell}
\usepackage[frozencache]{minted}
\usepackage[group-separator={,}, number-mode=text]{siunitx}
\usepackage{scalerel}
\UseTblrLibrary{booktabs}
\definecolor{myBlue}{HTML}{0F1A5F}
\definecolor{myRed}{HTML}{721010}
\hypersetup{
  colorlinks,
  linkcolor={myRed},
  citecolor={myBlue},
  urlcolor={blue!80!black}
  }
\glsdisablehyper
\hypersetup{breaklinks=true}
\usepgfplotslibrary{groupplots}
\usepgfplotslibrary{colormaps}
\usepgfplotslibrary{fillbetween}
\usetikzlibrary{plotmarks}

\pgfplotsset{compat=1.14}	 %
\pgfplotsset{compat/show suggested version=false}
\pgfplotsset{every mark/.append style={solid}}
\mleftright

\titlespacing*{\paragraph}{0pt}{0.35\baselineskip}{1em}
\newcommand{\myparallel}{{\mkern3mu\vphantom{\perp}\vrule depth 0pt\mkern2mu\vrule depth 0pt\mkern3mu}}

\usepackage{amsmath,amsfonts,bm, mathtools}

\newcommand{\mc}[1]{\mathcal{#1}}

\def\eqref#1{equation~\ref{#1}}

\def\1{\bm{1}}

\newcommand{\dd}{\mathrm{d}}

\def\vv{{\bm{v}}}

\def\vx{{\bm{x}}}
\def\vy{{\bm{y}}}
\def\vz{{\bm{z}}}

\def\mI{{\pmb{I}}}

\def\bmmu{{\bm{\mu}}}

\DeclareMathAlphabet{\mathsfit}{\encodingdefault}{\sfdefault}{m}{sl}
\SetMathAlphabet{\mathsfit}{bold}{\encodingdefault}{\sfdefault}{bx}{n}

\def\mepsilon{{\bm{\epsilon}}}

\def\mtheta{{\bm{\theta}}}

\newcommand{\pdata}{p_{\textnormal{data}}}

\newcommand{\norm}[1]{\left\lVert #1 \right\rVert}

\newcommand{\cond}{\left\vert\right.}

\newcommand{\inner}[2]{\left \langle #1 , #2 \right \rangle}
\newcommand{\normal}[2]{\mc{N}\prn{#1, #2}}

\DeclarePairedDelimiterX{\infdivx}[2]{(}{)}{%
  #1\delimsize\|#2%
}

\newcommand{\brk}[1]{\left[ #1 \right]}

\newcommand{\prn}[1]{\left( #1 \right)}

\newcommand{\zero}{\pmb{0}}

\DeclareDocumentCommand{\ex}{m o}{
   \mathbb{E}\IfValueT{#2}{_{#2}}\left[#1\right]
}

\DeclareMathOperator{\grad}{\nabla}

\DeclarePairedDelimiterX\Set[1]{\lbrace}{\rbrace}%
 {  #1 }

\def\ddefloop#1{\ifx\ddefloop#1\else\ddef{#1}\expandafter\ddefloop\fi}
\def\ddef#1{\expandafter\def\csname #1bb\endcsname{\ensuremath{\mathbb{#1}}}}
\ddefloop ABCDEFGHIJKLMNOPQRSTUVWXYZ\ddefloop
\def\ddefloop#1{\ifx\ddefloop#1\else\ddef{#1}\expandafter\ddefloop\fi}
\def\ddef#1{\expandafter\def\csname #1b\endcsname{\ensuremath{\mathbf{#1}}}}
\ddefloop ABCDEFGHIJKLMNOPQRSTUVWXYZ\ddefloop
\def\ddef#1{\expandafter\def\csname #1c\endcsname{\ensuremath{\mathcal{#1}}}}
\ddefloop ABCDEFGHIJKLMNOPQRSTUVWXYZ\ddefloop
\def\ddef#1{\expandafter\def\csname #1hat\endcsname{\ensuremath{\widehat{#1}}}}
\ddefloop ABCDEFGHIJKLMNOPQRSTUVWXYZ\ddefloop
\def\ddef#1{\expandafter\def\csname hc#1\endcsname{\ensuremath{\widehat{\mathcal{#1}}}}}
\ddefloop ABCDEFGHIJKLMNOPQRSTUVWXYZ\ddefloop
\def\ddef#1{\expandafter\def\csname #1til\endcsname{\ensuremath{\widetilde{#1}}}}
\ddefloop ABCDEFGHIJKLMNOPQRSTUVWXYZ\ddefloop
\def\ddef#1{\expandafter\def\csname tc#1\endcsname{\ensuremath{\widetilde{\mathcal{#1}}}}}
\ddefloop ABCDEFGHIJKLMNOPQRSTUVWXYZ\ddefloop
\def\ddef#1{\expandafter\def\csname #1Bar\endcsname{\ensuremath{\bar{#1}}}}
\ddefloop ABCDEFGHIJKLMNOPQRSTUVWXYZ\ddefloop

\definecolor{C0}{rgb}{0.121569, 0.466667, 0.705882}
\definecolor{C1}{rgb}{1.000000, 0.498039, 0.054902}
\definecolor{C2}{rgb}{0.172549, 0.627451, 0.172549}
\definecolor{C3}{rgb}{0.839216, 0.152941, 0.156863}
\definecolor{C4}{rgb}{0.580392, 0.403922, 0.741176}
\definecolor{C5}{rgb}{0.549020, 0.337255, 0.294118}
\definecolor{C6}{rgb}{0.890196, 0.466667, 0.760784}
\definecolor{C7}{rgb}{0.498039, 0.498039, 0.498039}
\definecolor{C8}{rgb}{0.737255, 0.741176, 0.133333}
\definecolor{C9}{rgb}{0.090196, 0.745098, 0.811765}

\newcolumntype{Y}{>{\centering\arraybackslash}X}
\newcolumntype{C}{>{\hsize=.0\hsize\centering\arraybackslash}X}

\colorlet{LightGoldenrod}{White!40!Goldenrod}
\colorlet{LightGray}{White!90!Periwinkle}
\definecolor{LG}{gray}{0.95}

\sethlcolor{LightGoldenrod}

\definecolor{codegreen}{rgb}{0,0.6,0}
  \definecolor{codegray}{rgb}{0.5,0.5,0.5}
  \definecolor{codepurple}{rgb}{0.58,0,0.82}
  \definecolor{backcolour}{rgb}{0.95,0.95,0.92}
  \lstdefinestyle{mystyle}{
    backgroundcolor=\color{backcolour},
    commentstyle=\color{codegreen},
    keywordstyle=\color{magenta},
    numberstyle=\tiny\color{codegray},
    stringstyle=\color{codepurple},
    basicstyle=\ttfamily\footnotesize,
    breakatwhitespace=false,
    breaklines=true,
    captionpos=b,
    keepspaces=true,
    numbers=left,
    numbersep=5pt,
    showspaces=false,
    showstringspaces=false,
    showtabs=false,
    tabsize=2
  }
  
  \lstnewenvironment{mylisting}{\lstset{style=mystyle}}{}

\newcommand{\mycc}{\cellcolor{LightGray}}

\newcommand{\edit}[1]{{#1}}

\newcommand{\wcfg}{w}
\newcommand{\gcfg}{\gamma}

\newcommand{\pred}{D_{\mtheta}(\vz_t, t, \vy)}
\newcommand{\prednoise}{\mepsilon_{\mtheta}(\vz_t, t, \vy)}
\newcommand{\predv}{\vv_{\mtheta}(\vz_t, t, \vy)}
\newcommand{\prednull}{D_{\mtheta}(\vz_t, t,\vy_{\textnormal{null}})}
\newcommand{\dpred}{\Delta D_t}
\newcommand{\dpredpar}{\Delta D_t^{\myparallel}}
\newcommand{\dpredorth}{\Delta D_t^{\perp}}
\newcommand{\predcfg}{\hat{D}_{\textrm{CFG}}(\vz_t, t, \vy)}

\title{Eliminating Oversaturation and Artifacts of High Guidance Scales in Diffusion Models}

\author{Seyedmorteza Sadat\textsuperscript{1}, Otmar Hilliges\textsuperscript{1}, Romann M.\ Weber\textsuperscript{2} \\
\textsuperscript{1}ETH Z\"urich, \textsuperscript{2}DisneyResearch\textbar{}Studios\\
\texttt{\{seyedmorteza.sadat,otmar.hilliges\}@inf.ethz.ch} \\
\texttt{\{romann.weber\}@disneyresearch.com}
}

\iclrfinalcopy

\newcommand{\figTeaser}{
    \vspace{-0.45cm}
    \begin{figure}[h!]
        \centering
        \begin{minipage}{\textwidth}
            \begin{subfigure}{0.493\textwidth}
                \begin{tikzpicture}
                    \node[anchor=south west, inner sep=0] (image) at (0,0) {
                            \includegraphics[width=\textwidth]{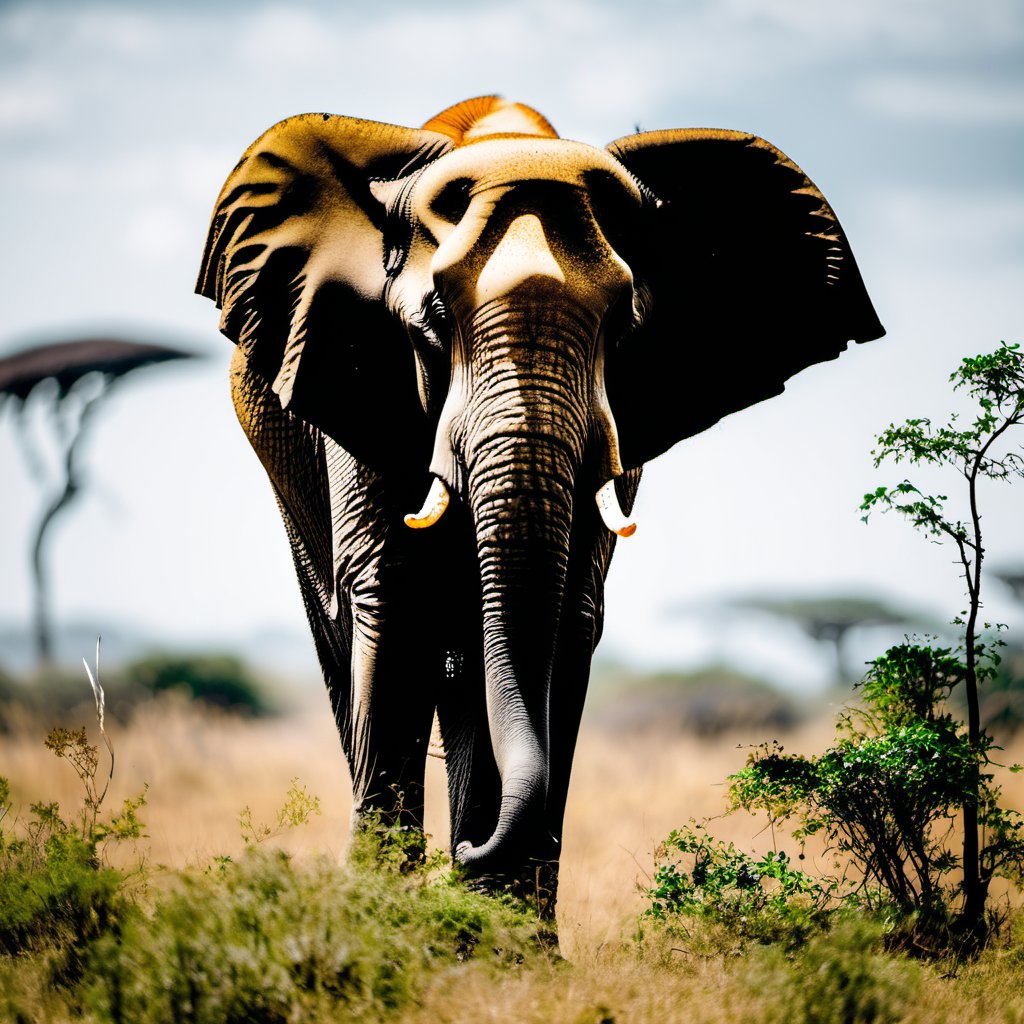}
                    };
                    \node[rotate=0, align=center, font=\normalsize, anchor=base, yshift=4pt] at (image.north) {CFG \strut};
                \end{tikzpicture}
            \end{subfigure}
            \hfill  
            \begin{subfigure}{0.493\textwidth}
                \begin{tikzpicture}
                    \node[anchor=south west, inner sep=0] (image) at (0,0) {
                \includegraphics[width=\textwidth]{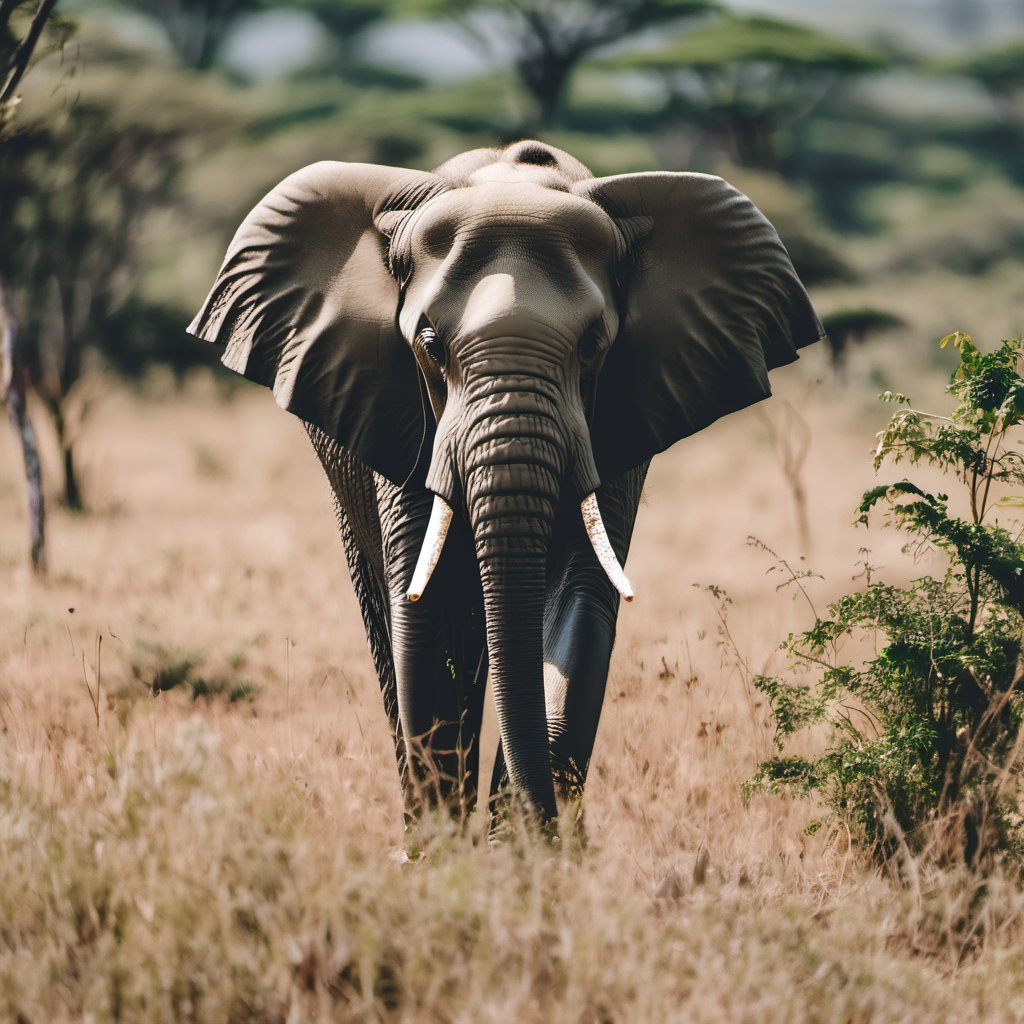}
                    };
                    \node[rotate=0, align=center, font=\normalsize, anchor=base, yshift=4pt] at (image.north) {\acrshort{ncg} (Ours) \strut};
                \end{tikzpicture}
            \end{subfigure}
        \end{minipage}
        \vspace{-0.1cm}
        \caption{Classifier-free guidance is essential for generating high-quality images but causes over-saturation and unrealistic artifacts in the outputs. We introduce \acrshort{ncg}, a novel method that keeps the quality of CFG but significantly reduces its harmful oversaturation. Both images are generated with Stable Diffusion XL \citep{sdxl} and a guidance scale of 15.}
        \label{fig:teaser}
    \vspace{-0.75cm}
    \end{figure}
}

\newcommand{\figMethod}{
    \begin{figure}[t]
    \vspace*{-0.5cm}
        \begin{minipage}{\textwidth}
            \centering
            \begin{subfigure}{0.245\textwidth}
                \includegraphics[width=\textwidth]{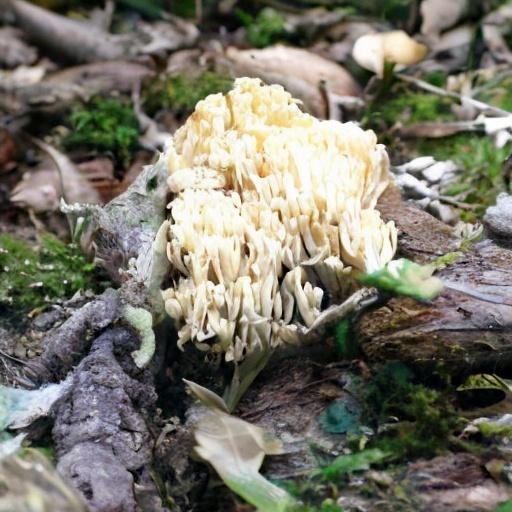}
                \caption{w/o CFG}
            \end{subfigure}
            \begin{subfigure}{0.245\textwidth}
                \includegraphics[width=\textwidth]{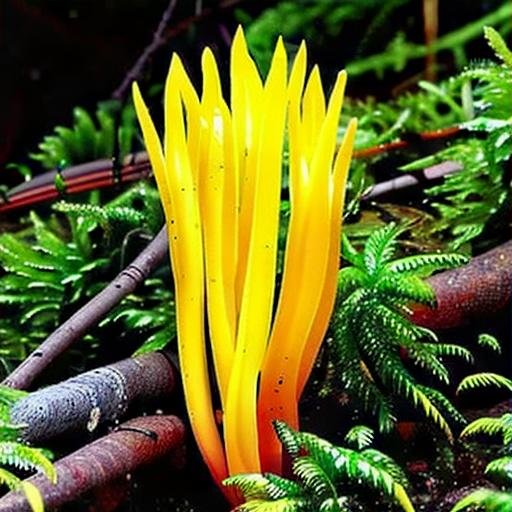}
                \caption{CFG}
            \end{subfigure}
            \begin{subfigure}{0.245\textwidth}
                \includegraphics[width=\textwidth]{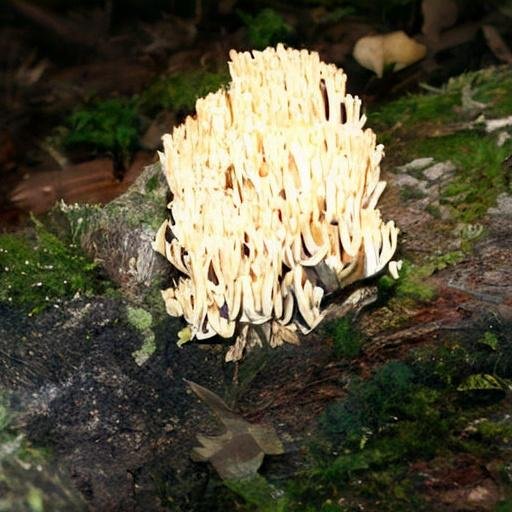}
                \caption{CFG with $\dpredpar$}
            \end{subfigure}
            \begin{subfigure}{0.245\textwidth}
                \includegraphics[width=\textwidth]{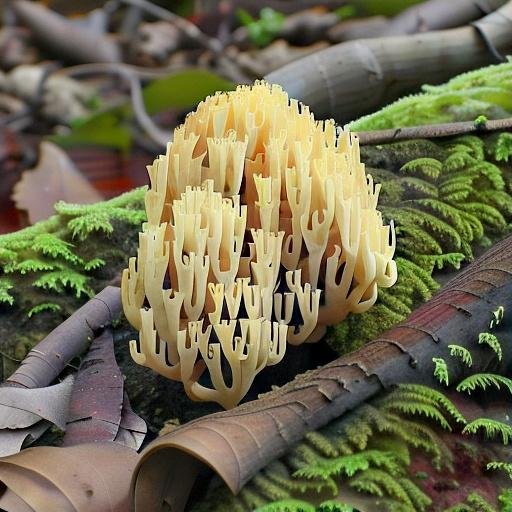}
                \caption{CFG with $\dpredorth$}
            \end{subfigure}
        \end{minipage}
        \caption{Influence of the parallel and orthogonal components ($\dpredpar$ and $\dpredorth$) in CFG. (a) The generation without CFG lacks quality and detail. (b) Applying CFG increases quality but introduces oversaturation. (c) Applying CFG only with the parallel component $\dpredpar$ barely changes the output quality compared to (a) and only increases saturation. (d) Applying CFG with only the orthogonal part $\dpredorth$ enhances image quality without causing oversaturation.}
        \label{fig:method-visual-projection}

        \vspace{0.2cm}

        \begin{minipage}{\textwidth}
            \centering
                \begin{subfigure}[b]{0.325\textwidth}
                    \includegraphics[width=\textwidth]{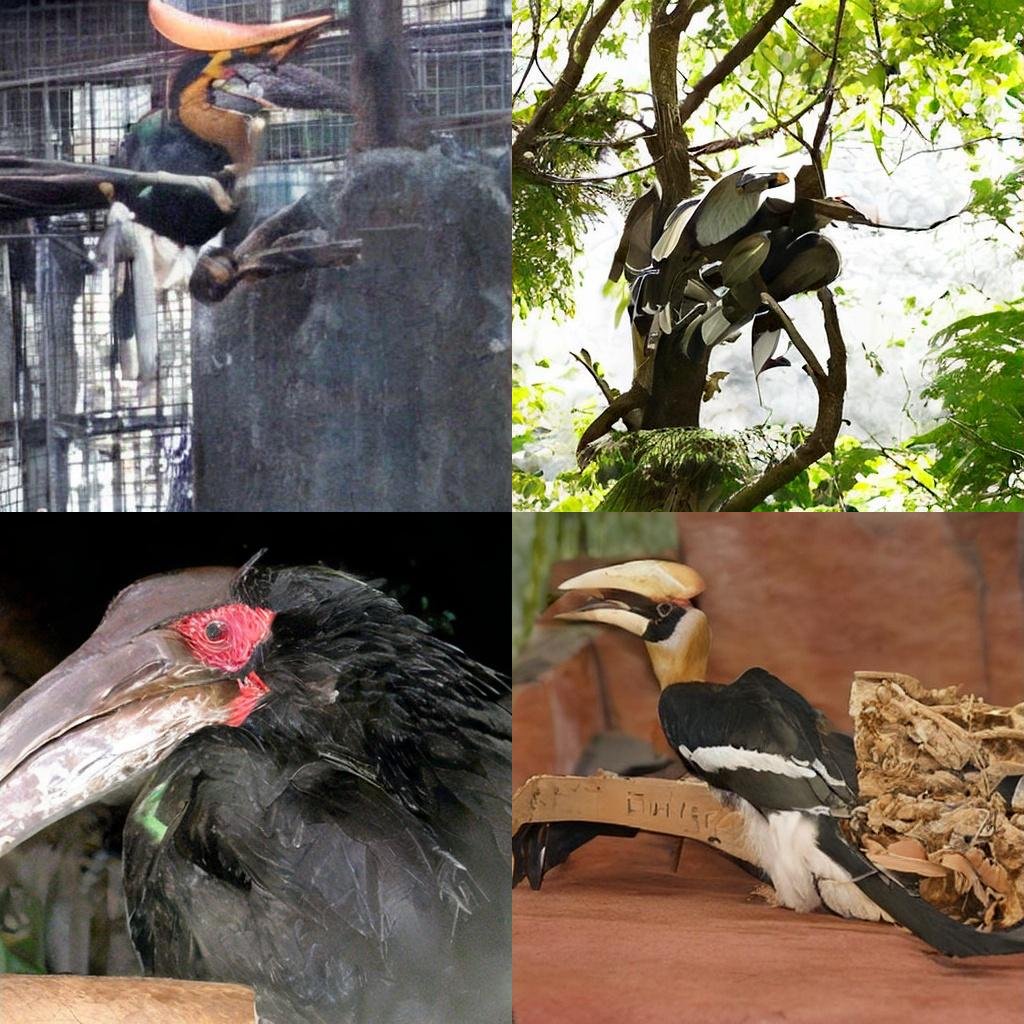}
                    \caption{w/o CFG}
                \end{subfigure}
                \begin{subfigure}[b]{0.325\textwidth}
                    \includegraphics[width=\textwidth]{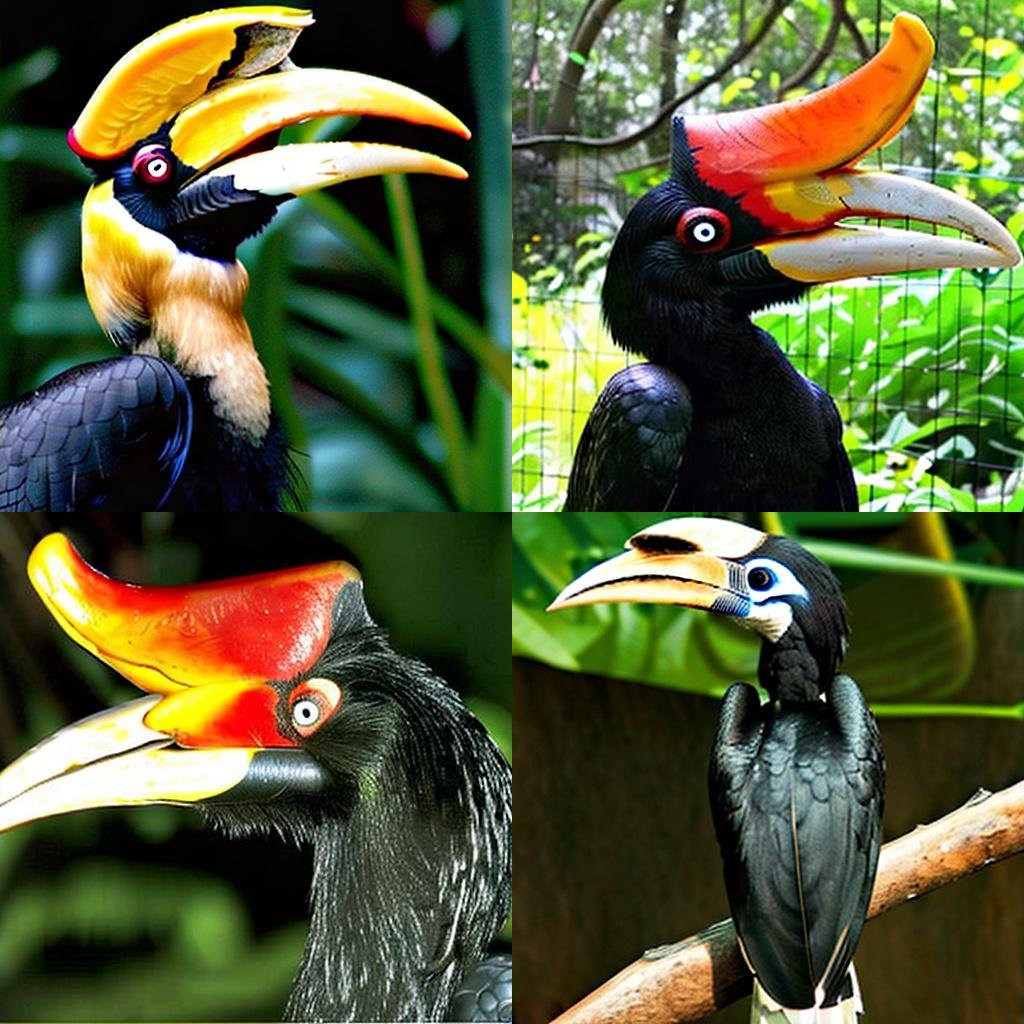}
                    \caption{CFG}
                \end{subfigure}
                \begin{subfigure}[b]{0.325\textwidth}
                    \includegraphics[width=\textwidth]{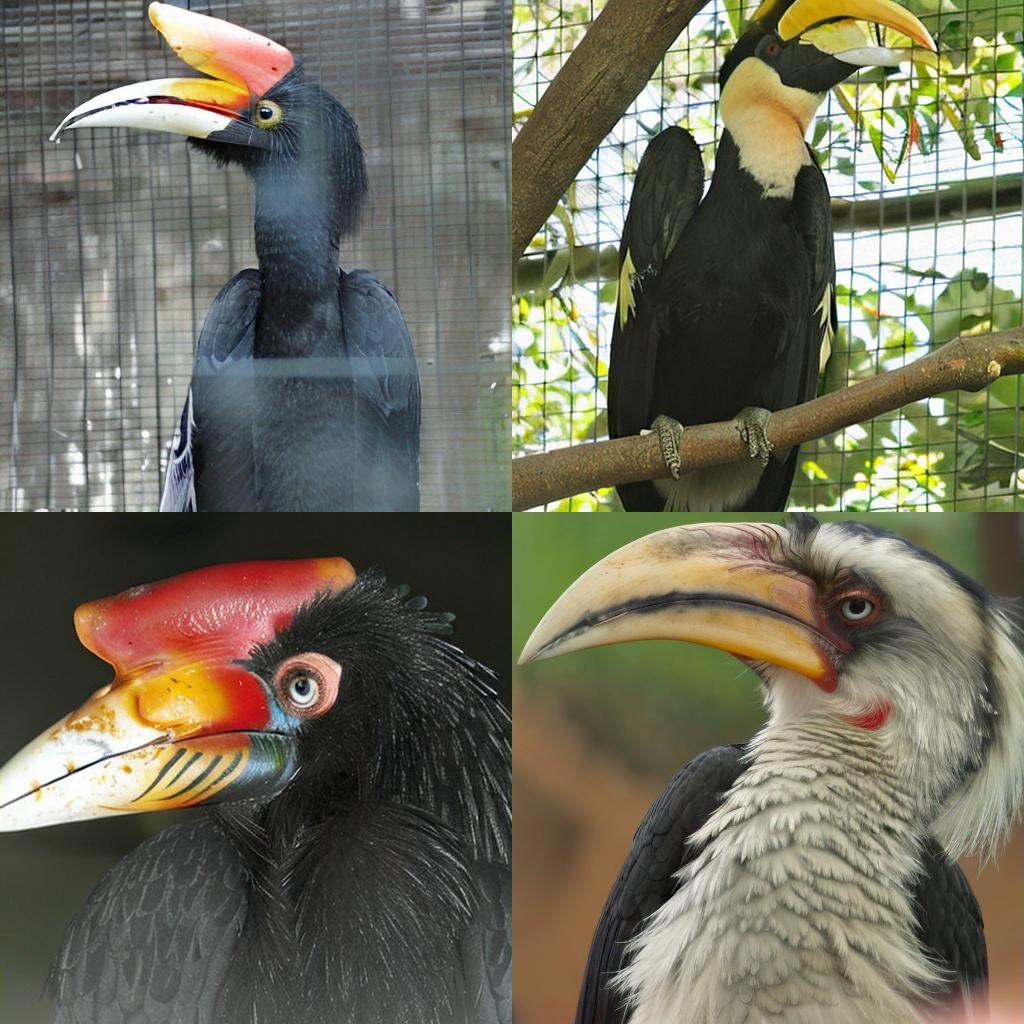}
                    \caption{\gls{ncg} (Ours)}
                \end{subfigure}
            \end{minipage}  
            \caption{Illustrating the effect of \gls{ncg} on generated images . (a) Sampling without guidance leads to low-quality generations. (b) CFG improves image quality but causes oversaturation. (c) Using \gls{ncg} instead of CFG results in high-quality generations without oversaturation.}
            \label{fig:method-visual}
            \vspace*{-0.5cm}
    \end{figure}
}

\newcommand{\figDiversity}{
    \begin{figure}[t]
        \centering
        \vspace*{-0.5cm}
        \begin{minipage}{\textwidth}
            \begin{subfigure}{0.495\textwidth}
                \begin{tikzpicture}
                    \node[anchor=south west, inner sep=0] (image) at (0,0) {
                        \includegraphics[width=\textwidth]{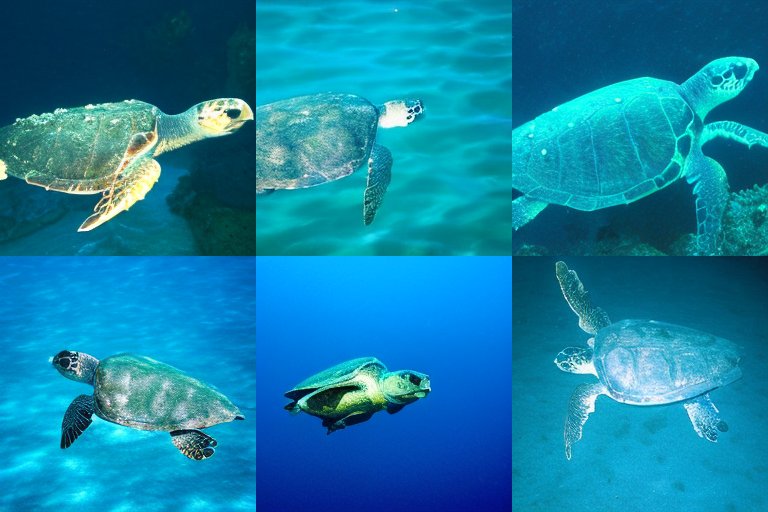}
                    };
                    \node[rotate=0, align=center, font=\normalsize, anchor=base, yshift=4pt] at (image.north) {CFG \strut};
                \end{tikzpicture}
            \end{subfigure}
            \begin{subfigure}{0.495\textwidth}
                \begin{tikzpicture}
                    \node[anchor=south west, inner sep=0] (image) at (0,0) {
                        \includegraphics[width=\textwidth]{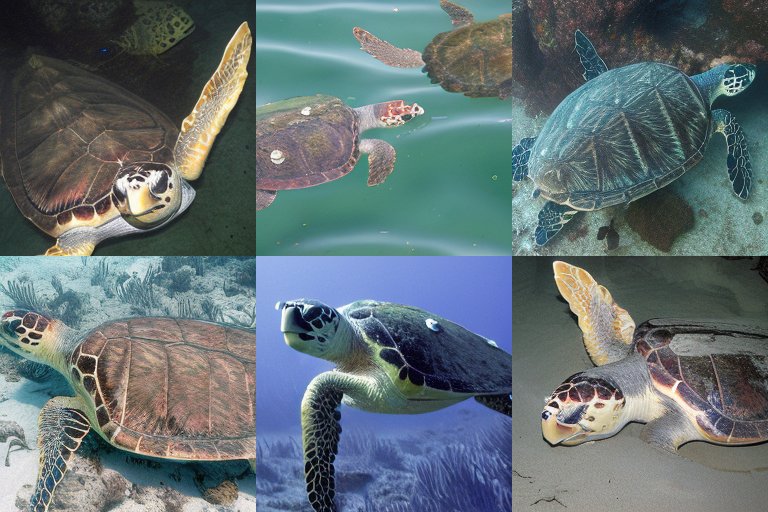}
                    };
                    \node[rotate=0, align=center, font=\normalsize, anchor=base, yshift=4pt] at (image.north) {\gls{ncg} (Ours) \strut};
                \end{tikzpicture}
            \end{subfigure}
            \begin{subfigure}{0.495\textwidth}
                \begin{tikzpicture}
                    \node[anchor=south west, inner sep=0] (image) at (0,0) {
                        \includegraphics[width=\textwidth]{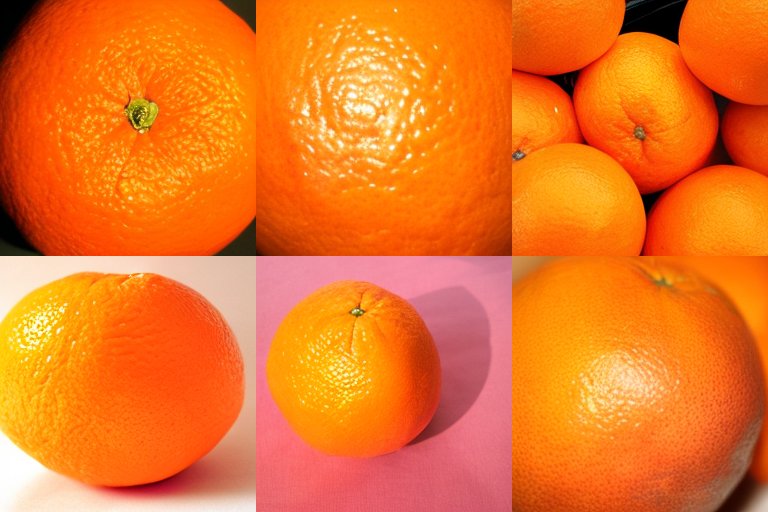}
                    };
                    \node[rotate=0, align=center, font=\normalsize, anchor=base, yshift=4pt] at (image.north) {CFG \strut};
                \end{tikzpicture}
            \end{subfigure}
            \begin{subfigure}{0.495\textwidth}
                \begin{tikzpicture}
                    \node[anchor=south west, inner sep=0] (image) at (0,0) {
                        \includegraphics[width=\textwidth]{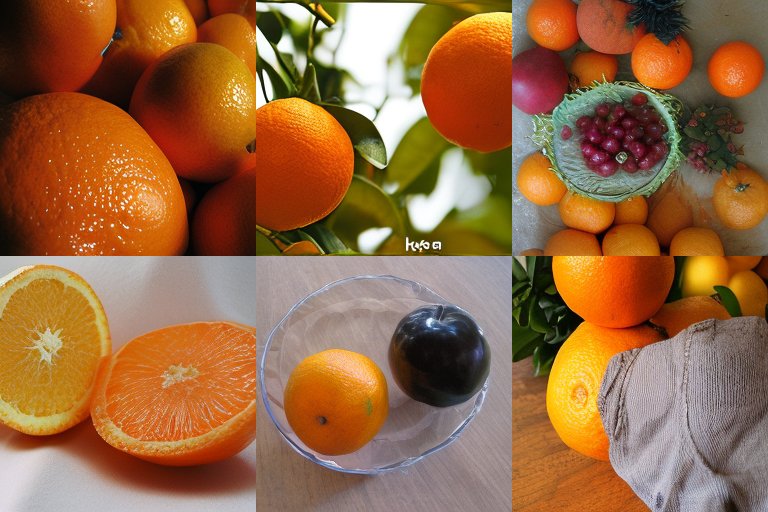}
                    };
                    \node[rotate=0, align=center, font=\normalsize, anchor=base, yshift=4pt] at (image.north) {\gls{ncg} (Ours) \strut};
                \end{tikzpicture}
            \end{subfigure}
        \end{minipage}
        \caption{Showcasing the diversity of generations after using \gls{ncg}. \gls{ncg} removes oversaturation issues while improving diversity w.r.t. the overall image composition.}
        \label{fig:diversity}
    \end{figure}
}

\newcommand{\figEDMMain}{
    \begin{figure}[t]
        \centering
        \vspace*{-0.75cm}
        \begin{minipage}{\textwidth}
            \begin{subfigure}{0.245\textwidth}
                \begin{tikzpicture}
                    \node[anchor=south west, inner sep=0] (image) at (0,0) {
                        \includegraphics[width=\textwidth]{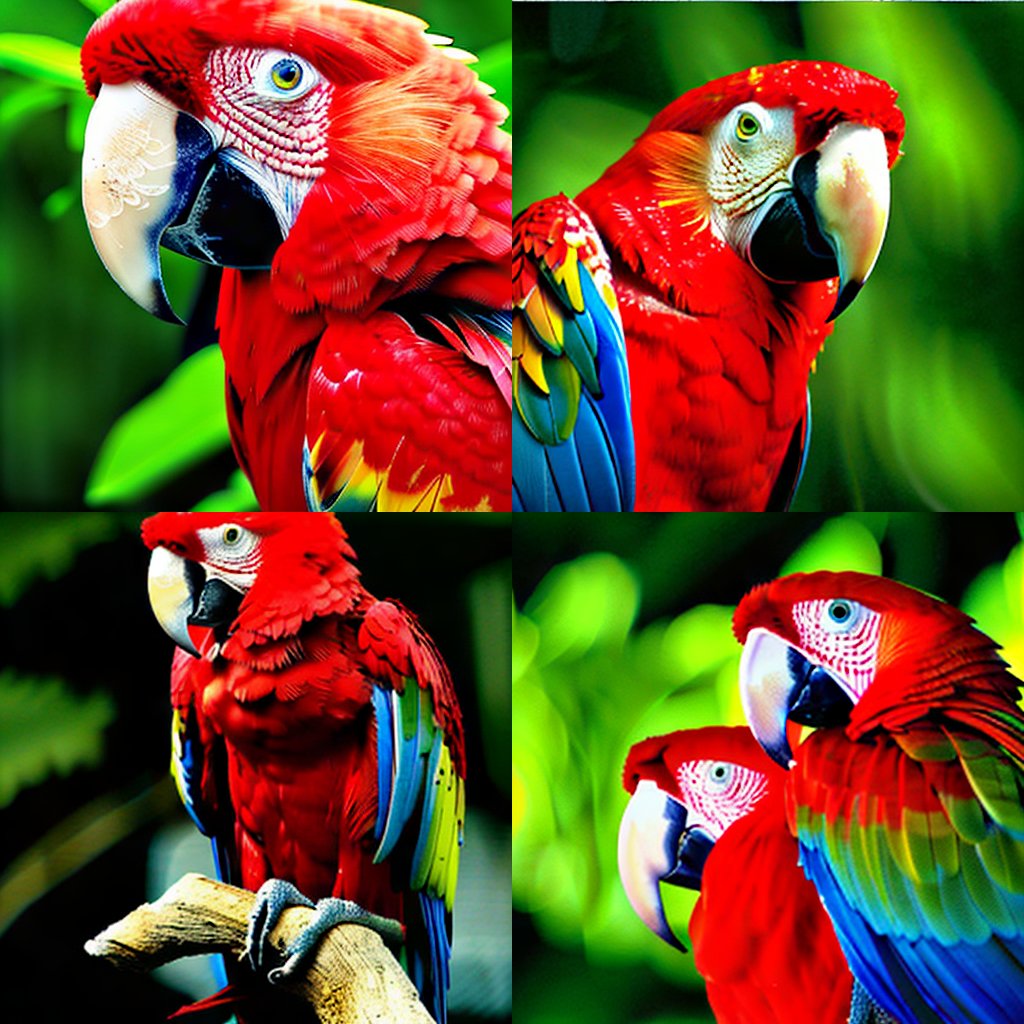}
                    };
                    \node[rotate=0, align=center, font=\normalsize, anchor=base, yshift=4pt] at (image.north) {CFG \strut};
                \end{tikzpicture}
            \end{subfigure}
            \begin{subfigure}{0.245\textwidth}
                \begin{tikzpicture}
                    \node[anchor=south west, inner sep=0] (image) at (0,0) {
                        \includegraphics[width=\textwidth]{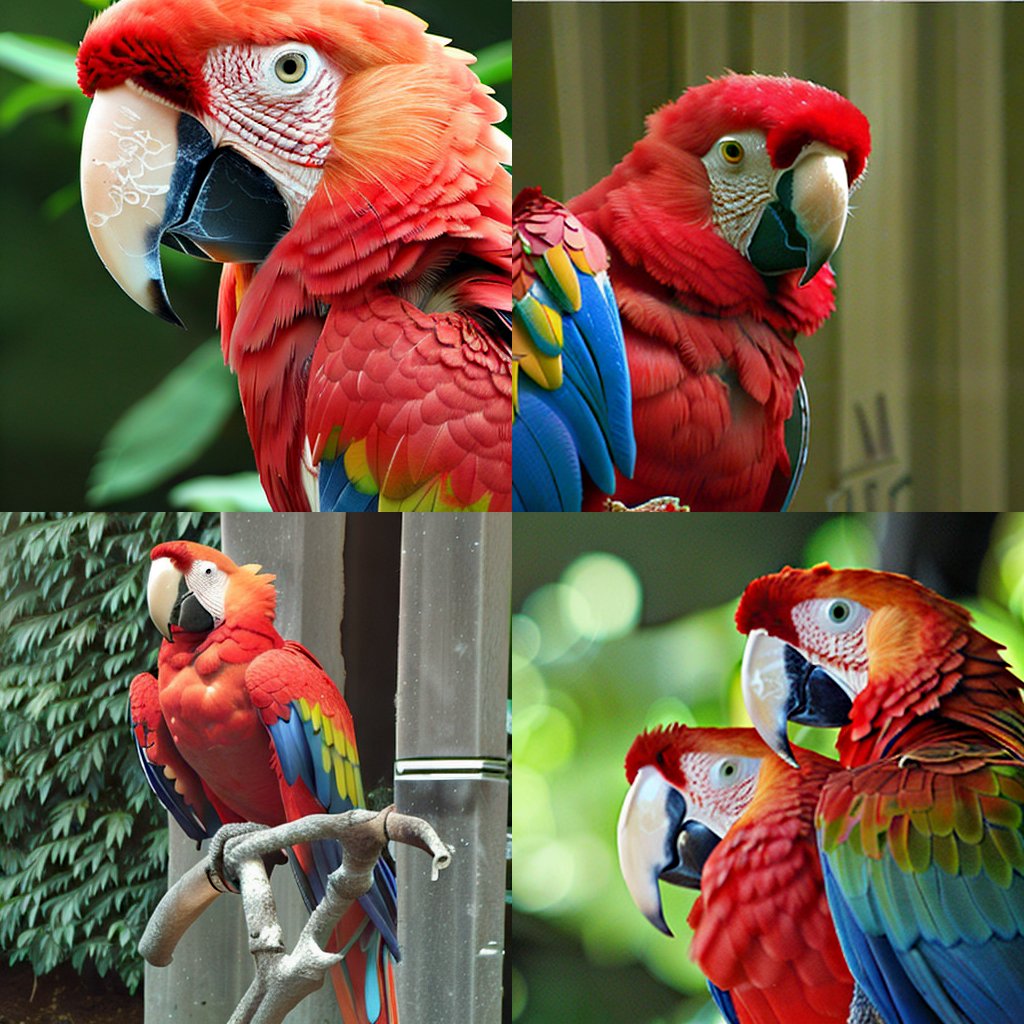}
                    };
                    \node[rotate=0, align=center, font=\normalsize, anchor=base, yshift=4pt] at (image.north) {\gls{ncg} (Ours) \strut};
                \end{tikzpicture}
            \end{subfigure}
            \begin{subfigure}{0.245\textwidth}
                \begin{tikzpicture}
                    \node[anchor=south west, inner sep=0] (image) at (0,0) {
                        \includegraphics[width=\textwidth]{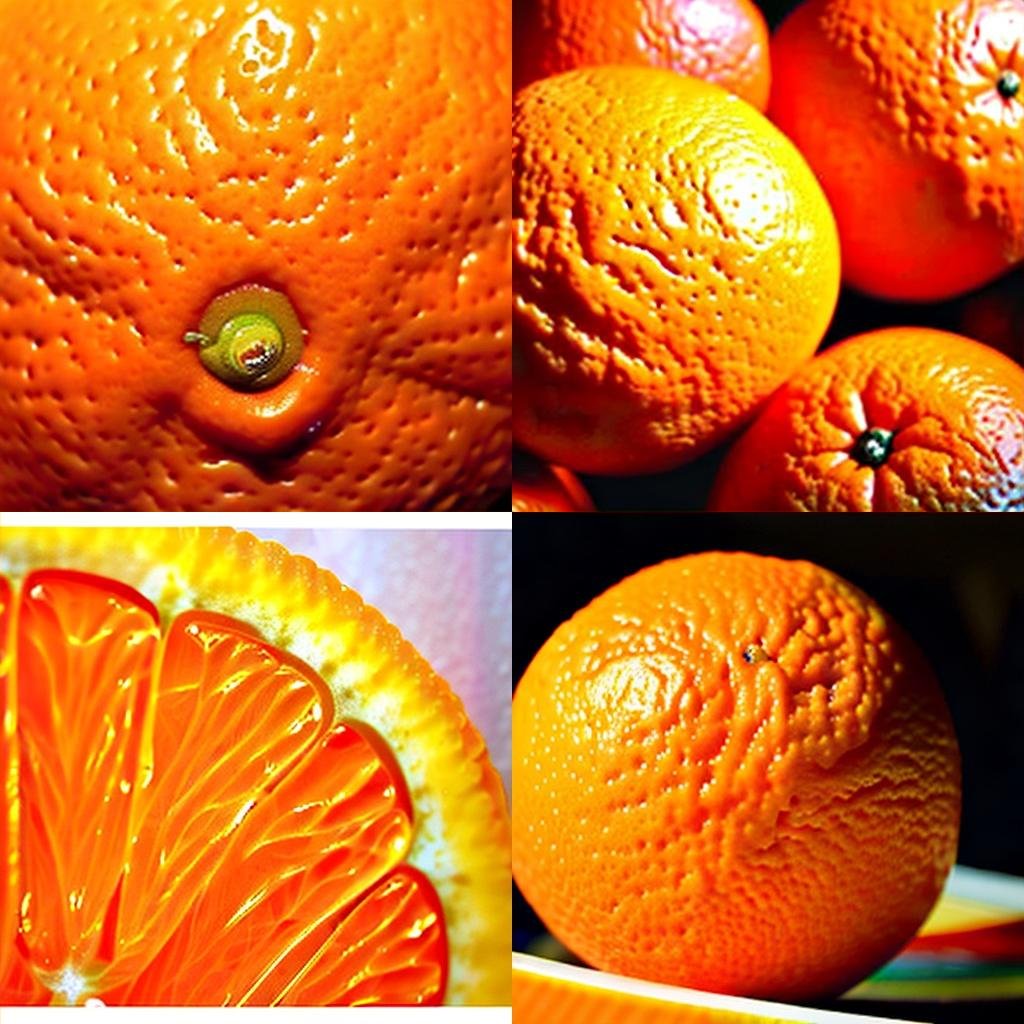}
                    };
                    \node[rotate=0, align=center, font=\normalsize, anchor=base, yshift=4pt] at (image.north) {CFG \strut};
                \end{tikzpicture}
            \end{subfigure}
            \begin{subfigure}{0.245\textwidth}
                \begin{tikzpicture}
                    \node[anchor=south west, inner sep=0] (image) at (0,0) {
                        \includegraphics[width=\textwidth]{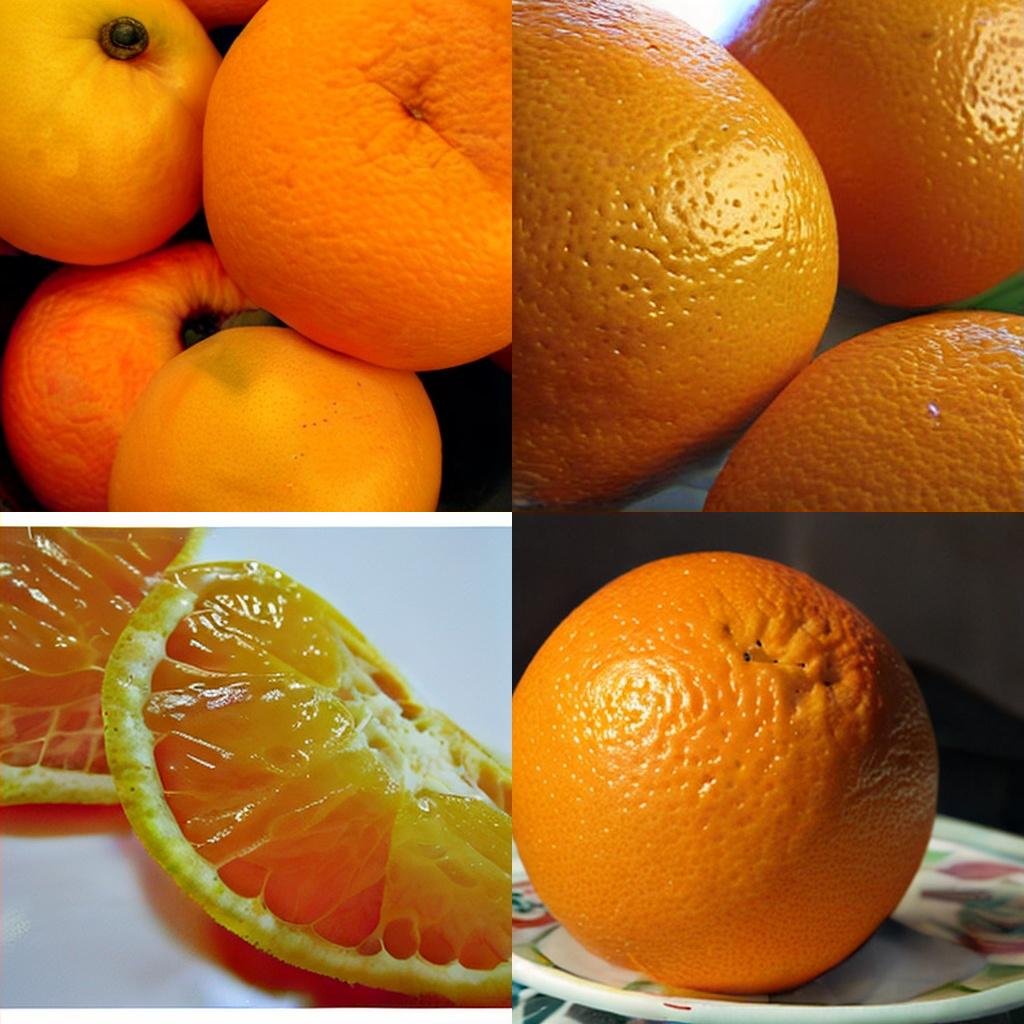}
                    };
                    \node[rotate=0, align=center, font=\normalsize, anchor=base, yshift=4pt] at (image.north) {\gls{ncg} (Ours) \strut};
                \end{tikzpicture}
            \end{subfigure}
            \begin{subfigure}{0.245\textwidth}
                \begin{tikzpicture}
                    \node[anchor=south west, inner sep=0] (image) at (0,0) {
                        \includegraphics[width=\textwidth]{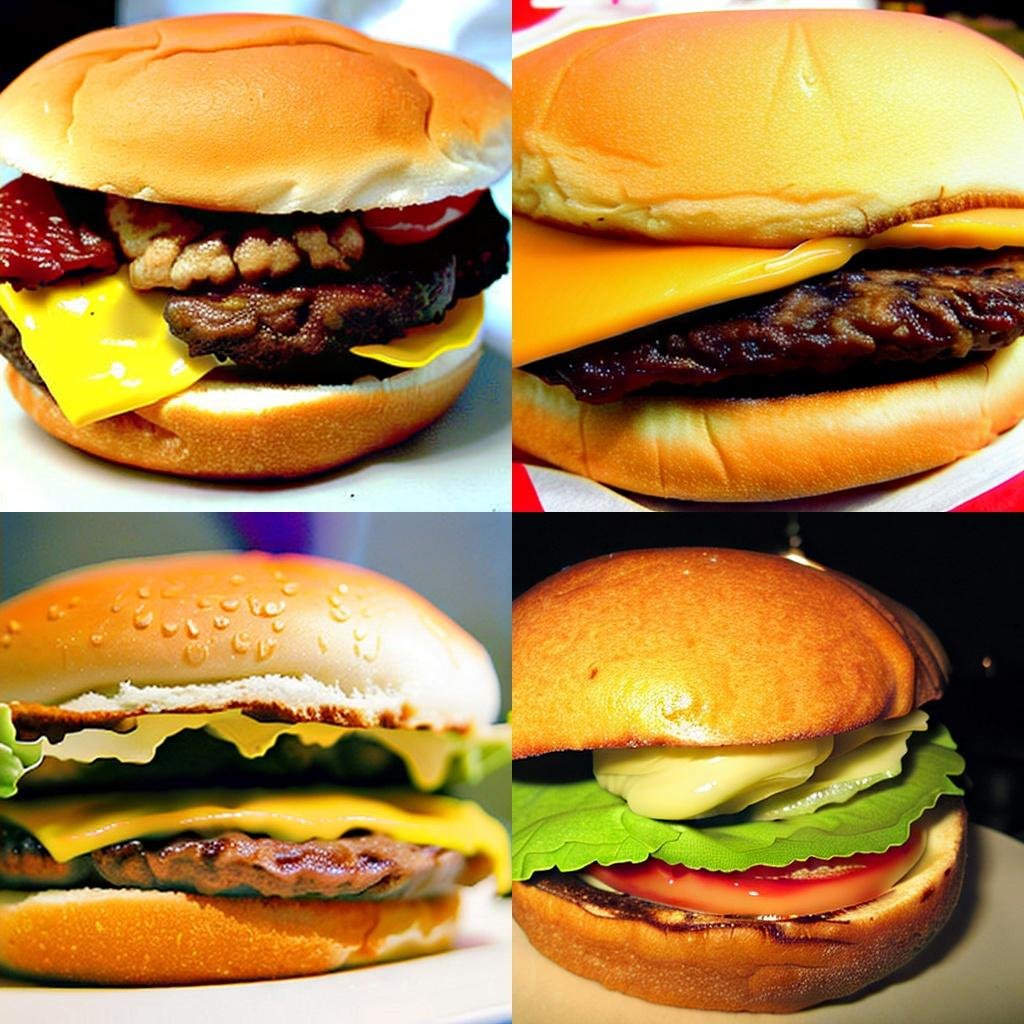}
                    };
                    \node[rotate=0, align=center, font=\normalsize, anchor=base, yshift=4pt] at (image.north) {CFG \strut};
                \end{tikzpicture}
            \end{subfigure}
            \begin{subfigure}{0.245\textwidth}
                \begin{tikzpicture}
                    \node[anchor=south west, inner sep=0] (image) at (0,0) {
                        \includegraphics[width=\textwidth]{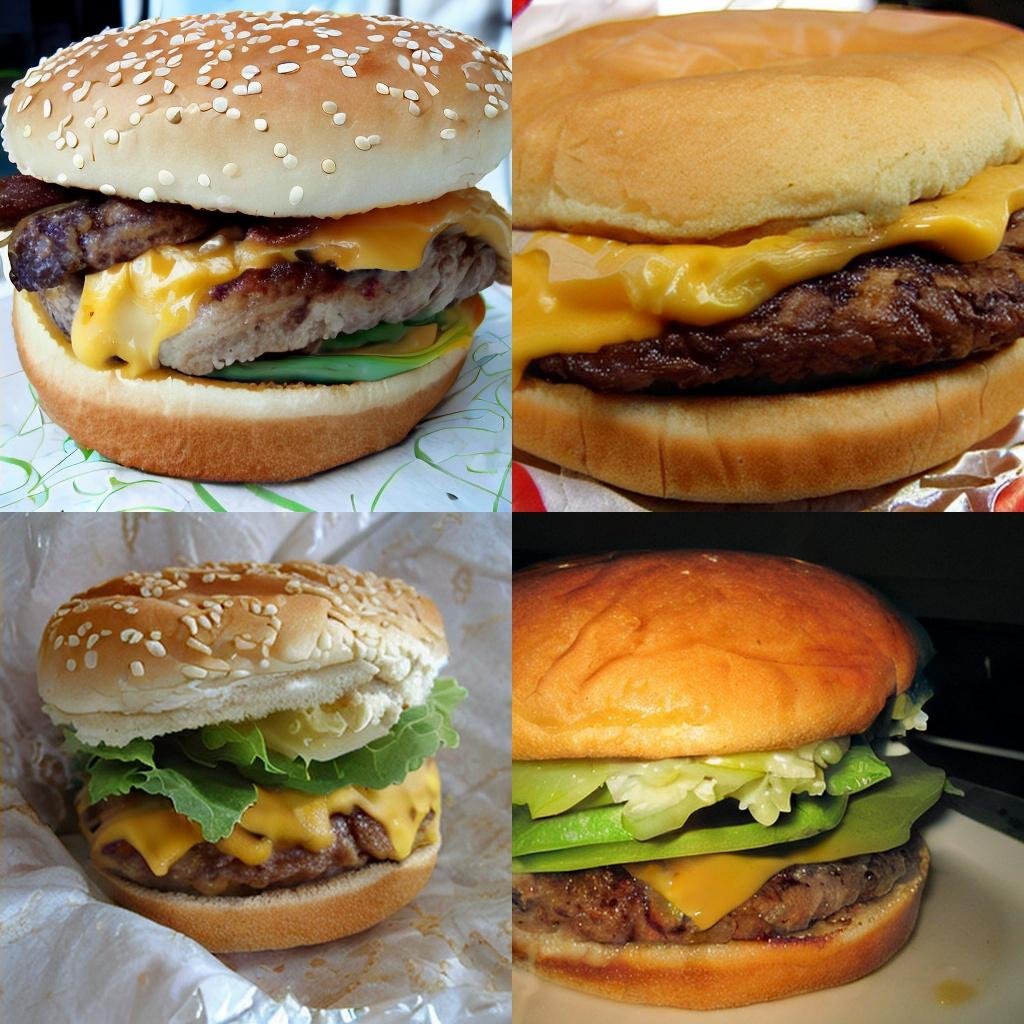}
                    };
                    \node[rotate=0, align=center, font=\normalsize, anchor=base, yshift=4pt] at (image.north) {\gls{ncg} (Ours) \strut};
                \end{tikzpicture}
            \end{subfigure}
            \begin{subfigure}{0.245\textwidth}
                \begin{tikzpicture}
                    \node[anchor=south west, inner sep=0] (image) at (0,0) {
                        \includegraphics[width=\textwidth]{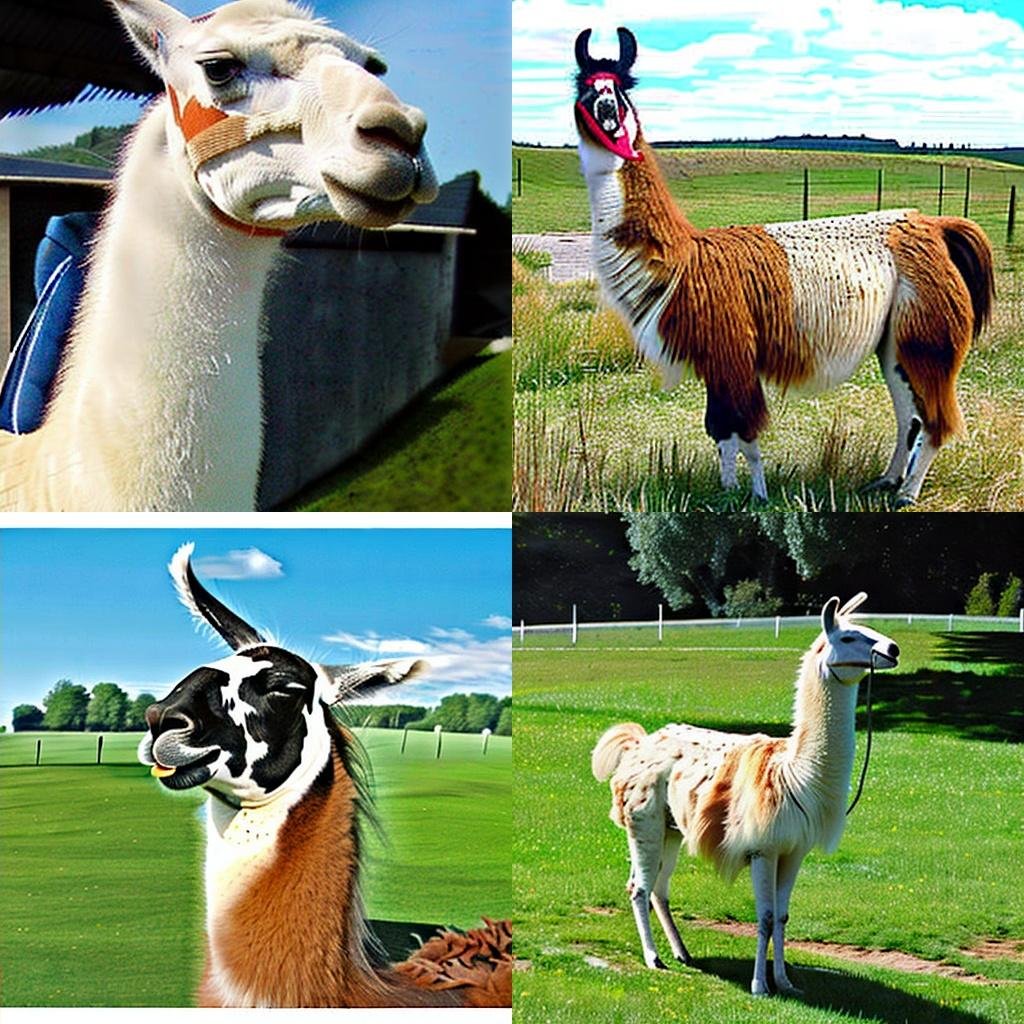}
                    };
                    \node[rotate=0, align=center, font=\normalsize, anchor=base, yshift=4pt] at (image.north) {CFG \strut};
                \end{tikzpicture}
            \end{subfigure}
            \begin{subfigure}{0.245\textwidth}
                \begin{tikzpicture}
                    \node[anchor=south west, inner sep=0] (image) at (0,0) {
                        \includegraphics[width=\textwidth]{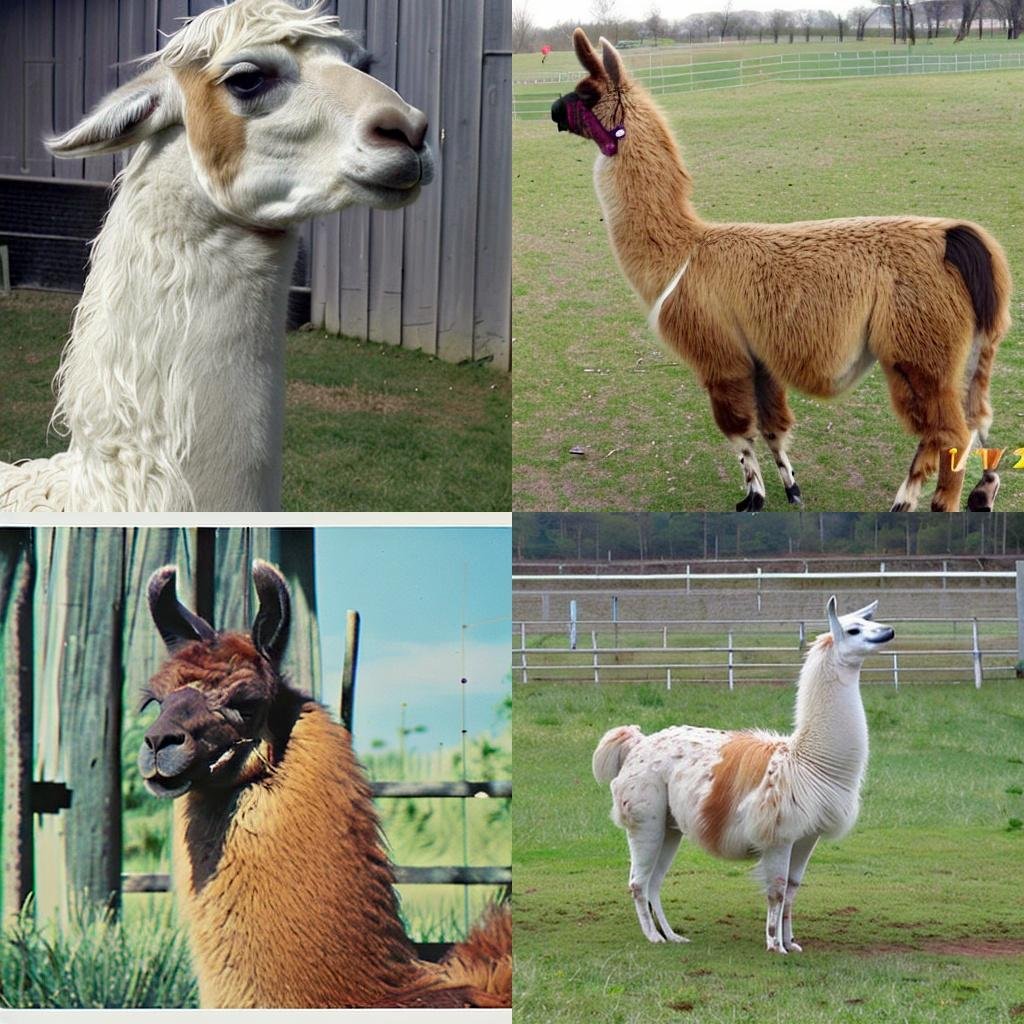}
                    };
                    \node[rotate=0, align=center, font=\normalsize, anchor=base, yshift=4pt] at (image.north) {\gls{ncg} (Ours) \strut};
                \end{tikzpicture}
            \end{subfigure}
        \end{minipage}
        \caption{Class-conditional generation results using EDM2 with $\wcfg=4$. \gls{ncg} significantly reduces saturation in the generations while keeping the high-quality global structure of each image.}
        \label{fig:edm-main}

        \begin{minipage}{\textwidth}
            \begin{subfigure}{0.245\textwidth}
                \begin{tikzpicture}
                    \node[anchor=south west, inner sep=0] (image) at (0,0) {
                        \includegraphics[width=\textwidth]{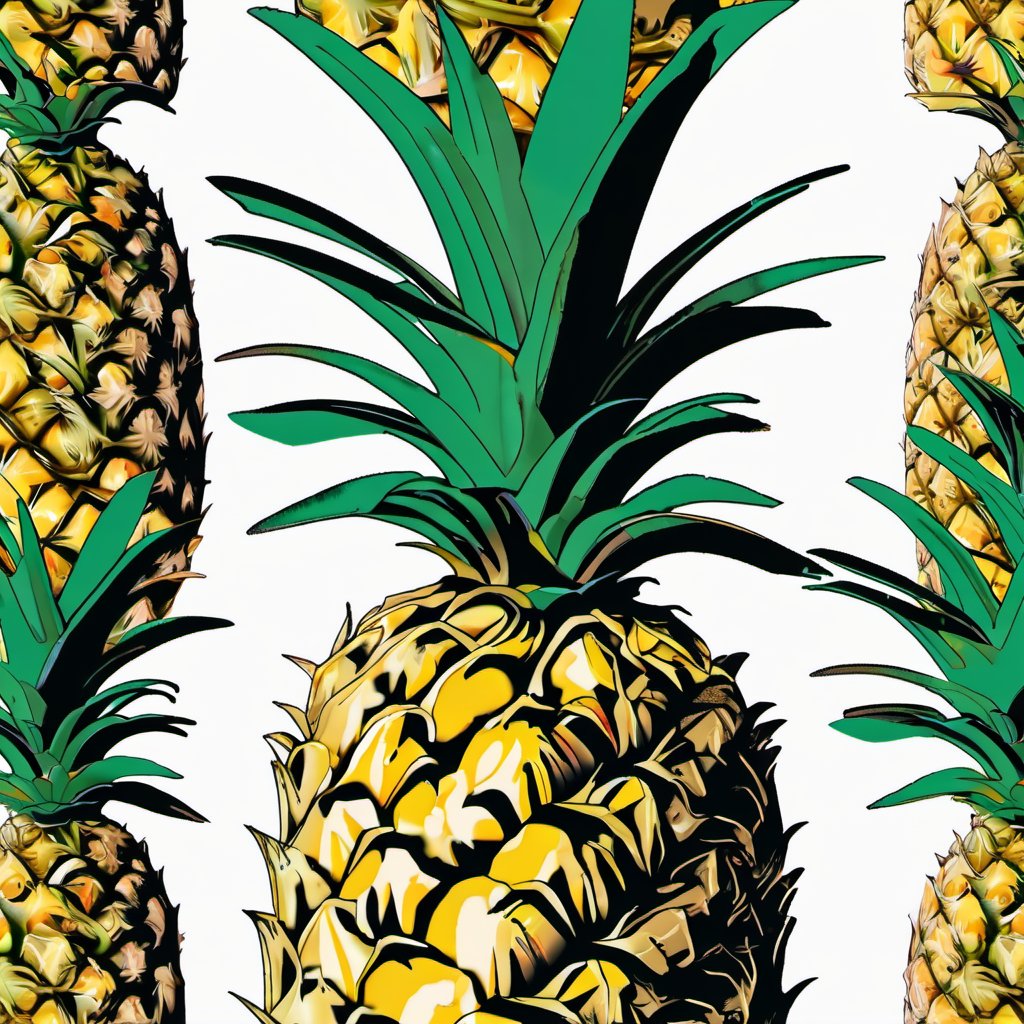}
                    };
                    \node[rotate=0, align=center, font=\normalsize, anchor=base, yshift=4pt] at (image.north) {CFG \strut};
                \end{tikzpicture}
            \end{subfigure}
            \begin{subfigure}{0.245\textwidth}
                \begin{tikzpicture}
                    \node[anchor=south west, inner sep=0] (image) at (0,0) {
                        \includegraphics[width=\textwidth]{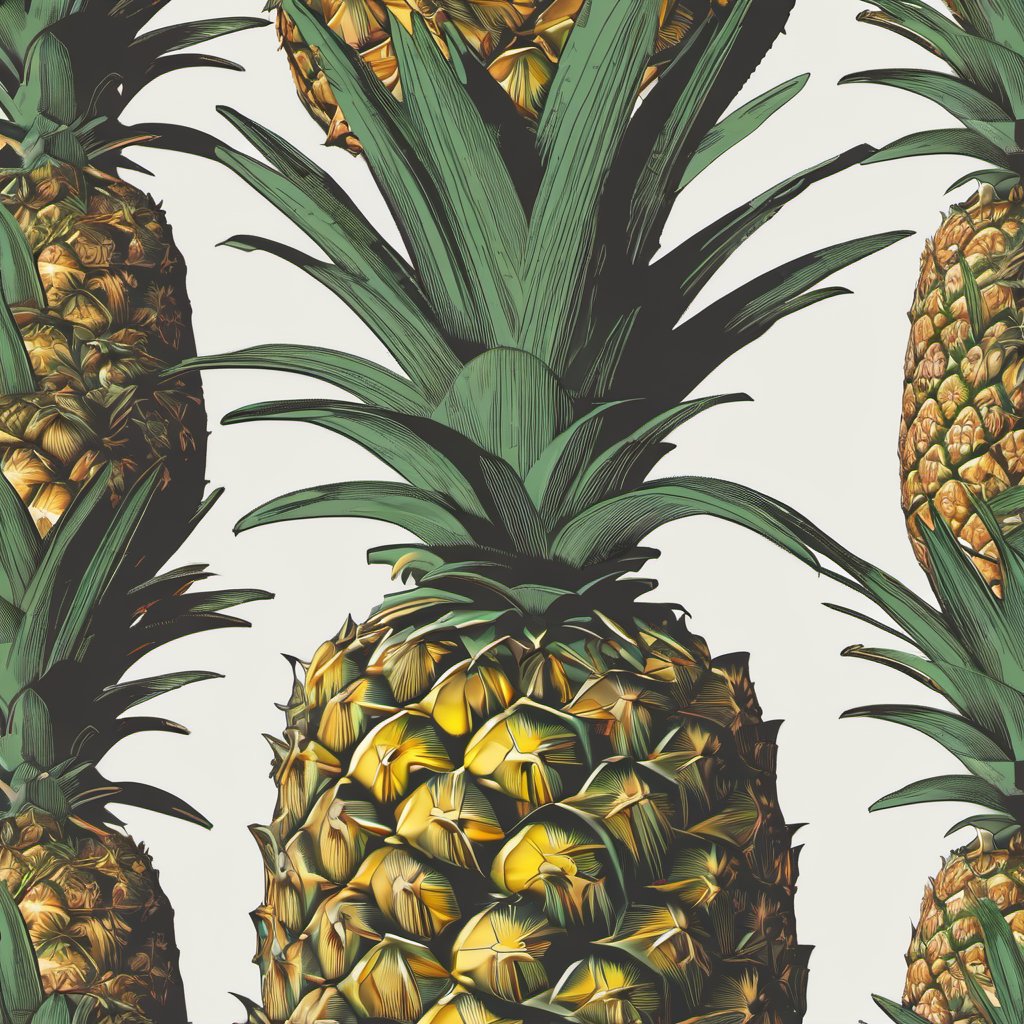}
                    };
                    \node[rotate=0, align=center, font=\normalsize, anchor=base, yshift=4pt] at (image.north) {\gls{ncg} (Ours) \strut};
                \end{tikzpicture}
            \end{subfigure}
            \begin{subfigure}{0.245\textwidth}
                \begin{tikzpicture}
                    \node[anchor=south west, inner sep=0] (image) at (0,0) {
                        \includegraphics[width=\textwidth]{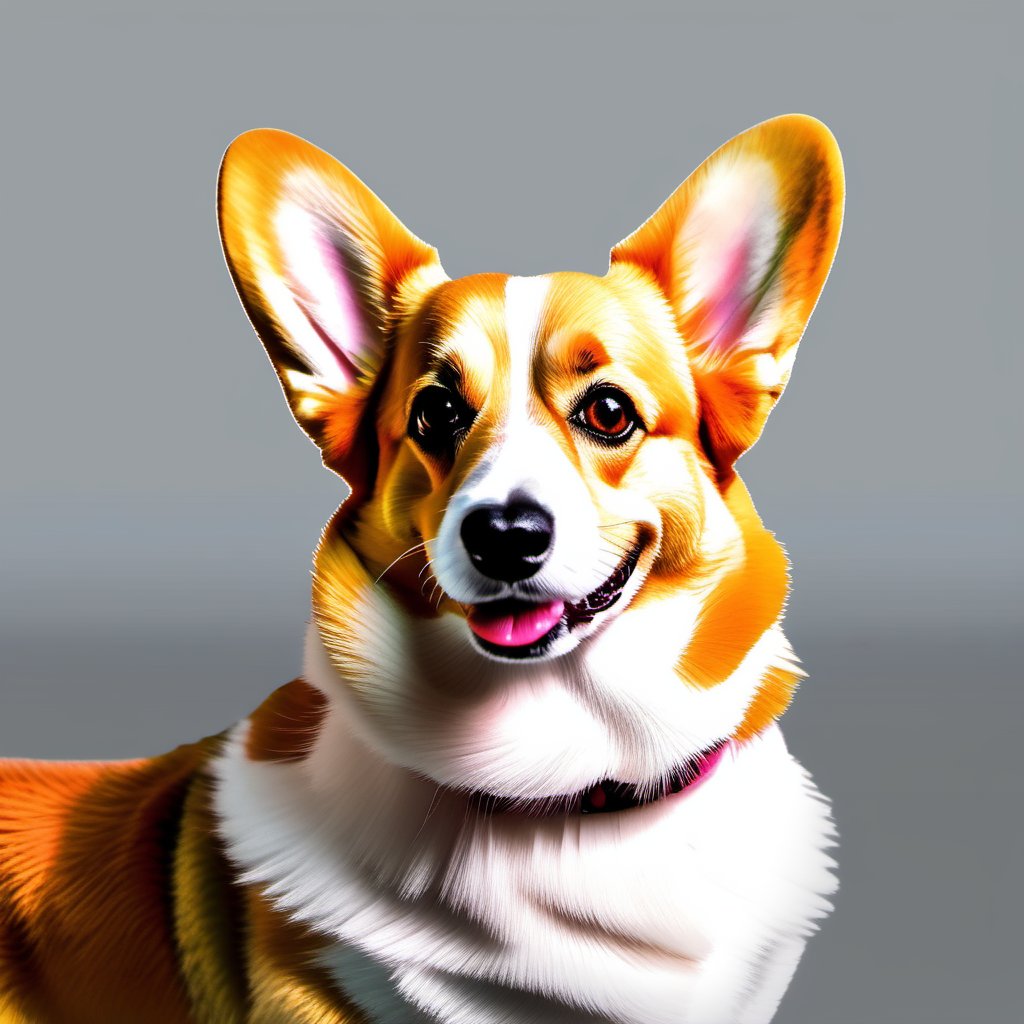}
                    };
                    \node[rotate=0, align=center, font=\normalsize, anchor=base, yshift=4pt] at (image.north) {CFG \strut};
                \end{tikzpicture}
            \end{subfigure}
            \begin{subfigure}{0.245\textwidth}
                \begin{tikzpicture}
                    \node[anchor=south west, inner sep=0] (image) at (0,0) {
                        \includegraphics[width=\textwidth]{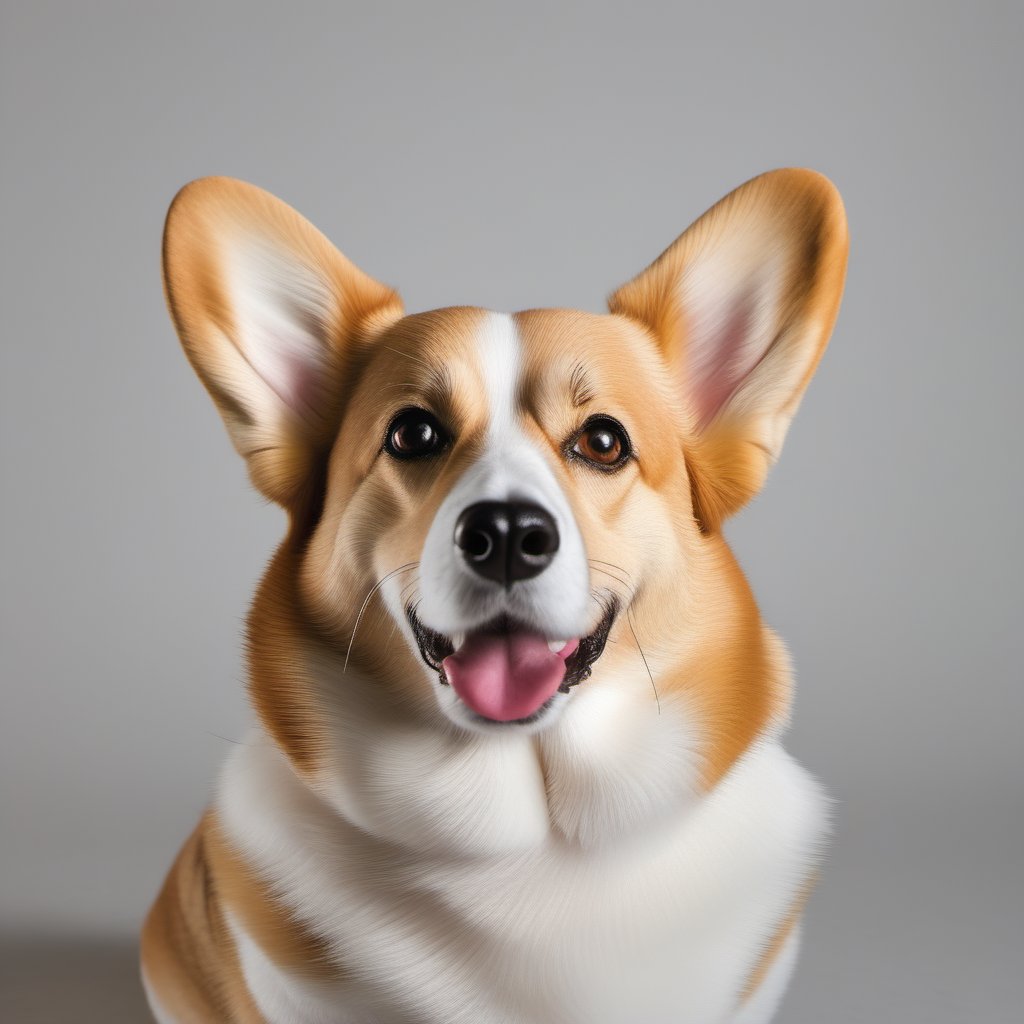}
                    };
                    \node[rotate=0, align=center, font=\normalsize, anchor=base, yshift=4pt] at (image.north) {\gls{ncg} (Ours) \strut};
                \end{tikzpicture}
            \end{subfigure}
            \begin{subfigure}{0.245\textwidth}
                \begin{tikzpicture}
                    \node[anchor=south west, inner sep=0] (image) at (0,0) {
                        \includegraphics[width=\textwidth]{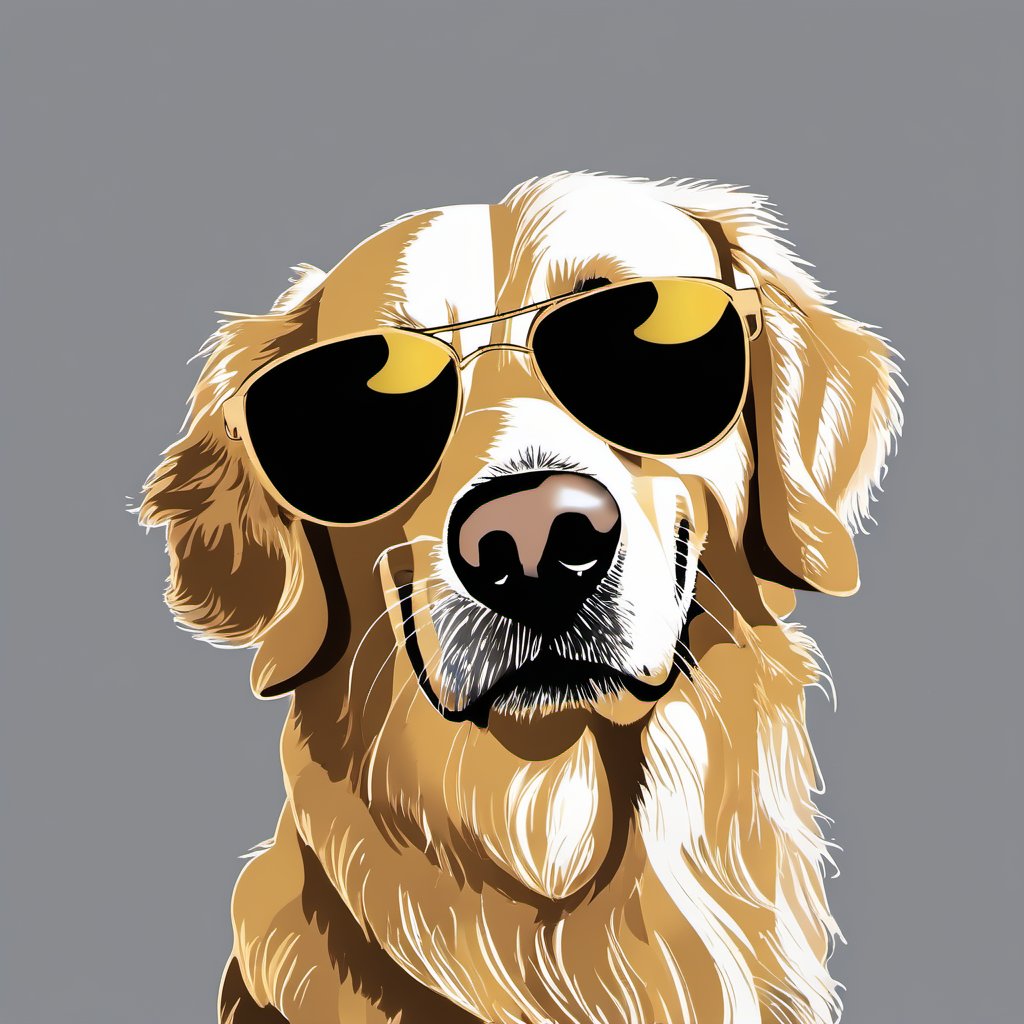}
                    };
                    \node[rotate=0, align=center, font=\normalsize, anchor=base, yshift=4pt] at (image.north) {CFG \strut};
                \end{tikzpicture}
            \end{subfigure}
            \begin{subfigure}{0.245\textwidth}
                \begin{tikzpicture}
                    \node[anchor=south west, inner sep=0] (image) at (0,0) {
                        \includegraphics[width=\textwidth]{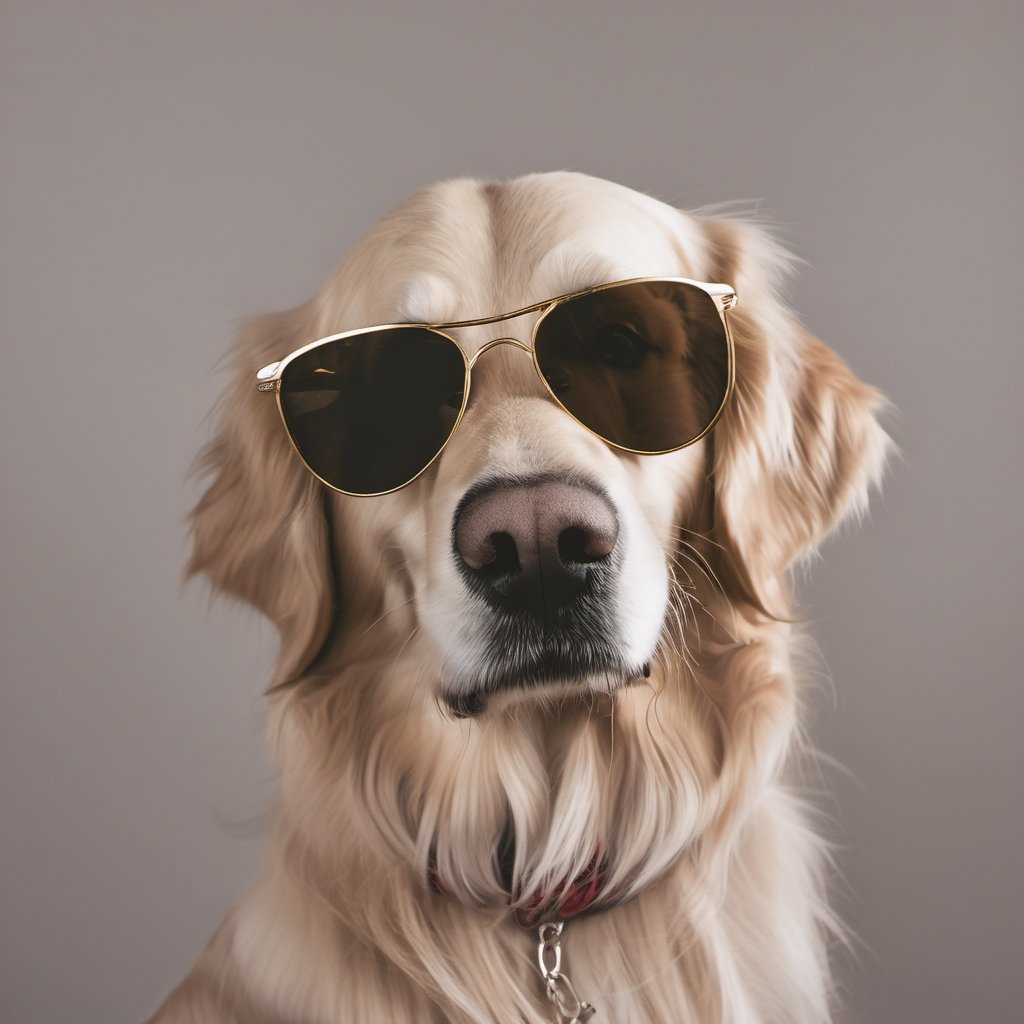}
                    };
                    \node[rotate=0, align=center, font=\normalsize, anchor=base, yshift=4pt] at (image.north) {\gls{ncg} (Ours) \strut};
                \end{tikzpicture}
            \end{subfigure}
            \begin{subfigure}{0.245\textwidth}
                \begin{tikzpicture}
                    \node[anchor=south west, inner sep=0] (image) at (0,0) {
                        \includegraphics[width=\textwidth]{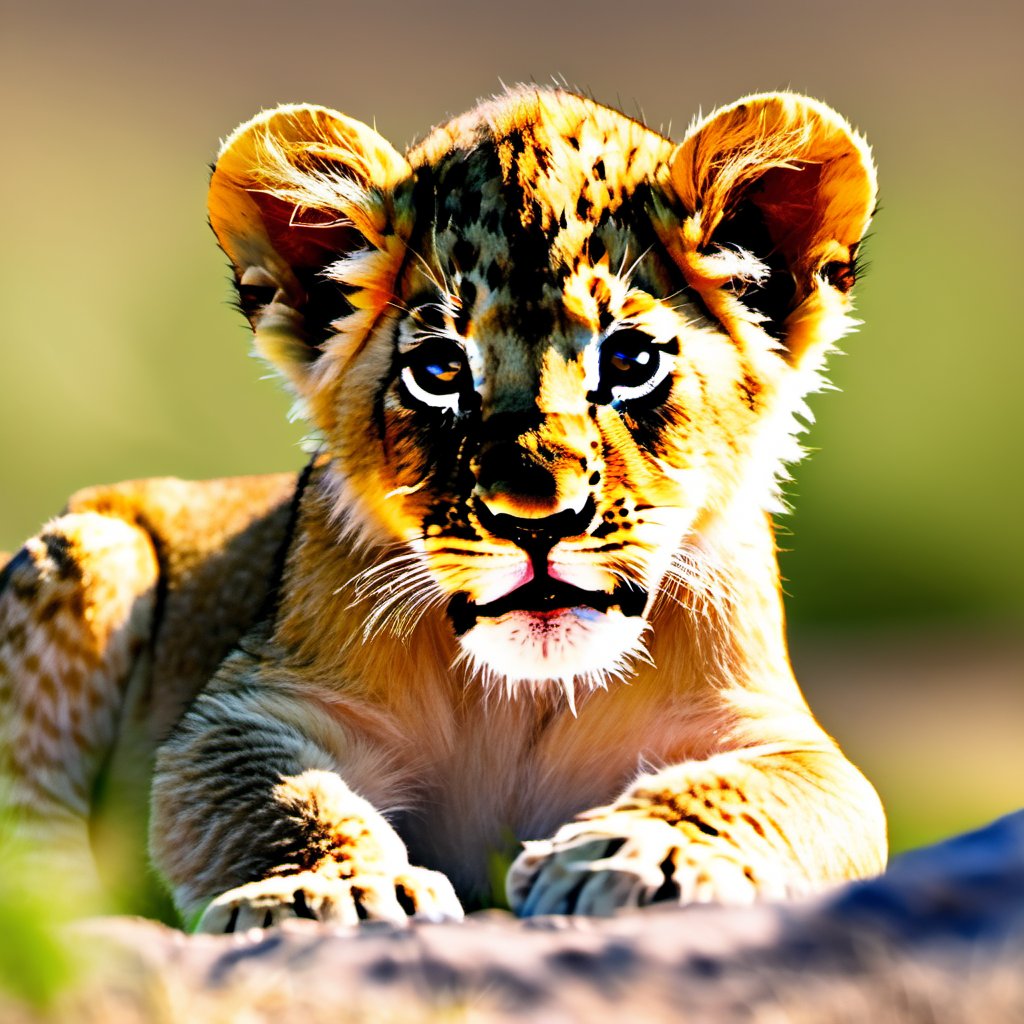}
                    };
                    \node[rotate=0, align=center, font=\normalsize, anchor=base, yshift=4pt] at (image.north) {CFG \strut};
                \end{tikzpicture}
            \end{subfigure}
            \begin{subfigure}{0.245\textwidth}
                \begin{tikzpicture}
                    \node[anchor=south west, inner sep=0] (image) at (0,0) {
                        \includegraphics[width=\textwidth]{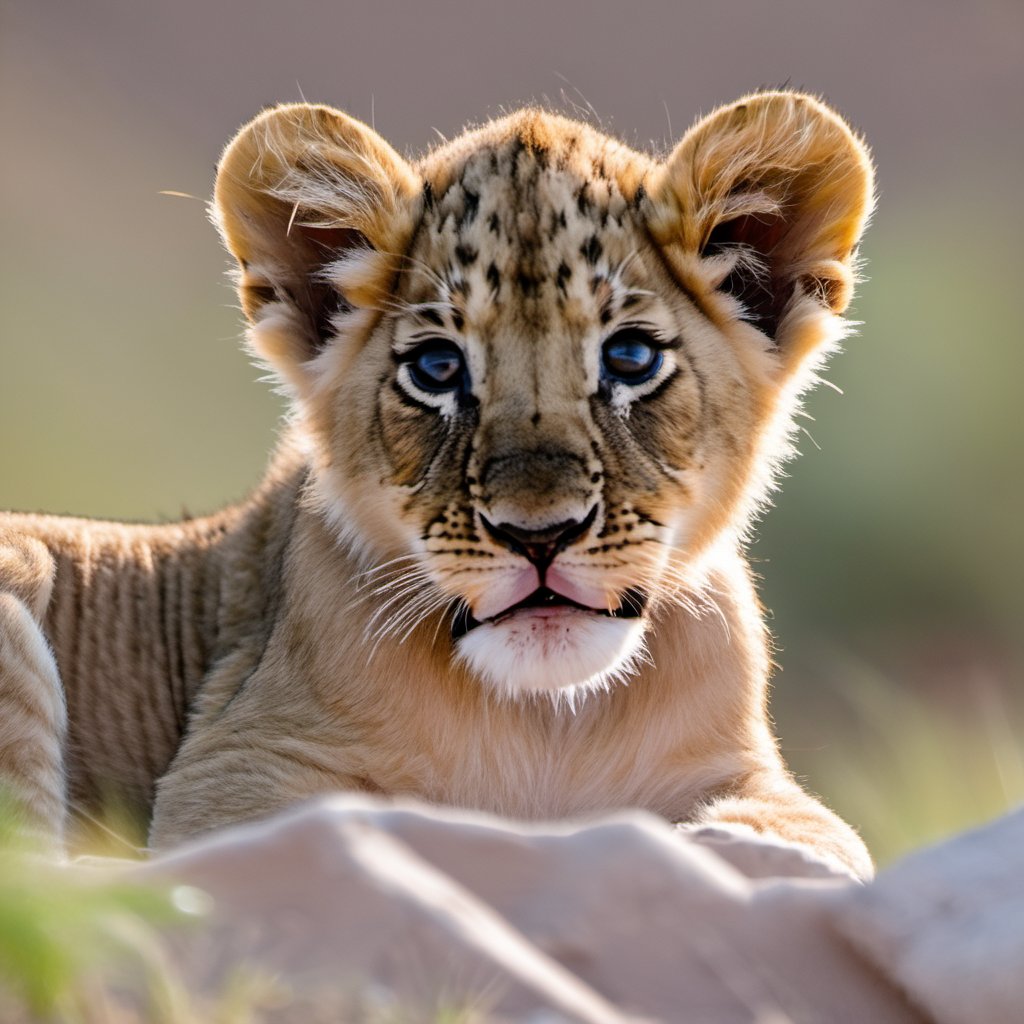}
                    };
                    \node[rotate=0, align=center, font=\normalsize, anchor=base, yshift=4pt] at (image.north) {\gls{ncg} (Ours) \strut};
                \end{tikzpicture}
            \end{subfigure}
        \end{minipage}
        \caption{Text-to-image generation results using Stable Diffusion XL with $\wcfg=15$. \gls{ncg} produces more realistic images compared to the oversaturated outputs of CFG.}
        \label{fig:sd-main}
        \vspace*{-0.5cm}

    \end{figure}
}

\newcommand{\figSDArtifacts}{
    \begin{figure}[t]
        \centering
        \vspace*{-0.75cm}
        \begin{minipage}{\textwidth}
            \begin{subfigure}{0.49\textwidth}
                \begin{subfigure}{0.495\textwidth}
                    \begin{tikzpicture}
                        \node[anchor=south west, inner sep=0] (image) at (0,0) {
                            \includegraphics[width=\textwidth]{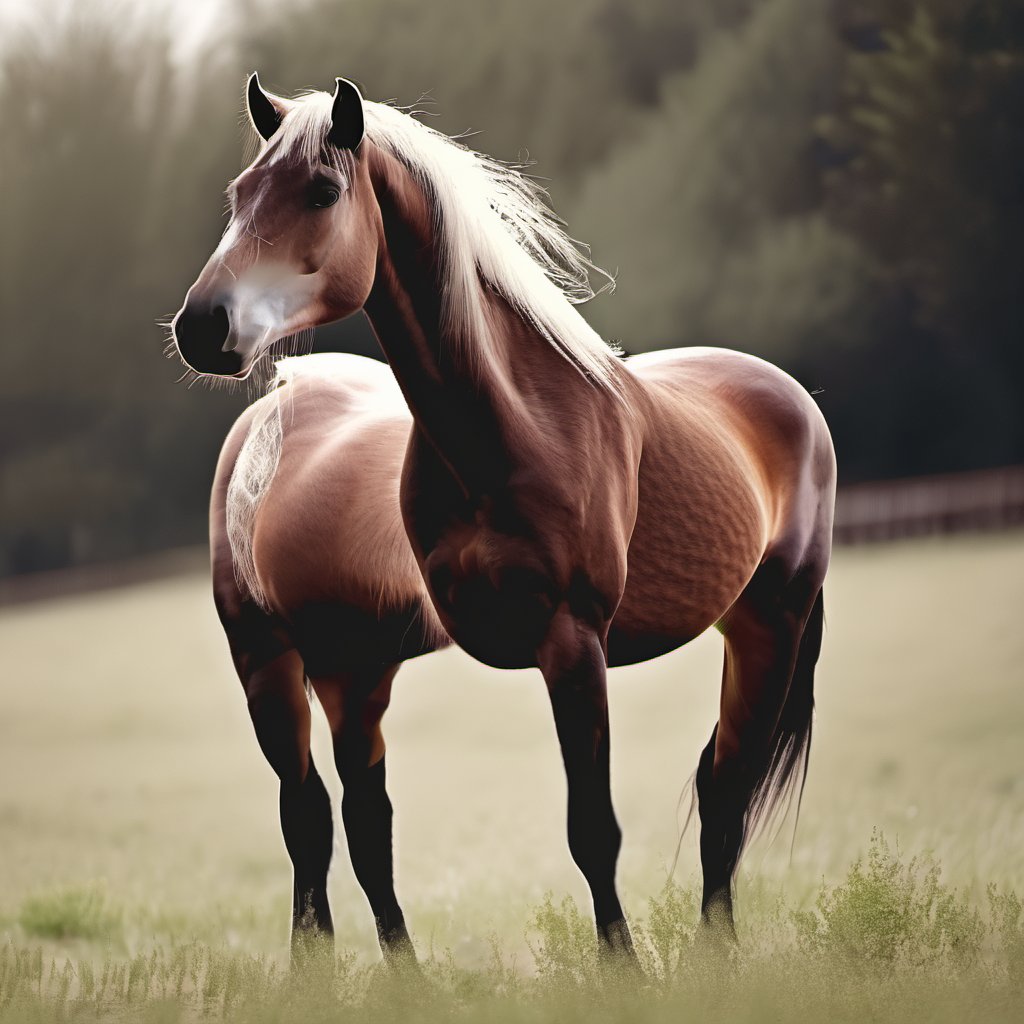}
                        };
                        \node[rotate=0, align=center, font=\normalsize, anchor=base, yshift=4pt] at (image.north) {CFG \strut};
                    \end{tikzpicture}
                \end{subfigure}
                \begin{subfigure}{0.495\textwidth}
                    \begin{tikzpicture}
                        \node[anchor=south west, inner sep=0] (image) at (0,0) {
                            \includegraphics[width=\textwidth]{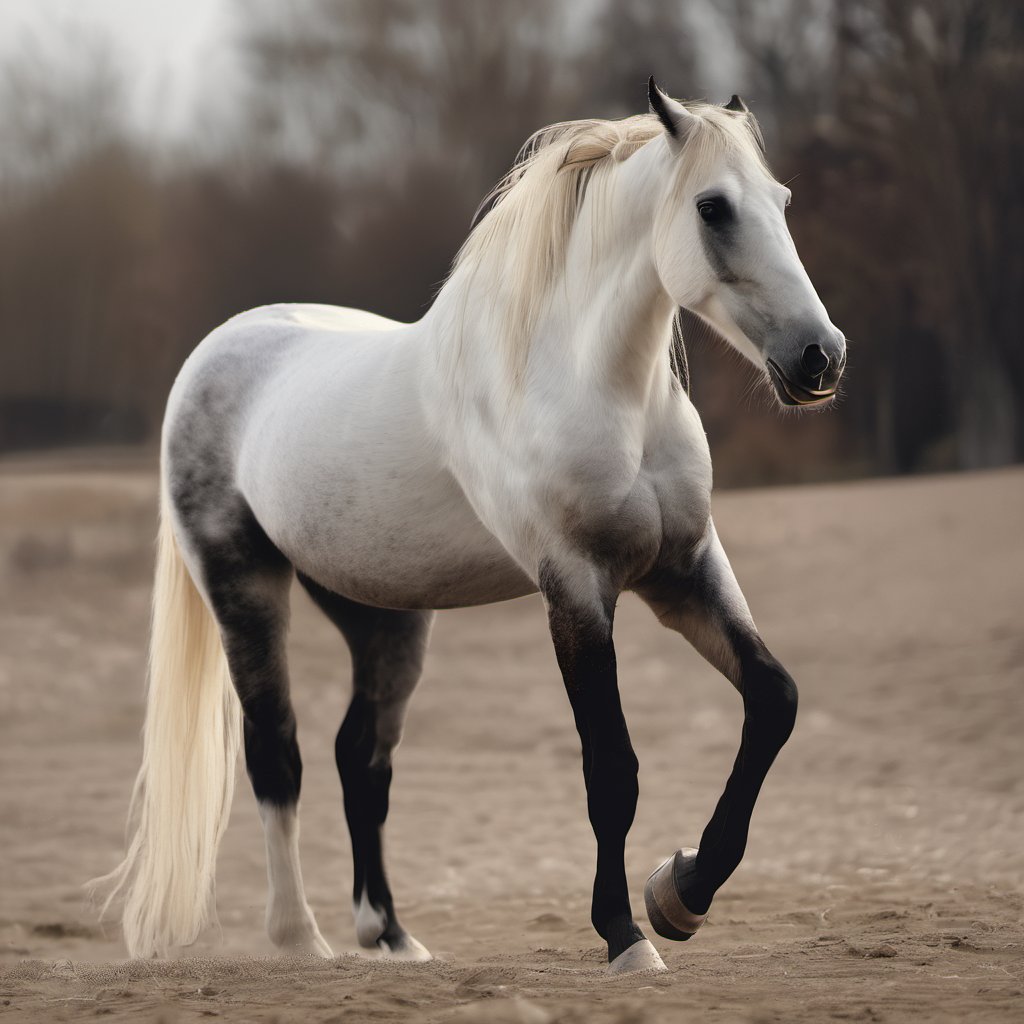}
                        };
                        \node[rotate=0, align=center, font=\normalsize, anchor=base, yshift=4pt] at (image.north) {\gls{ncg} (Ours) \strut};
                    \end{tikzpicture}
                \end{subfigure}
                \caption*{A horse}
            \end{subfigure}
            \begin{subfigure}{0.49\textwidth}
                \begin{subfigure}{0.495\textwidth}
                    \begin{tikzpicture}
                        \node[anchor=south west, inner sep=0] (image) at (0,0) {
                            \includegraphics[width=\textwidth]{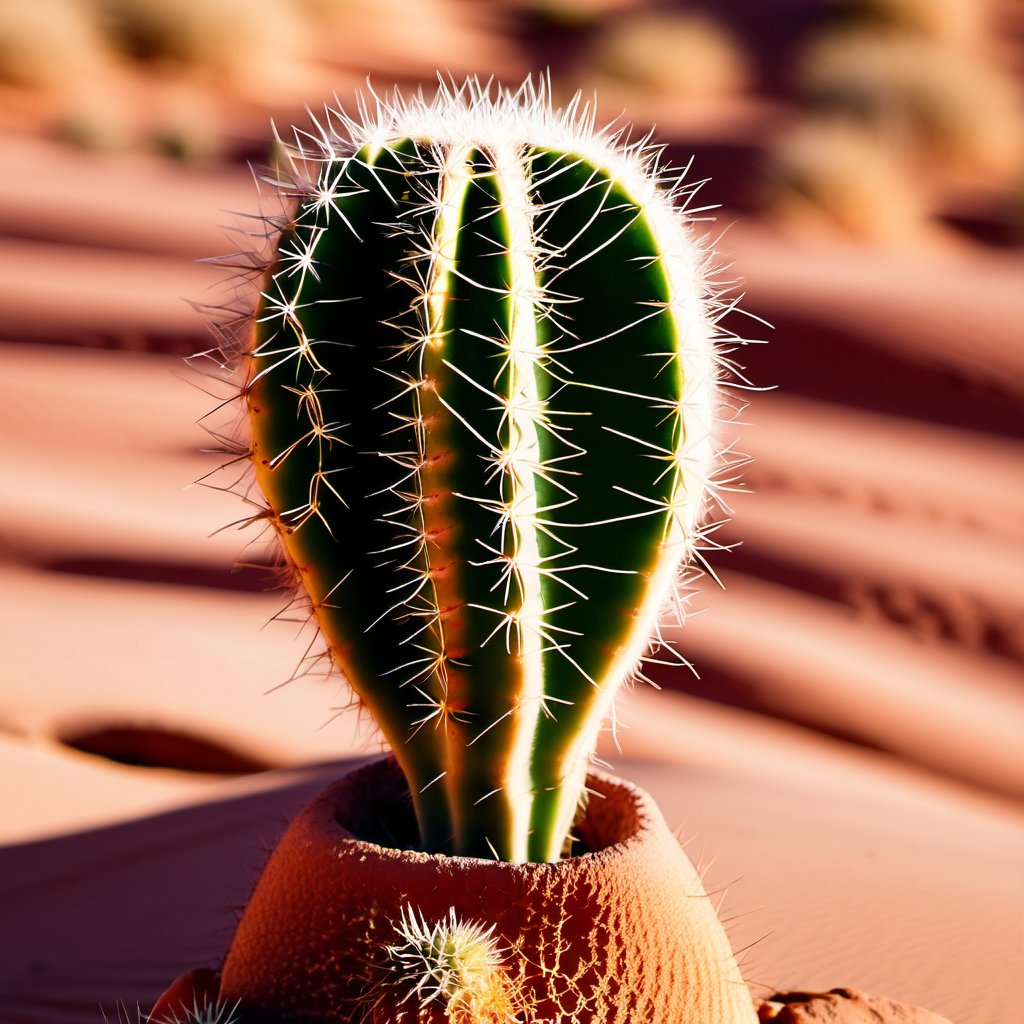}
                        };
                        \node[rotate=0, align=center, font=\normalsize, anchor=base, yshift=4pt] at (image.north) {CFG \strut};
                    \end{tikzpicture}
                \end{subfigure}
                \begin{subfigure}{0.495\textwidth}
                    \begin{tikzpicture}
                        \node[anchor=south west, inner sep=0] (image) at (0,0) {
                            \includegraphics[width=\textwidth]{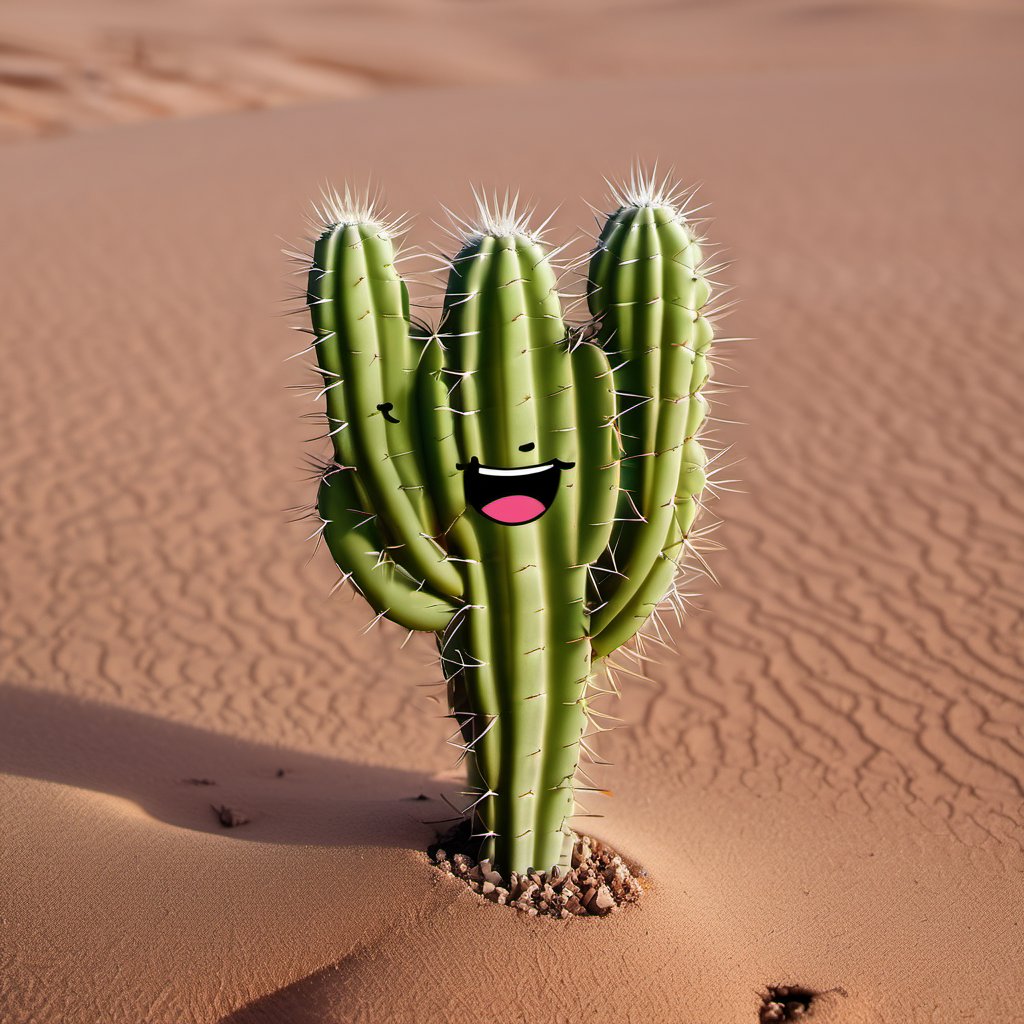}
                        };
                        \node[rotate=0, align=center, font=\normalsize, anchor=base, yshift=4pt] at (image.north) {\gls{ncg} (Ours) \strut};
                    \end{tikzpicture}
                \end{subfigure}
                \caption*{A small cactus with a happy face in the Sahara desert}
            \end{subfigure}

            \begin{subfigure}{0.49\textwidth}
                \begin{subfigure}{0.495\textwidth}
                    \begin{tikzpicture}
                        \node[anchor=south west, inner sep=0] (image) at (0,0) {
                            \includegraphics[width=\textwidth]{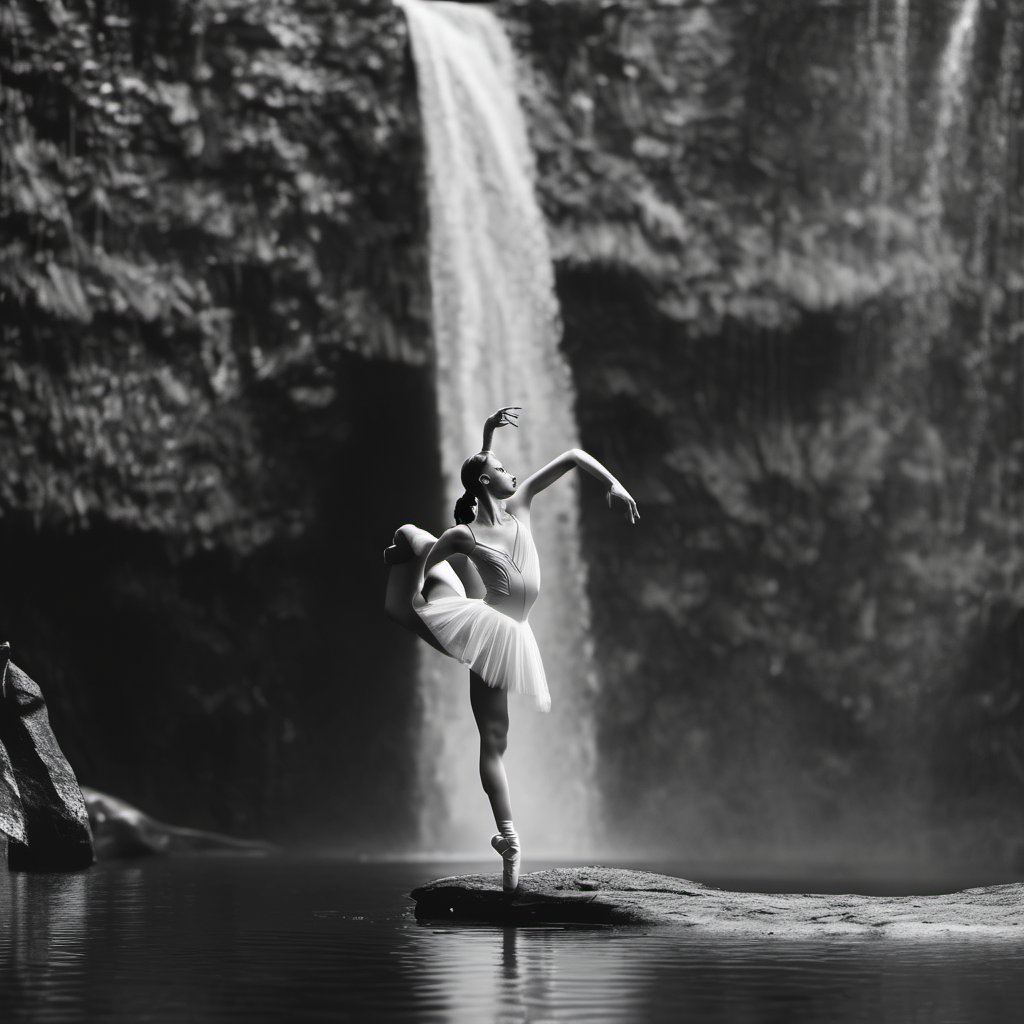}
                        };
                        \node[rotate=0, align=center, font=\normalsize, anchor=base, yshift=4pt] at (image.north) {CFG \strut};
                    \end{tikzpicture}
                \end{subfigure}
                \begin{subfigure}{0.495\textwidth}
                    \begin{tikzpicture}
                        \node[anchor=south west, inner sep=0] (image) at (0,0) {
                            \includegraphics[width=\textwidth]{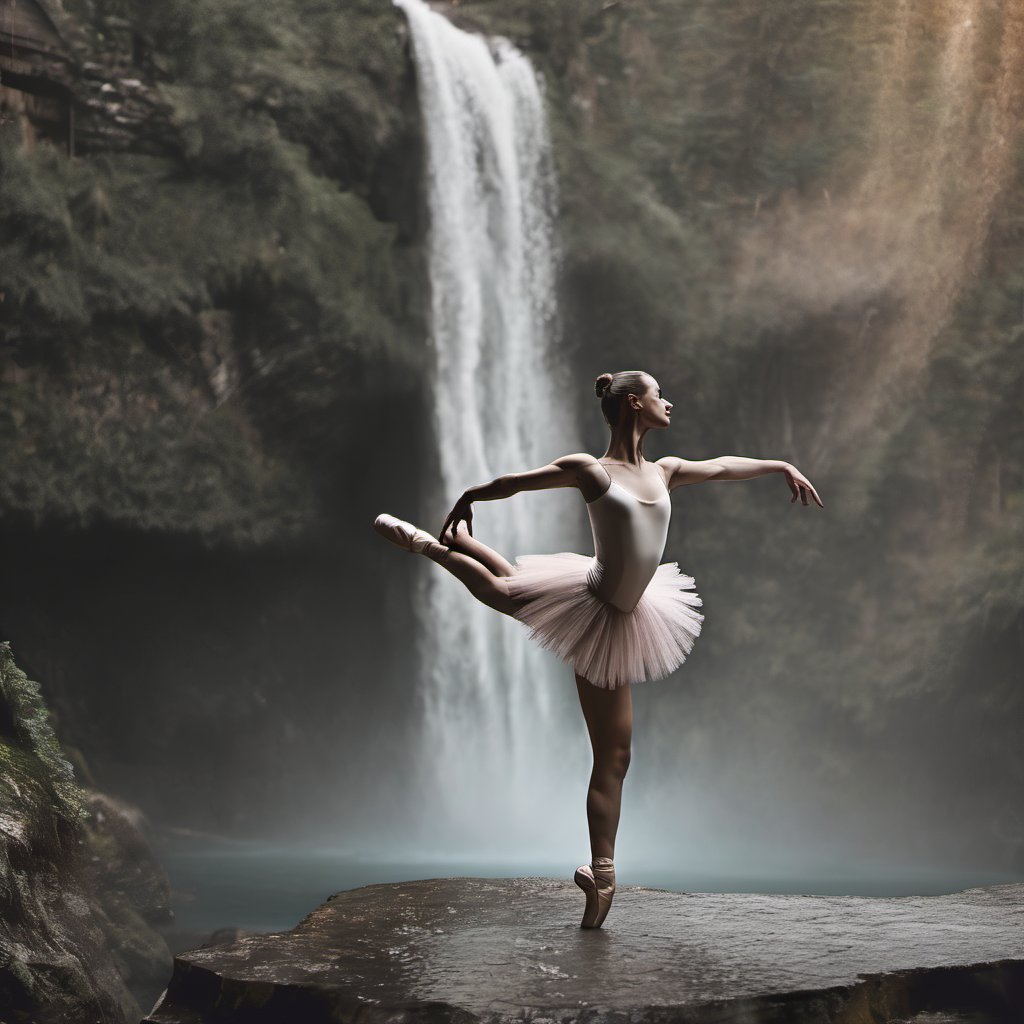}
                        };
                        \node[rotate=0, align=center, font=\normalsize, anchor=base, yshift=4pt] at (image.north) {\gls{ncg} (Ours) \strut};
                    \end{tikzpicture}
                \end{subfigure}
                \caption*{A ballet dancer next to a waterfall}
            \end{subfigure}
            \begin{subfigure}{0.49\textwidth}
                \begin{subfigure}{0.495\textwidth}
                    \begin{tikzpicture}
                        \node[anchor=south west, inner sep=0] (image) at (0,0) {
                            \includegraphics[width=\textwidth]{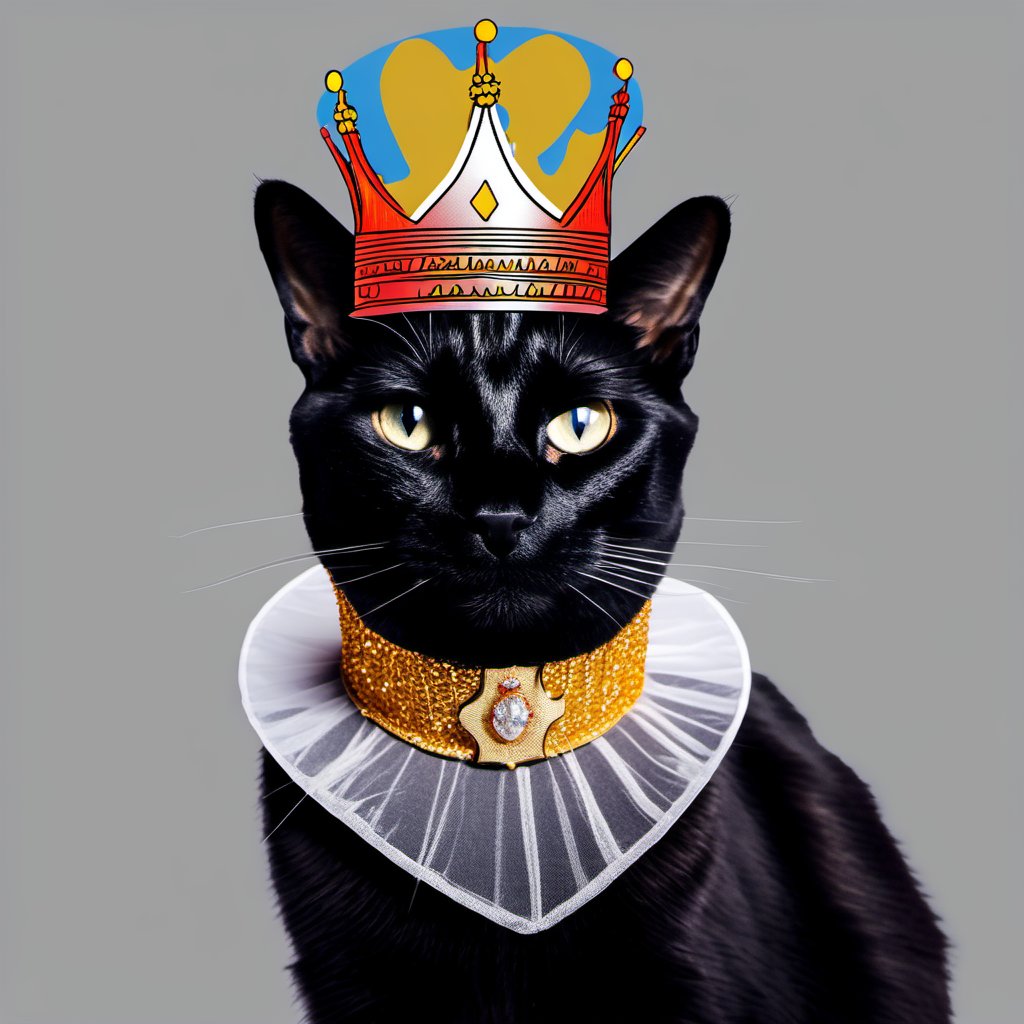}
                        };
                        \node[rotate=0, align=center, font=\normalsize, anchor=base, yshift=4pt] at (image.north) {CFG \strut};
                    \end{tikzpicture}
                \end{subfigure}
                \begin{subfigure}{0.495\textwidth}
                    \begin{tikzpicture}
                        \node[anchor=south west, inner sep=0] (image) at (0,0) {
                            \includegraphics[width=\textwidth]{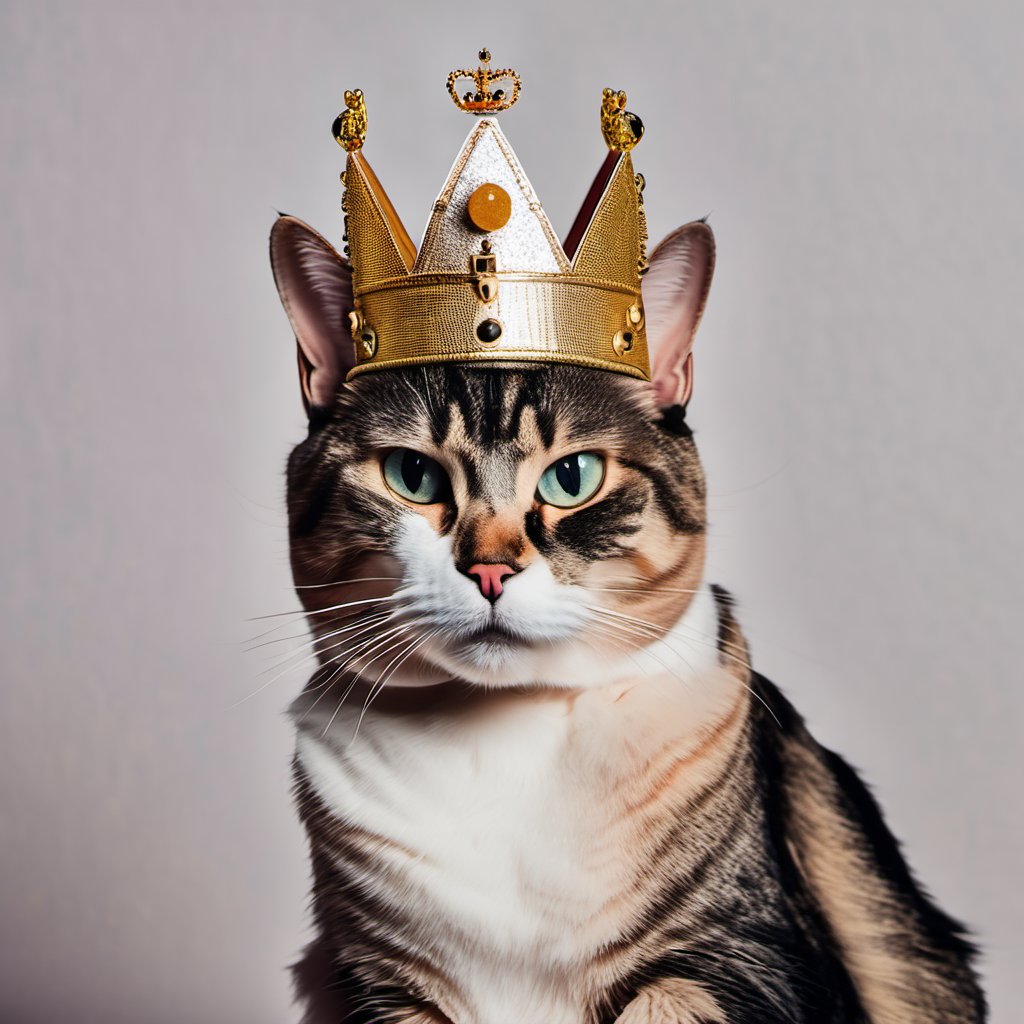}
                        };
                        \node[rotate=0, align=center, font=\normalsize, anchor=base, yshift=4pt] at (image.north) {\gls{ncg} (Ours) \strut};
                    \end{tikzpicture}
                \end{subfigure}
                \caption*{A cat with a queen hat}
            \end{subfigure}
        \end{minipage}
        \caption{Examples of artifacts in the CFG outputs that can be solved by using \gls{ncg}. We see that for all images, \gls{ncg} outputs follow the prompt with a more globally consistent generation.}
        \label{fig:artifacts}
        \vspace*{-0.25cm}
    \end{figure}
}

\newcommand{\figSDthreeMain}{
    \begin{figure}[t!]
        \centering
        \begin{minipage}{\textwidth}
            \begin{subfigure}{0.49\textwidth}
                \begin{subfigure}{0.495\textwidth}
                    \begin{tikzpicture}
                        \node[anchor=south west, inner sep=0] (image) at (0,0) {
                            \includegraphics[width=\textwidth]{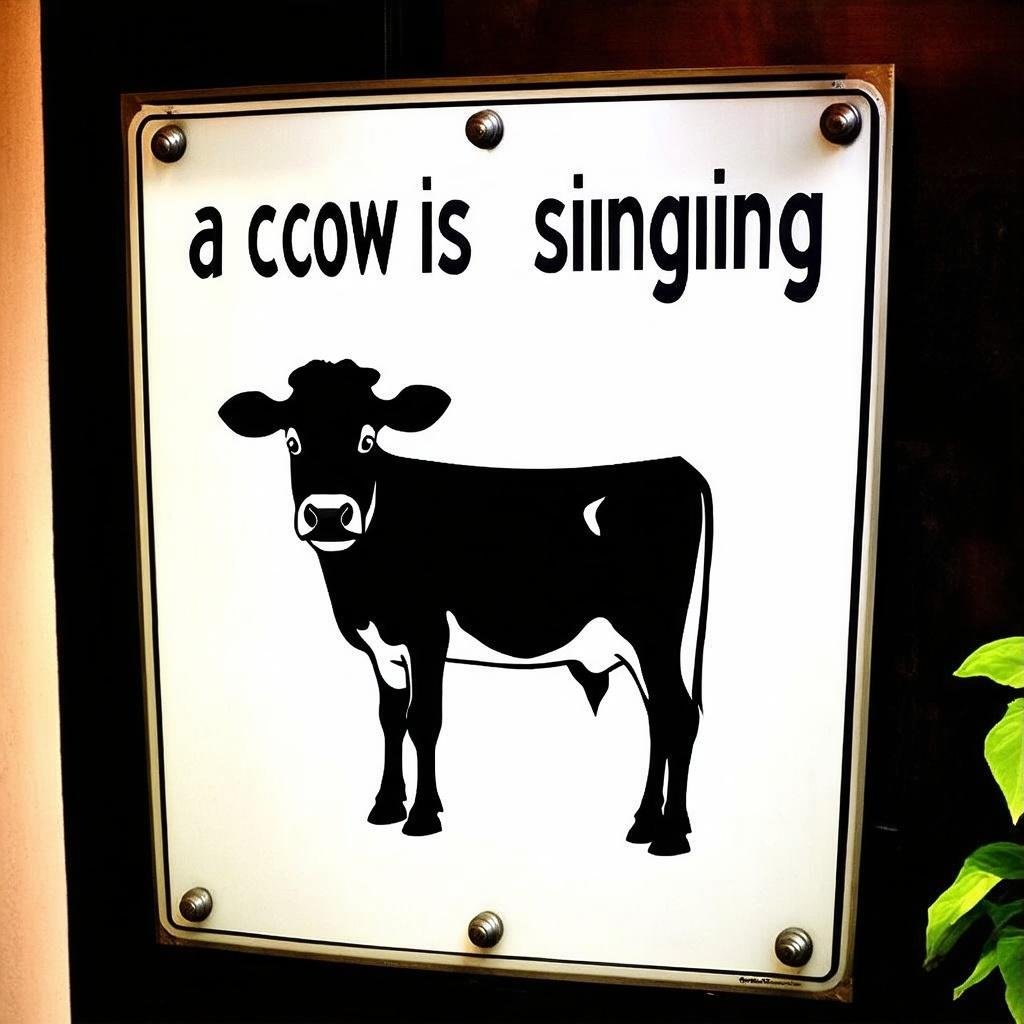}
                        };
                        \node[rotate=0, align=center, font=\normalsize, anchor=base, yshift=4pt] at (image.north) {CFG \strut};
                    \end{tikzpicture}
                \end{subfigure}
                \begin{subfigure}{0.495\textwidth}
                    \begin{tikzpicture}
                        \node[anchor=south west, inner sep=0] (image) at (0,0) {
                            \includegraphics[width=\textwidth]{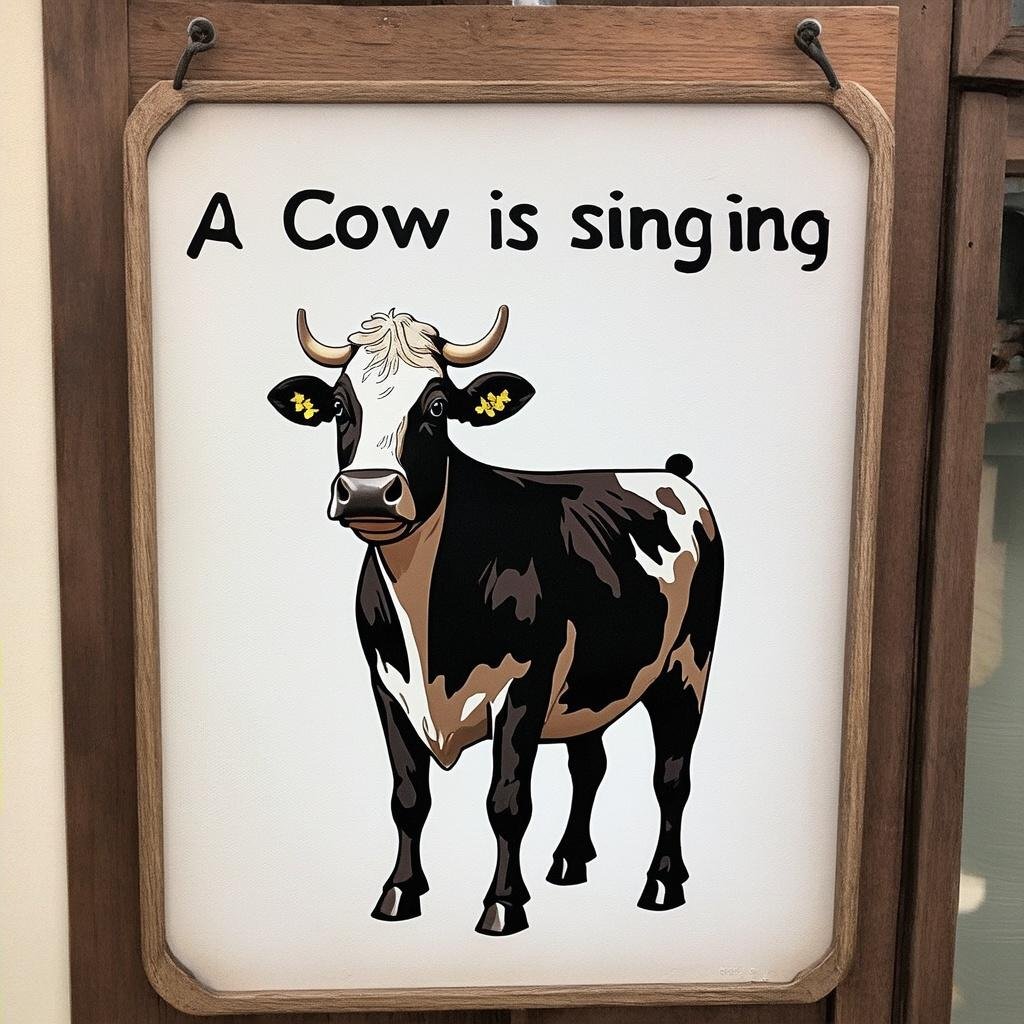}
                        };
                        \node[rotate=0, align=center, font=\normalsize, anchor=base, yshift=4pt] at (image.north) {\gls{ncg} (Ours) \strut};
                    \end{tikzpicture}
                \end{subfigure}
                \caption*{A sign that says ``A cow is singing''}
            \end{subfigure}
            \begin{subfigure}{0.49\textwidth}
                \begin{subfigure}{0.495\textwidth}
                    \begin{tikzpicture}
                        \node[anchor=south west, inner sep=0] (image) at (0,0) {
                            \includegraphics[width=\textwidth]{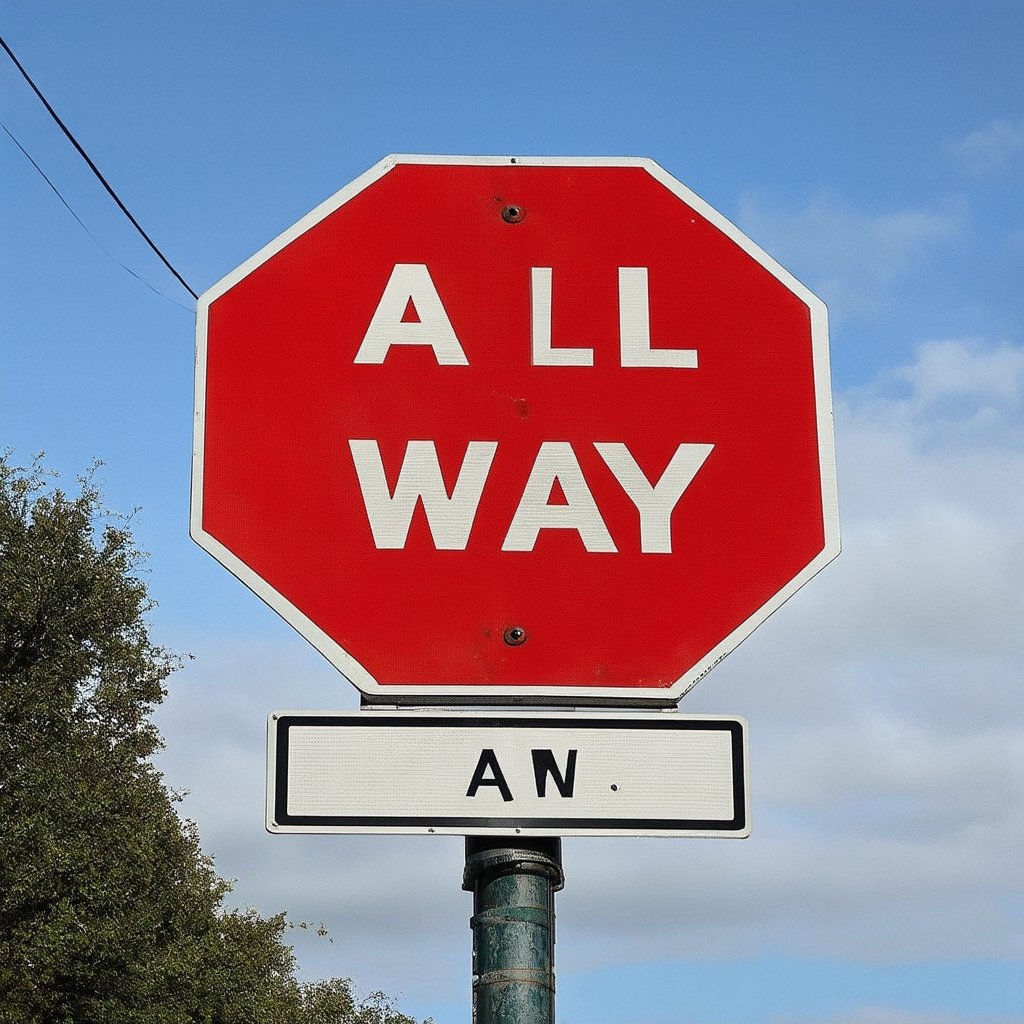}
                        };
                        \node[rotate=0, align=center, font=\normalsize, anchor=base, yshift=4pt] at (image.north) {CFG \strut};
                    \end{tikzpicture}
                \end{subfigure}
                \begin{subfigure}{0.495\textwidth}
                    \begin{tikzpicture}
                        \node[anchor=south west, inner sep=0] (image) at (0,0) {
                            \includegraphics[width=\textwidth]{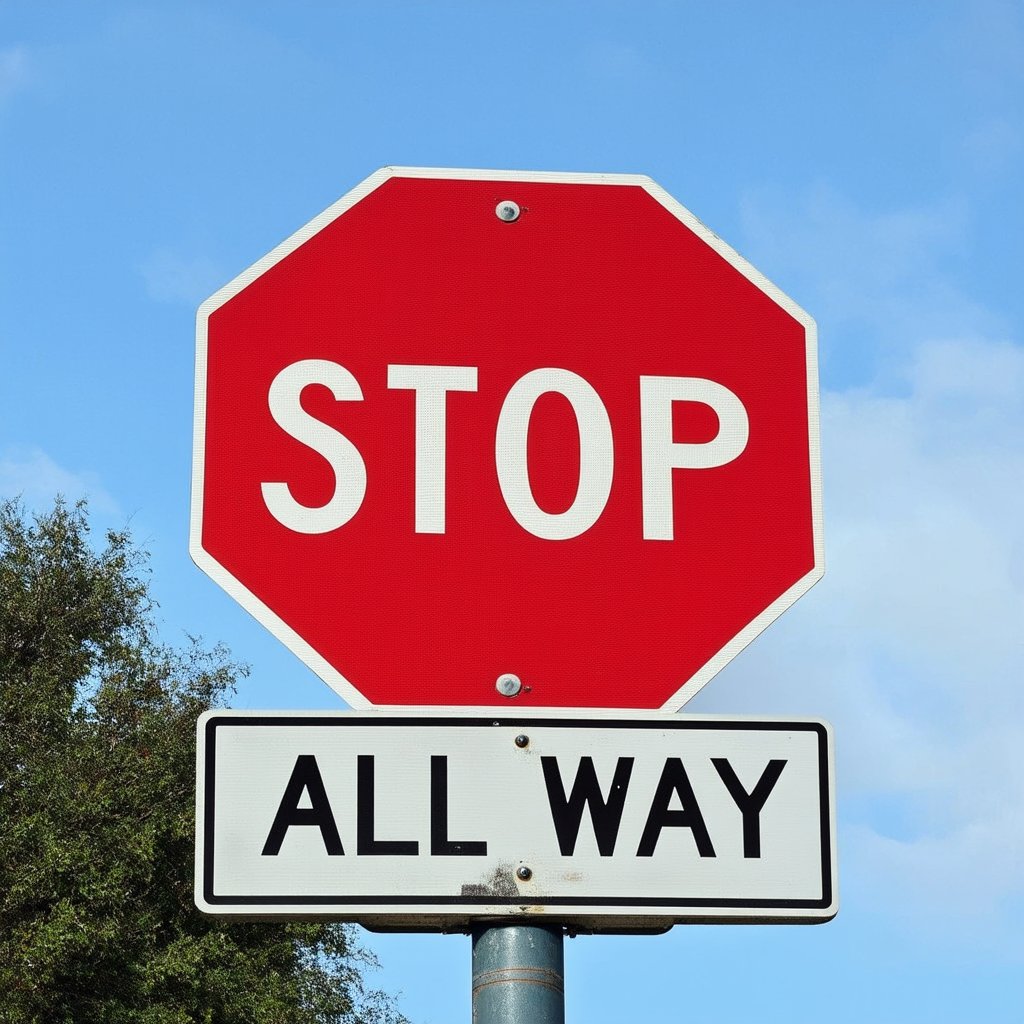}
                        };
                        \node[rotate=0, align=center, font=\normalsize, anchor=base, yshift=4pt] at (image.north) {\gls{ncg} (Ours) \strut};
                    \end{tikzpicture}
                \end{subfigure}
                \caption*{A stop sign with ``ALL WAY'' written below it}
            \end{subfigure}
            
            \begin{adjustbox}{valign=t}
            \begin{subfigure}{0.49\textwidth}
                \begin{subfigure}{0.495\textwidth}
                    \begin{tikzpicture}
                        \node[anchor=south west, inner sep=0] (image) at (0,0) {
                            \includegraphics[width=\textwidth]{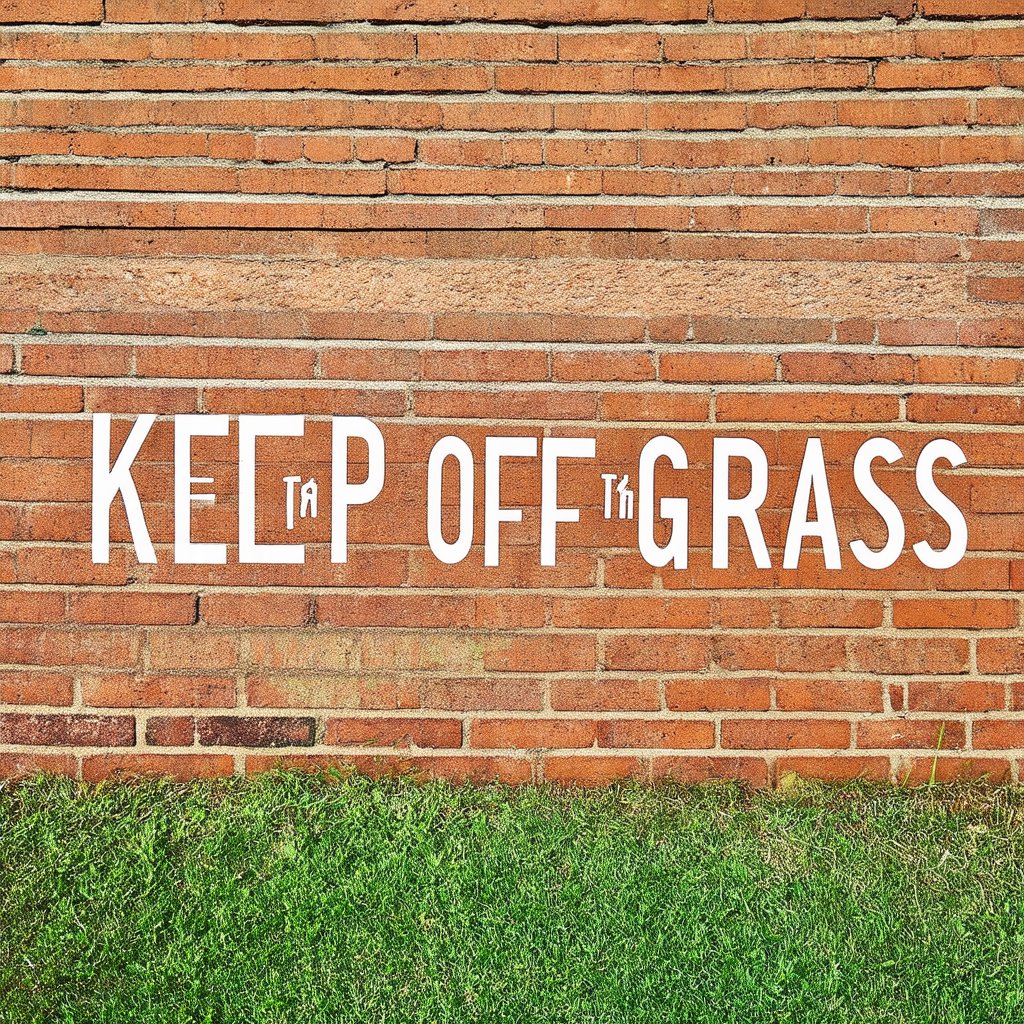}
                        };
                        \node[rotate=0, align=center, font=\normalsize, anchor=base, yshift=4pt] at (image.north) {CFG \strut};
                    \end{tikzpicture}
                \end{subfigure}
                \begin{subfigure}{0.495\textwidth}
                    \begin{tikzpicture}
                        \node[anchor=south west, inner sep=0] (image) at (0,0) {
                            \includegraphics[width=\textwidth]{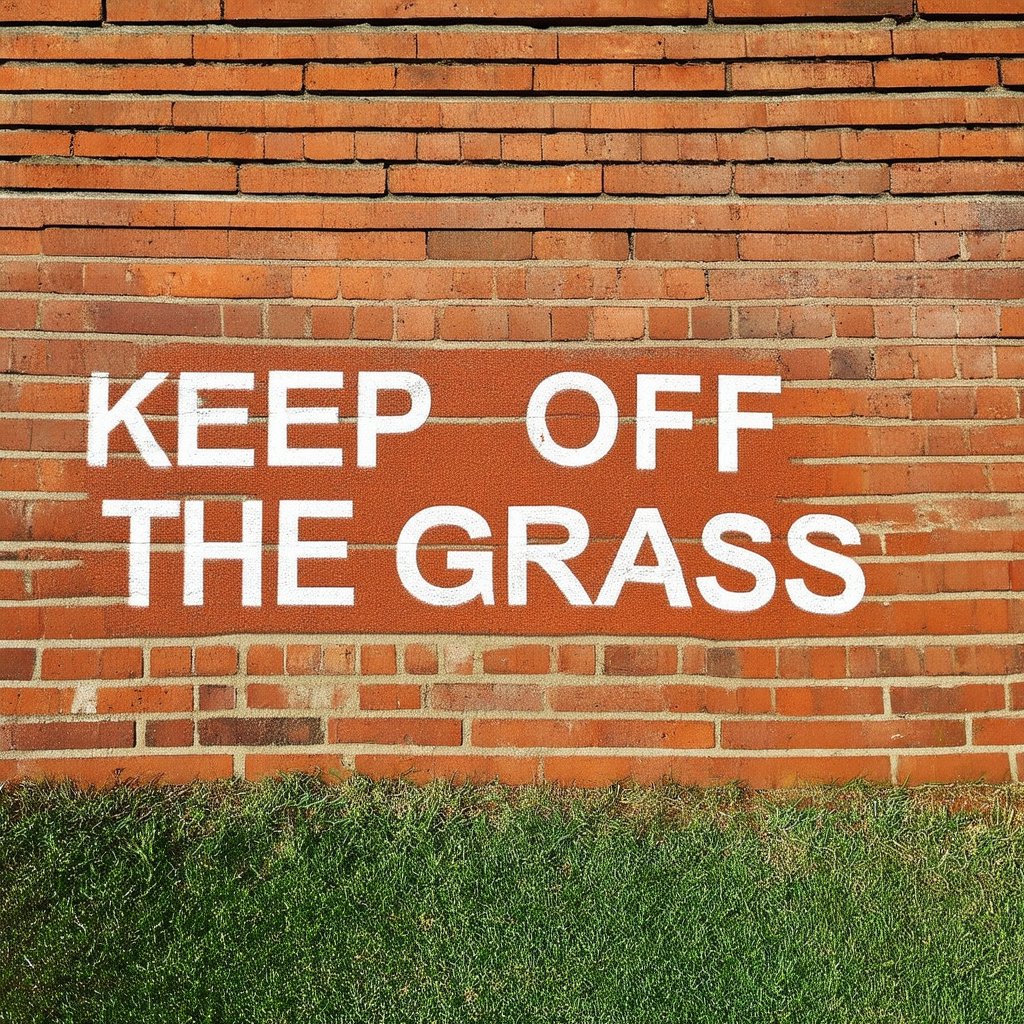}
                        };
                        \node[rotate=0, align=center, font=\normalsize, anchor=base, yshift=4pt] at (image.north) {\gls{ncg} (Ours) \strut};
                    \end{tikzpicture}
                \end{subfigure}
                \caption*{{The words ``KEEP OFF THE GRASS'' written on a brick wall}}
            \end{subfigure}
        \end{adjustbox}
            \begin{adjustbox}{valign=t}
                \begin{subfigure}{0.49\textwidth}
                    \begin{subfigure}{0.495\textwidth}
                        \begin{tikzpicture}
                            \node[anchor=south west, inner sep=0] (image) at (0,0) {
                                \includegraphics[width=\textwidth]{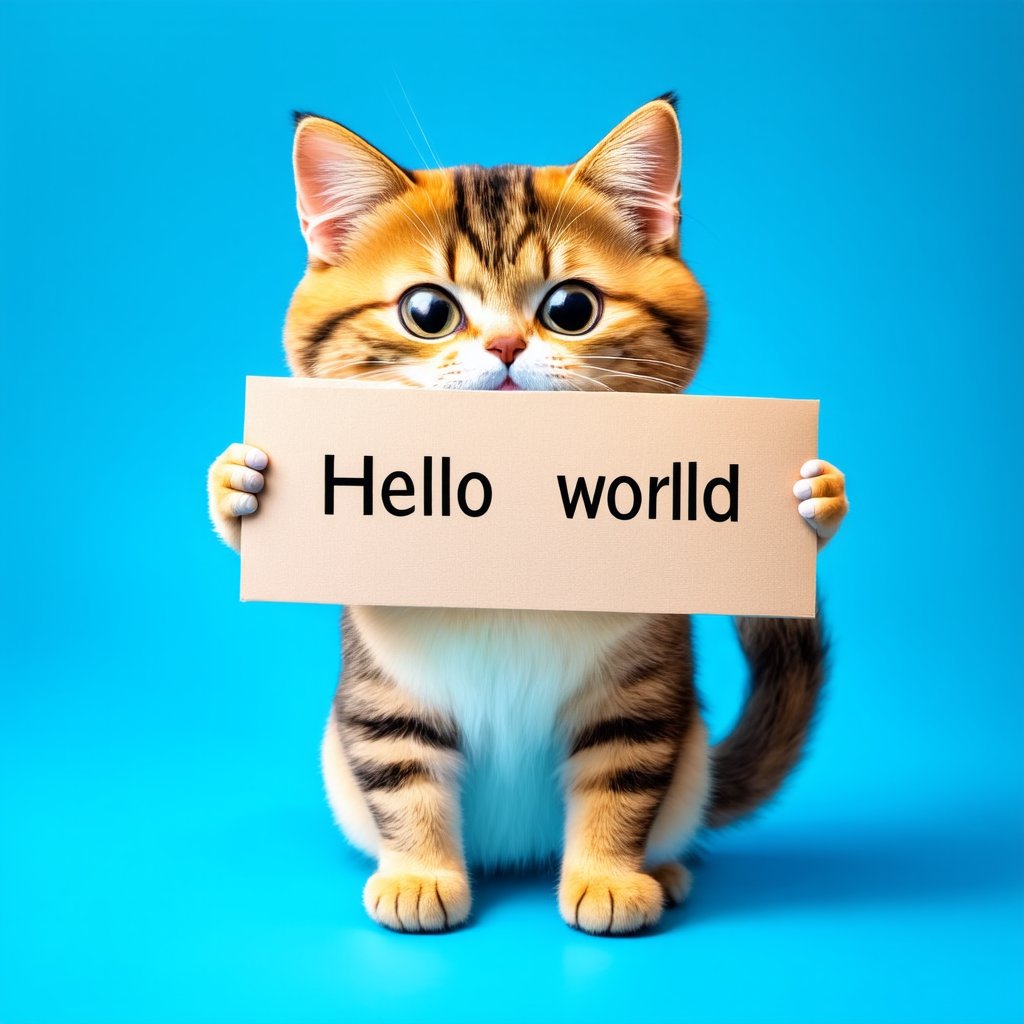}
                            };
                            \node[rotate=0, align=center, font=\normalsize, anchor=base, yshift=4pt] at (image.north) {CFG \strut};
                        \end{tikzpicture}
                    \end{subfigure}
                    \begin{subfigure}{0.495\textwidth}
                        \begin{tikzpicture}
                            \node[anchor=south west, inner sep=0] (image) at (0,0) {
                                \includegraphics[width=\textwidth]{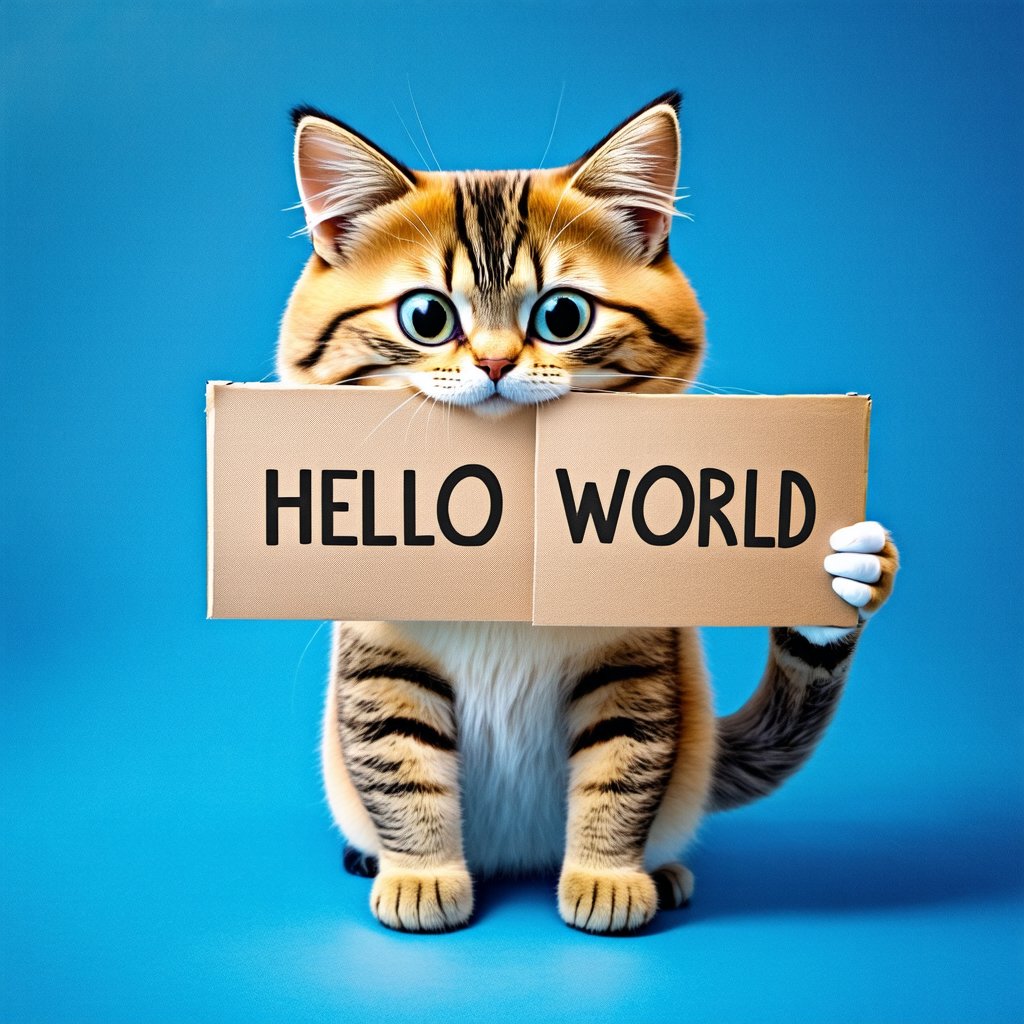}
                            };
                            \node[rotate=0, align=center, font=\normalsize, anchor=base, yshift=4pt] at (image.north) {\gls{ncg} (Ours) \strut};
                        \end{tikzpicture}
                    \end{subfigure}
                    \caption*{A cat holding a sign that says ``hello world''}
                \end{subfigure}
            \end{adjustbox}
        \end{minipage}
        \caption{Comparison of CFG and \gls{ncg} for text quality in generated images using Stable Diffusion 3 \citep{esser2024scaling}.  In contrast to CFG, \gls{ncg} consistently produces correct spellings.}
        \label{fig:sd3-text}
    \end{figure}
}

\newcommand{\figSDthreeAppendix}{
    \begin{figure}[t]
        \centering
        \begin{minipage}{\textwidth}
            \begin{subfigure}{0.49\textwidth}
                \begin{subfigure}{0.495\textwidth}
                    \begin{tikzpicture}
                        \node[anchor=south west, inner sep=0] (image) at (0,0) {
                            \includegraphics[width=\textwidth]{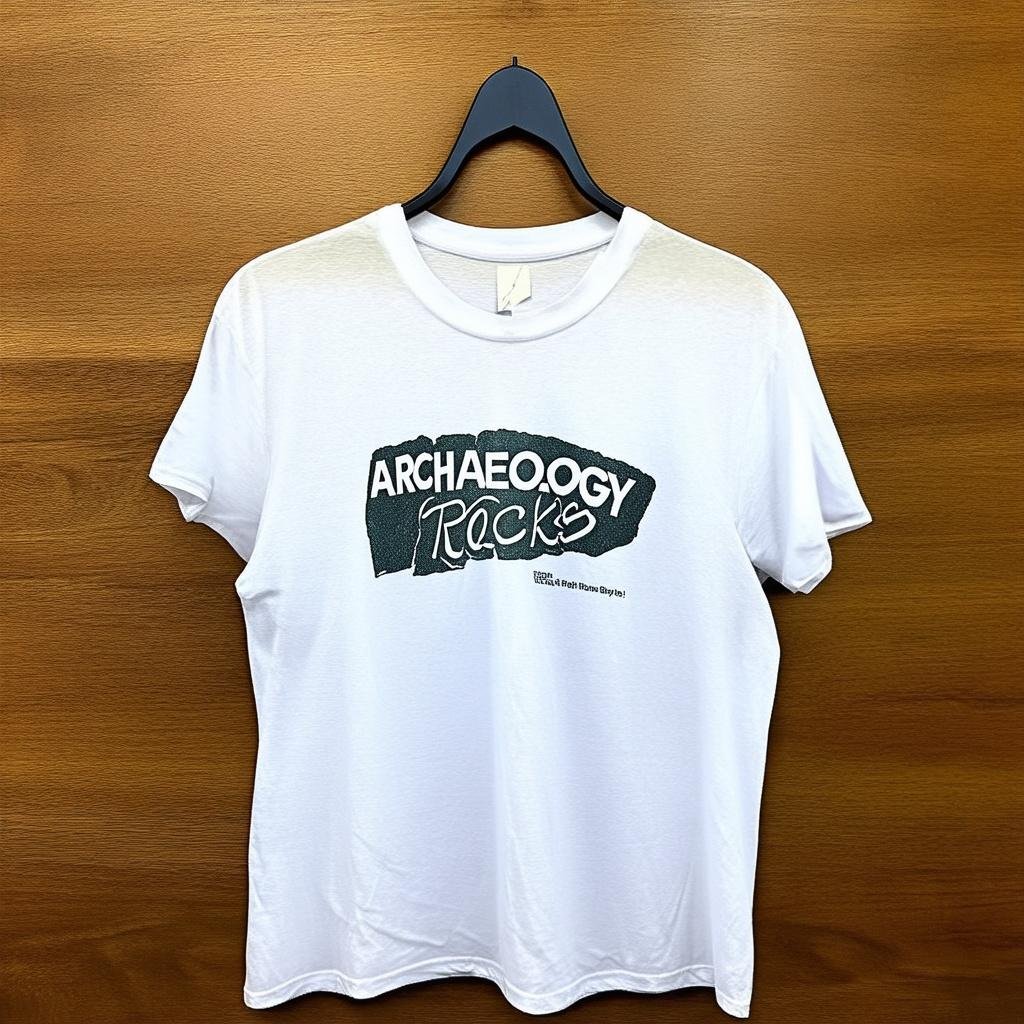}
                        };
                        \node[rotate=0, align=center, font=\normalsize, anchor=base, yshift=4pt] at (image.north) {CFG \strut};
                    \end{tikzpicture}
                \end{subfigure}
                \begin{subfigure}{0.495\textwidth}
                    \begin{tikzpicture}
                        \node[anchor=south west, inner sep=0] (image) at (0,0) {
                            \includegraphics[width=\textwidth]{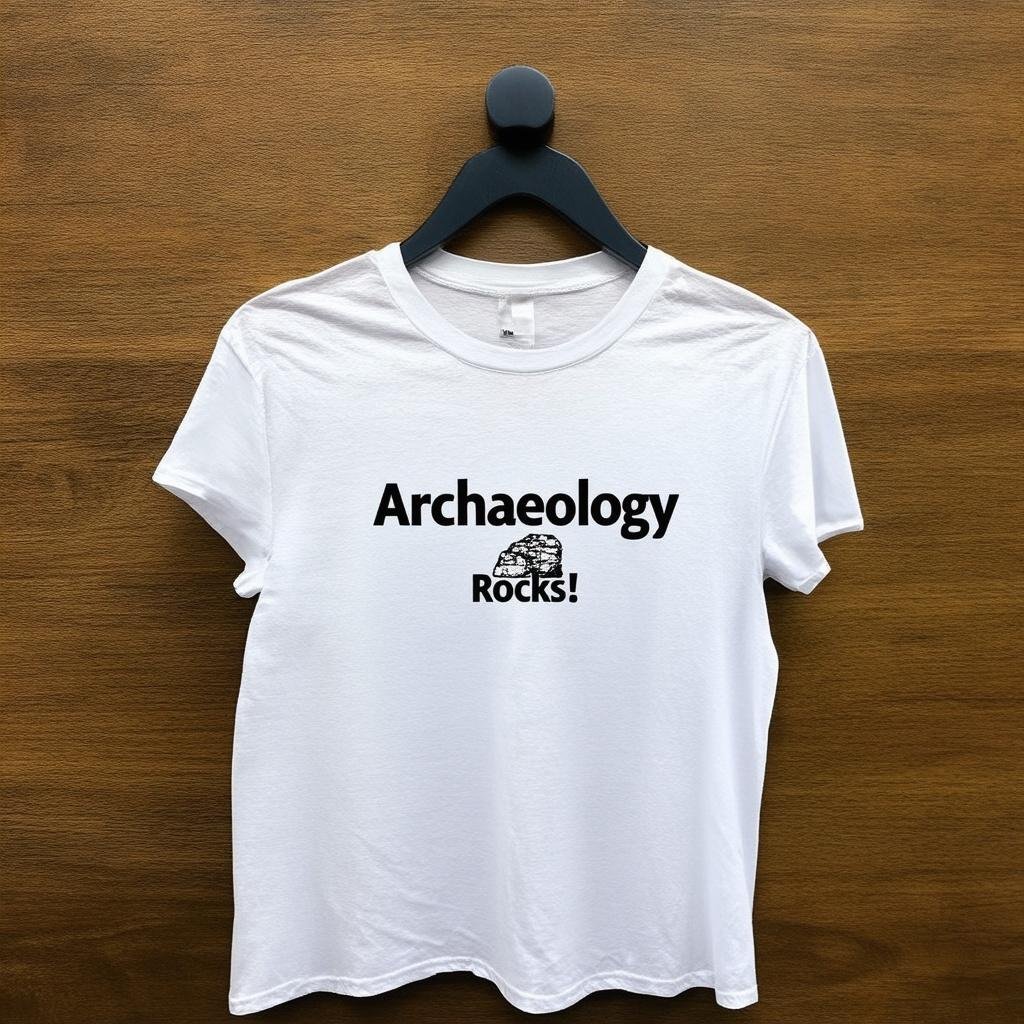}
                        };
                        \node[rotate=0, align=center, font=\normalsize, anchor=base, yshift=4pt] at (image.north) {\gls{ncg} (Ours) \strut};
                    \end{tikzpicture}
                \end{subfigure}
                \caption*{A t-shirt with ``Archaeology Rocks!'' written on it}
            \end{subfigure}
            \begin{subfigure}{0.49\textwidth}
                \begin{subfigure}{0.495\textwidth}
                    \begin{tikzpicture}
                        \node[anchor=south west, inner sep=0] (image) at (0,0) {
                            \includegraphics[width=\textwidth]{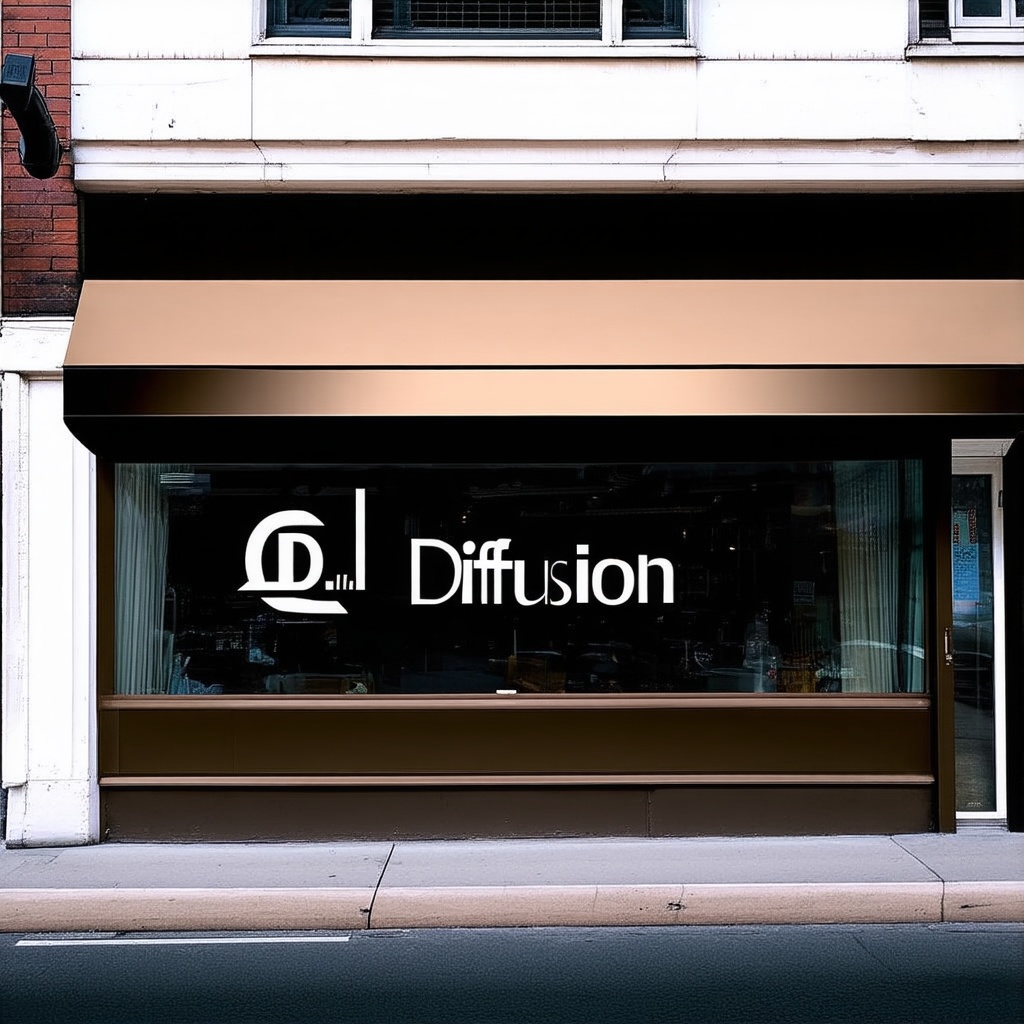}
                        };
                        \node[rotate=0, align=center, font=\normalsize, anchor=base, yshift=4pt] at (image.north) {CFG \strut};
                    \end{tikzpicture}
                \end{subfigure}
                \begin{subfigure}{0.495\textwidth}
                    \begin{tikzpicture}
                        \node[anchor=south west, inner sep=0] (image) at (0,0) {
                            \includegraphics[width=\textwidth]{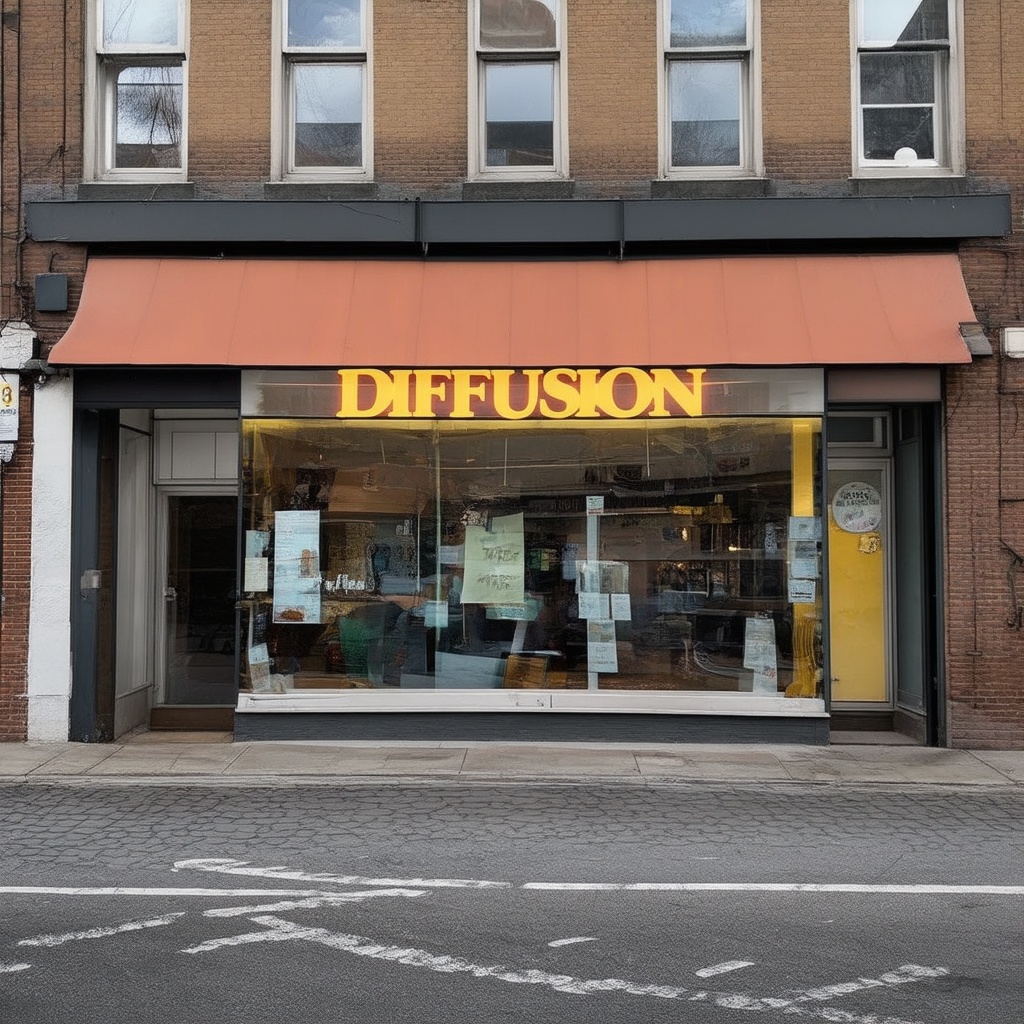}
                        };
                        \node[rotate=0, align=center, font=\normalsize, anchor=base, yshift=4pt] at (image.north) {\gls{ncg} (Ours) \strut};
                    \end{tikzpicture}
                \end{subfigure}
                \caption*{A storefront with ``Diffusion'' written on it}
            \end{subfigure}
            \begin{subfigure}{0.49\textwidth}
                \begin{subfigure}{0.495\textwidth}
                    \begin{tikzpicture}
                        \node[anchor=south west, inner sep=0] (image) at (0,0) {
                            \includegraphics[width=\textwidth]{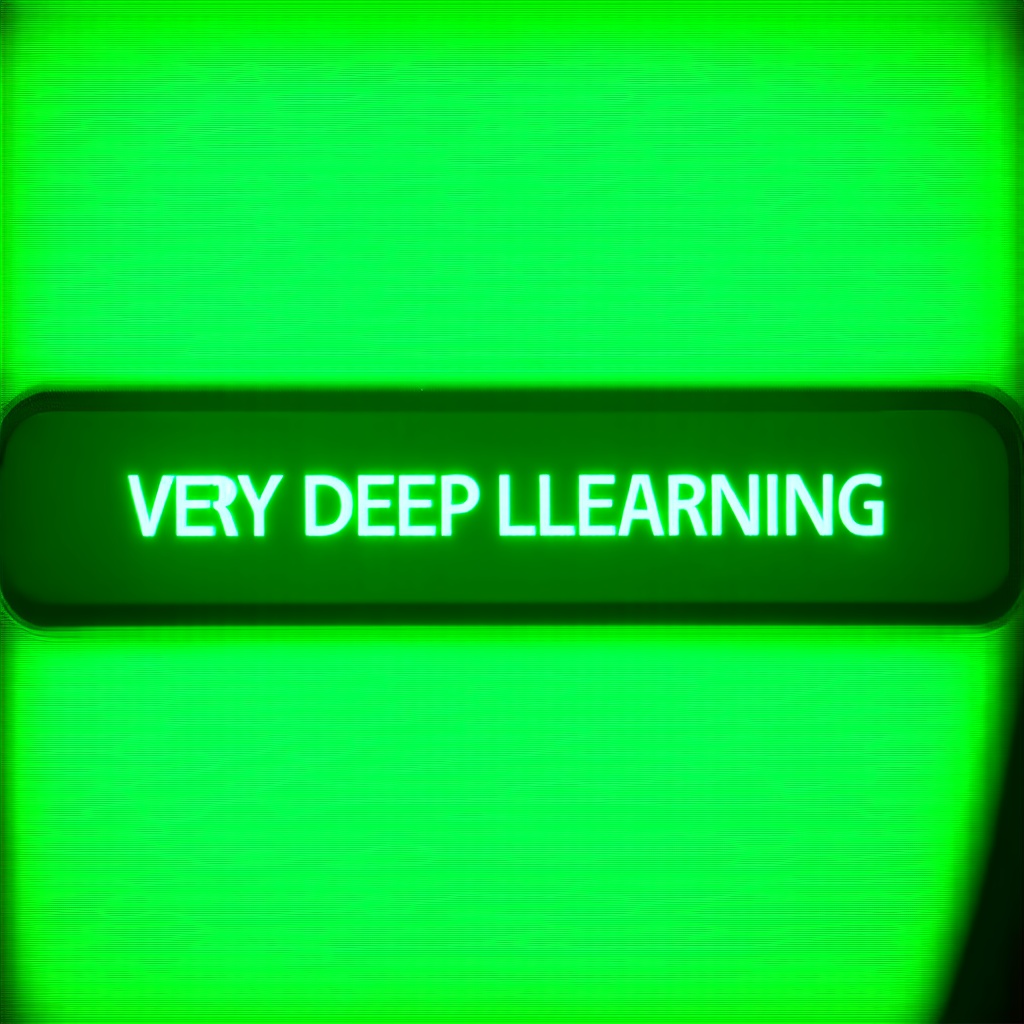}
                        };
                        \node[rotate=0, align=center, font=\normalsize, anchor=base, yshift=4pt] at (image.north) {CFG \strut};
                    \end{tikzpicture}
                \end{subfigure}
                \begin{subfigure}{0.495\textwidth}
                    \begin{tikzpicture}
                        \node[anchor=south west, inner sep=0] (image) at (0,0) {
                            \includegraphics[width=\textwidth]{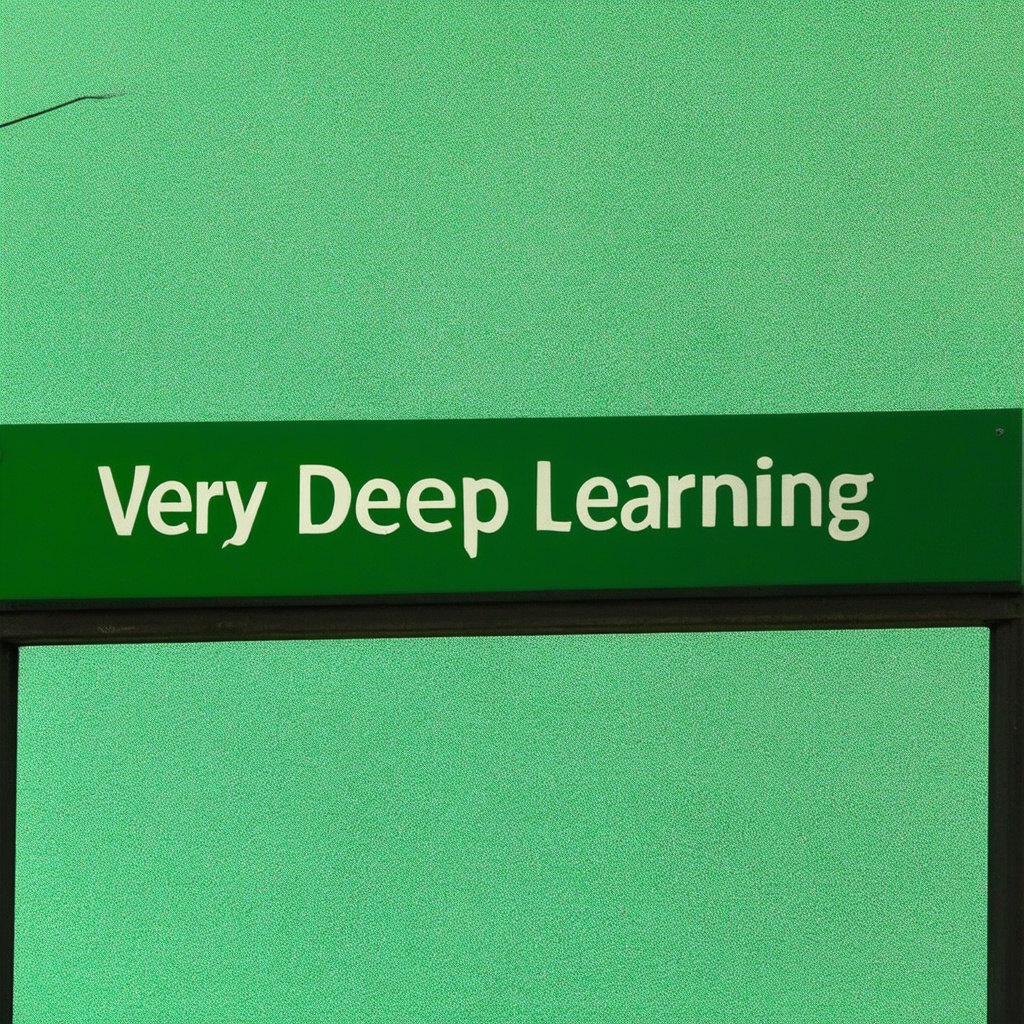}
                        };
                        \node[rotate=0, align=center, font=\normalsize, anchor=base, yshift=4pt] at (image.north) {\gls{ncg} (Ours) \strut};
                    \end{tikzpicture}
                \end{subfigure}
                \caption*{A green sign that says ``Very Deep Learning''}
            \end{subfigure}
            \begin{subfigure}{0.49\textwidth}
                \begin{subfigure}{0.495\textwidth}
                    \begin{tikzpicture}
                        \node[anchor=south west, inner sep=0] (image) at (0,0) {
                            \includegraphics[width=\textwidth]{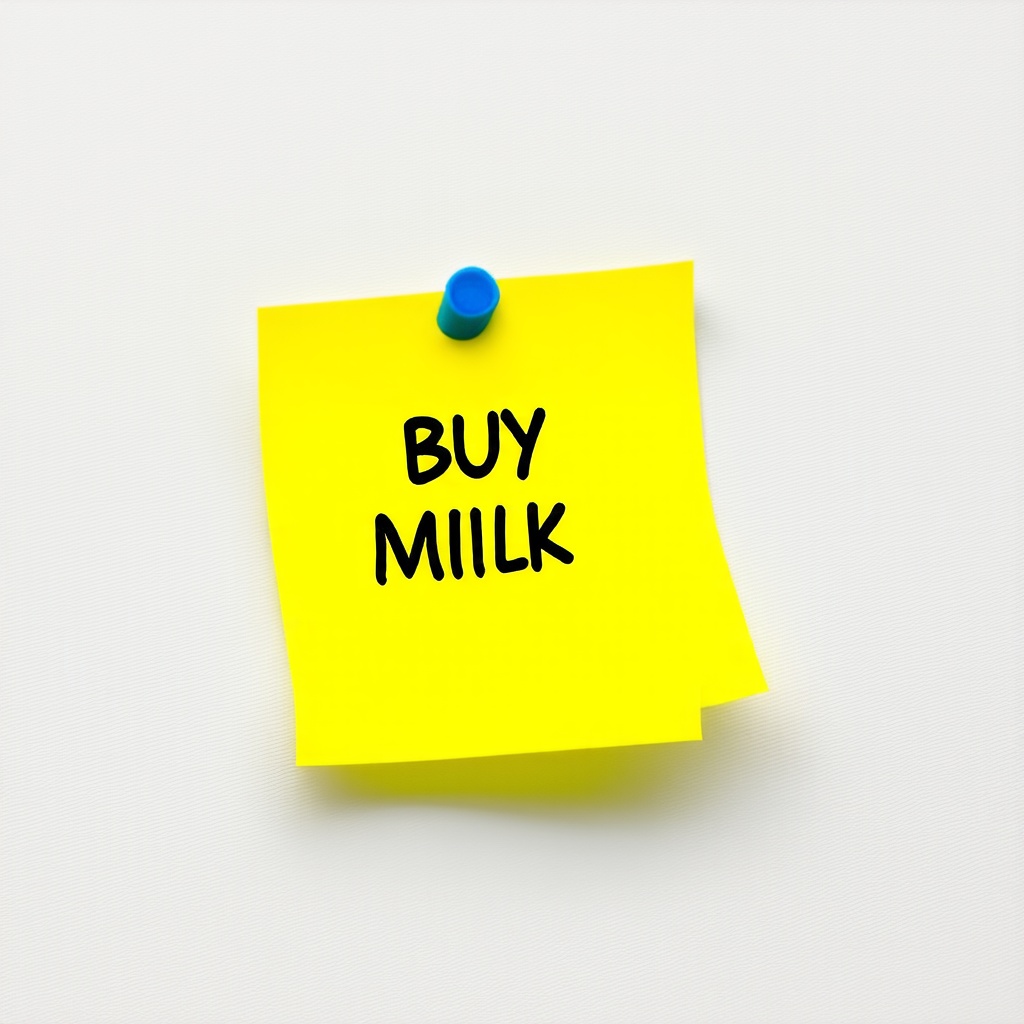}
                        };
                        \node[rotate=0, align=center, font=\normalsize, anchor=base, yshift=4pt] at (image.north) {CFG \strut};
                    \end{tikzpicture}
                \end{subfigure}
                \begin{subfigure}{0.495\textwidth}
                    \begin{tikzpicture}
                        \node[anchor=south west, inner sep=0] (image) at (0,0) {
                            \includegraphics[width=\textwidth]{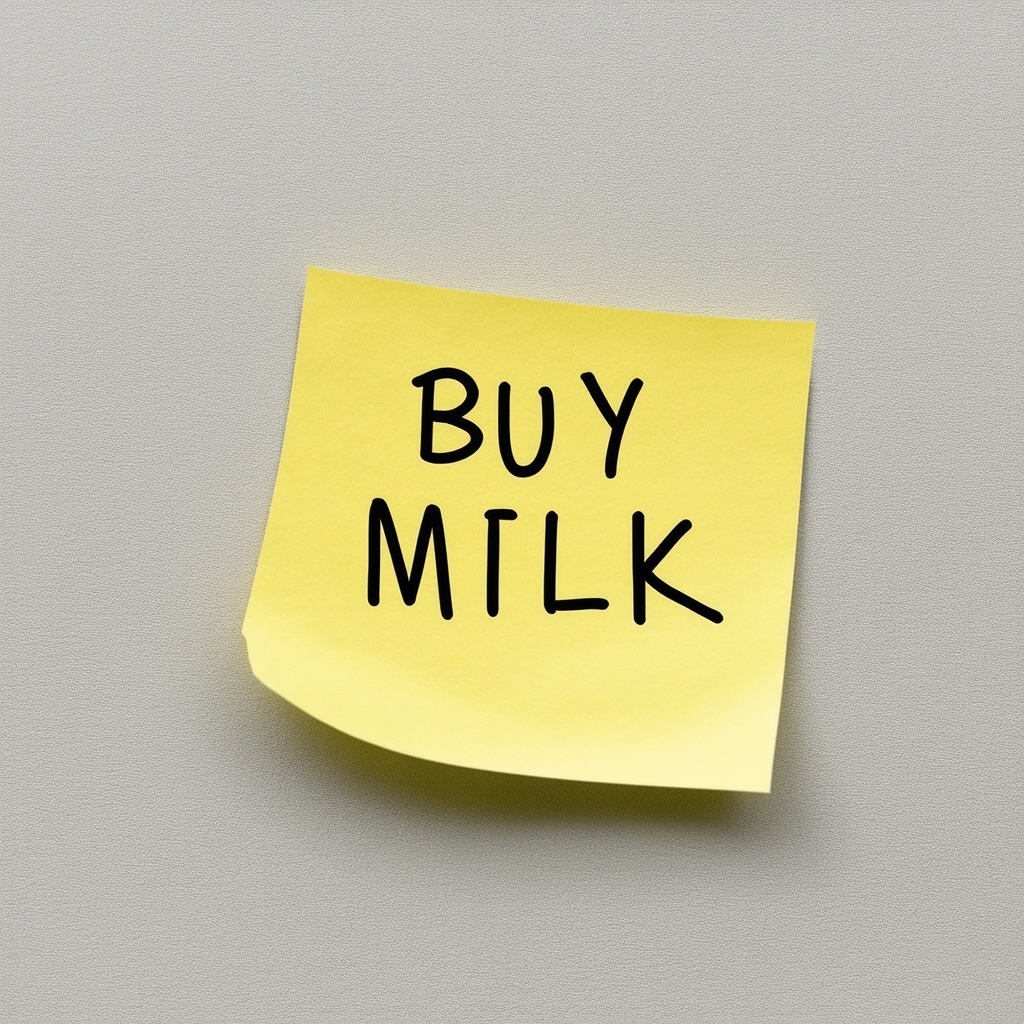}
                        };
                        \node[rotate=0, align=center, font=\normalsize, anchor=base, yshift=4pt] at (image.north) {\gls{ncg} (Ours) \strut};
                    \end{tikzpicture}
                \end{subfigure}
                \caption*{A yellow sticky note with ``BUY MILK'' written on it}
            \end{subfigure}
        \end{minipage}
        \caption{Additional visual examples on the quality of rendering text using Stable Diffusion 3. Compared to CFG, \gls{ncg} results show more consistent spellings.}
        \label{fig:sd3-appendix}
    \end{figure}
}

\newcommand{\figSDtwoAppendix}{
    \begin{figure}[t]
        \centering
        \begin{minipage}{\textwidth}
            \begin{subfigure}{0.49\textwidth}
                \begin{subfigure}{0.495\textwidth}
                    \begin{tikzpicture}
                        \node[anchor=south west, inner sep=0] (image) at (0,0) {
                            \includegraphics[width=\textwidth]{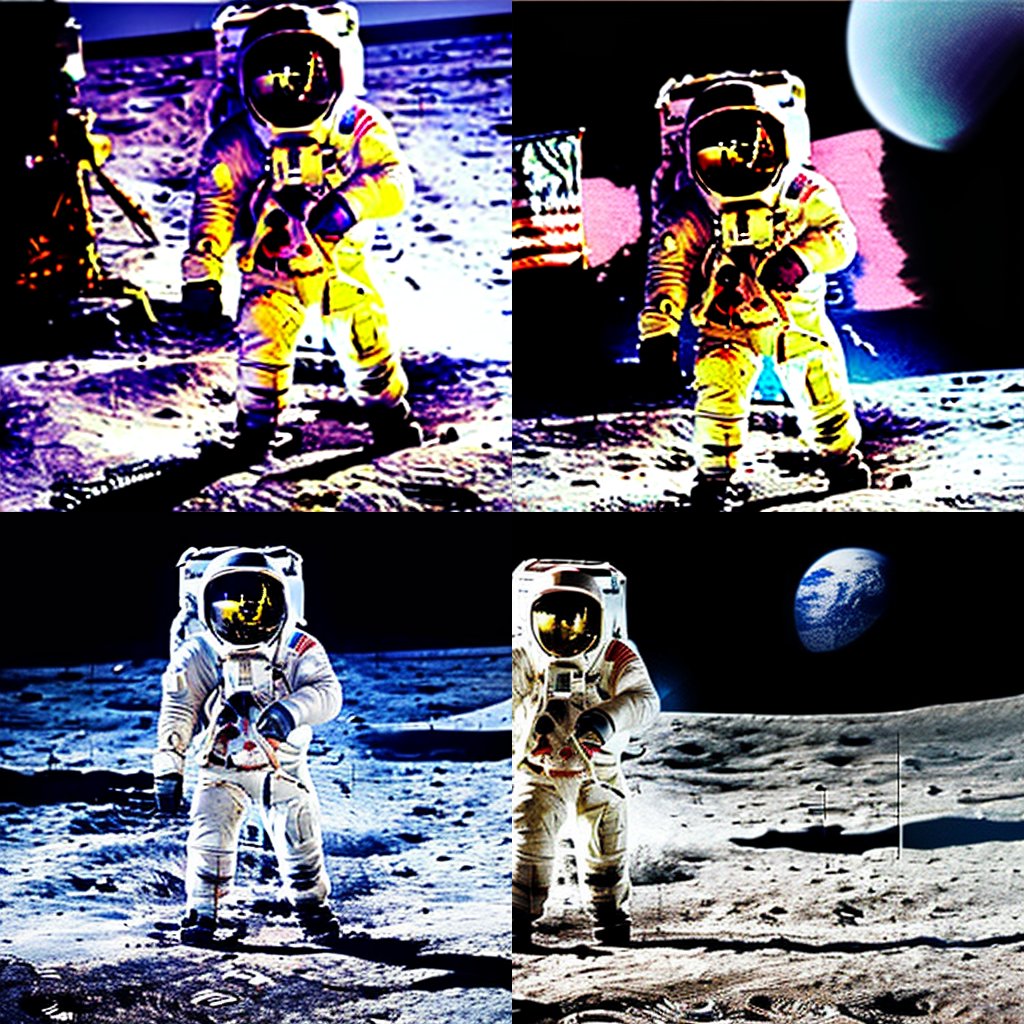}
                        };
                        \node[rotate=0, align=center, font=\normalsize, anchor=base, yshift=4pt] at (image.north) {CFG \strut};
                    \end{tikzpicture}
                \end{subfigure}
                \begin{subfigure}{0.495\textwidth}
                    \begin{tikzpicture}
                        \node[anchor=south west, inner sep=0] (image) at (0,0) {
                            \includegraphics[width=\textwidth]{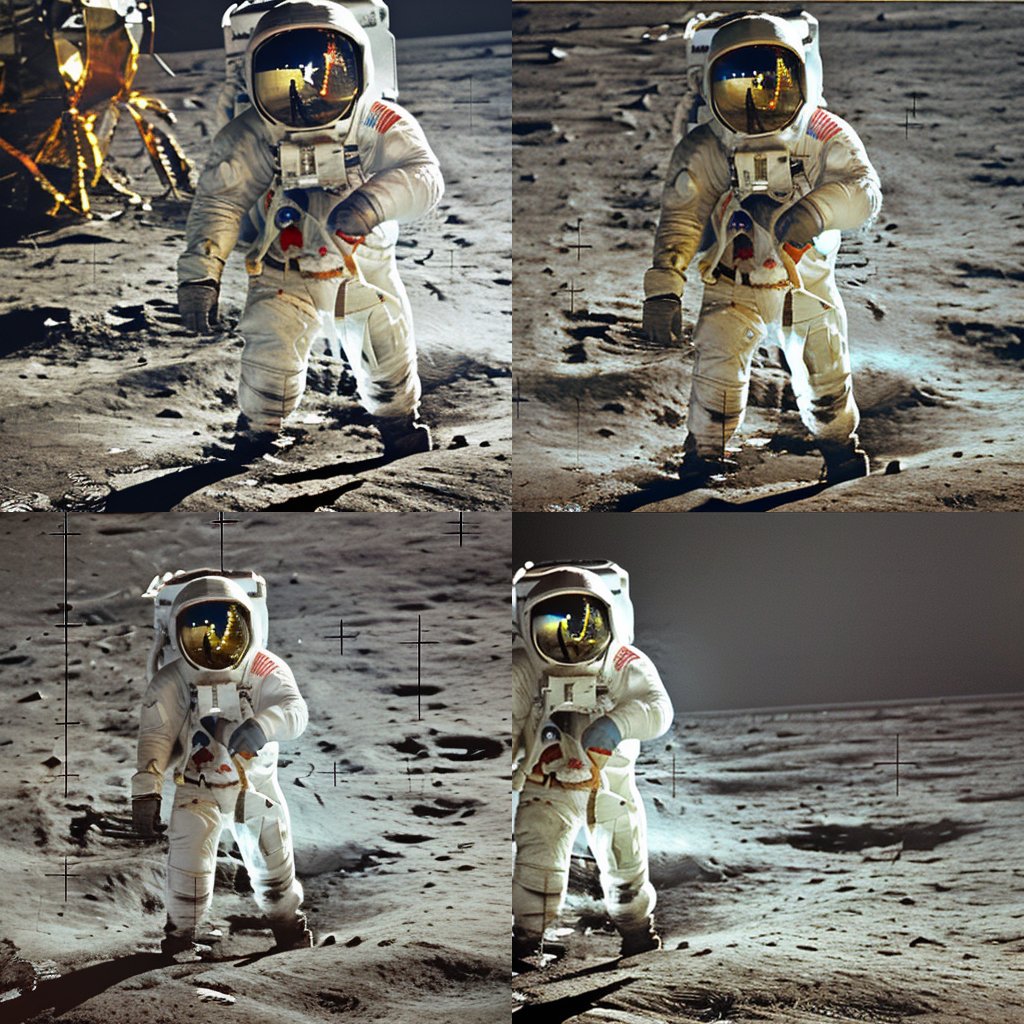}
                        };
                        \node[rotate=0, align=center, font=\normalsize, anchor=base, yshift=4pt] at (image.north) {\gls{ncg} (Ours) \strut};
                    \end{tikzpicture}
                \end{subfigure}
                \caption*{Astronaut on the Moon}
            \end{subfigure}
            \begin{subfigure}{0.49\textwidth}
                \begin{subfigure}{0.495\textwidth}
                    \begin{tikzpicture}
                        \node[anchor=south west, inner sep=0] (image) at (0,0) {
                            \includegraphics[width=\textwidth]{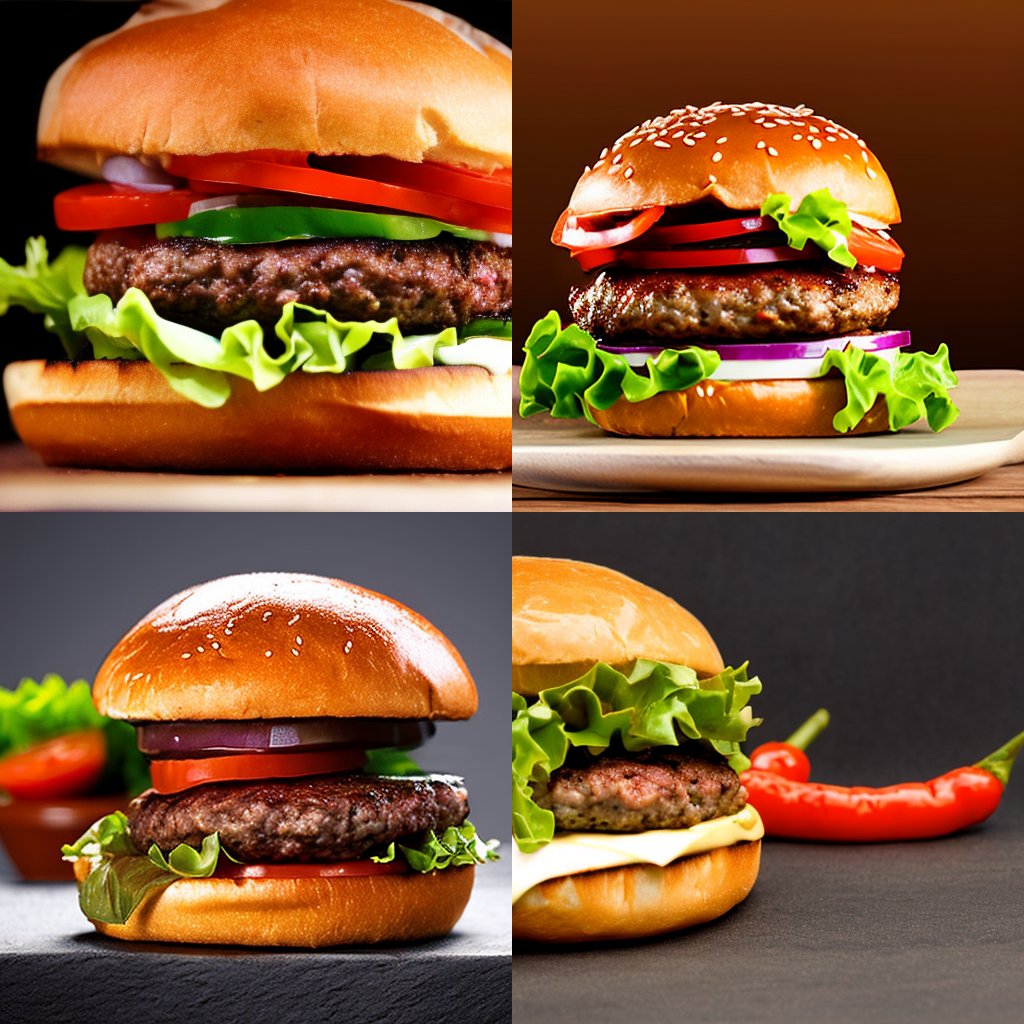}
                        };
                        \node[rotate=0, align=center, font=\normalsize, anchor=base, yshift=4pt] at (image.north) {CFG \strut};
                    \end{tikzpicture}
                \end{subfigure}
                \begin{subfigure}{0.495\textwidth}
                    \begin{tikzpicture}
                        \node[anchor=south west, inner sep=0] (image) at (0,0) {
                            \includegraphics[width=\textwidth]{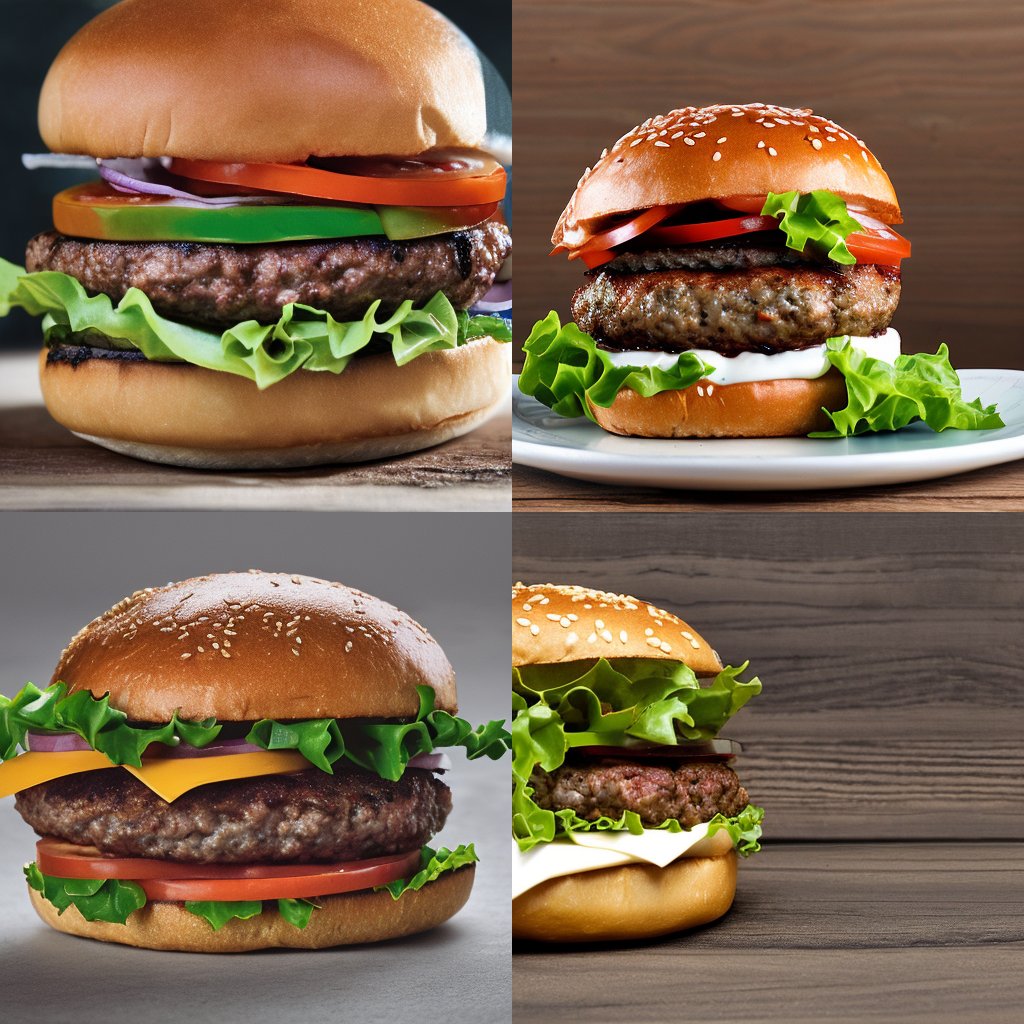}
                        };
                        \node[rotate=0, align=center, font=\normalsize, anchor=base, yshift=4pt] at (image.north) {\gls{ncg} (Ours) \strut};
                    \end{tikzpicture}
                \end{subfigure}
                \caption*{A burger}
            \end{subfigure}
        
            \begin{subfigure}{0.49\textwidth}
                \begin{subfigure}{0.495\textwidth}
                    \begin{tikzpicture}
                        \node[anchor=south west, inner sep=0] (image) at (0,0) {
                            \includegraphics[width=\textwidth]{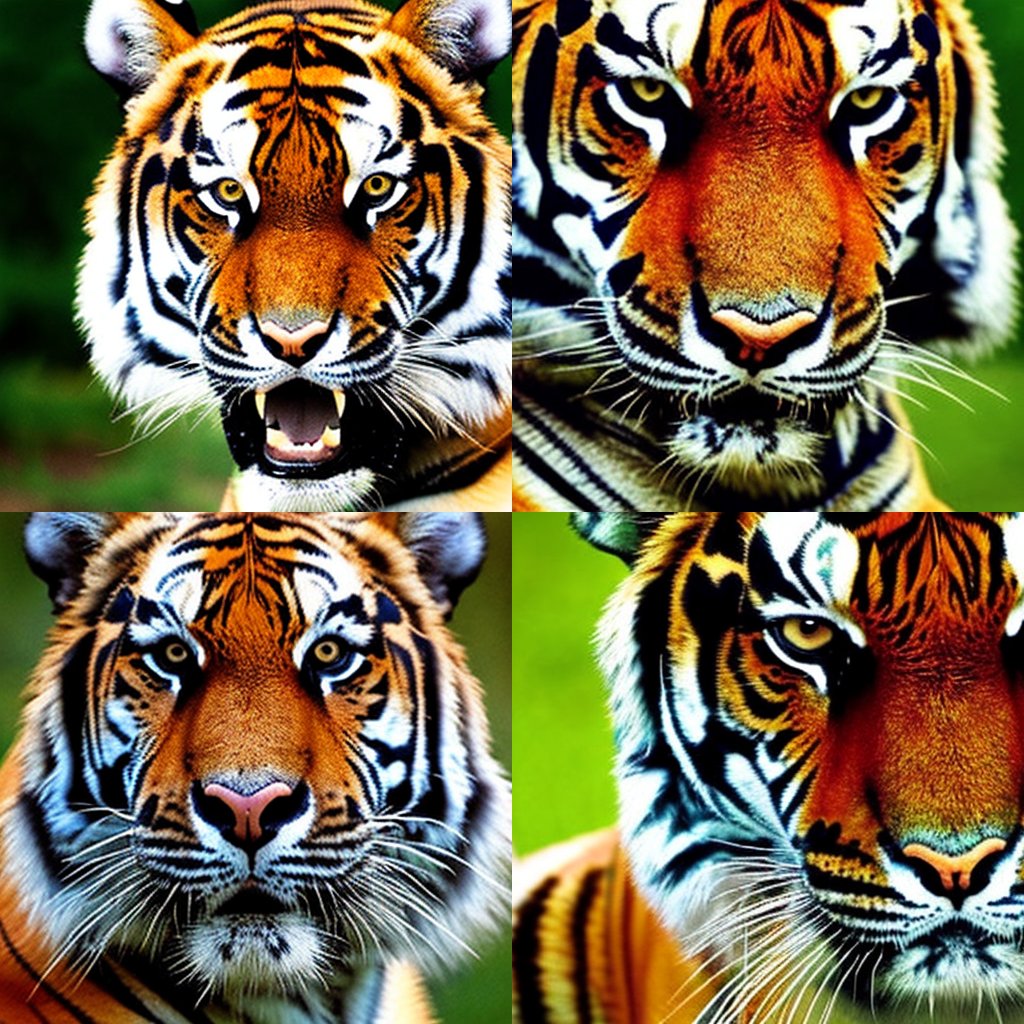}
                        };
                        \node[rotate=0, align=center, font=\normalsize, anchor=base, yshift=4pt] at (image.north) {CFG \strut};
                    \end{tikzpicture}
                \end{subfigure}
                \begin{subfigure}{0.495\textwidth}
                    \begin{tikzpicture}
                        \node[anchor=south west, inner sep=0] (image) at (0,0) {
                            \includegraphics[width=\textwidth]{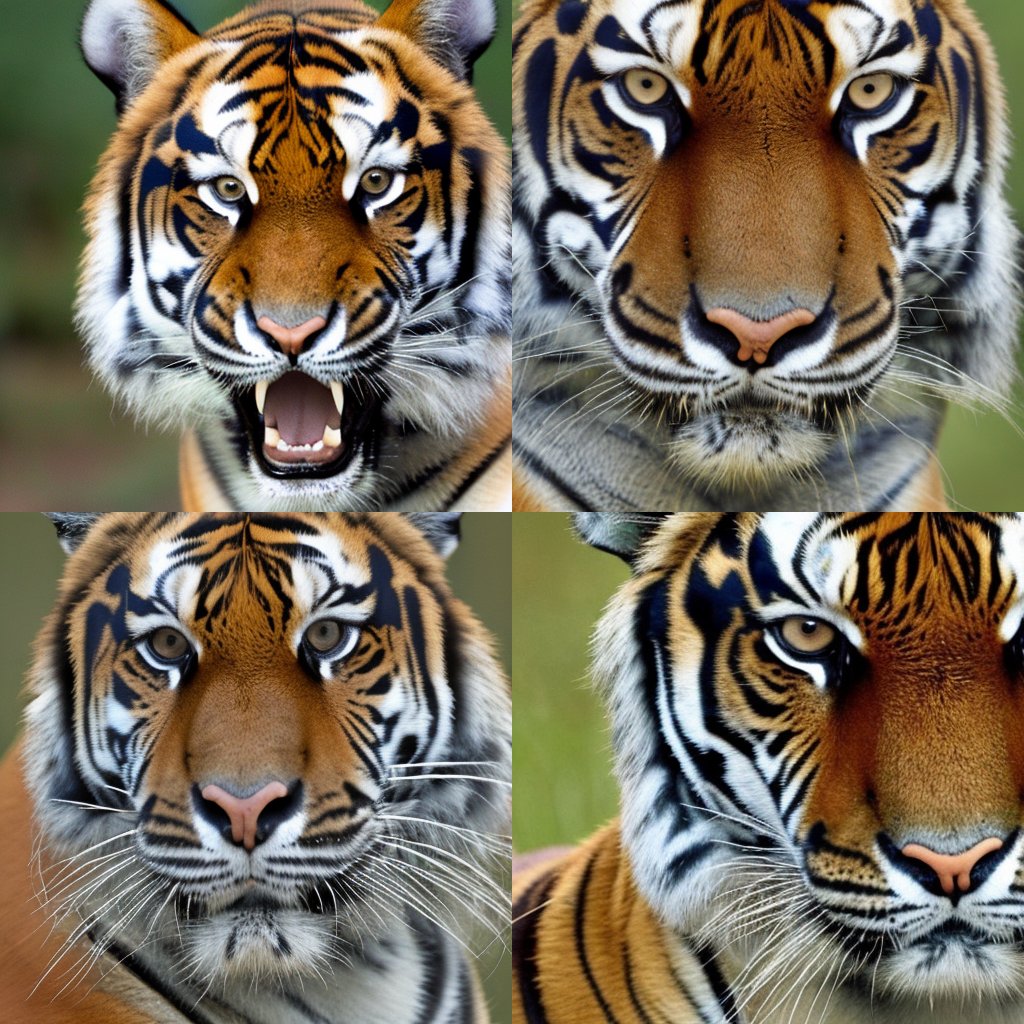}
                        };
                        \node[rotate=0, align=center, font=\normalsize, anchor=base, yshift=4pt] at (image.north) {\gls{ncg} (Ours) \strut};
                    \end{tikzpicture}
                \end{subfigure}
                \caption*{A tiger}
            \end{subfigure}
            \begin{subfigure}{0.49\textwidth}
                \begin{subfigure}{0.495\textwidth}
                    \begin{tikzpicture}
                        \node[anchor=south west, inner sep=0] (image) at (0,0) {
                            \includegraphics[width=\textwidth]{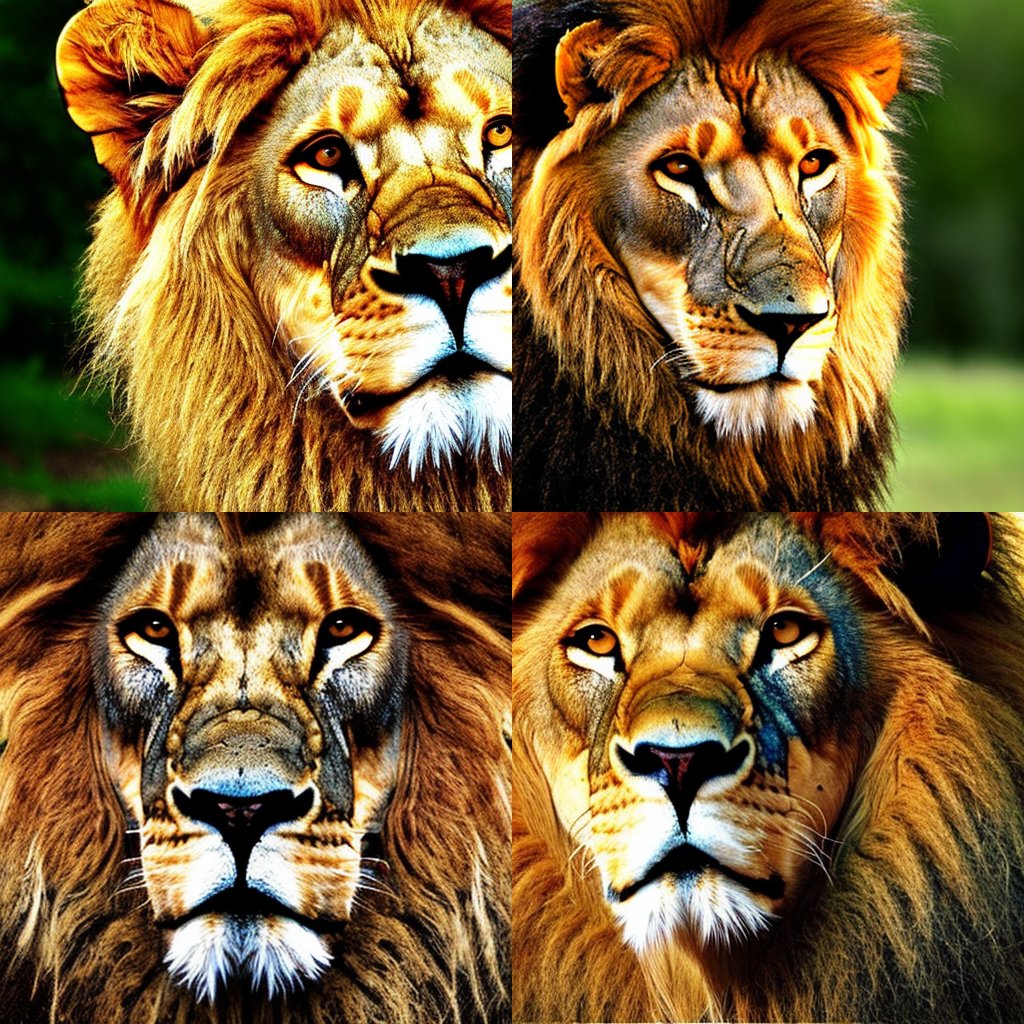}
                        };
                        \node[rotate=0, align=center, font=\normalsize, anchor=base, yshift=4pt] at (image.north) {CFG \strut};
                    \end{tikzpicture}
                \end{subfigure}
                \begin{subfigure}{0.495\textwidth}
                    \begin{tikzpicture}
                        \node[anchor=south west, inner sep=0] (image) at (0,0) {
                            \includegraphics[width=\textwidth]{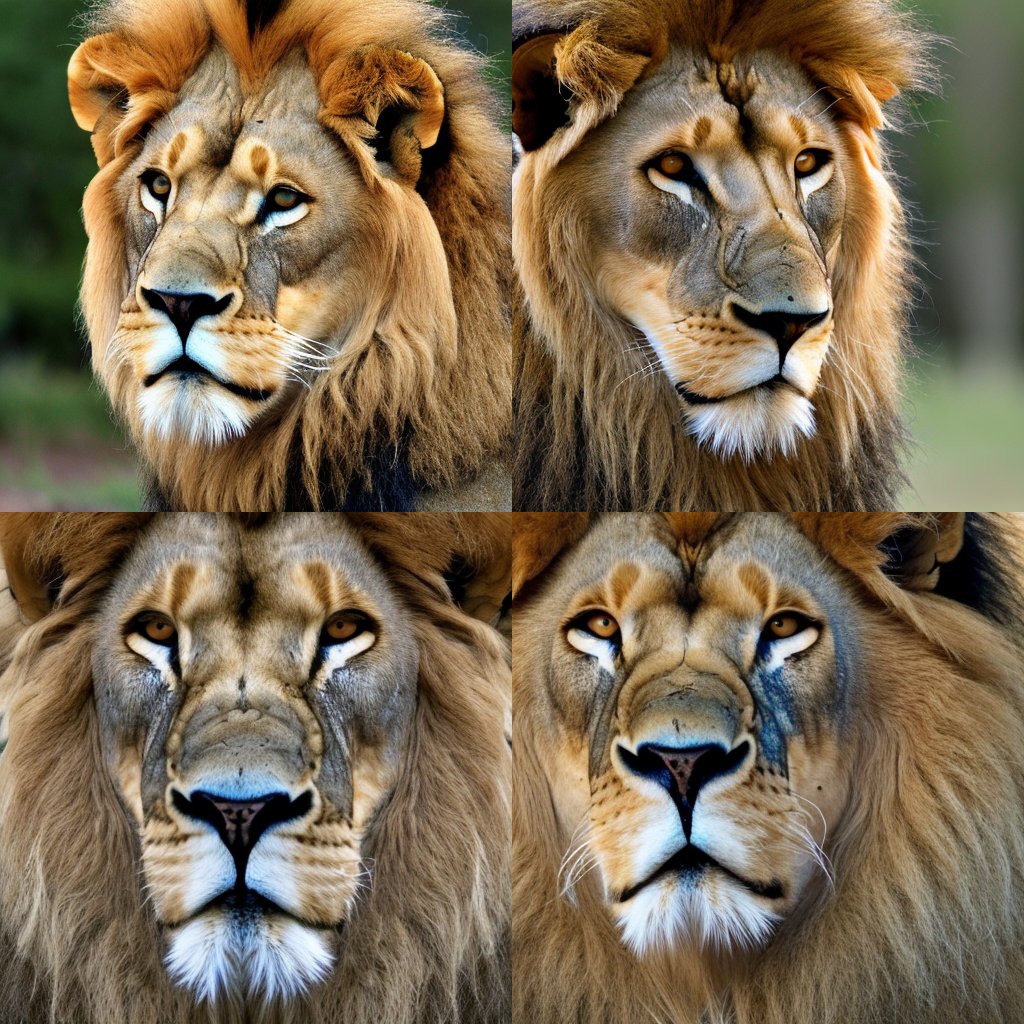}
                        };
                        \node[rotate=0, align=center, font=\normalsize, anchor=base, yshift=4pt] at (image.north) {\gls{ncg} (Ours) \strut};
                    \end{tikzpicture}
                \end{subfigure}
                \caption*{A lion}
            \end{subfigure}
        \end{minipage}
        \caption{More visual examples comparing CFG and \gls{ncg} using Stable Diffusion 2.1.}
        \label{fig:sd2-appendix}
    \end{figure}
}

\newcommand{\figSDXLAppendix}{
    \begin{figure}[t]
        \centering
        \begin{minipage}{\textwidth}
            \begin{subfigure}{0.49\textwidth}
                \begin{subfigure}{0.495\textwidth}
                    \begin{tikzpicture}
                        \node[anchor=south west, inner sep=0] (image) at (0,0) {
                            \includegraphics[width=\textwidth]{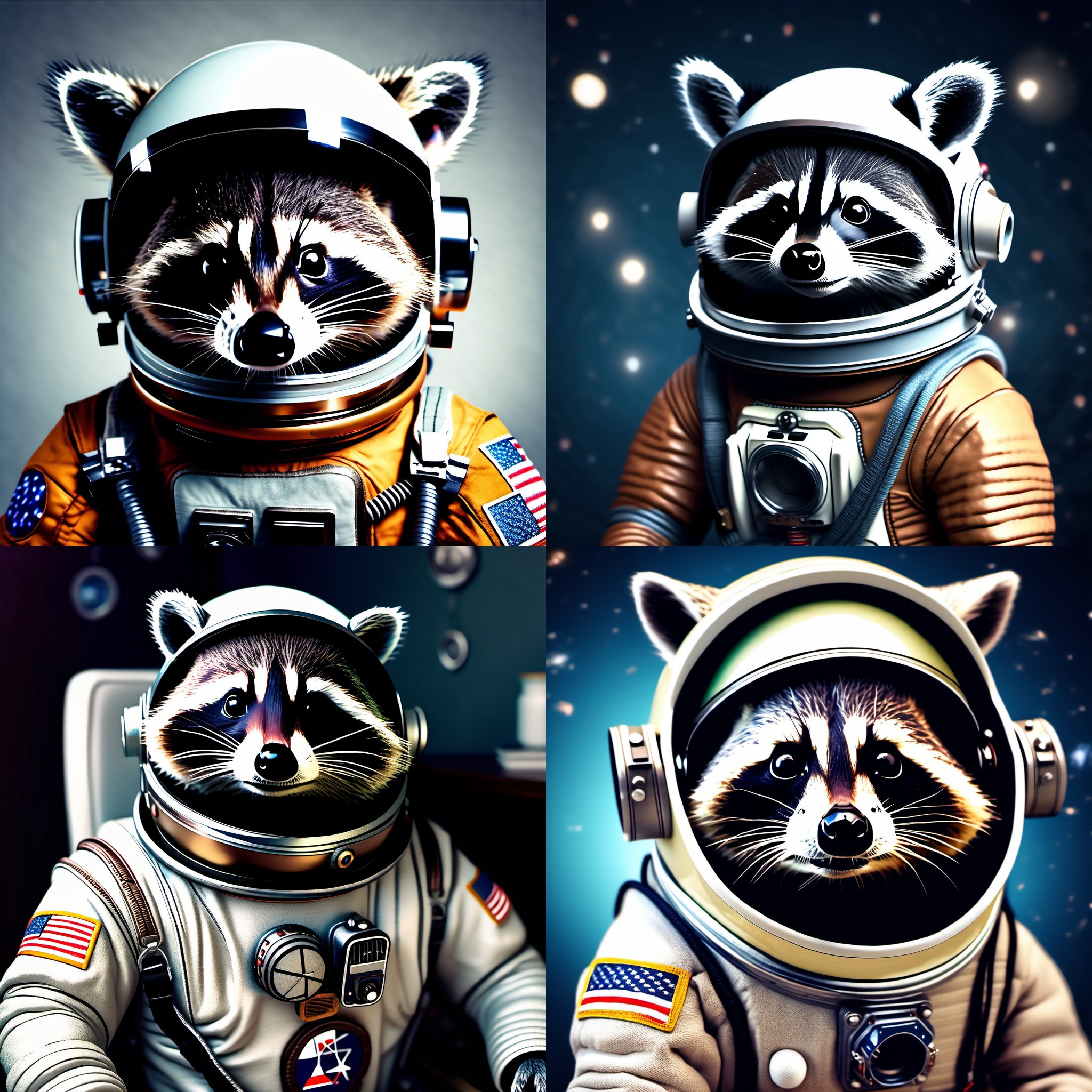}
                        };
                        \node[rotate=0, align=center, font=\normalsize, anchor=base, yshift=4pt] at (image.north) {CFG \strut};
                    \end{tikzpicture}
                \end{subfigure}
                \begin{subfigure}{0.495\textwidth}
                    \begin{tikzpicture}
                        \node[anchor=south west, inner sep=0] (image) at (0,0) {
                            \includegraphics[width=\textwidth]{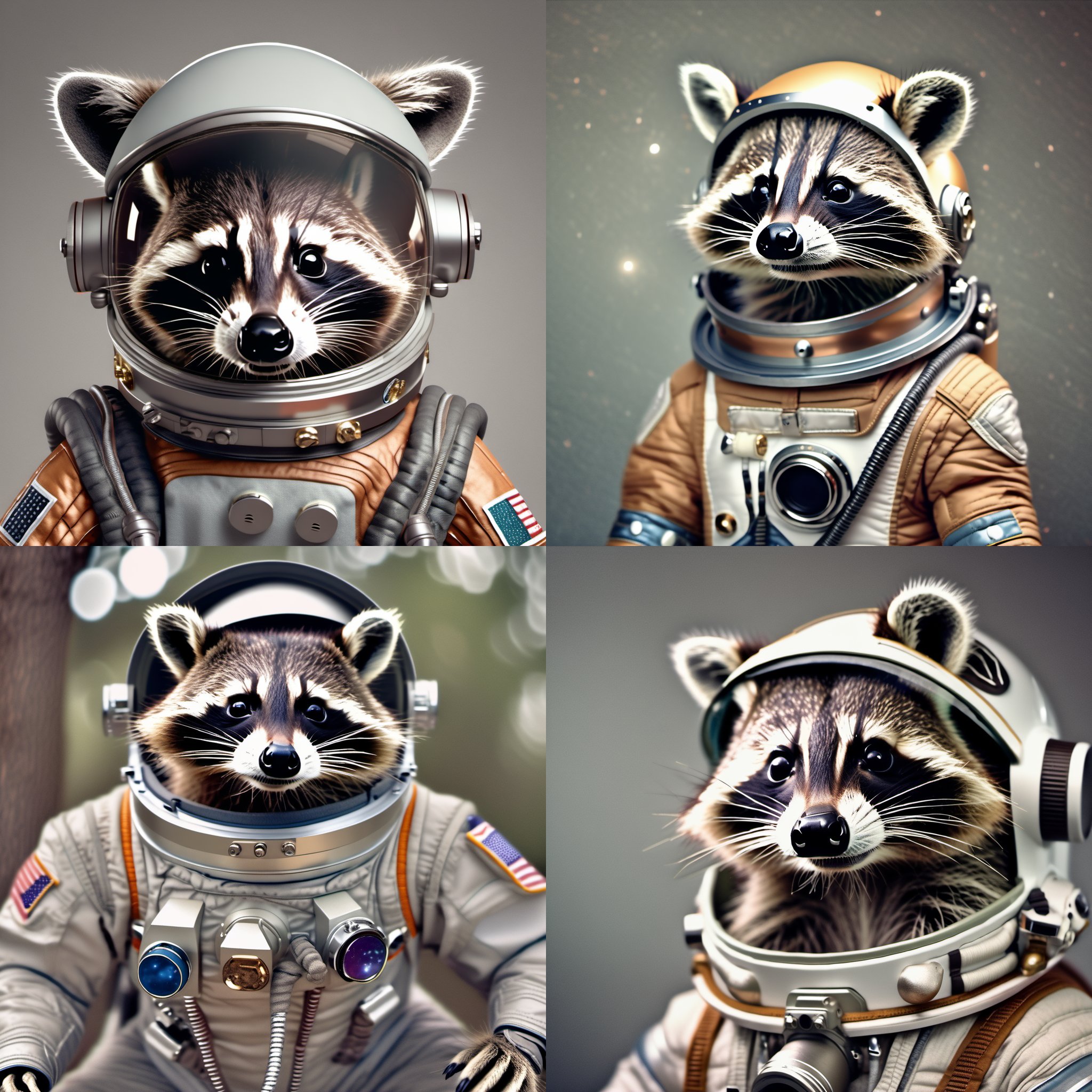}
                        };
                        \node[rotate=0, align=center, font=\normalsize, anchor=base, yshift=4pt] at (image.north) {\gls{ncg} (Ours) \strut};
                    \end{tikzpicture}
                \end{subfigure}
                \caption*{A 4k dslr photo of a raccoon wearing an astronaut helmet, photorealistic.}
            \end{subfigure}
            \begin{subfigure}{0.49\textwidth}
                \begin{subfigure}{0.495\textwidth}
                    \begin{tikzpicture}
                        \node[anchor=south west, inner sep=0] (image) at (0,0) {
                            \includegraphics[width=\textwidth]{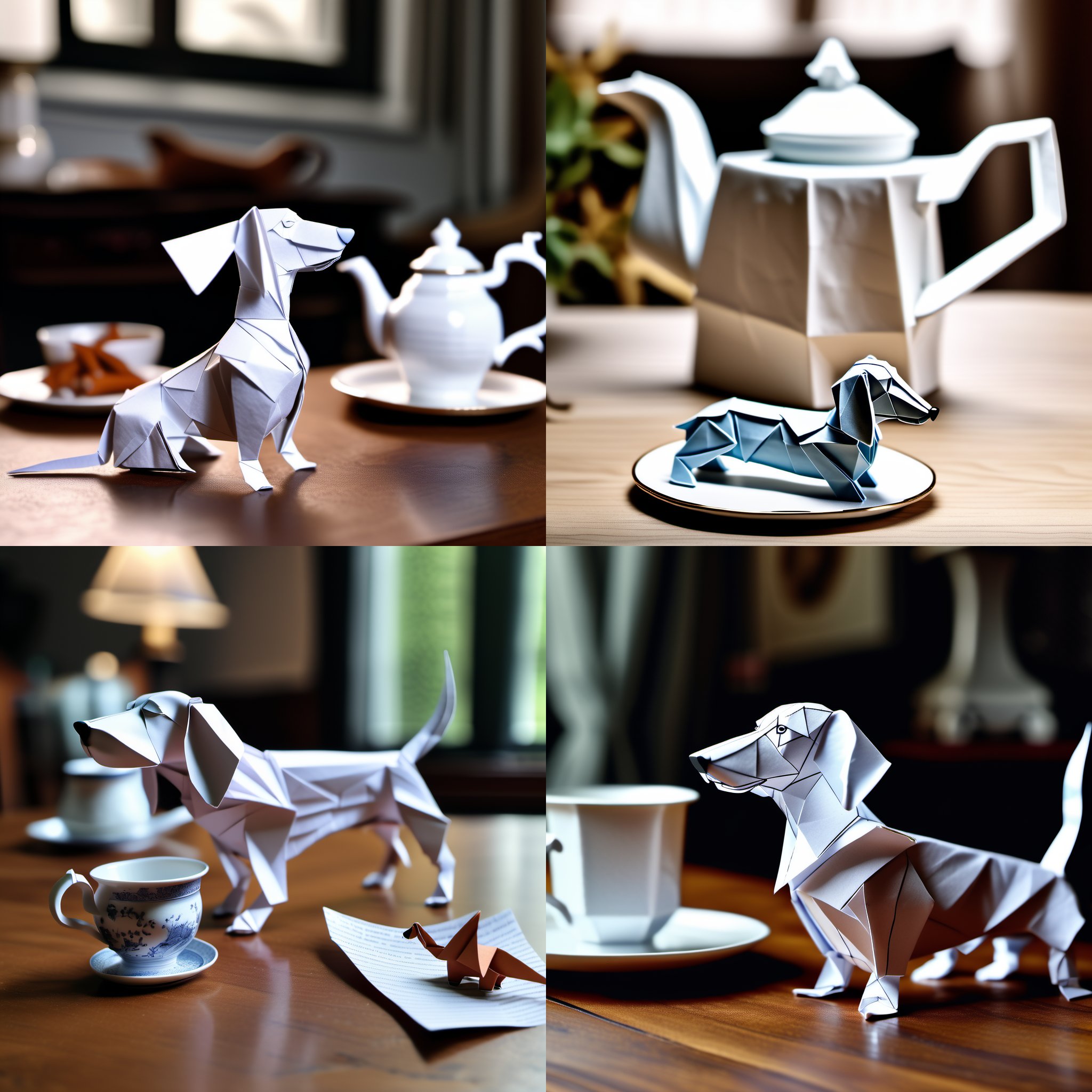}
                        };
                        \node[rotate=0, align=center, font=\normalsize, anchor=base, yshift=4pt] at (image.north) {CFG \strut};
                    \end{tikzpicture}
                \end{subfigure}
                \begin{subfigure}{0.495\textwidth}
                    \begin{tikzpicture}
                        \node[anchor=south west, inner sep=0] (image) at (0,0) {
                            \includegraphics[width=\textwidth]{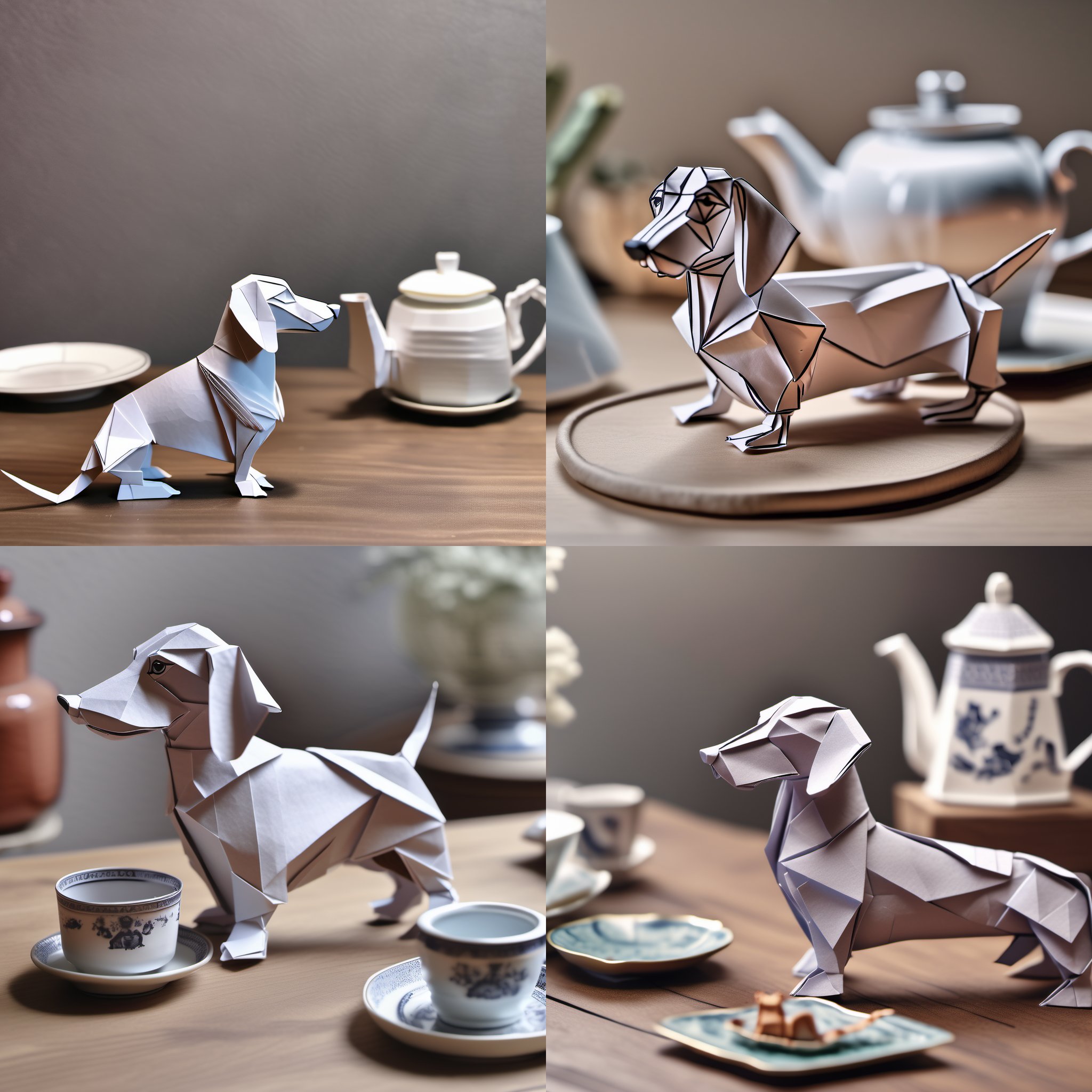}
                        };
                        \node[rotate=0, align=center, font=\normalsize, anchor=base, yshift=4pt] at (image.north) {\gls{ncg} (Ours) \strut};
                    \end{tikzpicture}
                \end{subfigure}
                \caption*{A highly detailed paper origami of a Dachshund on a table next to a porcelain teapot, 4k dslr.}
            \end{subfigure}
        
            \begin{subfigure}{0.49\textwidth}
                \begin{subfigure}{0.495\textwidth}
                    \begin{tikzpicture}
                        \node[anchor=south west, inner sep=0] (image) at (0,0) {
                            \includegraphics[width=\textwidth]{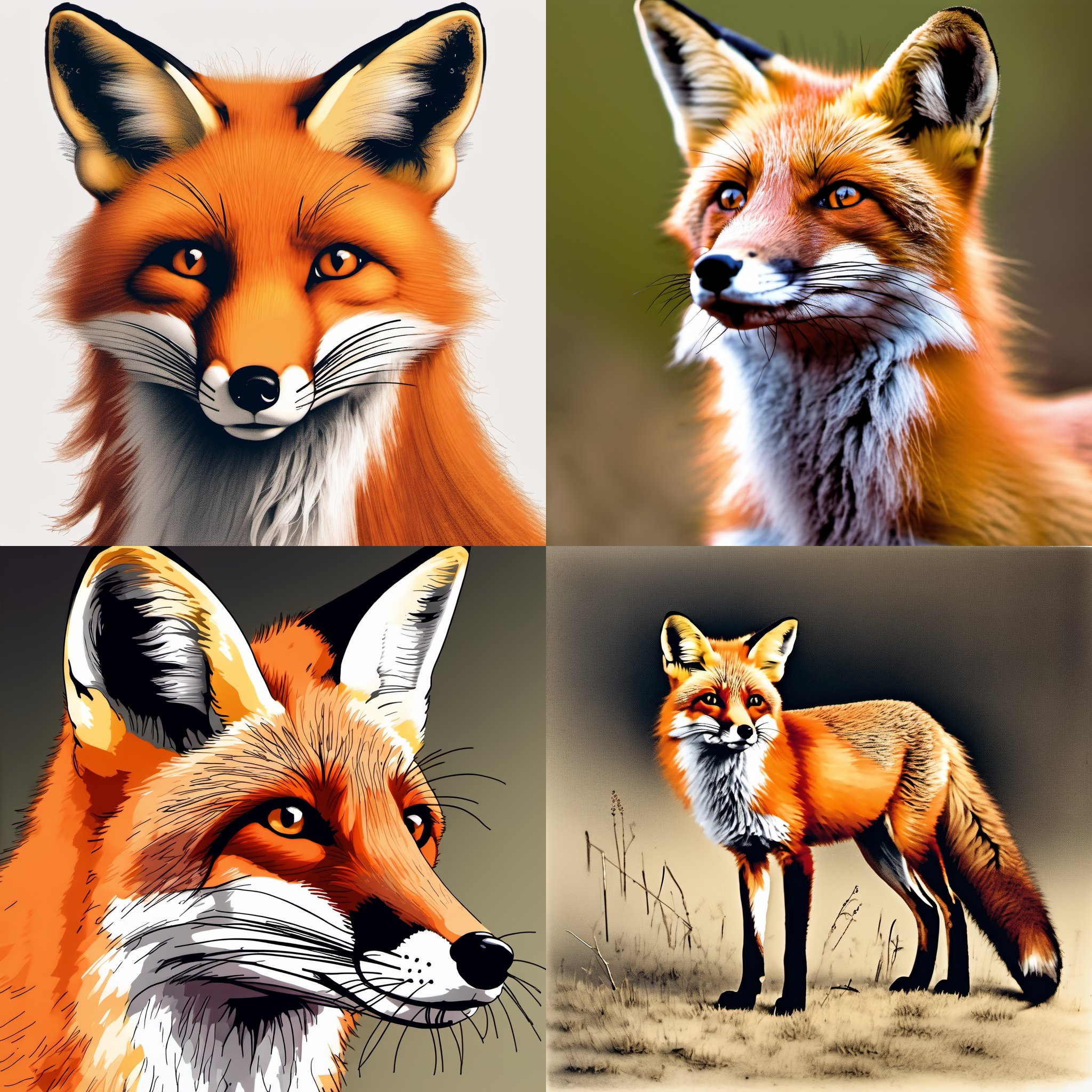}
                        };
                        \node[rotate=0, align=center, font=\normalsize, anchor=base, yshift=4pt] at (image.north) {CFG \strut};
                    \end{tikzpicture}
                \end{subfigure}
                \begin{subfigure}{0.495\textwidth}
                    \begin{tikzpicture}
                        \node[anchor=south west, inner sep=0] (image) at (0,0) {
                            \includegraphics[width=\textwidth]{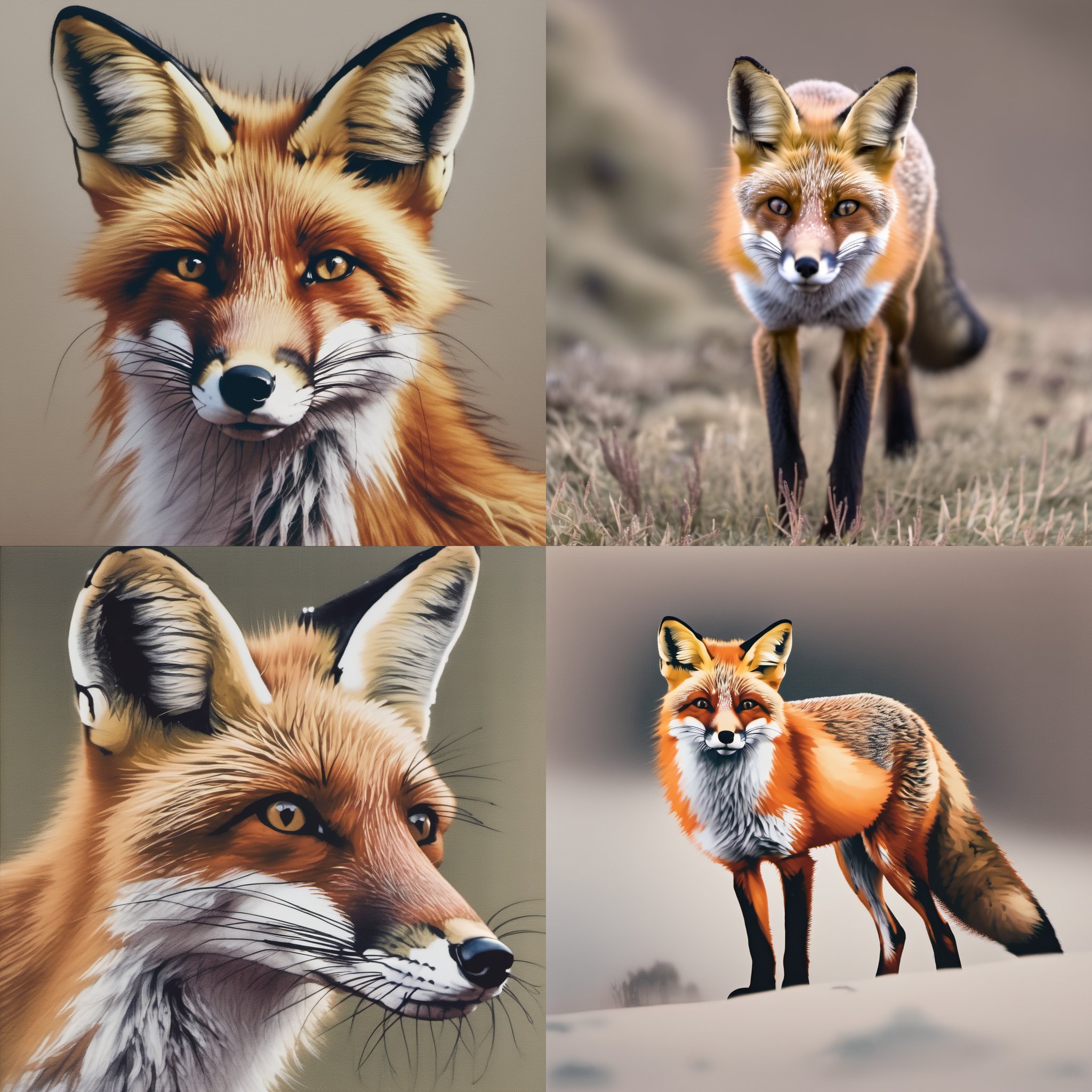}
                        };
                        \node[rotate=0, align=center, font=\normalsize, anchor=base, yshift=4pt] at (image.north) {\gls{ncg} (Ours) \strut};
                    \end{tikzpicture}
                \end{subfigure}
                \caption*{A fox}
            \end{subfigure}
            \begin{subfigure}{0.49\textwidth}
                \begin{subfigure}{0.495\textwidth}
                    \begin{tikzpicture}
                        \node[anchor=south west, inner sep=0] (image) at (0,0) {
                            \includegraphics[width=\textwidth]{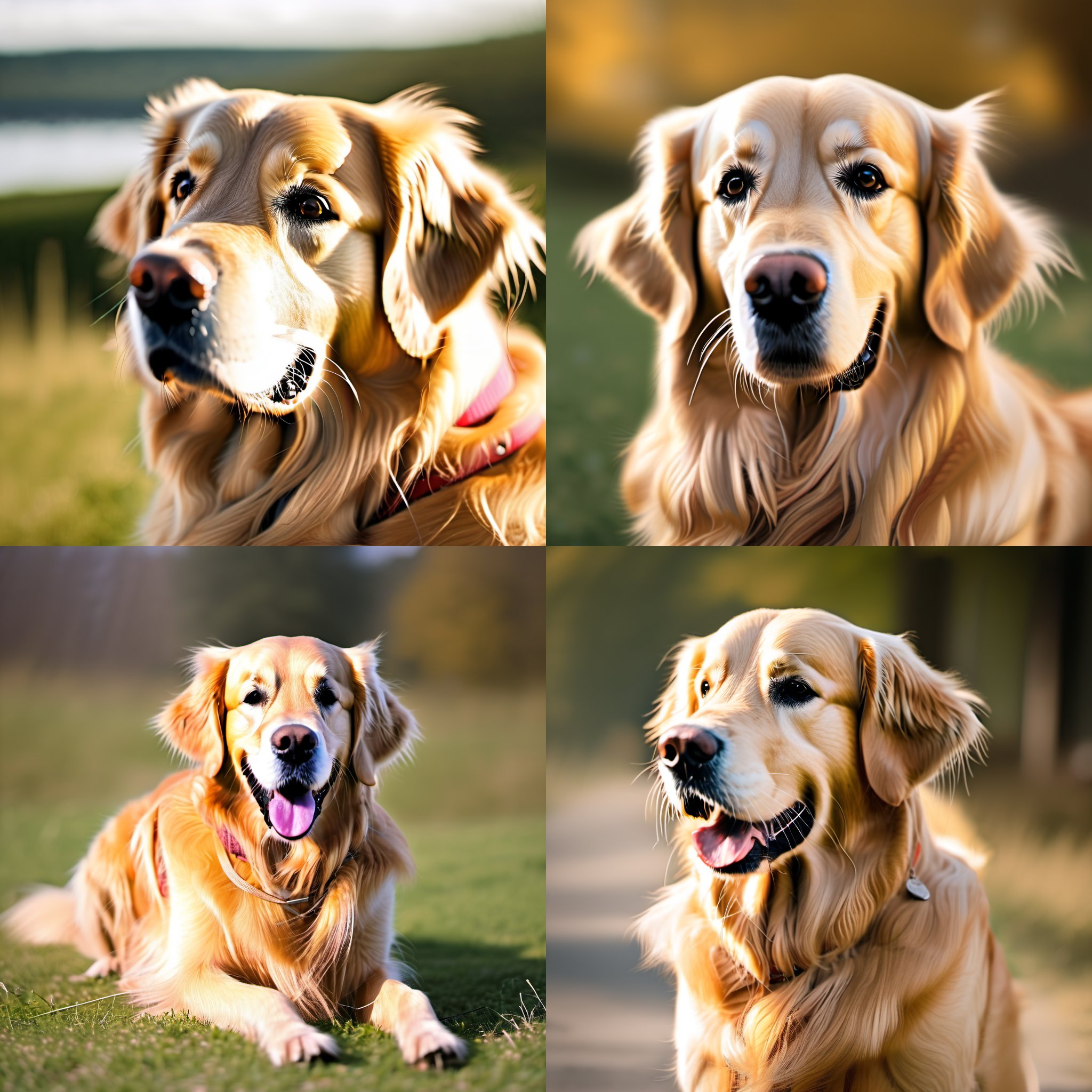}
                        };
                        \node[rotate=0, align=center, font=\normalsize, anchor=base, yshift=4pt] at (image.north) {CFG \strut};
                    \end{tikzpicture}
                \end{subfigure}
                \begin{subfigure}{0.495\textwidth}
                    \begin{tikzpicture}
                        \node[anchor=south west, inner sep=0] (image) at (0,0) {
                            \includegraphics[width=\textwidth]{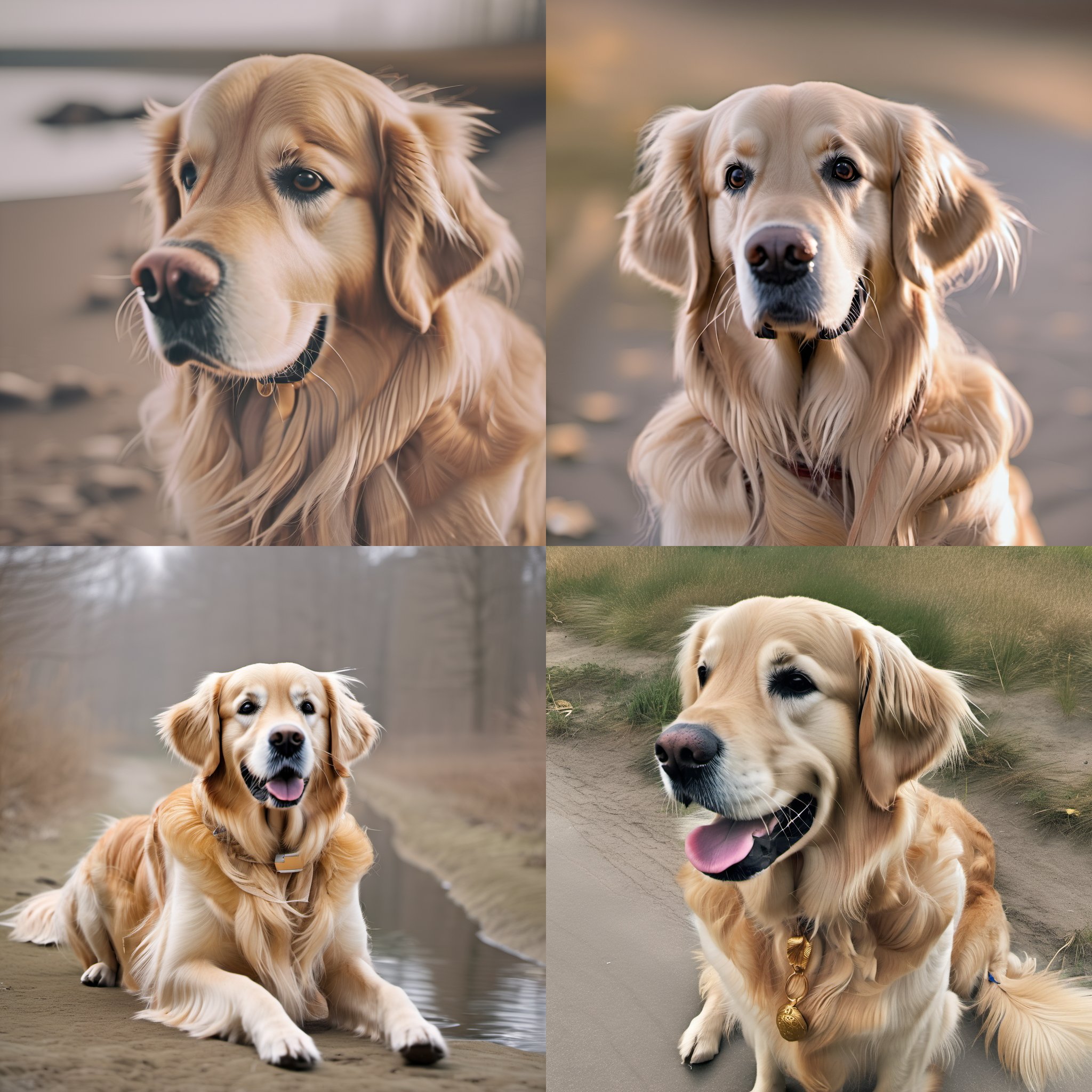}
                        };
                        \node[rotate=0, align=center, font=\normalsize, anchor=base, yshift=4pt] at (image.north) {\gls{ncg} (Ours) \strut};
                    \end{tikzpicture}
                \end{subfigure}
                \caption*{A golden retriever}
            \end{subfigure}
        \end{minipage}
        \caption{More visual examples comparing CFG and \gls{ncg} using Stable Diffusion XL.}
        \label{fig:sdxl-appendix}
    \end{figure}
}

\newcommand{\figPixartLCMAppendix}{
    \begin{figure}[t]
        \centering
        \begin{minipage}{\textwidth}
            \begin{subfigure}{0.49\textwidth}
                \begin{subfigure}{0.495\textwidth}
                    \begin{tikzpicture}
                        \node[anchor=south west, inner sep=0] (image) at (0,0) {
                            \includegraphics[width=\textwidth]{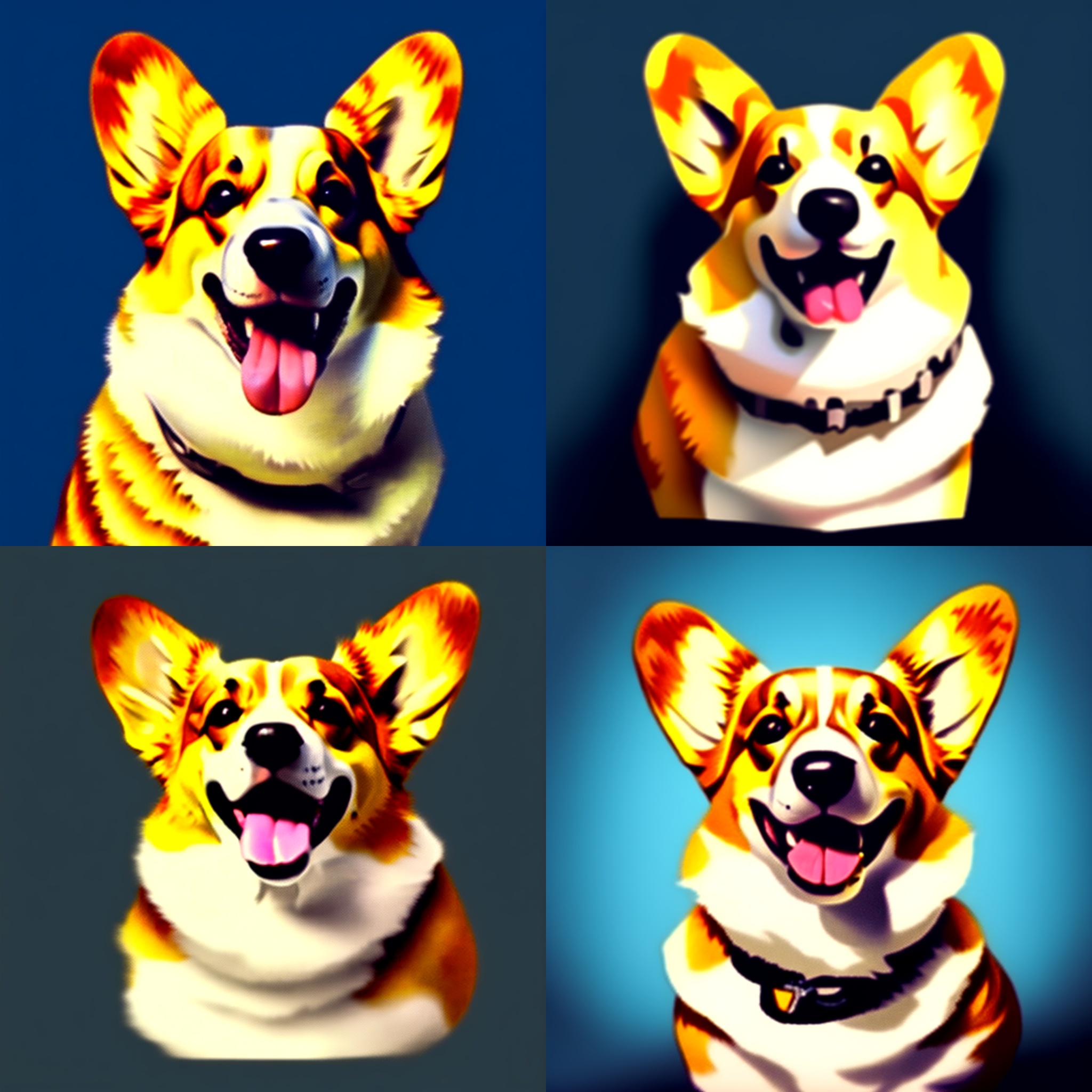}
                        };
                        \node[rotate=0, align=center, font=\normalsize, anchor=base, yshift=4pt] at (image.north) {CFG \strut};
                    \end{tikzpicture}
                \end{subfigure}
                \begin{subfigure}{0.495\textwidth}
                    \begin{tikzpicture}
                        \node[anchor=south west, inner sep=0] (image) at (0,0) {
                            \includegraphics[width=\textwidth]{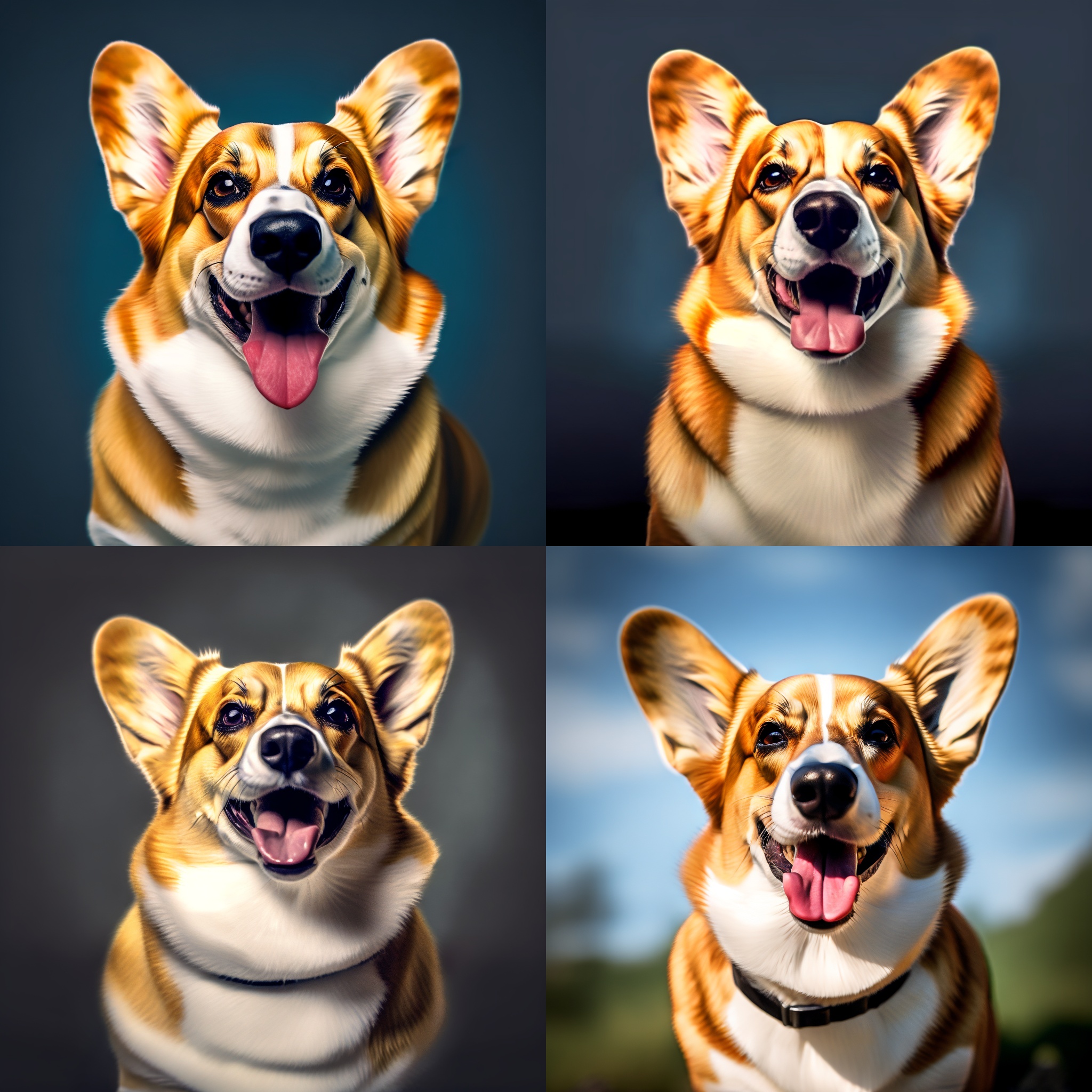}
                        };
                        \node[rotate=0, align=center, font=\normalsize, anchor=base, yshift=4pt] at (image.north) {\gls{ncg} (Ours) \strut};
                    \end{tikzpicture}
                \end{subfigure}
                \caption*{Image of a corgi}
            \end{subfigure}
            \begin{subfigure}{0.49\textwidth}
                \begin{subfigure}{0.495\textwidth}
                    \begin{tikzpicture}
                        \node[anchor=south west, inner sep=0] (image) at (0,0) {
                            \includegraphics[width=\textwidth]{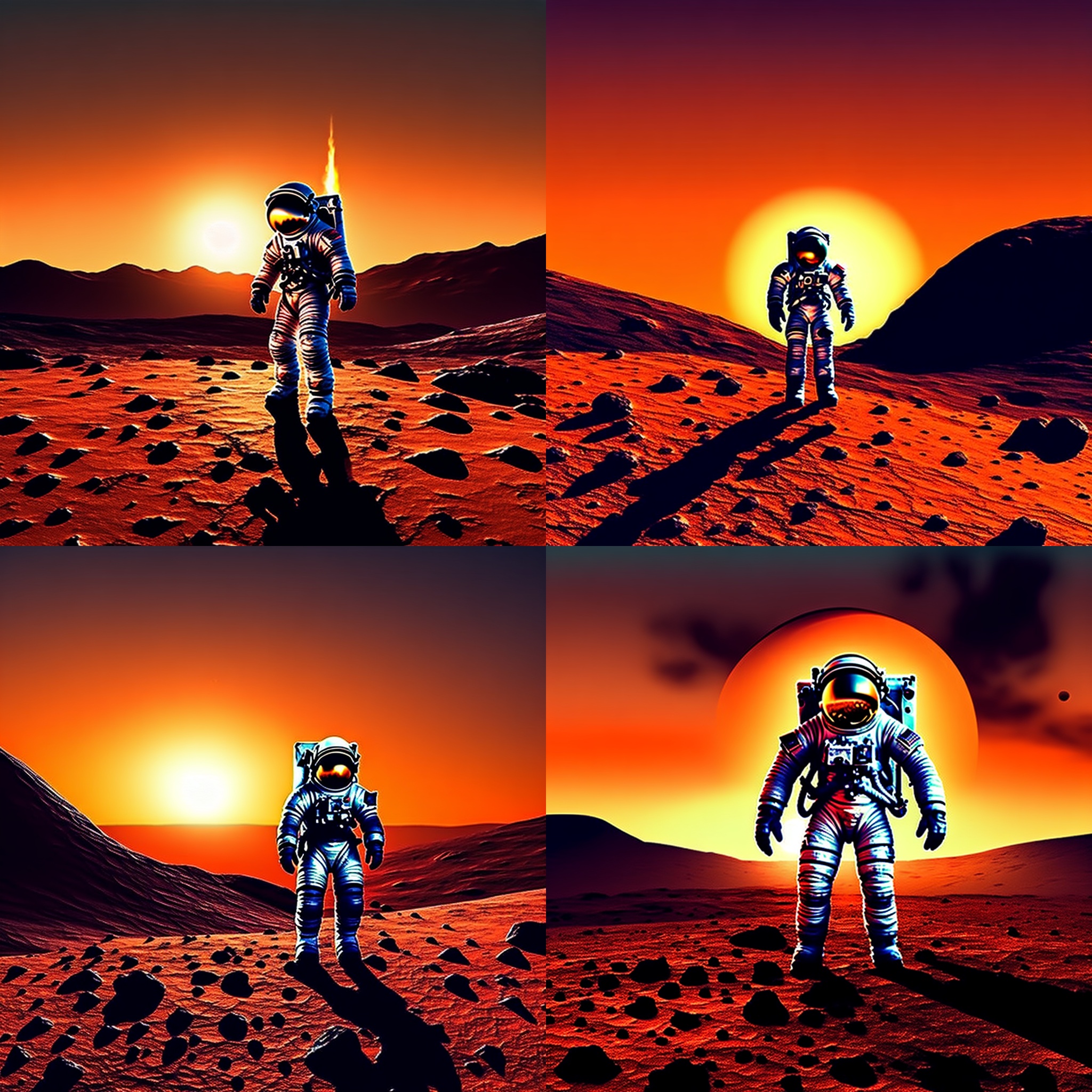}
                        };
                        \node[rotate=0, align=center, font=\normalsize, anchor=base, yshift=4pt] at (image.north) {CFG \strut};
                    \end{tikzpicture}
                \end{subfigure}
                \begin{subfigure}{0.495\textwidth}
                    \begin{tikzpicture}
                        \node[anchor=south west, inner sep=0] (image) at (0,0) {
                            \includegraphics[width=\textwidth]{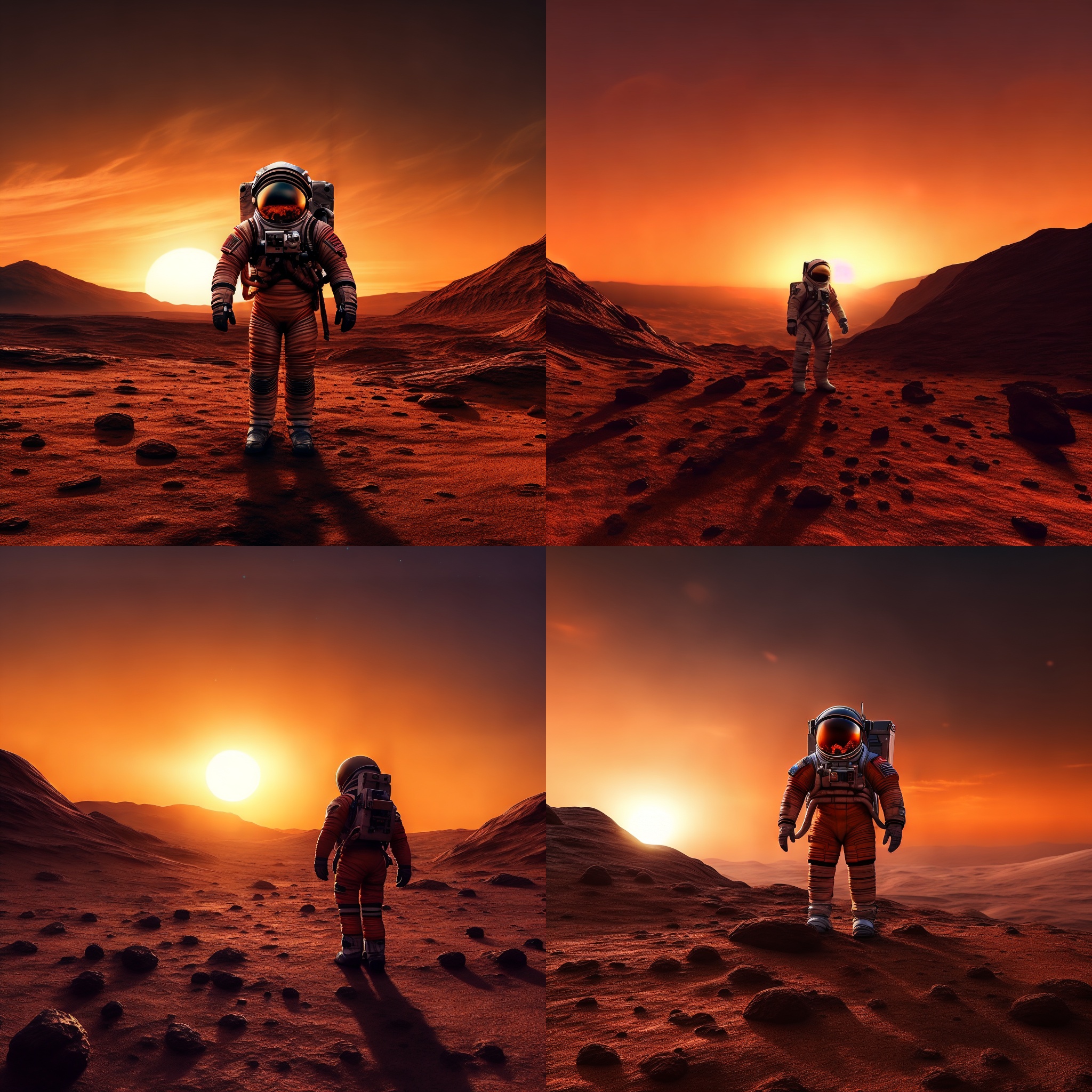}
                        };
                        \node[rotate=0, align=center, font=\normalsize, anchor=base, yshift=4pt] at (image.north) {\gls{ncg} (Ours) \strut};
                    \end{tikzpicture}
                \end{subfigure}
                \caption*{Astronaut on Mars during sunset}
            \end{subfigure}
        
            \begin{subfigure}{0.49\textwidth}
                \begin{subfigure}{0.495\textwidth}
                    \begin{tikzpicture}
                        \node[anchor=south west, inner sep=0] (image) at (0,0) {
                            \includegraphics[width=\textwidth]{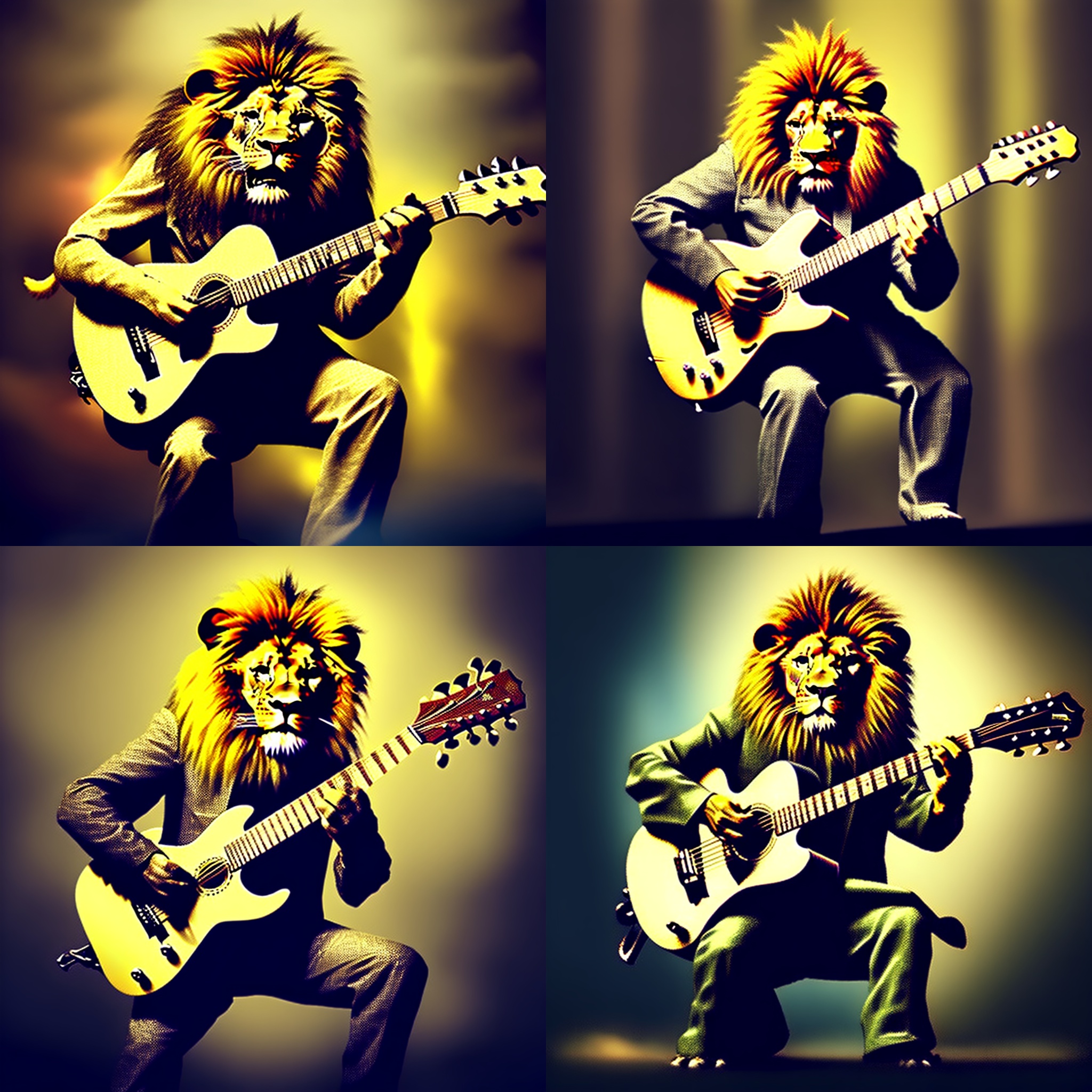}
                        };
                        \node[rotate=0, align=center, font=\normalsize, anchor=base, yshift=4pt] at (image.north) {CFG \strut};
                    \end{tikzpicture}
                \end{subfigure}
                \begin{subfigure}{0.495\textwidth}
                    \begin{tikzpicture}
                        \node[anchor=south west, inner sep=0] (image) at (0,0) {
                            \includegraphics[width=\textwidth]{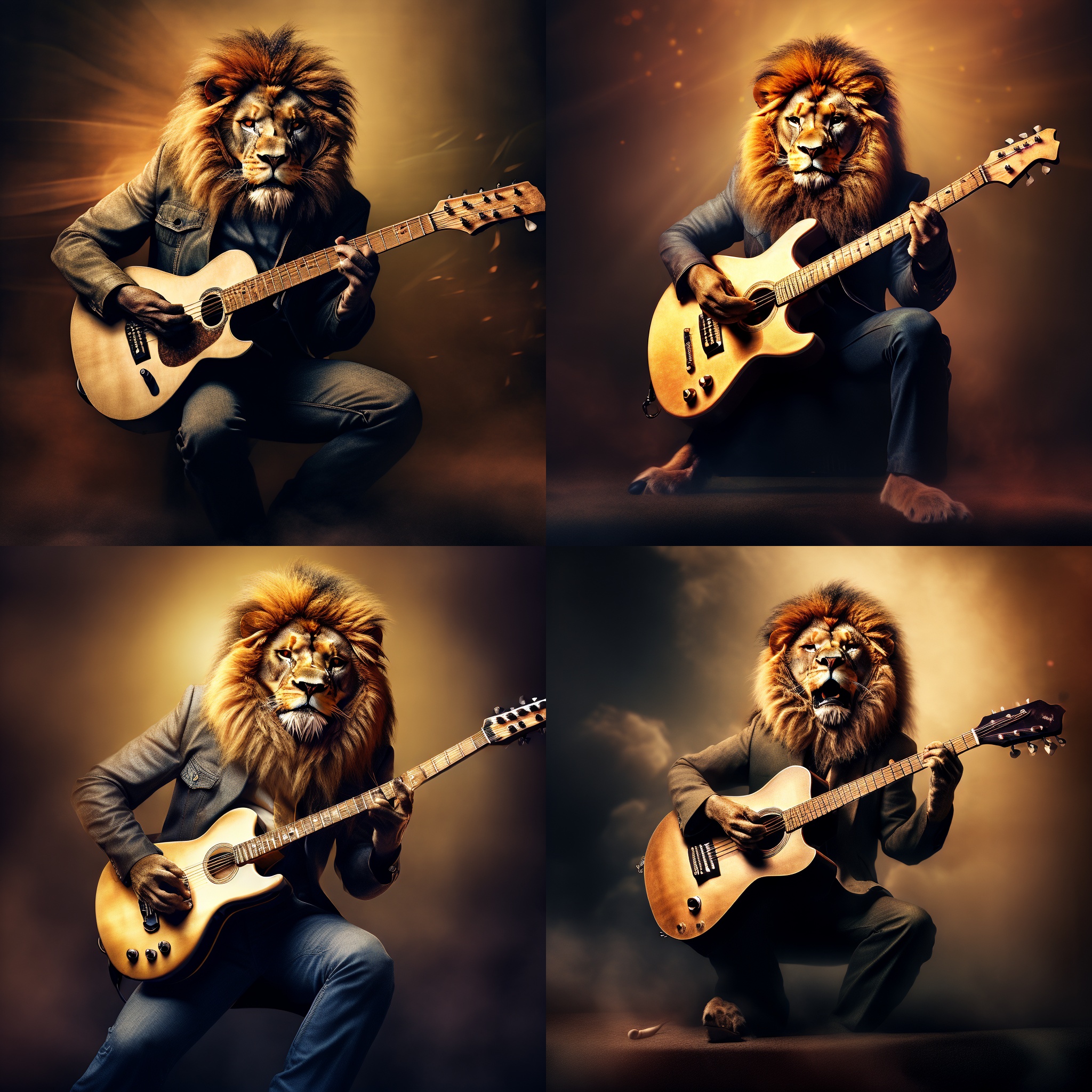}
                        };
                        \node[rotate=0, align=center, font=\normalsize, anchor=base, yshift=4pt] at (image.north) {\gls{ncg} (Ours) \strut};
                    \end{tikzpicture}
                \end{subfigure}
                \caption*{A lion is playing guitar}
            \end{subfigure}
            \begin{subfigure}{0.49\textwidth}
                \begin{subfigure}{0.495\textwidth}
                    \begin{tikzpicture}
                        \node[anchor=south west, inner sep=0] (image) at (0,0) {
                            \includegraphics[width=\textwidth]{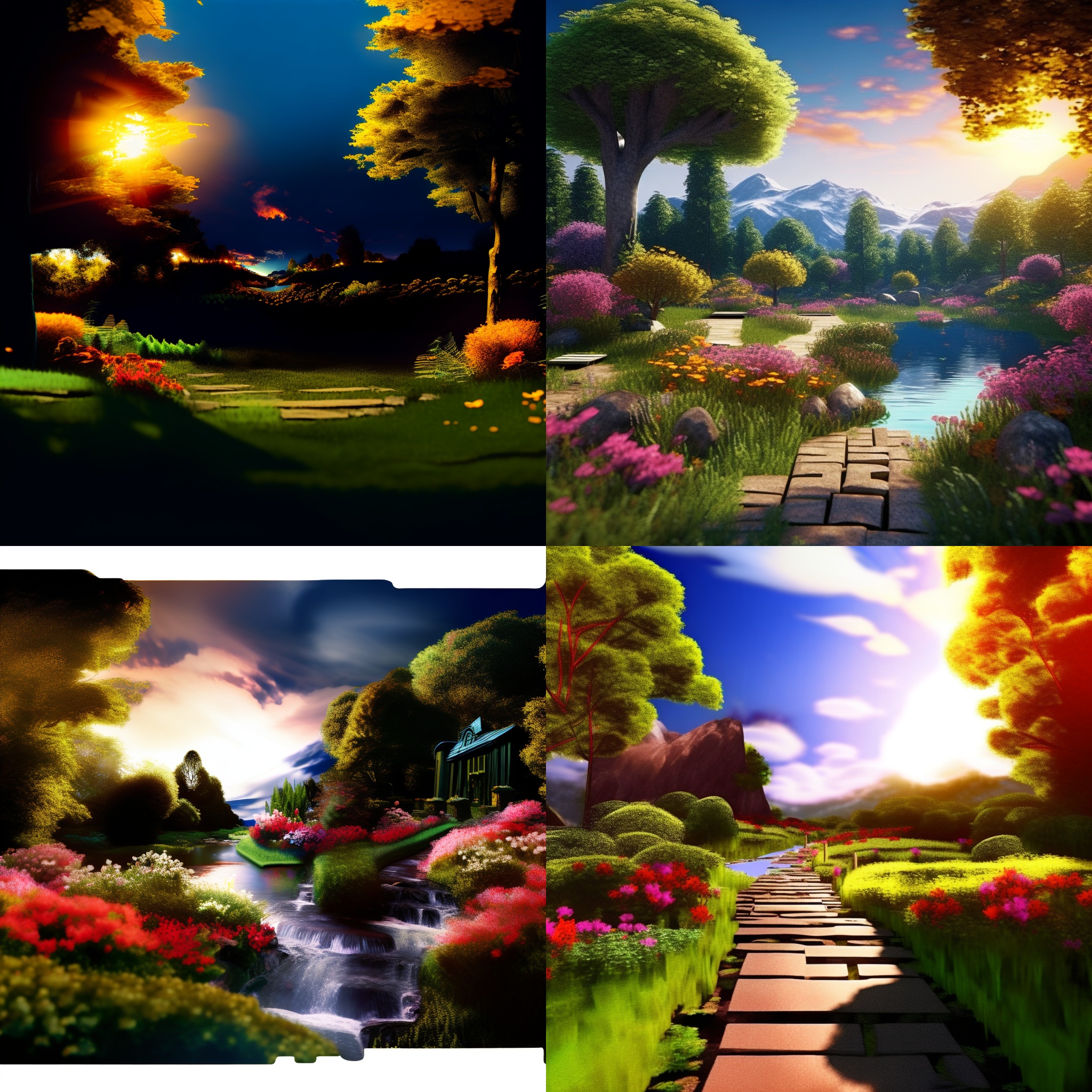}
                        };
                        \node[rotate=0, align=center, font=\normalsize, anchor=base, yshift=4pt] at (image.north) {CFG \strut};
                    \end{tikzpicture}
                \end{subfigure}
                \begin{subfigure}{0.495\textwidth}
                    \begin{tikzpicture}
                        \node[anchor=south west, inner sep=0] (image) at (0,0) {
                            \includegraphics[width=\textwidth]{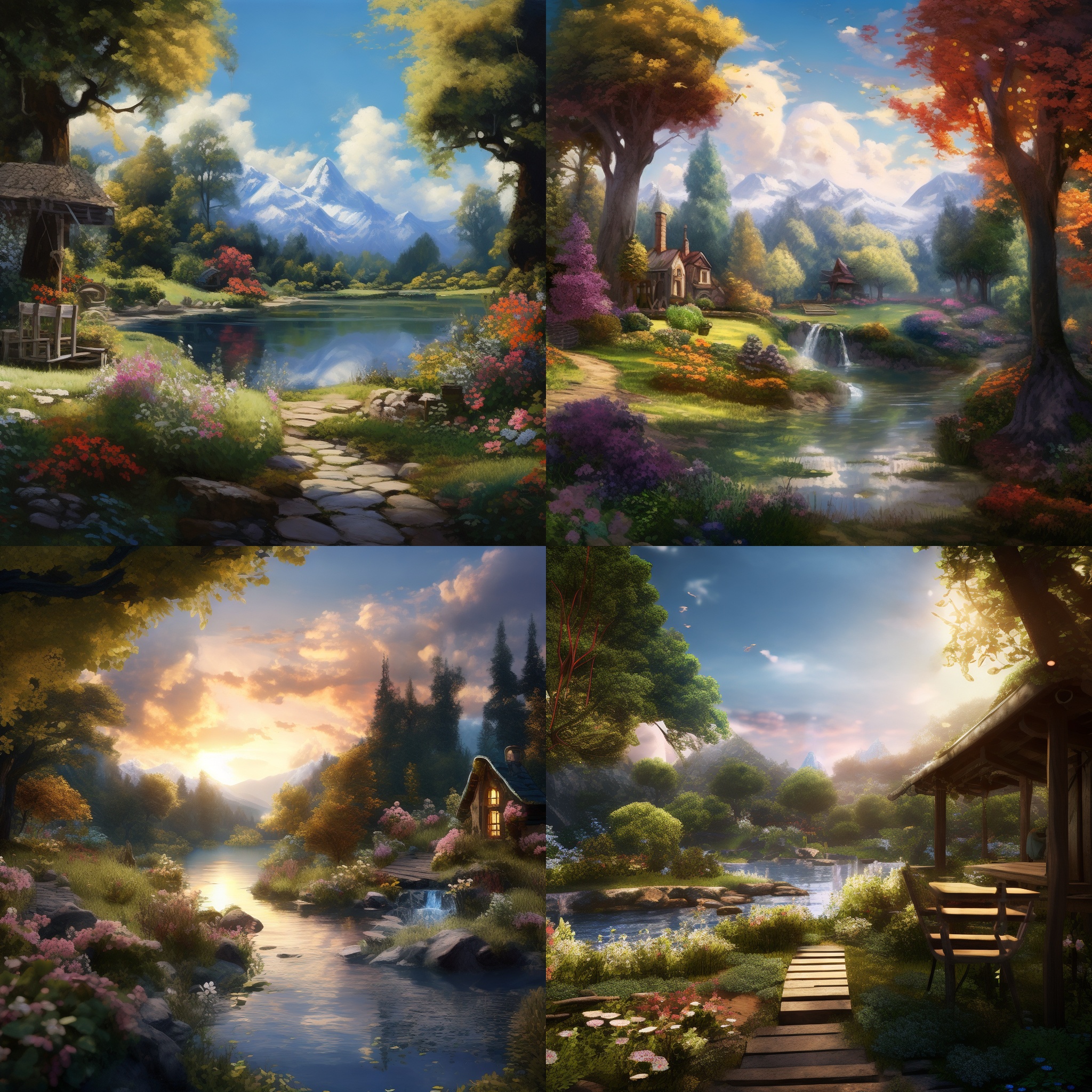} 
                        };
                        \node[rotate=0, align=center, font=\normalsize, anchor=base, yshift=4pt] at (image.north) {\gls{ncg} (Ours) \strut};
                    \end{tikzpicture}
                \end{subfigure}
                \caption*{A beautiful outdoor scene}
            \end{subfigure}
        \end{minipage}
        \caption{More visual examples comparing CFG and \gls{ncg} using \textsc{PixArt}-$\delta$.}
        \label{fig:pixart-lcm-appendix}
    \end{figure}
}

\newcommand{\figSDXLFlashAppendix}{
    \begin{figure}[t]
        \centering
        \begin{minipage}{\textwidth}
            \begin{subfigure}{0.49\textwidth}
                \begin{subfigure}{0.495\textwidth}
                    \begin{tikzpicture}
                        \node[anchor=south west, inner sep=0] (image) at (0,0) {
                            \includegraphics[width=\textwidth]{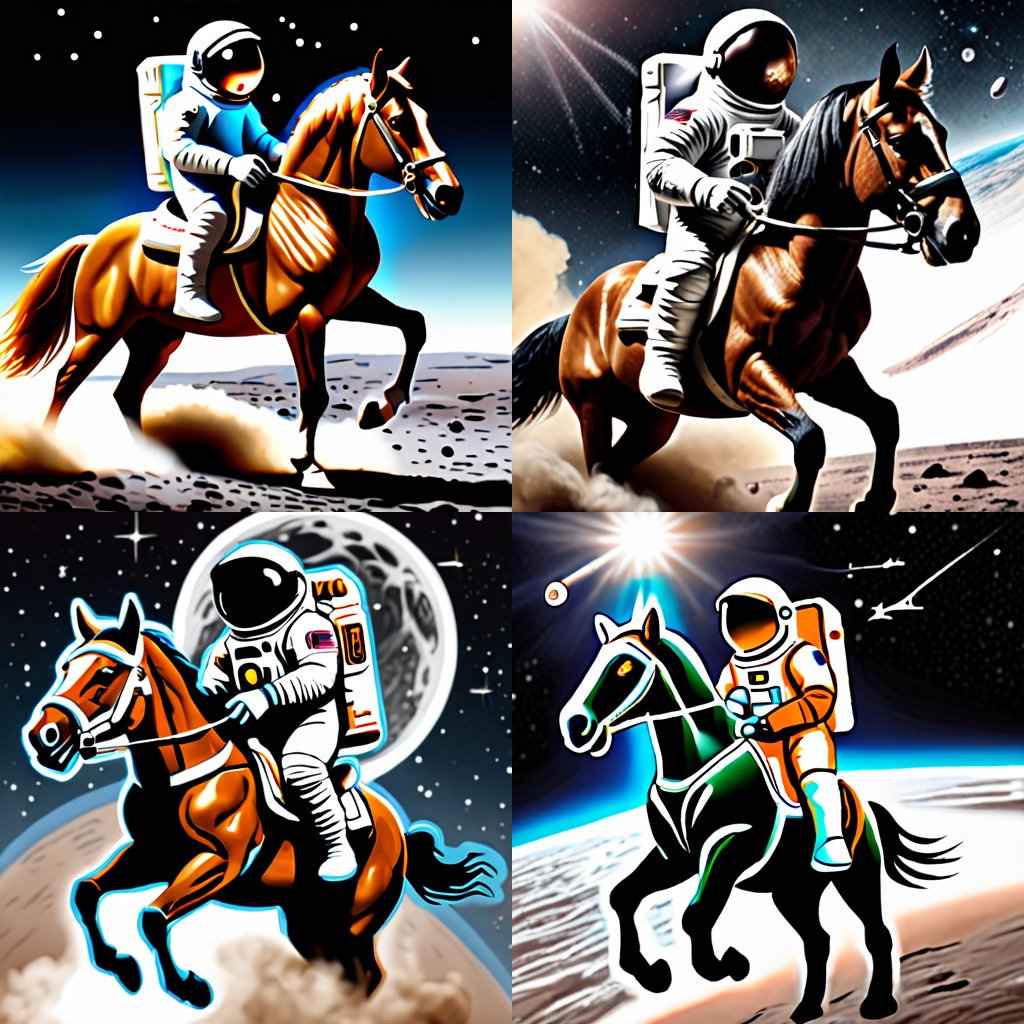}
                        };
                        \node[rotate=0, align=center, font=\normalsize, anchor=base, yshift=4pt] at (image.north) {CFG \strut};
                    \end{tikzpicture}
                \end{subfigure}
                \begin{subfigure}{0.495\textwidth}
                    \begin{tikzpicture}
                        \node[anchor=south west, inner sep=0] (image) at (0,0) {
                            \includegraphics[width=\textwidth]{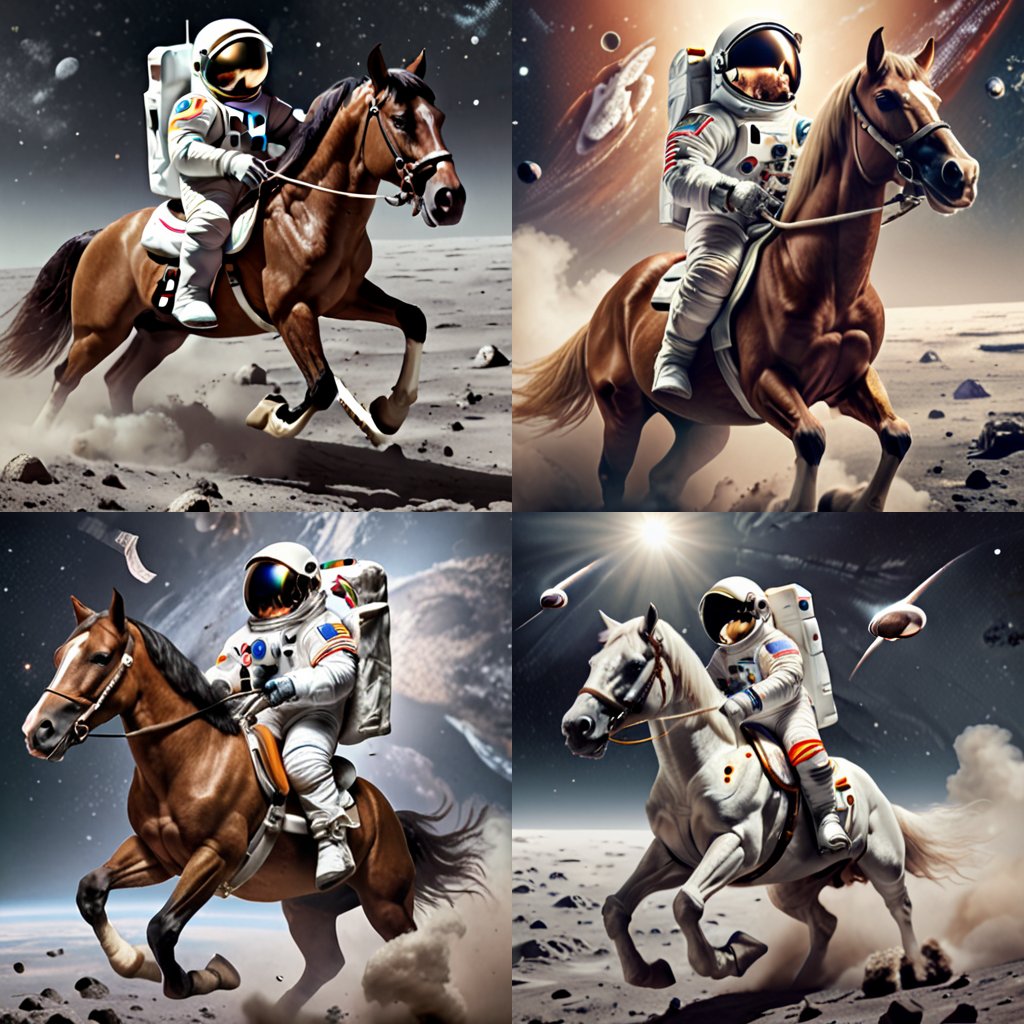}
                        };
                        \node[rotate=0, align=center, font=\normalsize, anchor=base, yshift=4pt] at (image.north) {\gls{ncg} (Ours) \strut};
                    \end{tikzpicture}
                \end{subfigure}
                \caption*{Astronaut riding a horse}
            \end{subfigure}
            \begin{subfigure}{0.49\textwidth}
                \begin{subfigure}{0.495\textwidth}
                    \begin{tikzpicture}
                        \node[anchor=south west, inner sep=0] (image) at (0,0) {
                            \includegraphics[width=\textwidth]{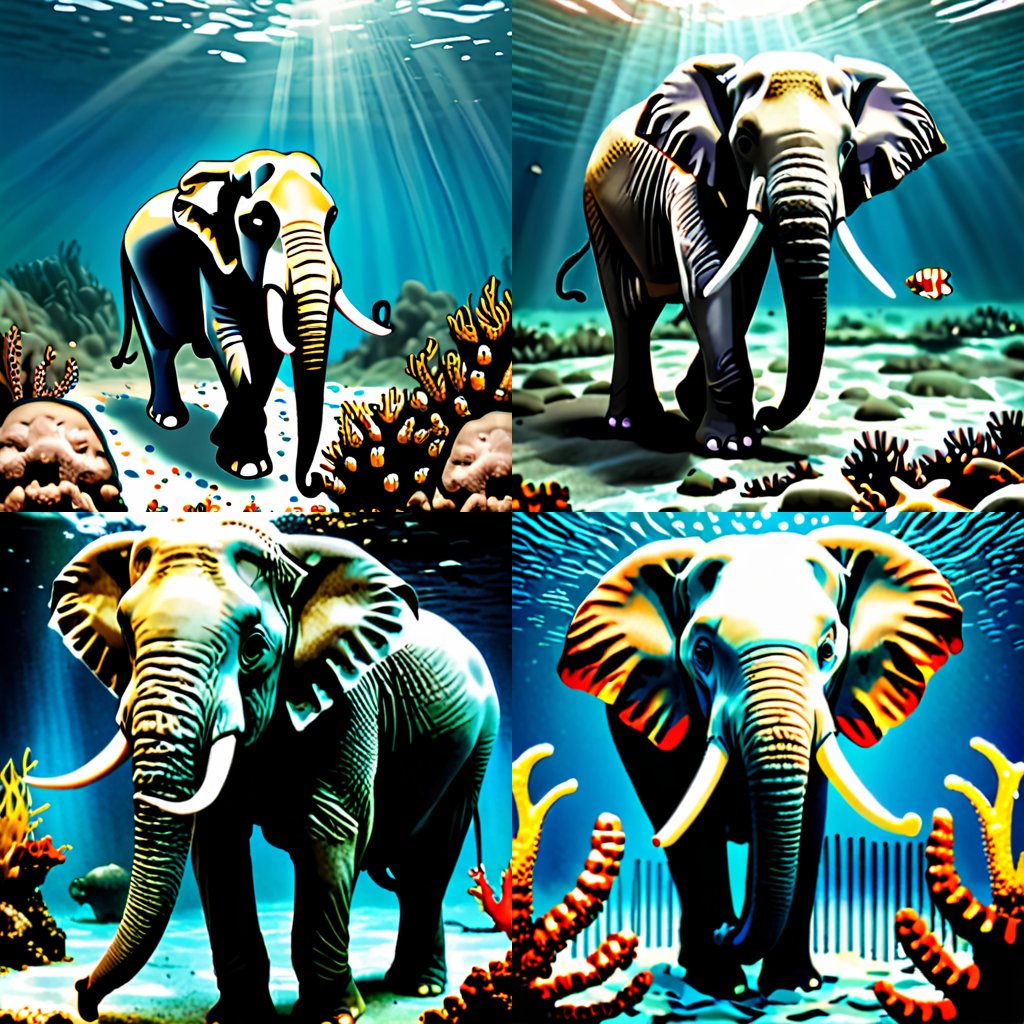}
                        };
                        \node[rotate=0, align=center, font=\normalsize, anchor=base, yshift=4pt] at (image.north) {CFG \strut};
                    \end{tikzpicture}
                \end{subfigure}
                \begin{subfigure}{0.495\textwidth}
                    \begin{tikzpicture}
                        \node[anchor=south west, inner sep=0] (image) at (0,0) {
                            \includegraphics[width=\textwidth]{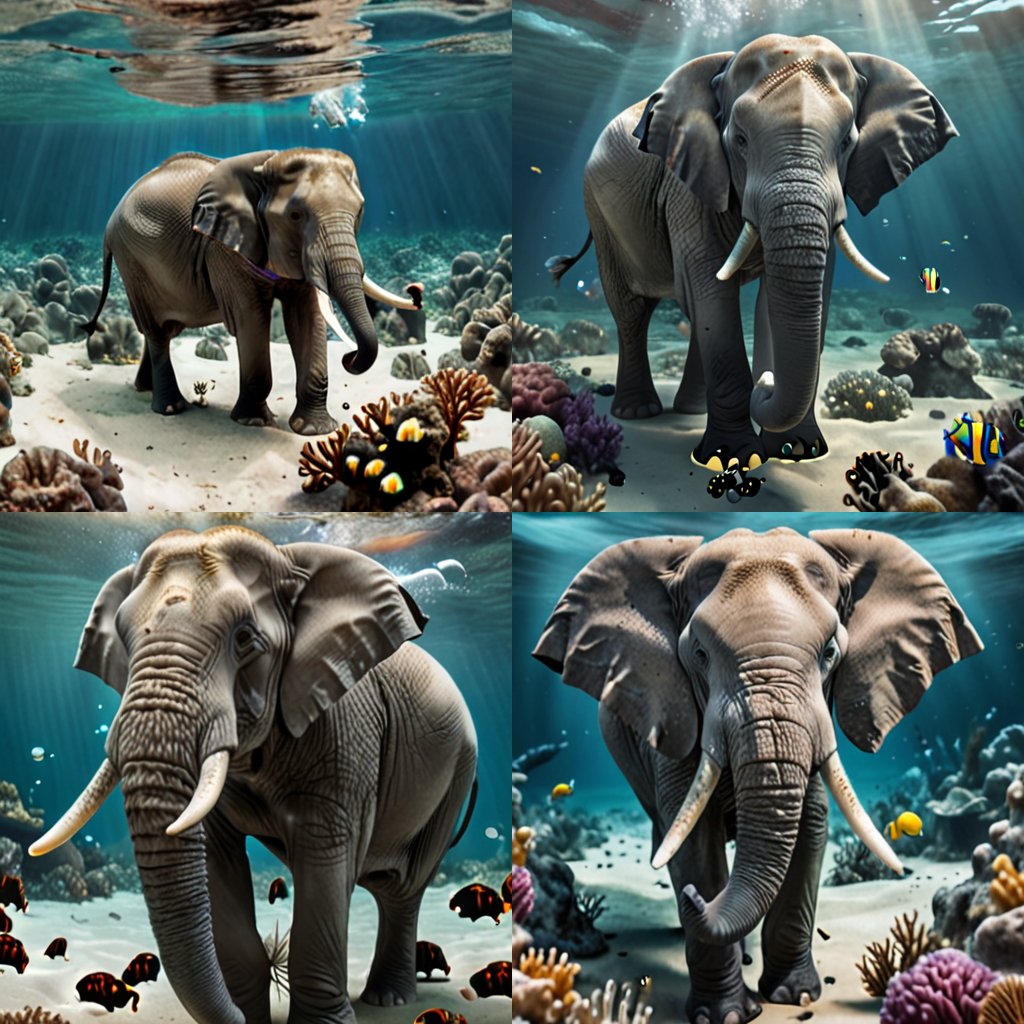}
                        };
                        \node[rotate=0, align=center, font=\normalsize, anchor=base, yshift=4pt] at (image.north) {\gls{ncg} (Ours) \strut};
                    \end{tikzpicture}
                \end{subfigure}
                \caption*{An elephant under the sea}
            \end{subfigure}
        
            \begin{subfigure}{0.49\textwidth}
                \begin{subfigure}{0.495\textwidth}
                    \begin{tikzpicture}
                        \node[anchor=south west, inner sep=0] (image) at (0,0) {
                            \includegraphics[width=\textwidth]{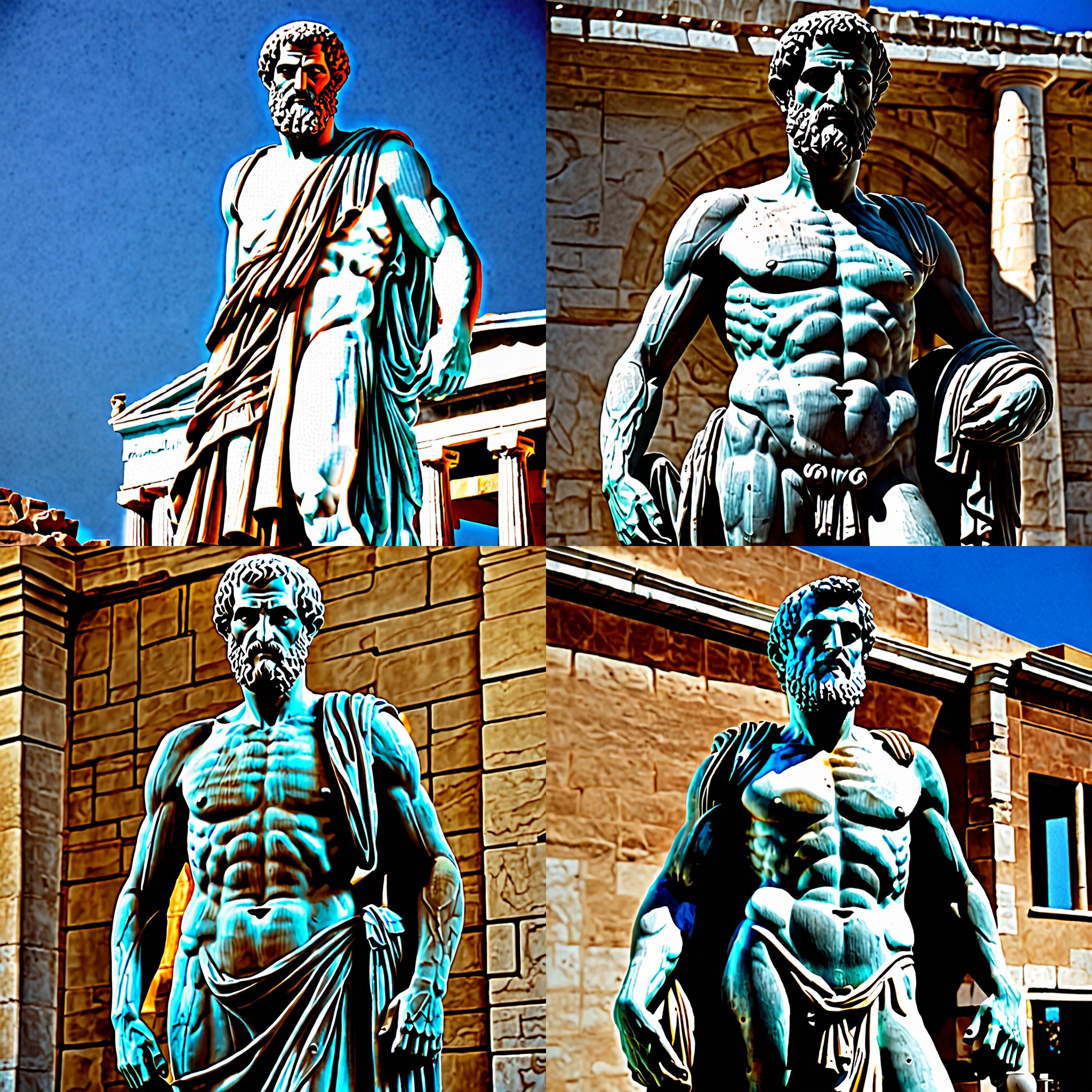}
                        };
                        \node[rotate=0, align=center, font=\normalsize, anchor=base, yshift=4pt] at (image.north) {CFG \strut};
                    \end{tikzpicture}
                \end{subfigure}
                \begin{subfigure}{0.495\textwidth}
                    \begin{tikzpicture}
                        \node[anchor=south west, inner sep=0] (image) at (0,0) {
                            \includegraphics[width=\textwidth]{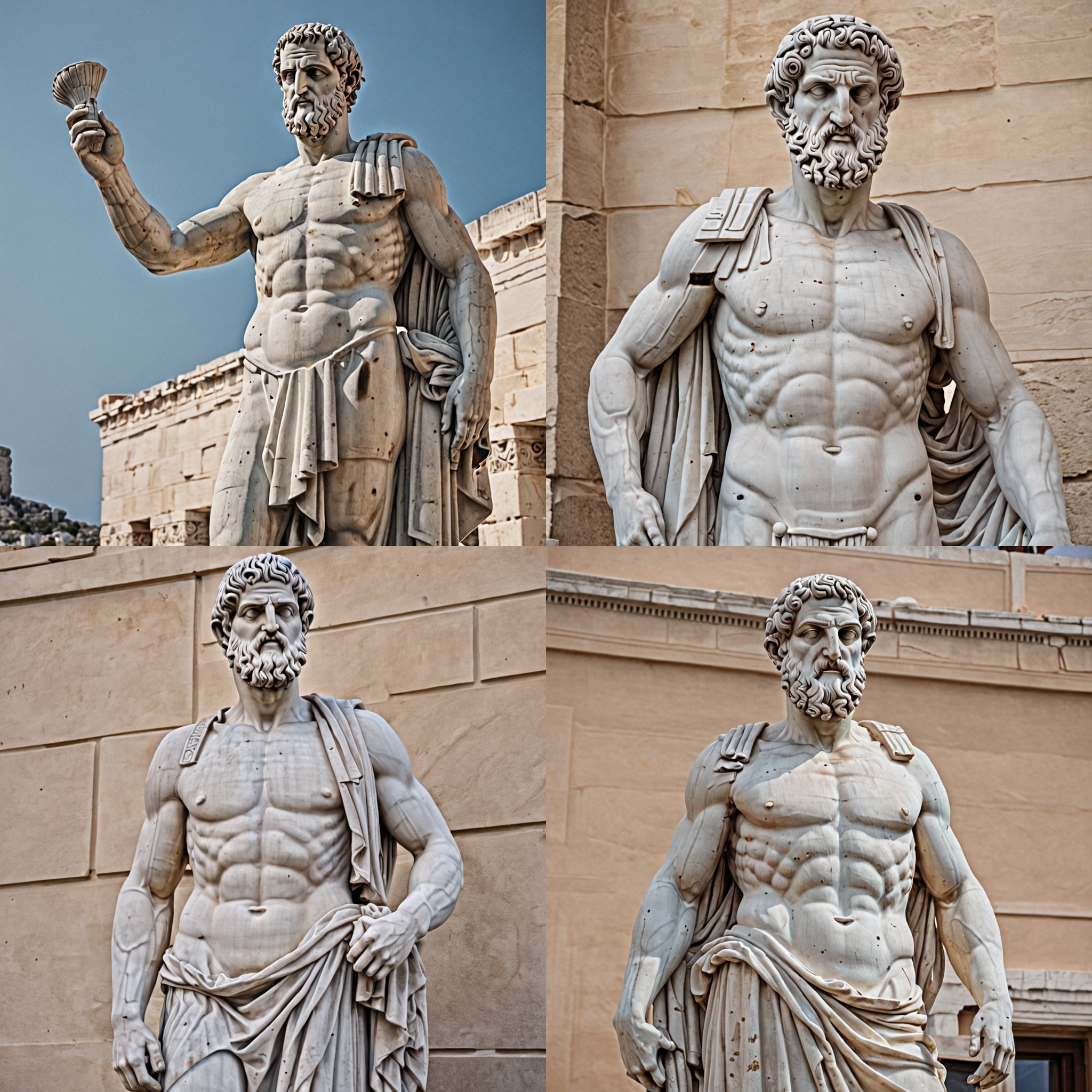}
                        };
                        \node[rotate=0, align=center, font=\normalsize, anchor=base, yshift=4pt] at (image.north) {\gls{ncg} (Ours) \strut};
                    \end{tikzpicture}
                \end{subfigure}
                \caption*{A statue of a greek man}
            \end{subfigure}
            \begin{subfigure}{0.49\textwidth}
                \begin{subfigure}{0.495\textwidth}
                    \begin{tikzpicture}
                        \node[anchor=south west, inner sep=0] (image) at (0,0) {
                            \includegraphics[width=\textwidth]{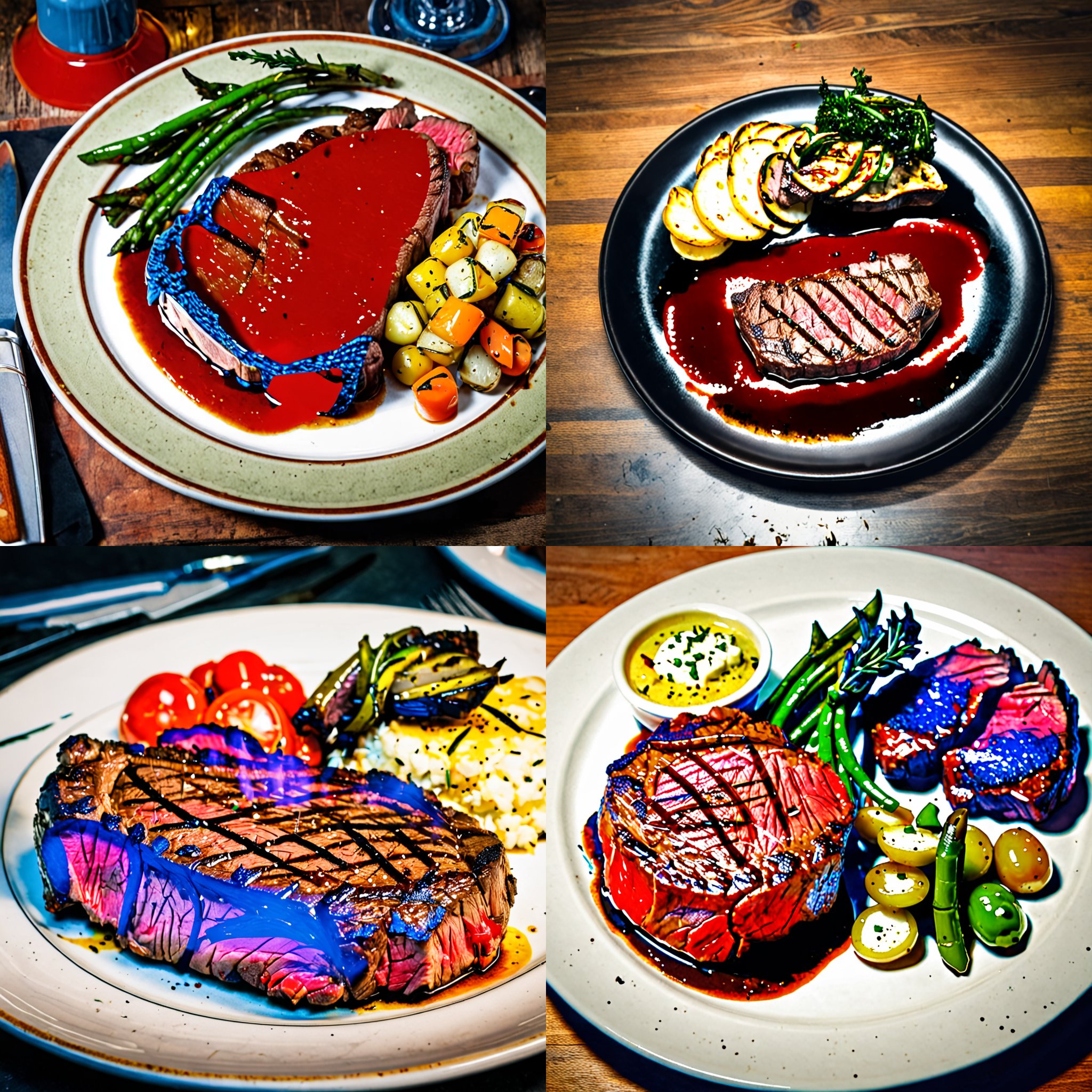}
                        };
                        \node[rotate=0, align=center, font=\normalsize, anchor=base, yshift=4pt] at (image.north) {CFG \strut};
                    \end{tikzpicture}
                \end{subfigure}
                \begin{subfigure}{0.495\textwidth}
                    \begin{tikzpicture}
                        \node[anchor=south west, inner sep=0] (image) at (0,0) {
                            \includegraphics[width=\textwidth]{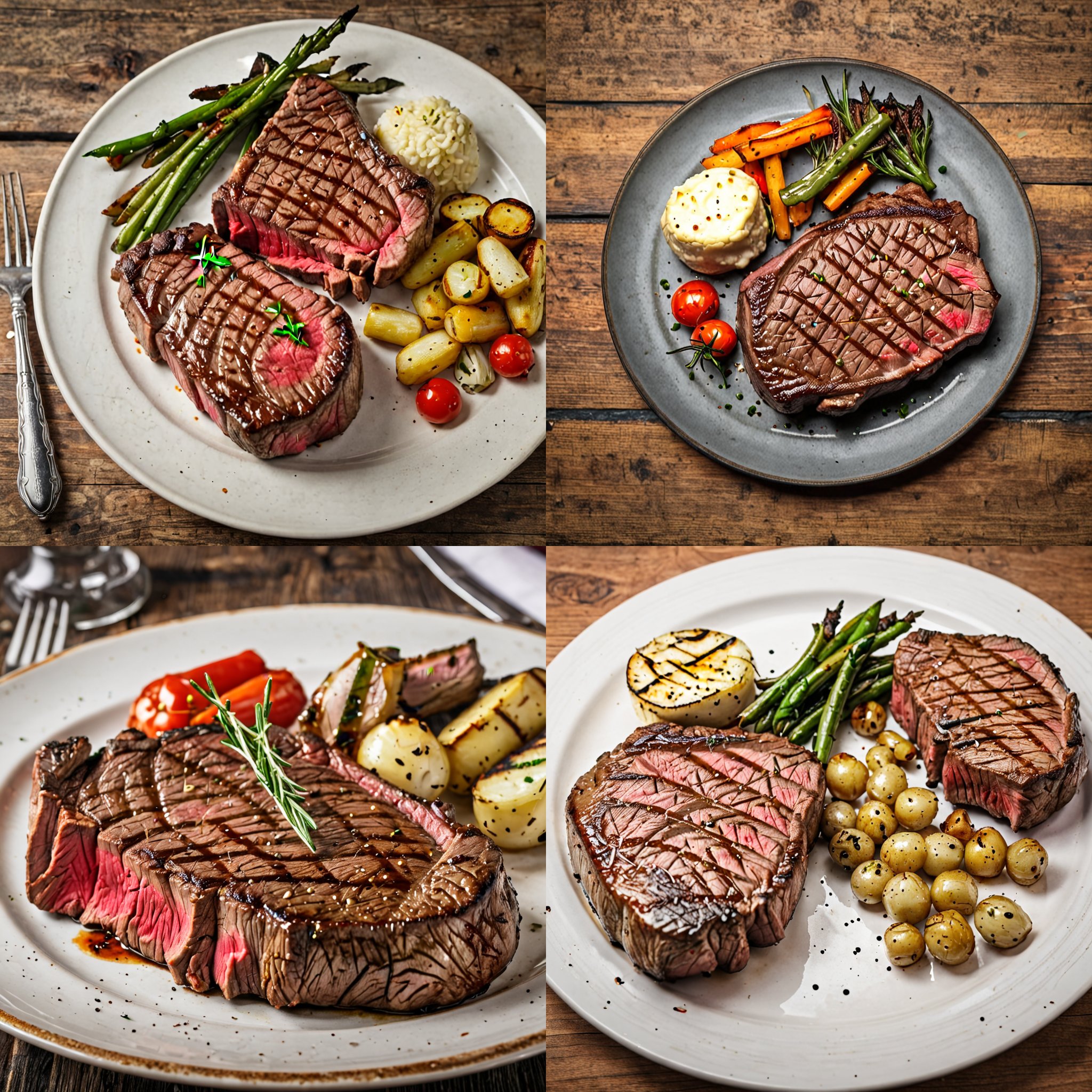}
                        };
                        \node[rotate=0, align=center, font=\normalsize, anchor=base, yshift=4pt] at (image.north) {\gls{ncg} (Ours) \strut};
                    \end{tikzpicture}
                \end{subfigure}
                \caption*{A plate of steak}
            \end{subfigure}
        \end{minipage}
        \caption{More visual examples comparing CFG and \gls{ncg} using SDXL-Flash.}
        \label{fig:sdxl-flash-appendix}
    \end{figure}
}

\newcommand{\figSDXLLightningAppendix}{
    \begin{figure}[t]
        \centering
        \begin{minipage}{\textwidth}
            \begin{subfigure}{0.49\textwidth}
                \begin{subfigure}{0.495\textwidth}
                    \begin{tikzpicture}
                        \node[anchor=south west, inner sep=0] (image) at (0,0) {
                            \includegraphics[width=\textwidth]{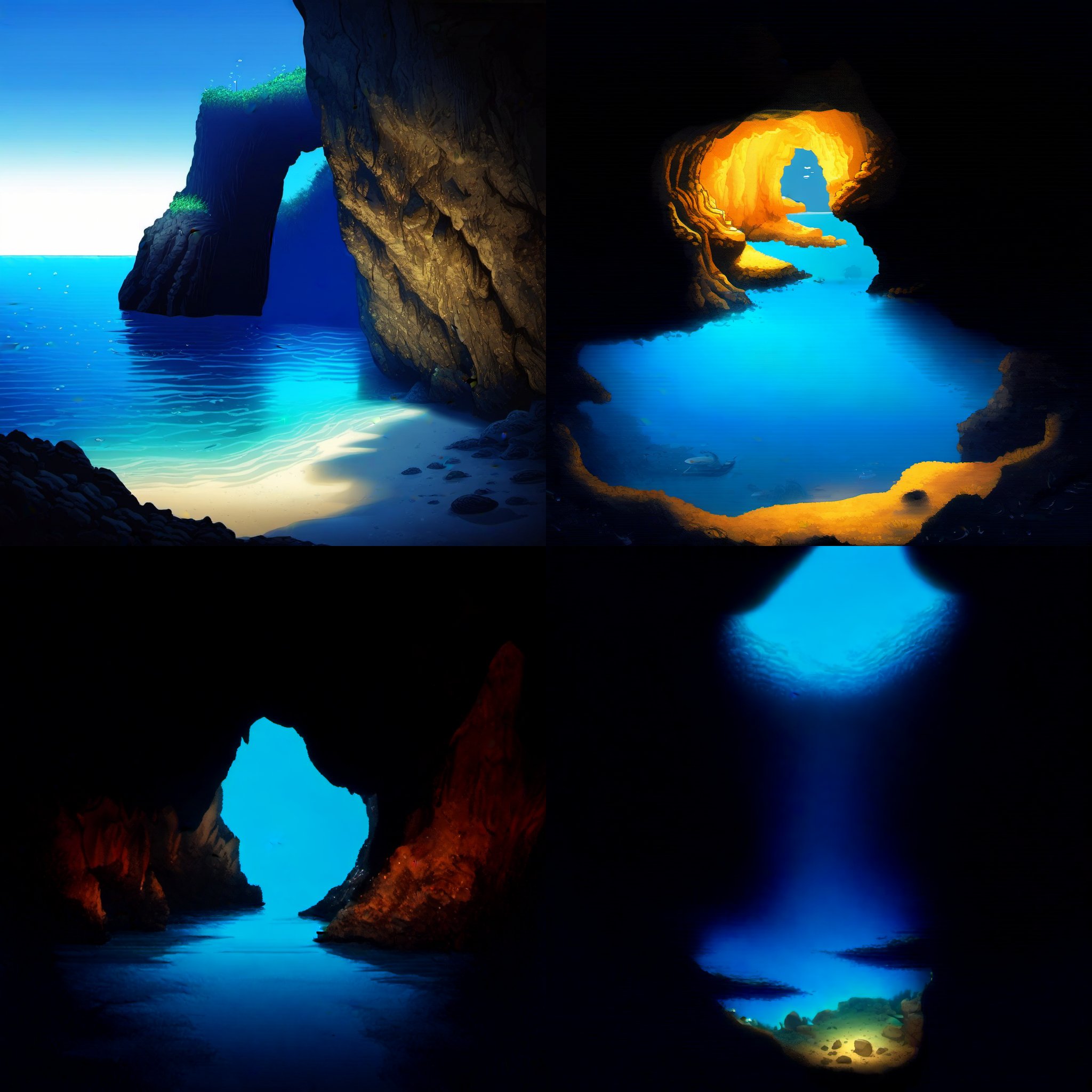}
                        };
                        \node[rotate=0, align=center, font=\normalsize, anchor=base, yshift=4pt] at (image.north) {CFG \strut};
                    \end{tikzpicture}
                \end{subfigure}
                \begin{subfigure}{0.495\textwidth}
                    \begin{tikzpicture}
                        \node[anchor=south west, inner sep=0] (image) at (0,0) {
                            \includegraphics[width=\textwidth]{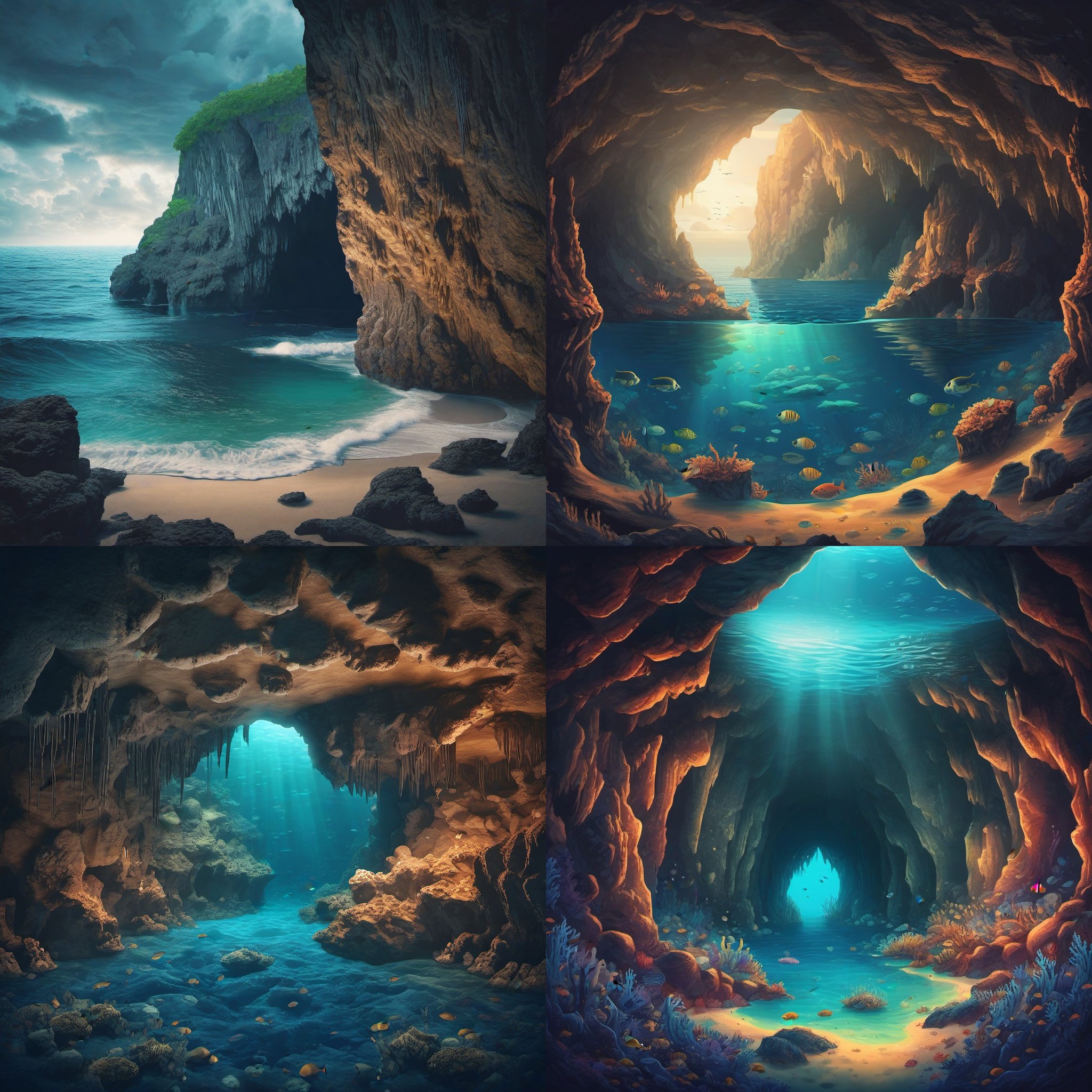}
                        };
                        \node[rotate=0, align=center, font=\normalsize, anchor=base, yshift=4pt] at (image.north) {\gls{ncg} (Ours) \strut};
                    \end{tikzpicture}
                \end{subfigure}
                \caption*{A cave under the ocean}
            \end{subfigure}
            \begin{subfigure}{0.49\textwidth}
                \begin{subfigure}{0.495\textwidth}
                    \begin{tikzpicture}
                        \node[anchor=south west, inner sep=0] (image) at (0,0) {
                            \includegraphics[width=\textwidth]{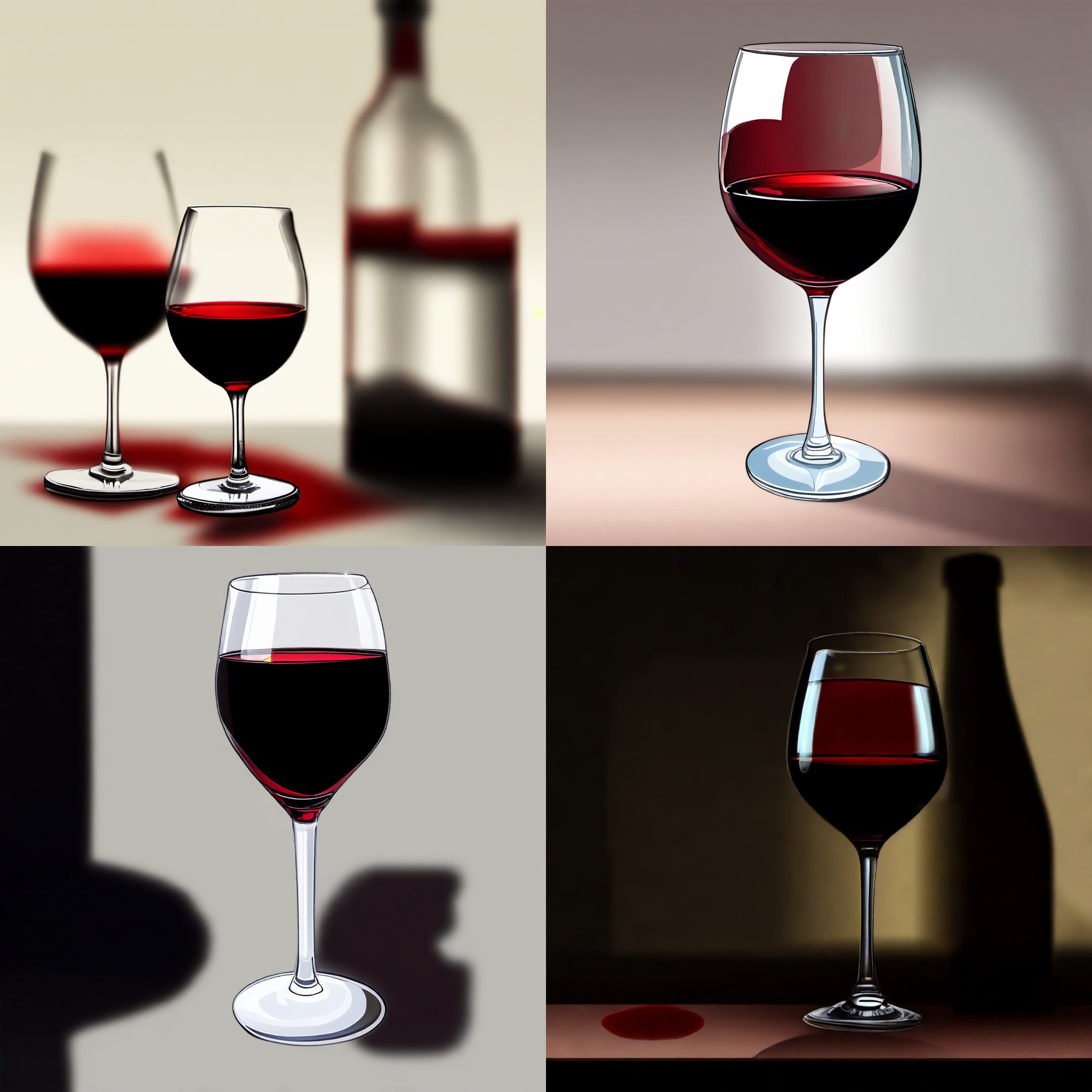}
                        };
                        \node[rotate=0, align=center, font=\normalsize, anchor=base, yshift=4pt] at (image.north) {CFG \strut};
                    \end{tikzpicture}
                \end{subfigure}
                \begin{subfigure}{0.495\textwidth}
                    \begin{tikzpicture}
                        \node[anchor=south west, inner sep=0] (image) at (0,0) {
                            \includegraphics[width=\textwidth]{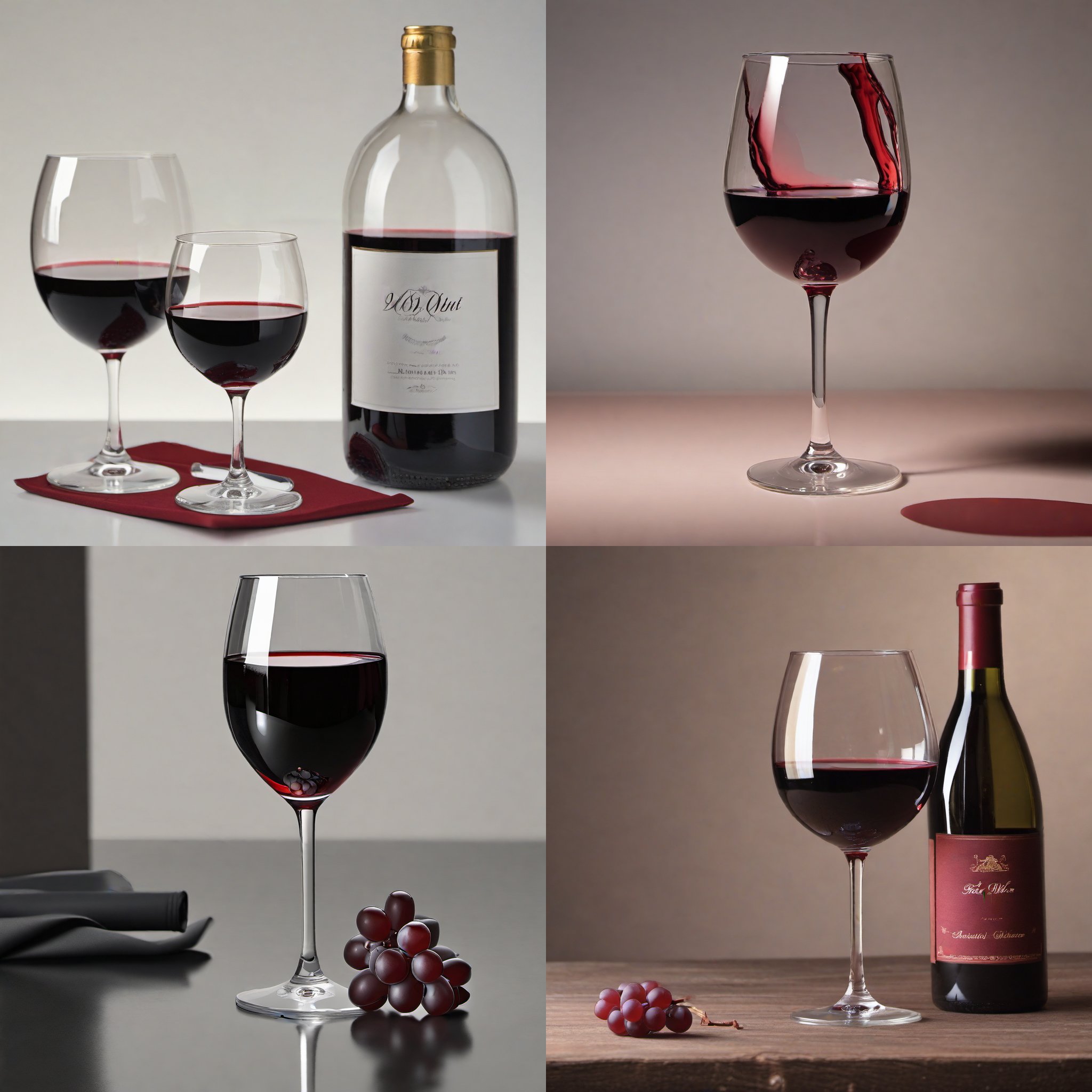}
                        };
                        \node[rotate=0, align=center, font=\normalsize, anchor=base, yshift=4pt] at (image.north) {\gls{ncg} (Ours) \strut};
                    \end{tikzpicture}
                \end{subfigure}
                \caption*{A glass of red wine}
            \end{subfigure}

            \begin{subfigure}{0.49\textwidth}
                \begin{subfigure}{0.495\textwidth}
                    \begin{tikzpicture}
                        \node[anchor=south west, inner sep=0] (image) at (0,0) {
                            \includegraphics[width=\textwidth]{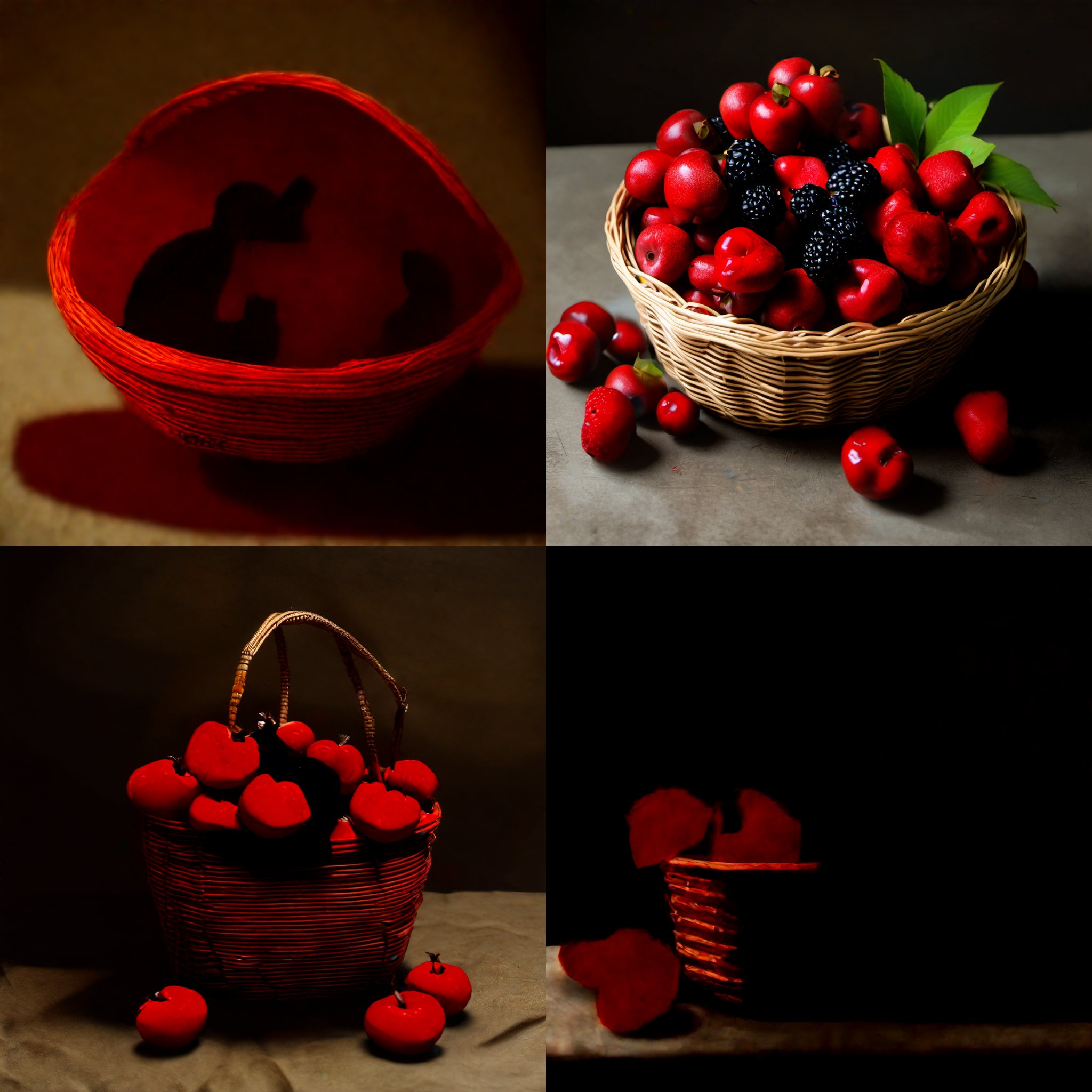}
                        };
                        \node[rotate=0, align=center, font=\normalsize, anchor=base, yshift=4pt] at (image.north) {CFG \strut};
                    \end{tikzpicture}
                \end{subfigure}
                \begin{subfigure}{0.495\textwidth}
                    \begin{tikzpicture}
                        \node[anchor=south west, inner sep=0] (image) at (0,0) {
                            \includegraphics[width=\textwidth]{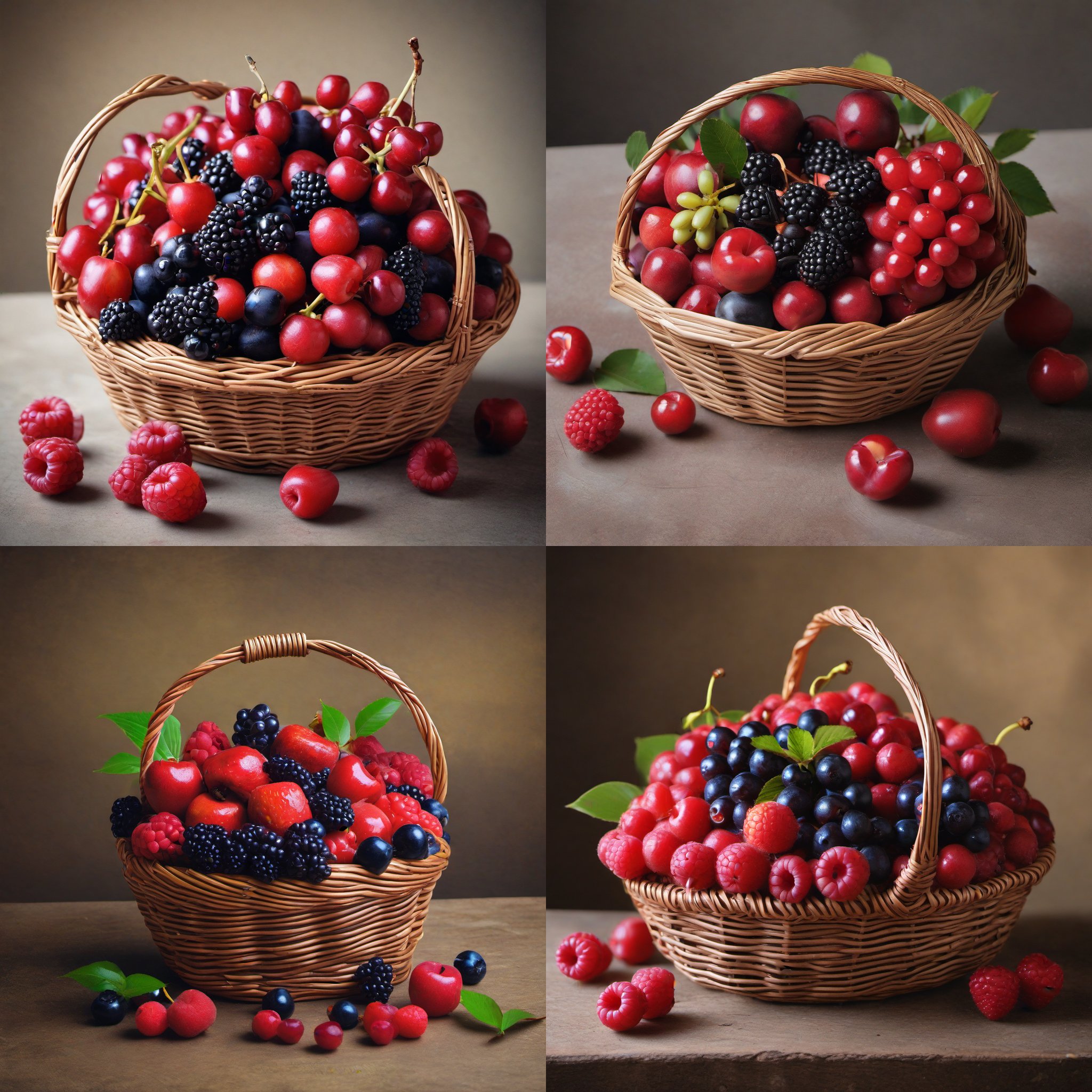}
                        };
                        \node[rotate=0, align=center, font=\normalsize, anchor=base, yshift=4pt] at (image.north) {\gls{ncg} (Ours) \strut};
                    \end{tikzpicture}
                \end{subfigure}
                \caption*{A basket of red fruits}
            \end{subfigure}
            \begin{subfigure}{0.49\textwidth}
                \begin{subfigure}{0.495\textwidth}
                    \begin{tikzpicture}
                        \node[anchor=south west, inner sep=0] (image) at (0,0) {
                            \includegraphics[width=\textwidth]{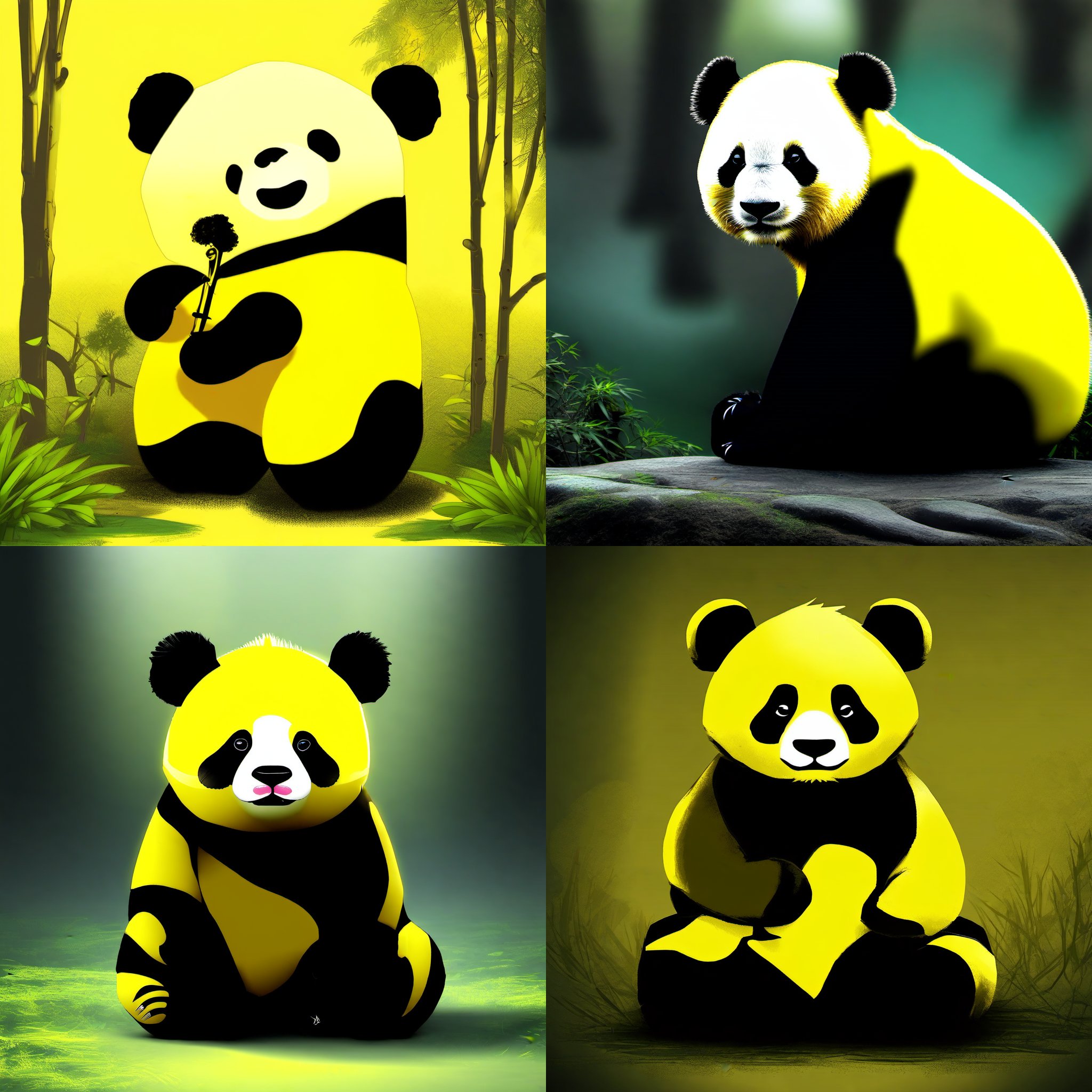}
                        };
                        \node[rotate=0, align=center, font=\normalsize, anchor=base, yshift=4pt] at (image.north) {CFG \strut};
                    \end{tikzpicture}
                \end{subfigure}
                \begin{subfigure}{0.495\textwidth}
                    \begin{tikzpicture}
                        \node[anchor=south west, inner sep=0] (image) at (0,0) {
                            \includegraphics[width=\textwidth]{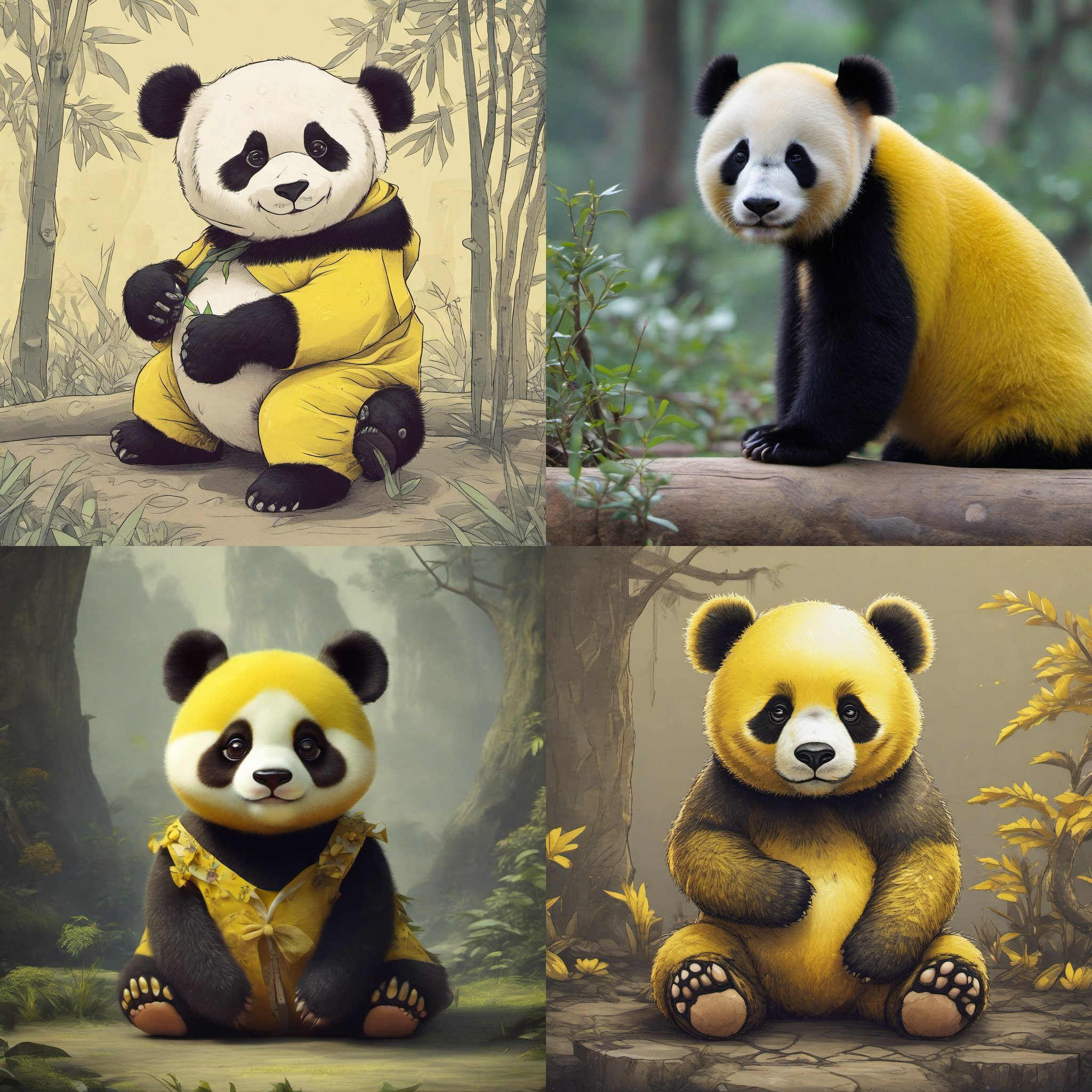}
                        };
                        \node[rotate=0, align=center, font=\normalsize, anchor=base, yshift=4pt] at (image.north) {\gls{ncg} (Ours) \strut};
                    \end{tikzpicture}
                \end{subfigure}
                \caption*{A yellow panda}
            \end{subfigure}
        \end{minipage}
        \caption{More visual examples comparing CFG and \gls{ncg} using SDXL-Lightning.}
        \label{fig:sdxl-lightning-appendix}
    \end{figure}
}

\newcommand{\figRescale}{
    \begin{figure}[t!]
        \centering
        \begin{minipage}{\textwidth}
            \begin{subfigure}{0.245\textwidth}
                \begin{tikzpicture}
                    \node[anchor=south west, inner sep=0] (image) at (0,0) {
                        \includegraphics[width=\textwidth]{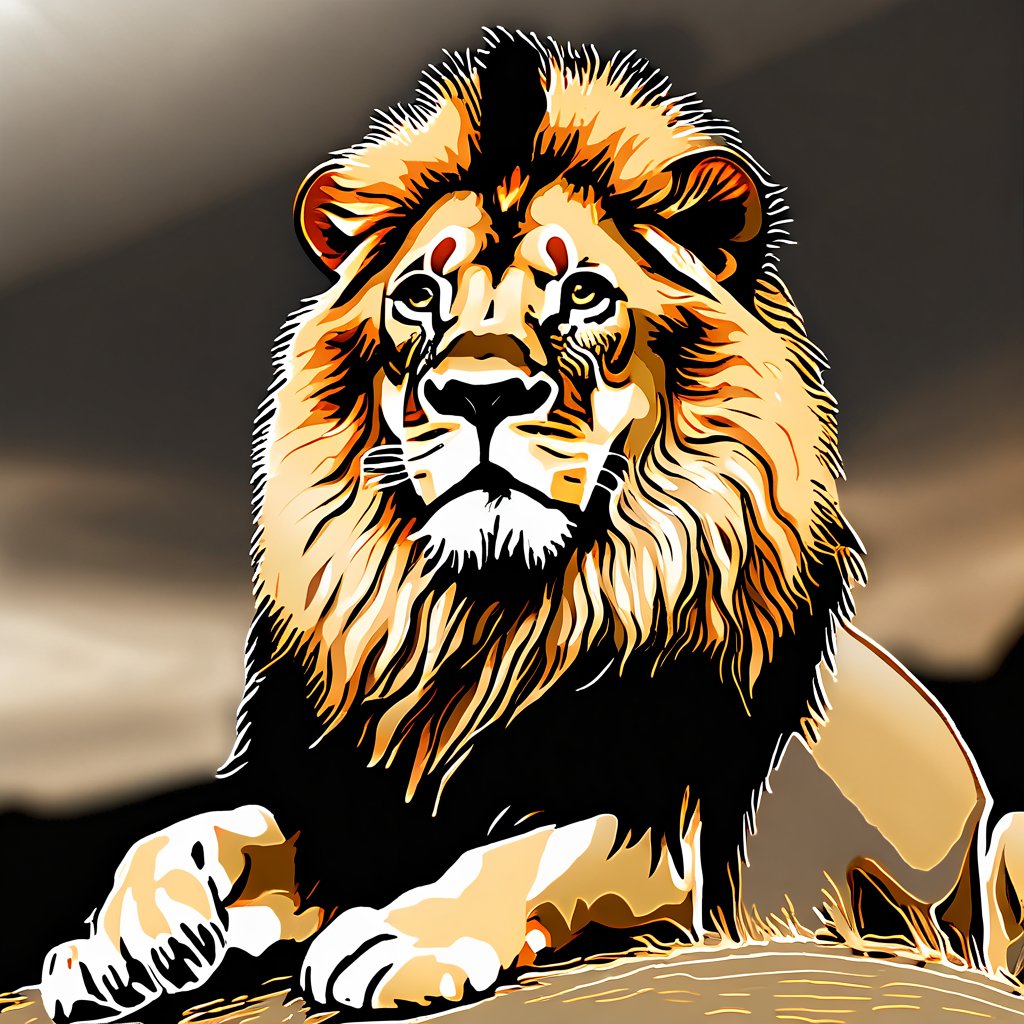}
                    };
                    \node[rotate=0, align=center, font=\normalsize, anchor=base, yshift=4pt] at (image.north) {CFG Rescale \strut};
                \end{tikzpicture}
            \end{subfigure}
            \begin{subfigure}{0.245\textwidth}
                \begin{tikzpicture}
                    \node[anchor=south west, inner sep=0] (image) at (0,0) {
                        \includegraphics[width=\textwidth]{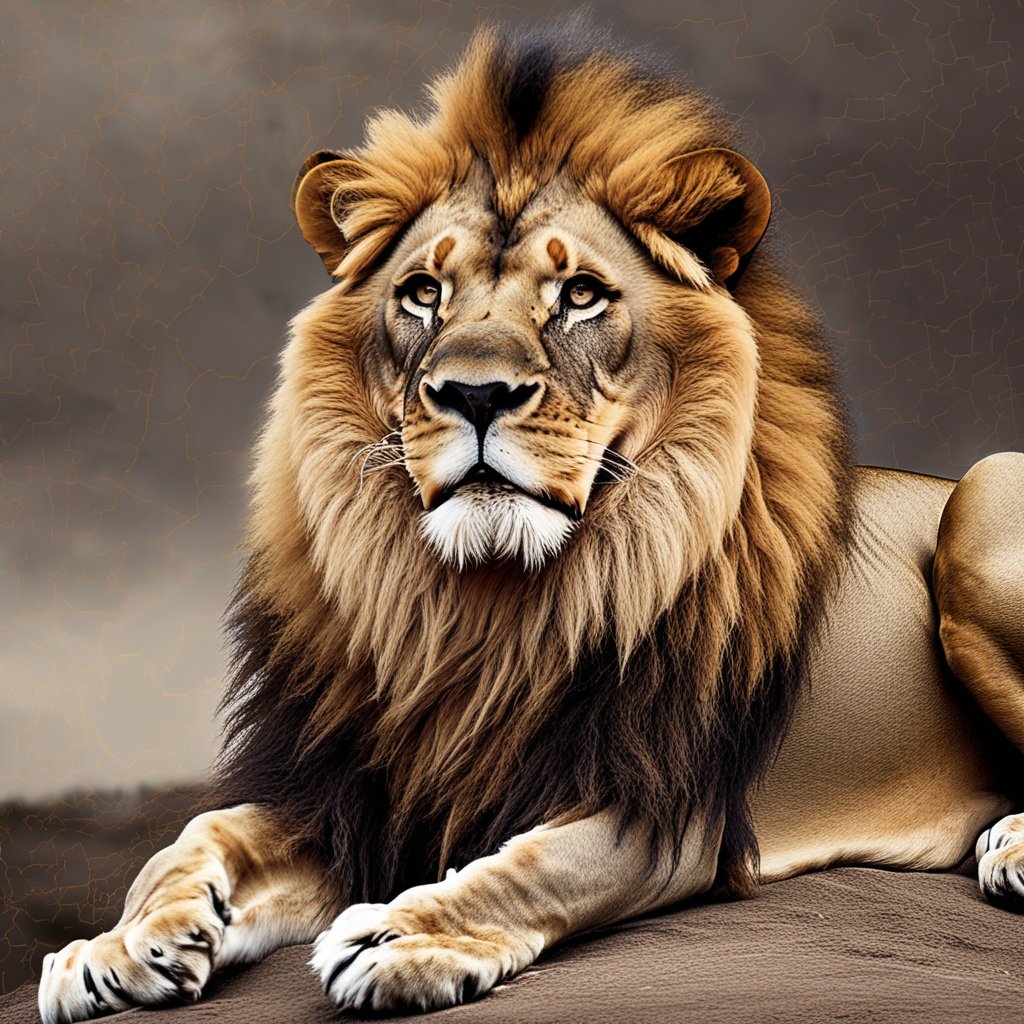}
                    };
                    \node[rotate=0, align=center, font=\normalsize, anchor=base, yshift=4pt] at (image.north) {\gls{ncg} (Ours) \strut};
                \end{tikzpicture}
            \end{subfigure}
            \begin{subfigure}{0.245\textwidth}
                \begin{tikzpicture}
                    \node[anchor=south west, inner sep=0] (image) at (0,0) {
                        \includegraphics[width=\textwidth]{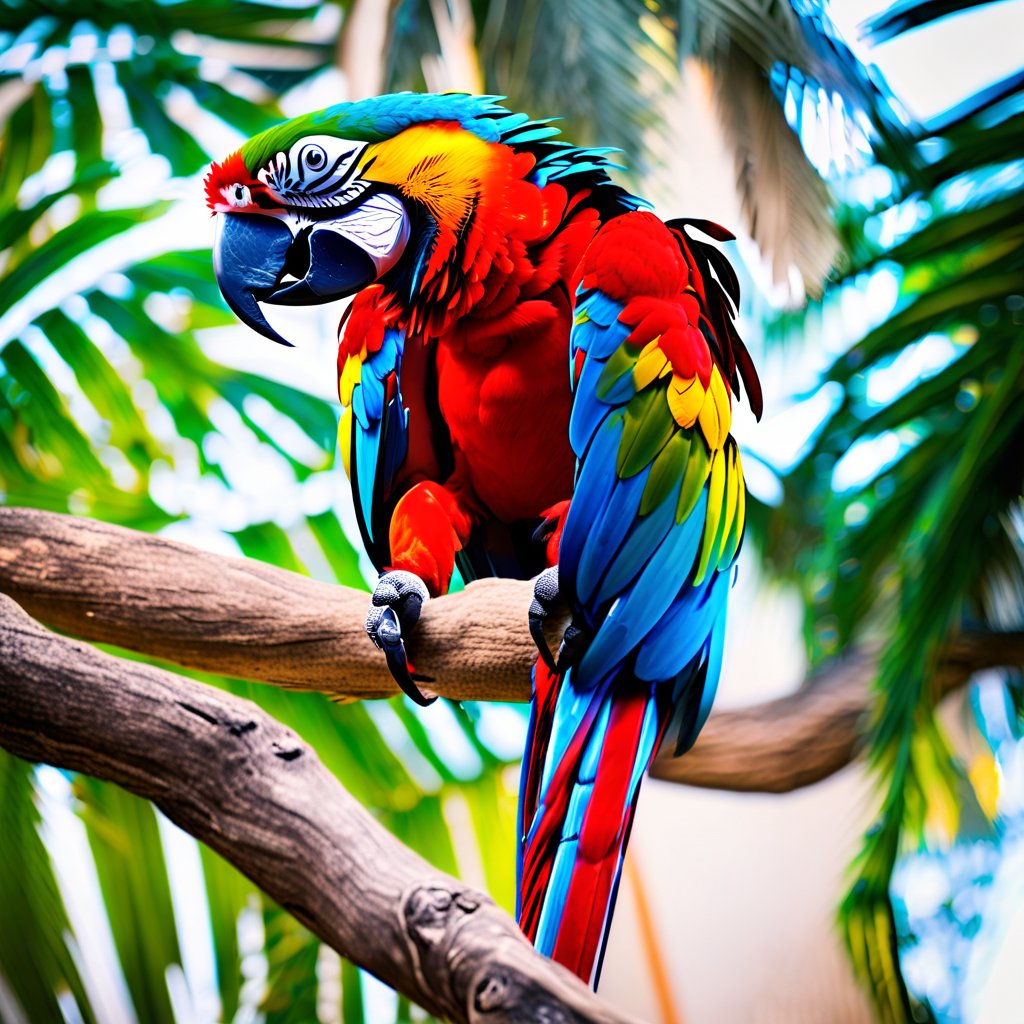}
                    };
                    \node[rotate=0, align=center, font=\normalsize, anchor=base, yshift=4pt] at (image.north) {CFG Rescale \strut};
                \end{tikzpicture}
            \end{subfigure}
            \begin{subfigure}{0.245\textwidth}
                \begin{tikzpicture}
                    \node[anchor=south west, inner sep=0] (image) at (0,0) {
                        \includegraphics[width=\textwidth]{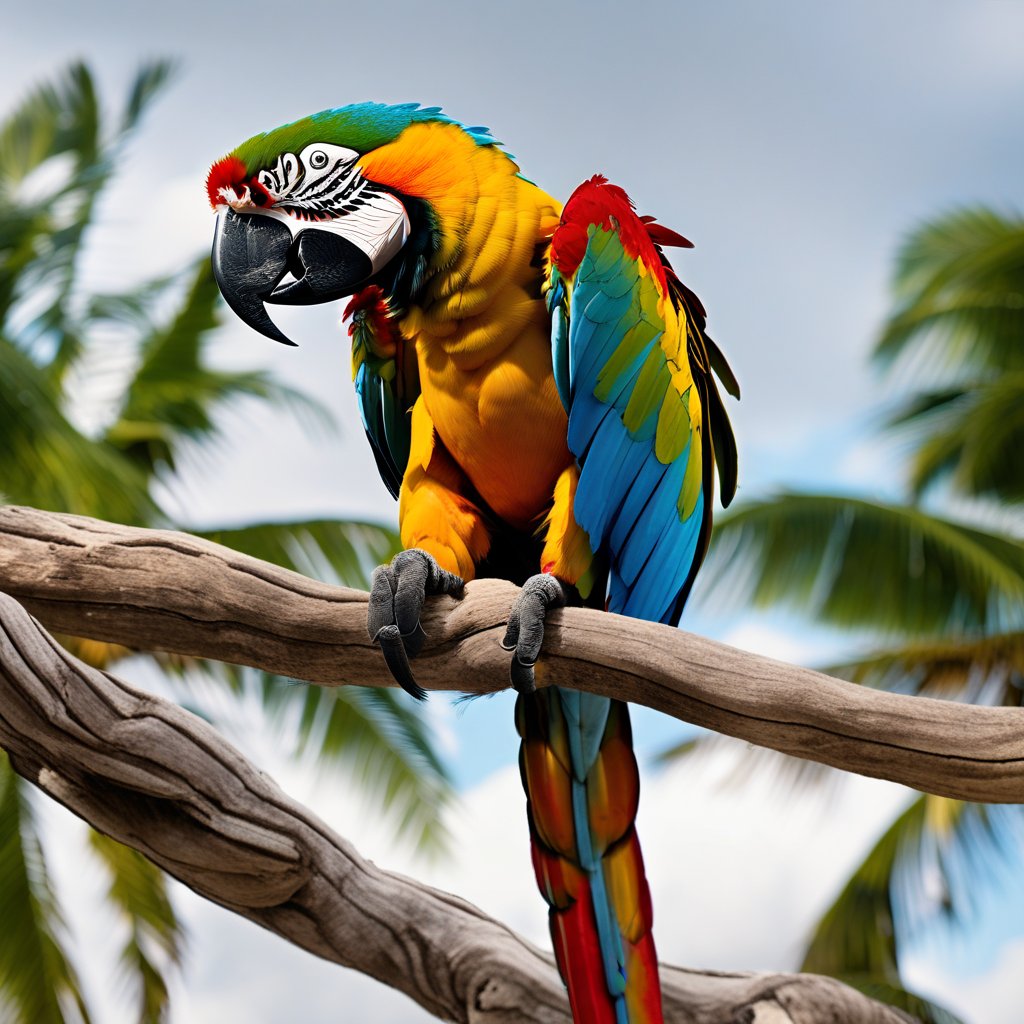}
                    };
                    \node[rotate=0, align=center, font=\normalsize, anchor=base, yshift=4pt] at (image.north) {\gls{ncg} (Ours) \strut};
                \end{tikzpicture}
            \end{subfigure}
            \begin{subfigure}{0.245\textwidth}
                \begin{tikzpicture}
                    \node[anchor=south west, inner sep=0] (image) at (0,0) {
                        \includegraphics[width=\textwidth]{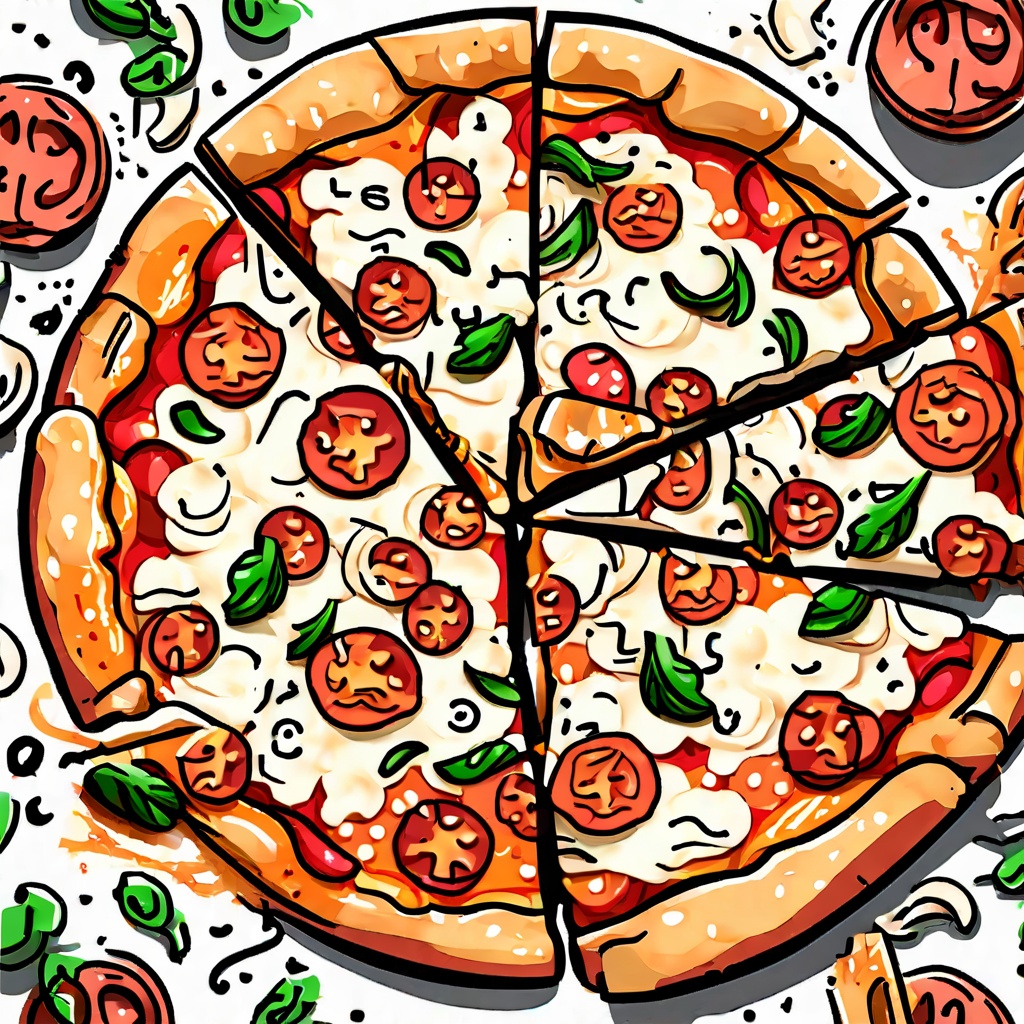}
                    };
                    \node[rotate=0, align=center, font=\normalsize, anchor=base, yshift=4pt] at (image.north) {CFG Rescale \strut};
                \end{tikzpicture}
            \end{subfigure}
            \begin{subfigure}{0.245\textwidth}
                \begin{tikzpicture}
                    \node[anchor=south west, inner sep=0] (image) at (0,0) {
                        \includegraphics[width=\textwidth]{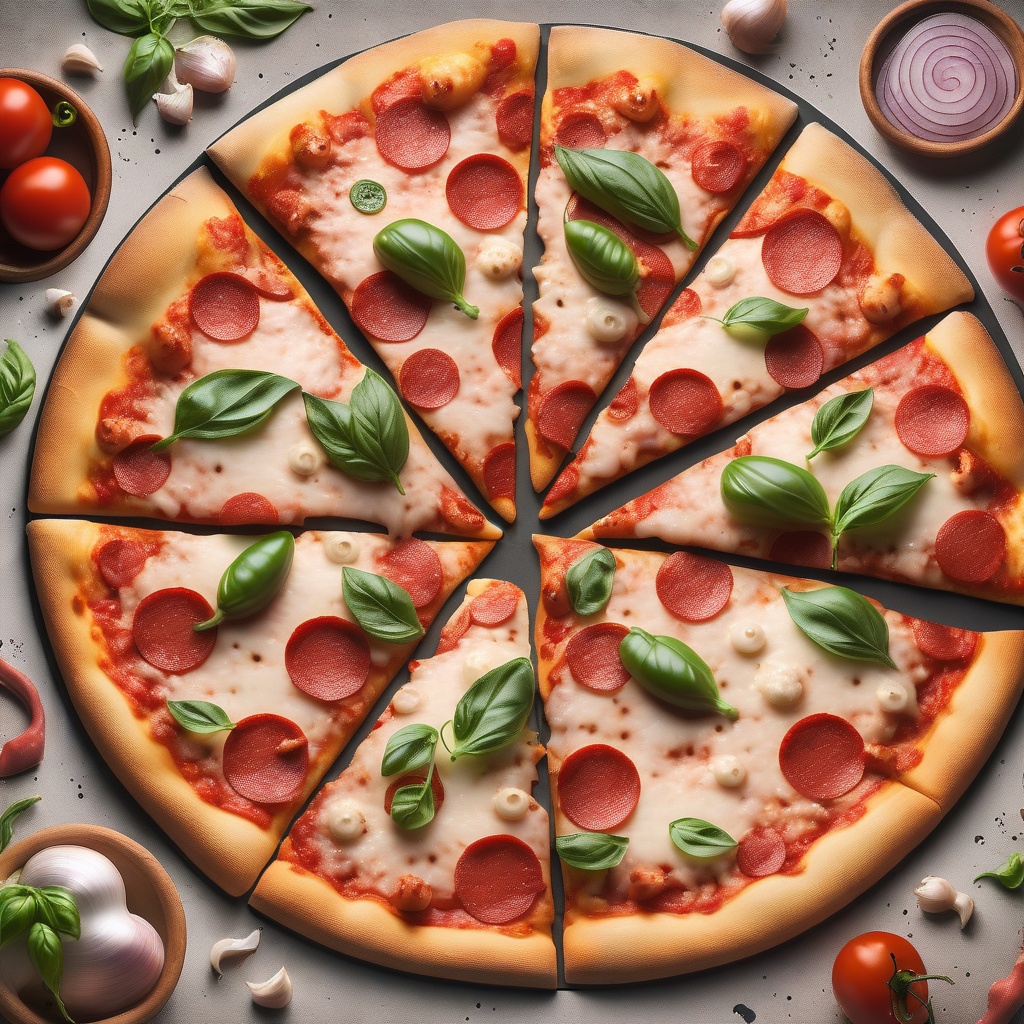}
                    };
                    \node[rotate=0, align=center, font=\normalsize, anchor=base, yshift=4pt] at (image.north) {\gls{ncg} (Ours) \strut};
                \end{tikzpicture}
            \end{subfigure}
            \begin{subfigure}{0.245\textwidth}
                \begin{tikzpicture}
                    \node[anchor=south west, inner sep=0] (image) at (0,0) {
                        \includegraphics[width=\textwidth]{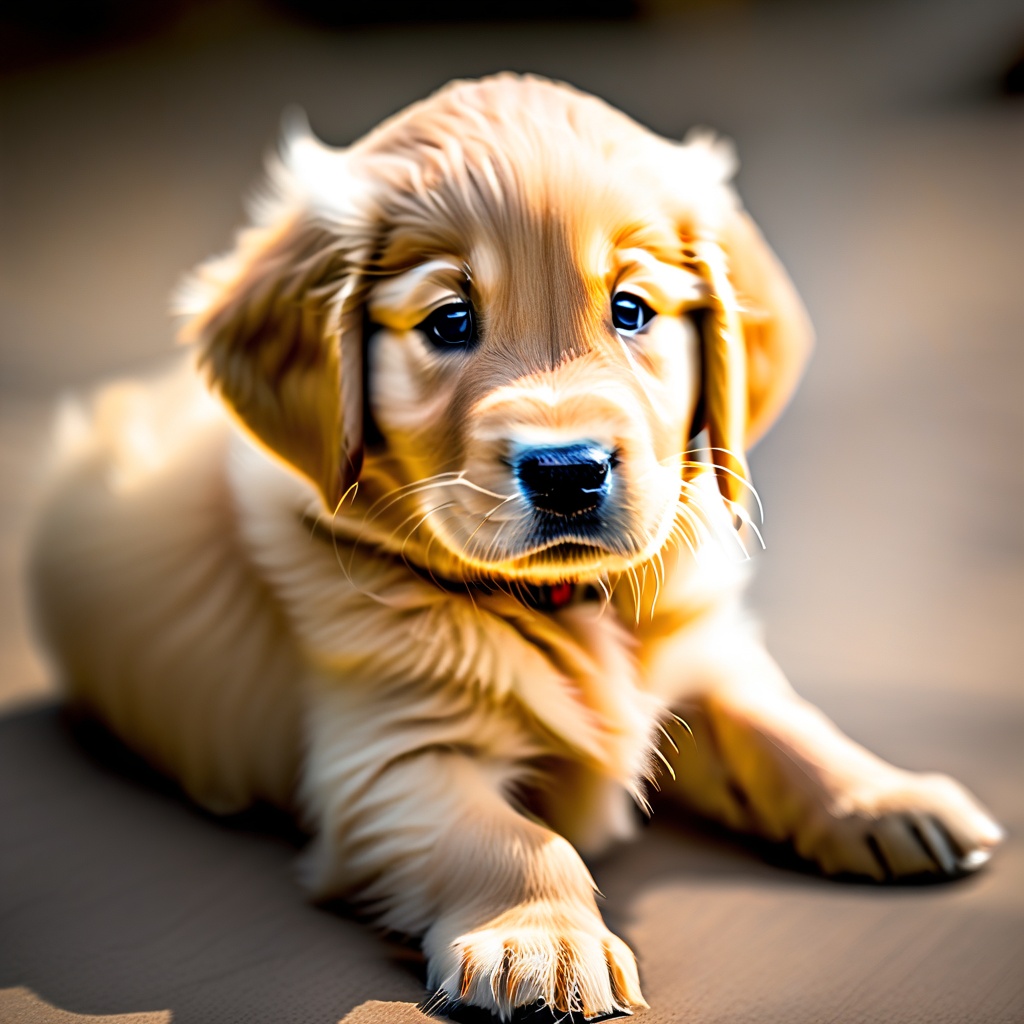}
                    };
                    \node[rotate=0, align=center, font=\normalsize, anchor=base, yshift=4pt] at (image.north) {CFG Rescale \strut};
                \end{tikzpicture}
            \end{subfigure}
            \begin{subfigure}{0.245\textwidth}
                \begin{tikzpicture}
                    \node[anchor=south west, inner sep=0] (image) at (0,0) {
                        \includegraphics[width=\textwidth]{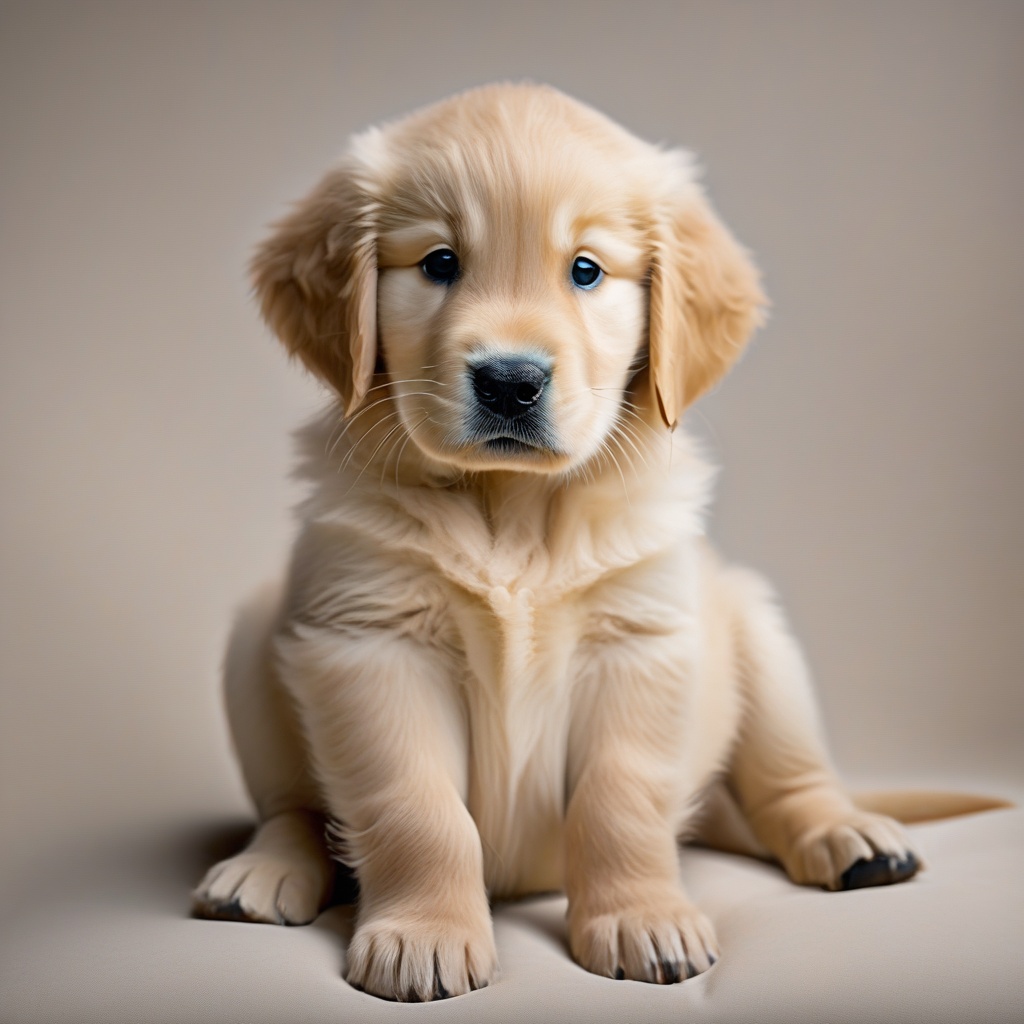}
                    };
                    \node[rotate=0, align=center, font=\normalsize, anchor=base, yshift=4pt] at (image.north) {\gls{ncg} (Ours) \strut};
                \end{tikzpicture}
            \end{subfigure}
        \end{minipage}
        \caption{Comparison between \gls{ncg} and CFG Rescale using Stable Diffusion XL. CFG Rescale is unable to solve the saturation issue at high guidance scales compared with \gls{ncg}.}
        \label{fig:cfg-rescale}
    \end{figure}
}

\newcommand{\figDistilledMain}{
    \begin{figure}[t!]
        \centering
        \begin{minipage}{\textwidth}
            \begin{subfigure}{0.49\textwidth}
                    \begin{subfigure}{0.495\textwidth}
                        \begin{tikzpicture}
                            \node[anchor=south west, inner sep=0] (image) at (0,0) {
                                \includegraphics[width=\textwidth]{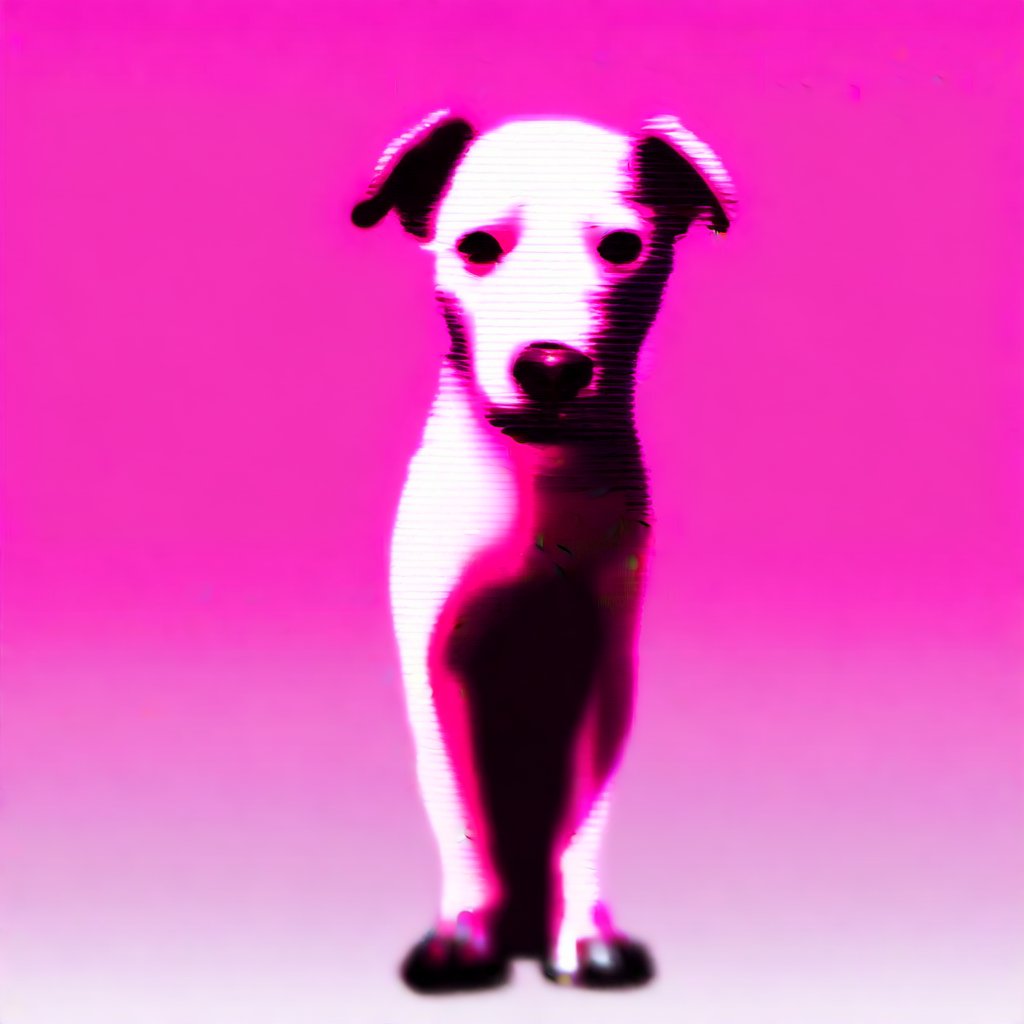}
                            };
                            \node[rotate=0, align=center, font=\normalsize, anchor=base, yshift=4pt] at (image.north) {CFG \strut};
                        \end{tikzpicture}
                    \end{subfigure}
                    \begin{subfigure}{0.495\textwidth}
                        \begin{tikzpicture}
                            \node[anchor=south west, inner sep=0] (image) at (0,0) {
                                \includegraphics[width=\textwidth]{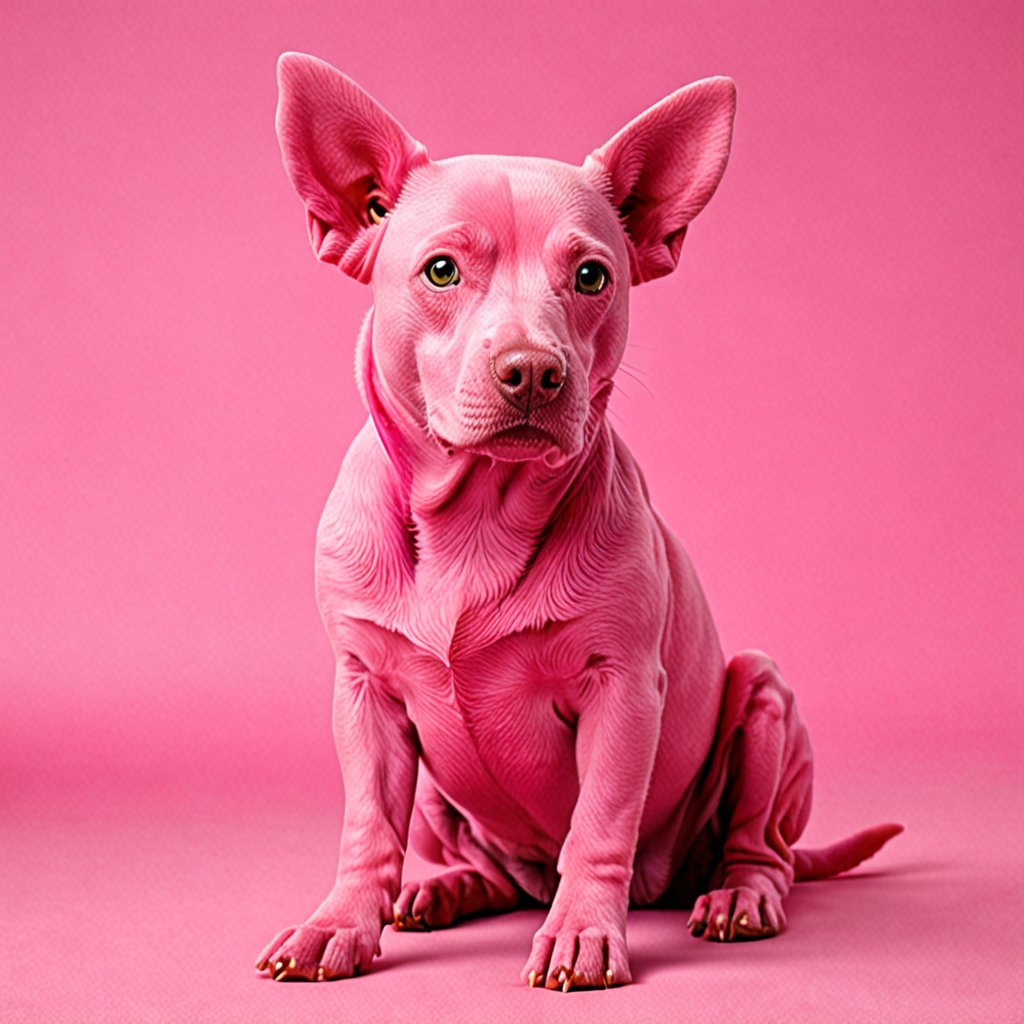}
                            };
                            \node[rotate=0, align=center, font=\normalsize, anchor=base, yshift=4pt] at (image.north) {\gls{ncg} (Ours) \strut};
                        \end{tikzpicture}
                    \end{subfigure}
                    \caption*{A pink dog}
                \end{subfigure}
                \begin{subfigure}{0.49\textwidth}
                    \begin{subfigure}{0.495\textwidth}
                        \begin{tikzpicture}
                            \node[anchor=south west, inner sep=0] (image) at (0,0) {
                                \includegraphics[width=\textwidth]{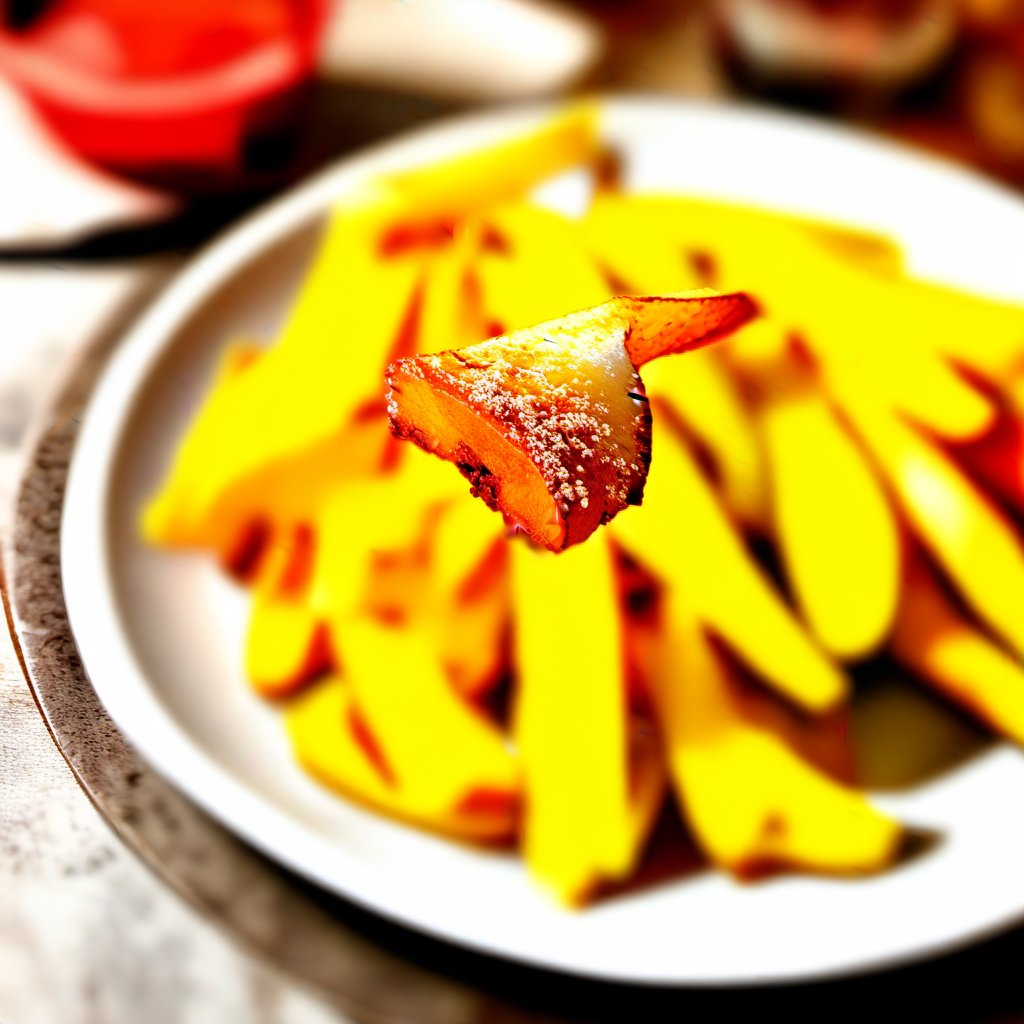}
                            };
                            \node[rotate=0, align=center, font=\normalsize, anchor=base, yshift=4pt] at (image.north) {CFG \strut};
                        \end{tikzpicture}
                    \end{subfigure}
                    \begin{subfigure}{0.495\textwidth}
                        \begin{tikzpicture}
                            \node[anchor=south west, inner sep=0] (image) at (0,0) {
                                \includegraphics[width=\textwidth]{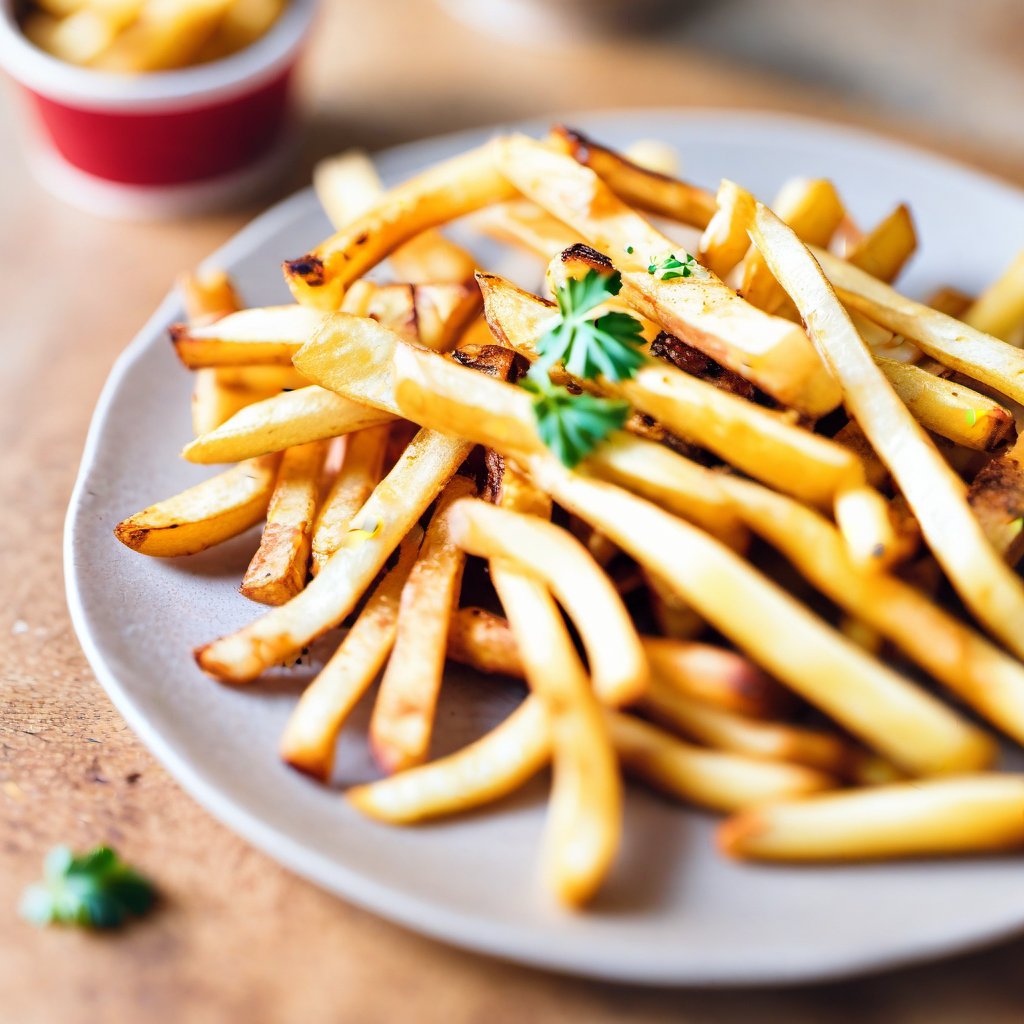}
                            };
                            \node[rotate=0, align=center, font=\normalsize, anchor=base, yshift=4pt] at (image.north) {\gls{ncg} (Ours) \strut};
                        \end{tikzpicture}
                    \end{subfigure}
                    \caption*{A plate of fries}
                \end{subfigure}
        \end{minipage}
        \caption{Showcasing the compatibility of \gls{ncg} with distilled diffusion models using SDXL-Lightning. Compared to CFG, using \gls{ncg} does not result in degradation in the output quality.}
        \label{fig:distilled}
    \end{figure}
}

\newcommand{\figDistilled}{
    \begin{figure}[t]
        \centering
        \begin{minipage}{\textwidth}
            \begin{booktabs}{c}
                \toprule
                \textsc{PixArt}-$\delta$ \\
                \midrule
                \begin{subfigure}{0.49\textwidth}
                    \begin{subfigure}{0.495\textwidth}
                        \begin{tikzpicture}
                            \node[anchor=south west, inner sep=0] (image) at (0,0) {
                                \includegraphics[width=\textwidth]{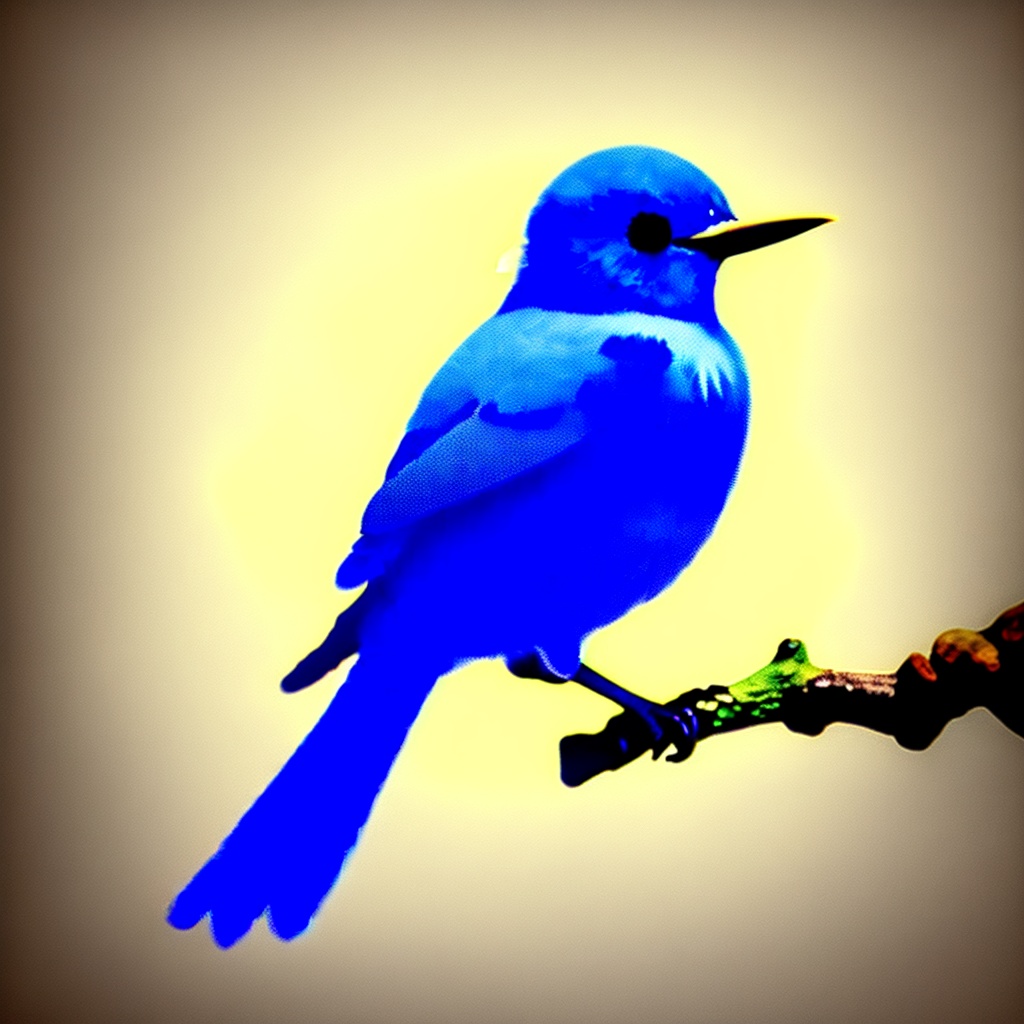}
                            };
                            \node[rotate=0, align=center, font=\normalsize, anchor=base, yshift=4pt] at (image.north) {CFG \strut};
                        \end{tikzpicture}
                    \end{subfigure}
                    \begin{subfigure}{0.495\textwidth}
                        \begin{tikzpicture}
                            \node[anchor=south west, inner sep=0] (image) at (0,0) {
                                \includegraphics[width=\textwidth]{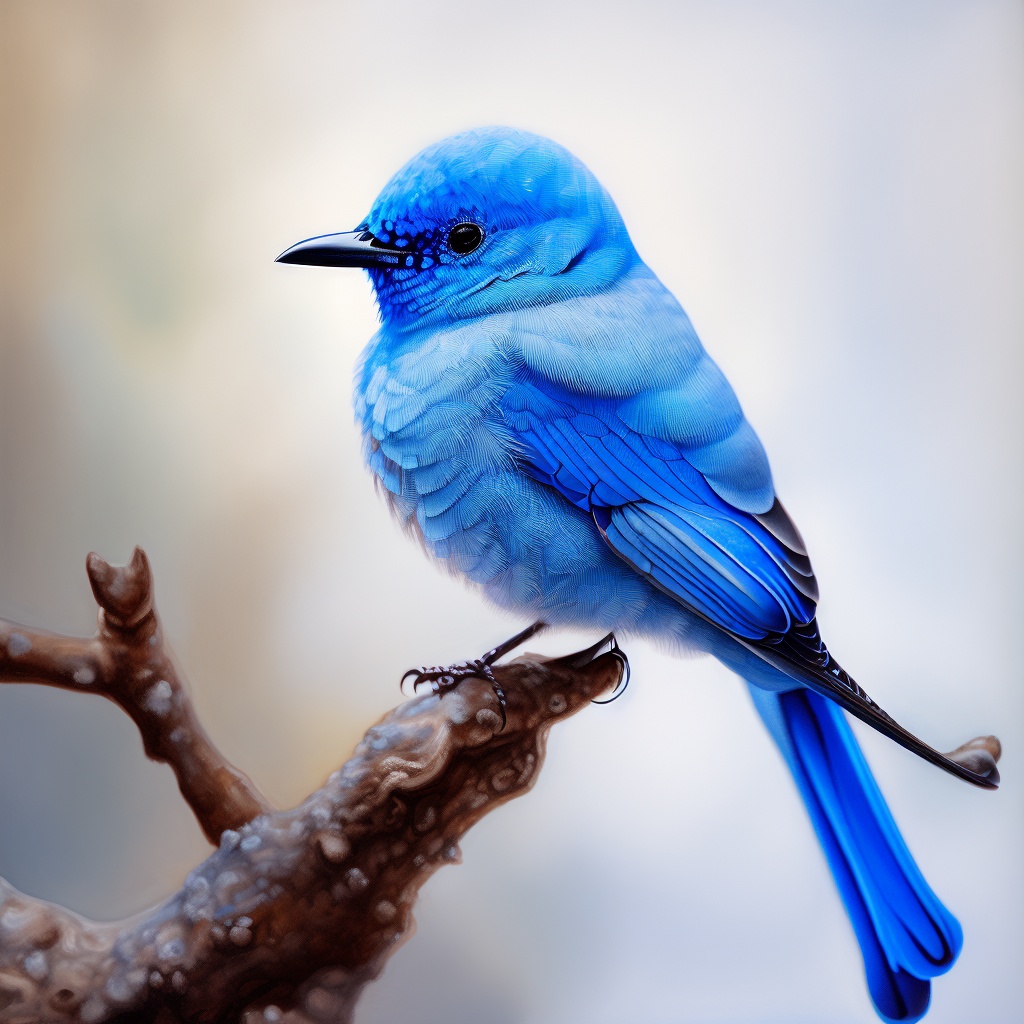}
                            };
                            \node[rotate=0, align=center, font=\normalsize, anchor=base, yshift=4pt] at (image.north) {\gls{ncg} (Ours) \strut};
                        \end{tikzpicture}
                    \end{subfigure}
                    \caption*{A blue bird}
                \end{subfigure}
                \begin{subfigure}{0.49\textwidth}
                    \begin{subfigure}{0.495\textwidth}
                        \begin{tikzpicture}
                            \node[anchor=south west, inner sep=0] (image) at (0,0) {
                                \includegraphics[width=\textwidth]{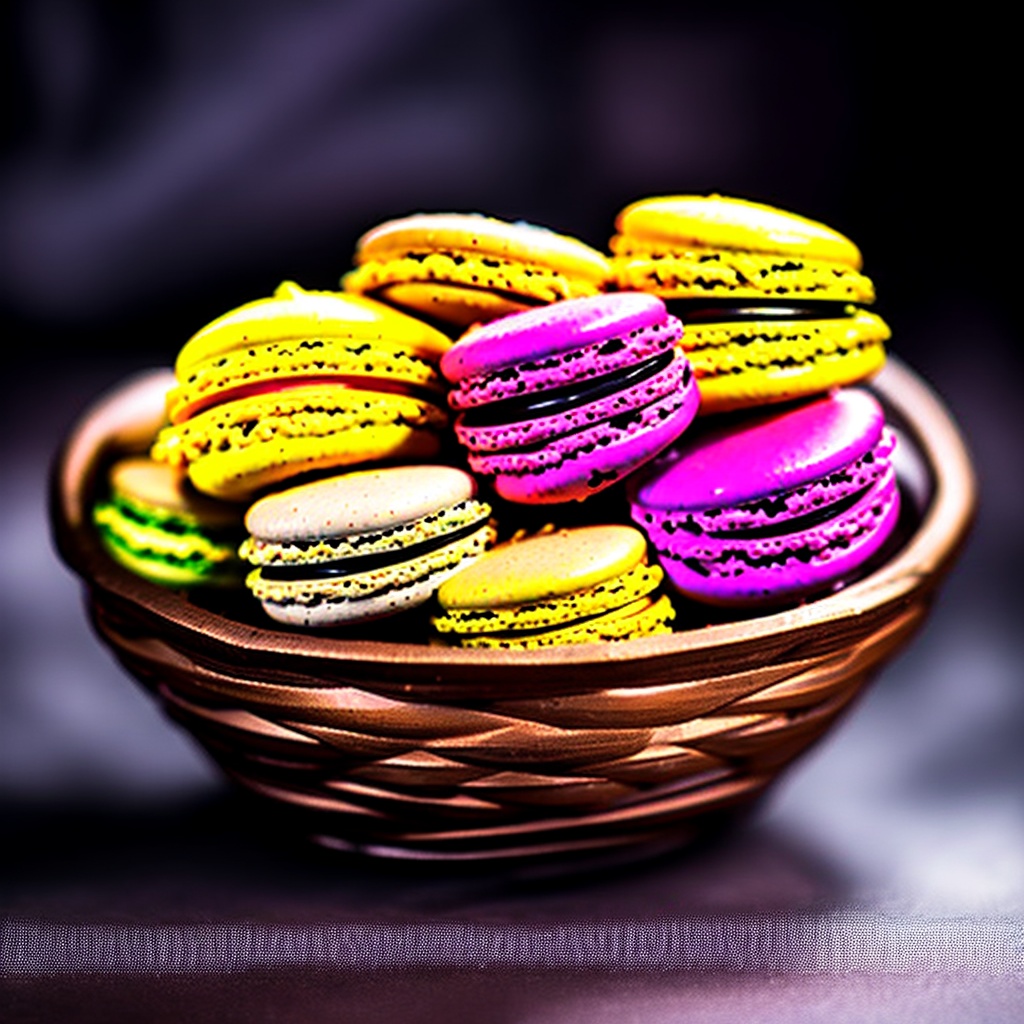}
                            };
                            \node[rotate=0, align=center, font=\normalsize, anchor=base, yshift=4pt] at (image.north) {CFG \strut};
                        \end{tikzpicture}
                    \end{subfigure}
                    \begin{subfigure}{0.495\textwidth}
                        \begin{tikzpicture}
                            \node[anchor=south west, inner sep=0] (image) at (0,0) {
                                \includegraphics[width=\textwidth]{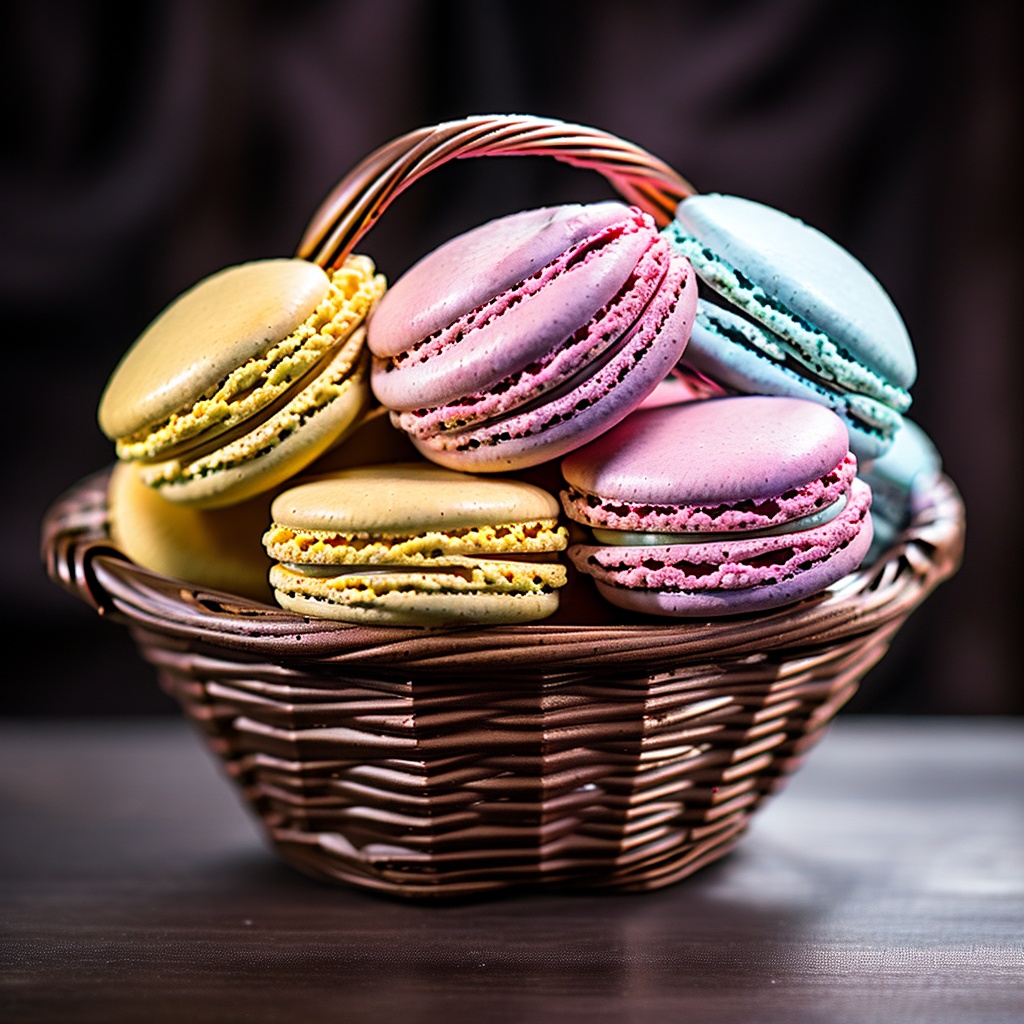}
                            };
                            \node[rotate=0, align=center, font=\normalsize, anchor=base, yshift=4pt] at (image.north) {\gls{ncg} (Ours) \strut};
                        \end{tikzpicture}
                    \end{subfigure}
                    \caption*{A basket of macarons}
                \end{subfigure} \\
                \midrule
                SDXL-Lightning \\
                \midrule
                \begin{subfigure}{0.49\textwidth}
                    \begin{subfigure}{0.495\textwidth}
                        \begin{tikzpicture}
                            \node[anchor=south west, inner sep=0] (image) at (0,0) {
                                \includegraphics[width=\textwidth]{figures/sdxl_lightning/pink_dog_base.jpg}
                            };
                            \node[rotate=0, align=center, font=\normalsize, anchor=base, yshift=4pt] at (image.north) {CFG \strut};
                        \end{tikzpicture}
                    \end{subfigure}
                    \begin{subfigure}{0.495\textwidth}
                        \begin{tikzpicture}
                            \node[anchor=south west, inner sep=0] (image) at (0,0) {
                                \includegraphics[width=\textwidth]{figures/sdxl_lightning/pink_dog_ours.jpg}
                            };
                            \node[rotate=0, align=center, font=\normalsize, anchor=base, yshift=4pt] at (image.north) {\gls{ncg} (Ours) \strut};
                        \end{tikzpicture}
                    \end{subfigure}
                    \caption*{A pink dog}
                \end{subfigure}
                \begin{subfigure}{0.49\textwidth}
                    \begin{subfigure}{0.495\textwidth}
                        \begin{tikzpicture}
                            \node[anchor=south west, inner sep=0] (image) at (0,0) {
                                \includegraphics[width=\textwidth]{figures/sdxl_lightning/fries_base.jpg}
                            };
                            \node[rotate=0, align=center, font=\normalsize, anchor=base, yshift=4pt] at (image.north) {CFG \strut};
                        \end{tikzpicture}
                    \end{subfigure}
                    \begin{subfigure}{0.495\textwidth}
                        \begin{tikzpicture}
                            \node[anchor=south west, inner sep=0] (image) at (0,0) {
                                \includegraphics[width=\textwidth]{figures/sdxl_lightning/fries_ours.jpg}
                            };
                            \node[rotate=0, align=center, font=\normalsize, anchor=base, yshift=4pt] at (image.north) {\gls{ncg} (Ours) \strut};
                        \end{tikzpicture}
                    \end{subfigure}
                    \caption*{A plate of fries}
                \end{subfigure} \\
                \midrule
                SDXL-Flash \\
                \midrule
                \begin{subfigure}{0.49\textwidth}
                    \begin{subfigure}{0.495\textwidth}
                        \begin{tikzpicture}
                            \node[anchor=south west, inner sep=0] (image) at (0,0) {
                                \includegraphics[width=\textwidth]{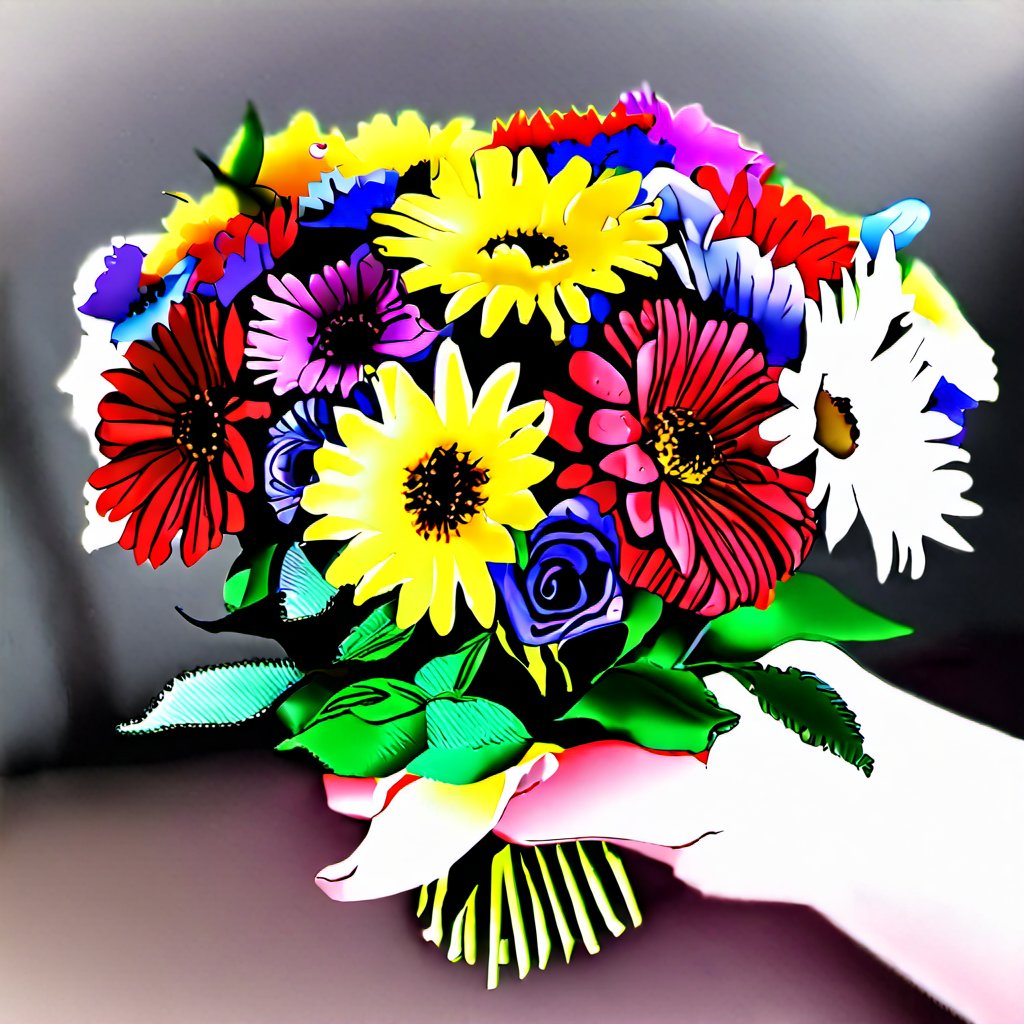}
                            };
                            \node[rotate=0, align=center, font=\normalsize, anchor=base, yshift=4pt] at (image.north) {CFG \strut};
                        \end{tikzpicture}
                    \end{subfigure}
                    \begin{subfigure}{0.495\textwidth}
                        \begin{tikzpicture}
                            \node[anchor=south west, inner sep=0] (image) at (0,0) {
                                \includegraphics[width=\textwidth]{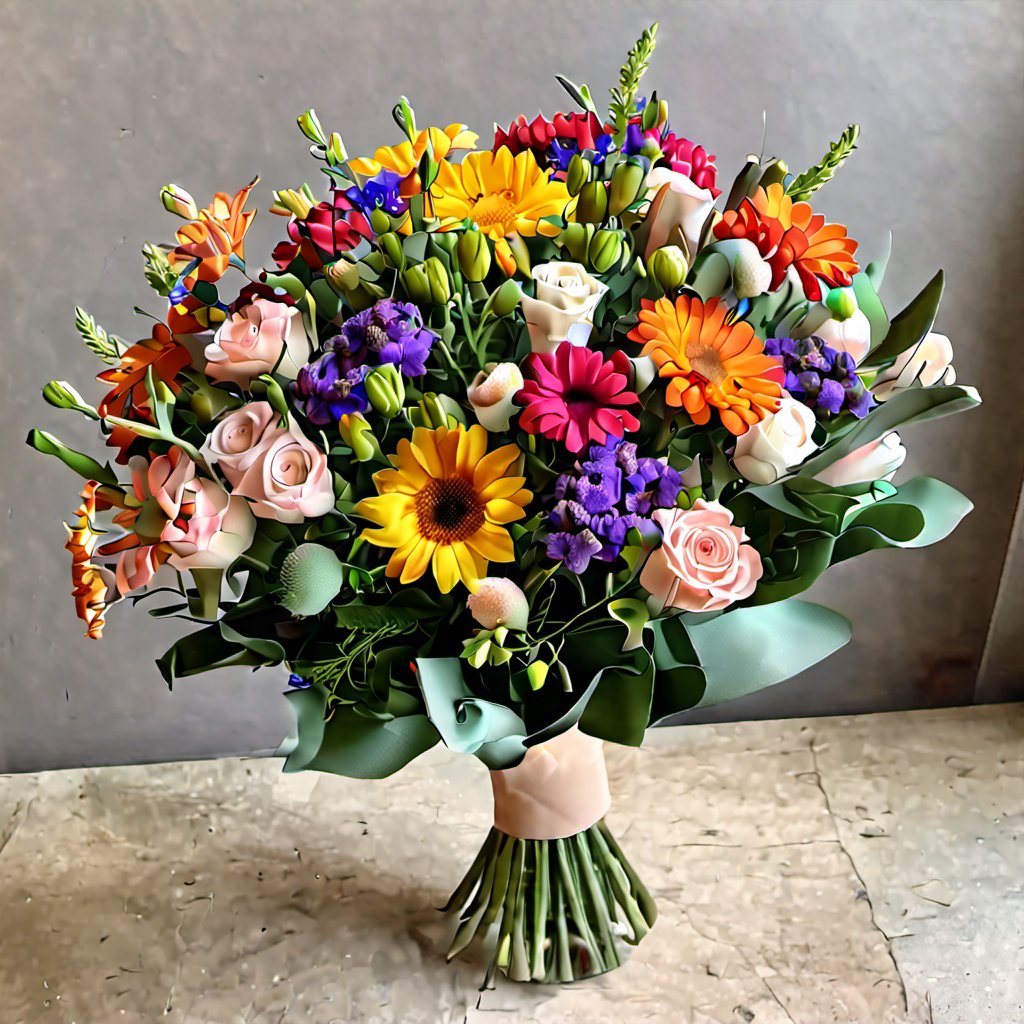}
                            };
                            \node[rotate=0, align=center, font=\normalsize, anchor=base, yshift=4pt] at (image.north) {\gls{ncg} (Ours) \strut};
                        \end{tikzpicture}
                    \end{subfigure}
                    \caption*{A bouquet of flowers}
                \end{subfigure}
                \begin{subfigure}{0.49\textwidth}
                    \begin{subfigure}{0.495\textwidth}
                        \begin{tikzpicture}
                            \node[anchor=south west, inner sep=0] (image) at (0,0) {
                                \includegraphics[width=\textwidth]{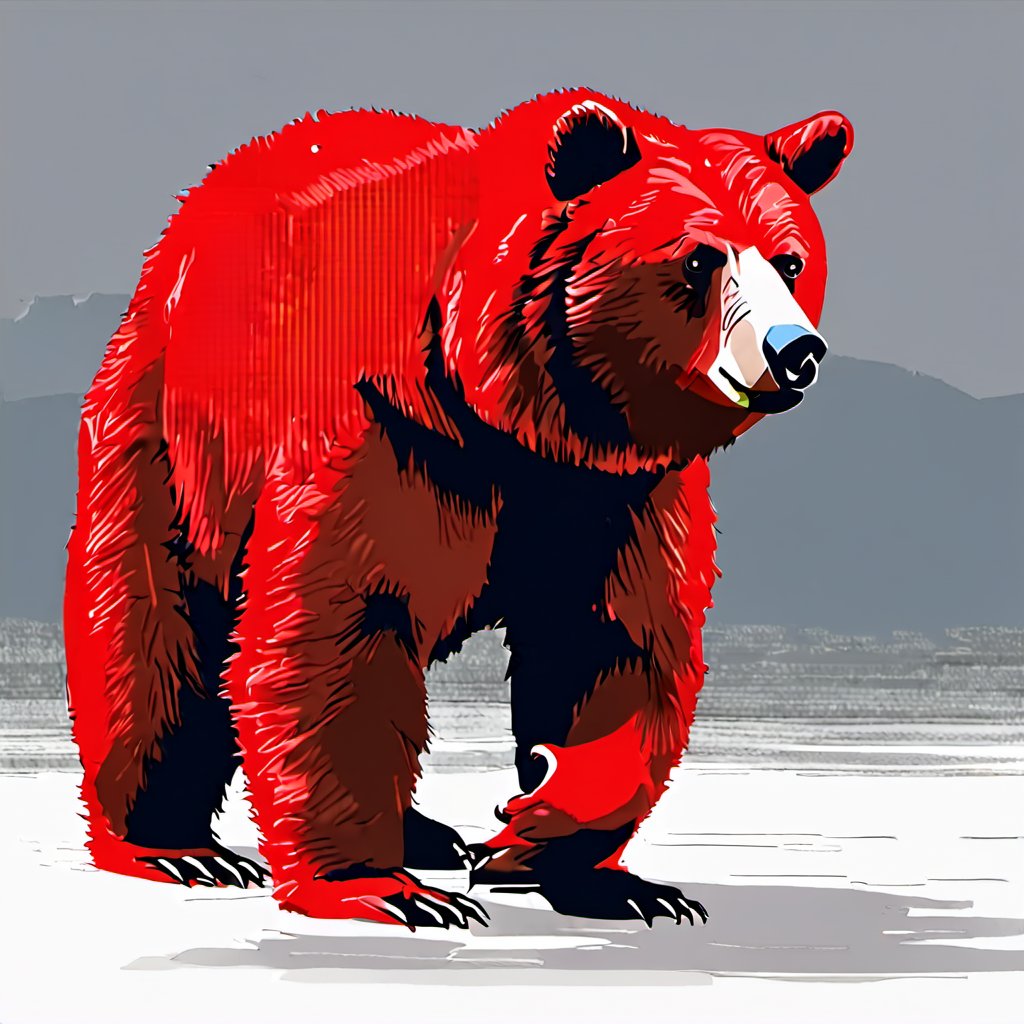}
                            };
                            \node[rotate=0, align=center, font=\normalsize, anchor=base, yshift=4pt] at (image.north) {CFG \strut};
                        \end{tikzpicture}
                    \end{subfigure}
                    \begin{subfigure}{0.495\textwidth}
                        \begin{tikzpicture}
                            \node[anchor=south west, inner sep=0] (image) at (0,0) {
                                \includegraphics[width=\textwidth]{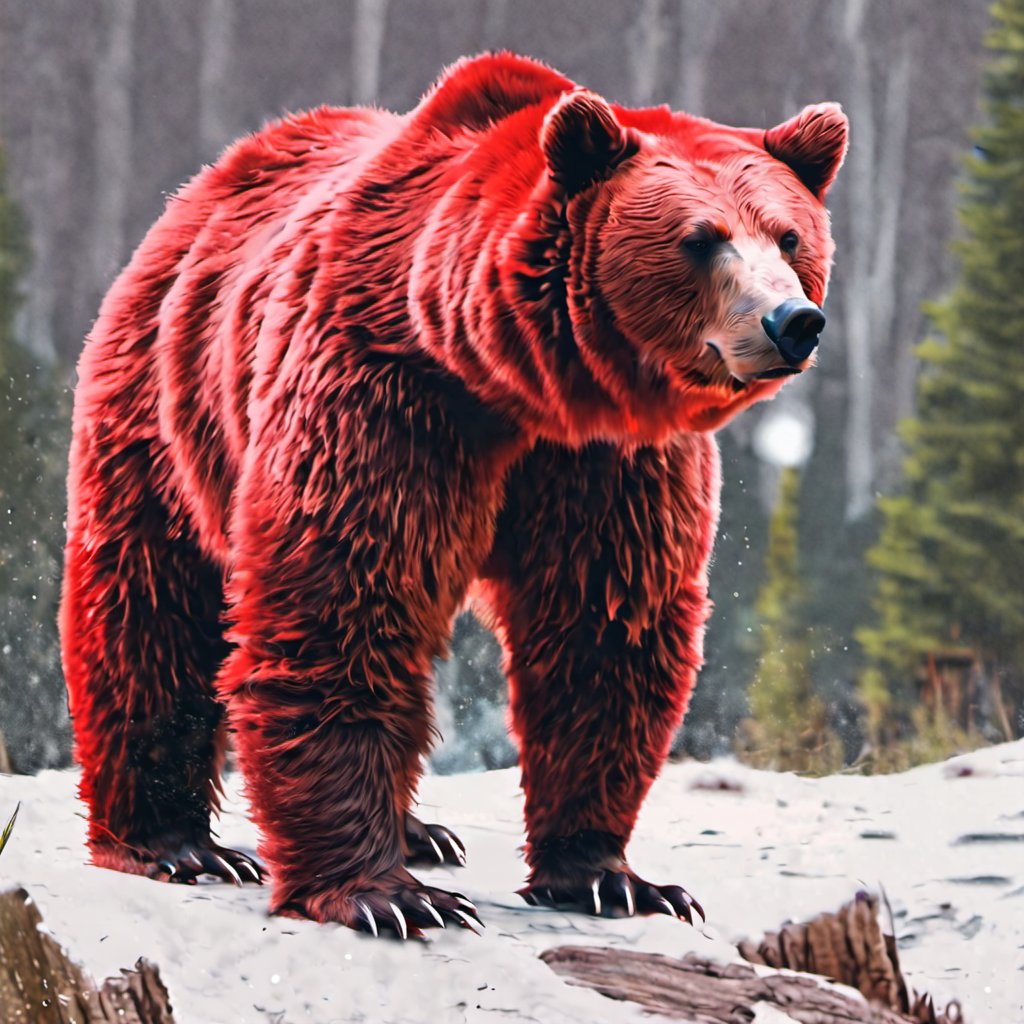}
                            };
                            \node[rotate=0, align=center, font=\normalsize, anchor=base, yshift=4pt] at (image.north) {\gls{ncg} (Ours) \strut};
                        \end{tikzpicture}
                    \end{subfigure}
                    \caption*{A red bear}
                \end{subfigure} \\
                \bottomrule
            \end{booktabs}
            
        \end{minipage}
        \caption{Extended results showcasing the compatibility of \gls{ncg} with distilled diffusion models. Compared to CFG, using \gls{ncg} does not lead to degradation in the output quality.}
        \label{fig:distilled-extended}
    \end{figure}
}

\newcommand{\figAblationProjection}{
    \vspace{-0.25cm}
    \begin{figure}[t]
        \centering
        \begin{minipage}{\textwidth}
            \begin{subfigure}{0.32\textwidth}
                \includegraphics[width=\textwidth]{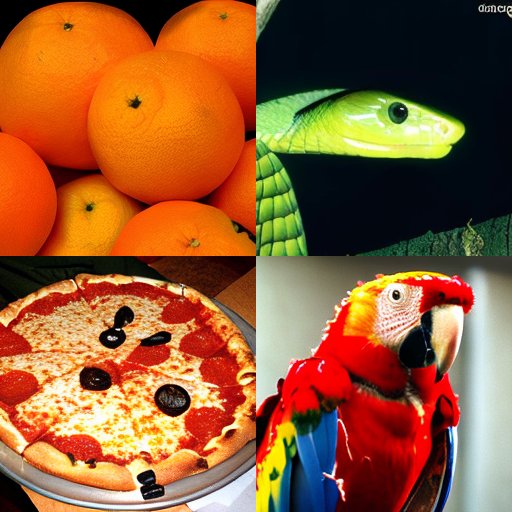}
                \caption{CFG}
            \end{subfigure}
            \begin{subfigure}{0.32\textwidth}
                \includegraphics[width=\textwidth]{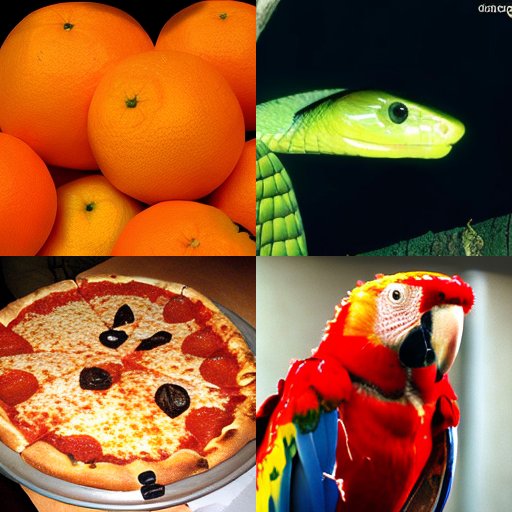}
                \caption{+Projection ($\prednoise$)}
            \end{subfigure}
            \begin{subfigure}{0.32\textwidth}
                \includegraphics[width=\textwidth]{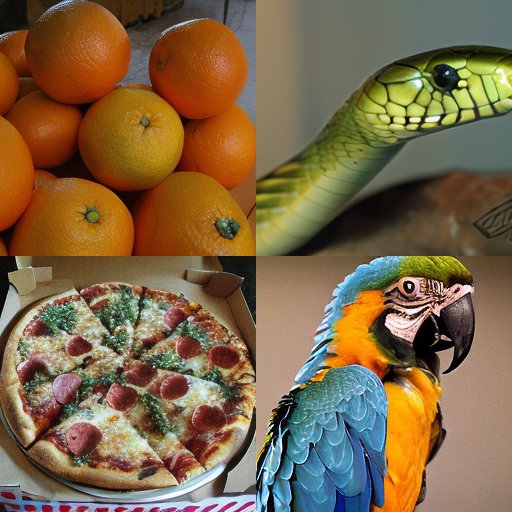}
                \caption{+Projection ($\pred$)}
            \end{subfigure}
        \end{minipage}
        \caption{The importance of projecting onto the denoised samples. When performing projection w.r.t.\ the predicted noise (b), the outputs are barely different than standard CFG (a). However, projecting onto denoised samples (c) more effectively reduces saturation.}
        \label{fig:projection-ablation}
        \vspace{-0.1cm}
    \end{figure}
}

\newcommand{\figEDMAppendix}{
    \begin{figure}[t]
        \centering
        \begin{minipage}{\textwidth}
            \begin{subfigure}{0.49\textwidth}
                \begin{tikzpicture}
                    \node[anchor=south west, inner sep=0] (image) at (0,0) {
                        \includegraphics[width=\textwidth]{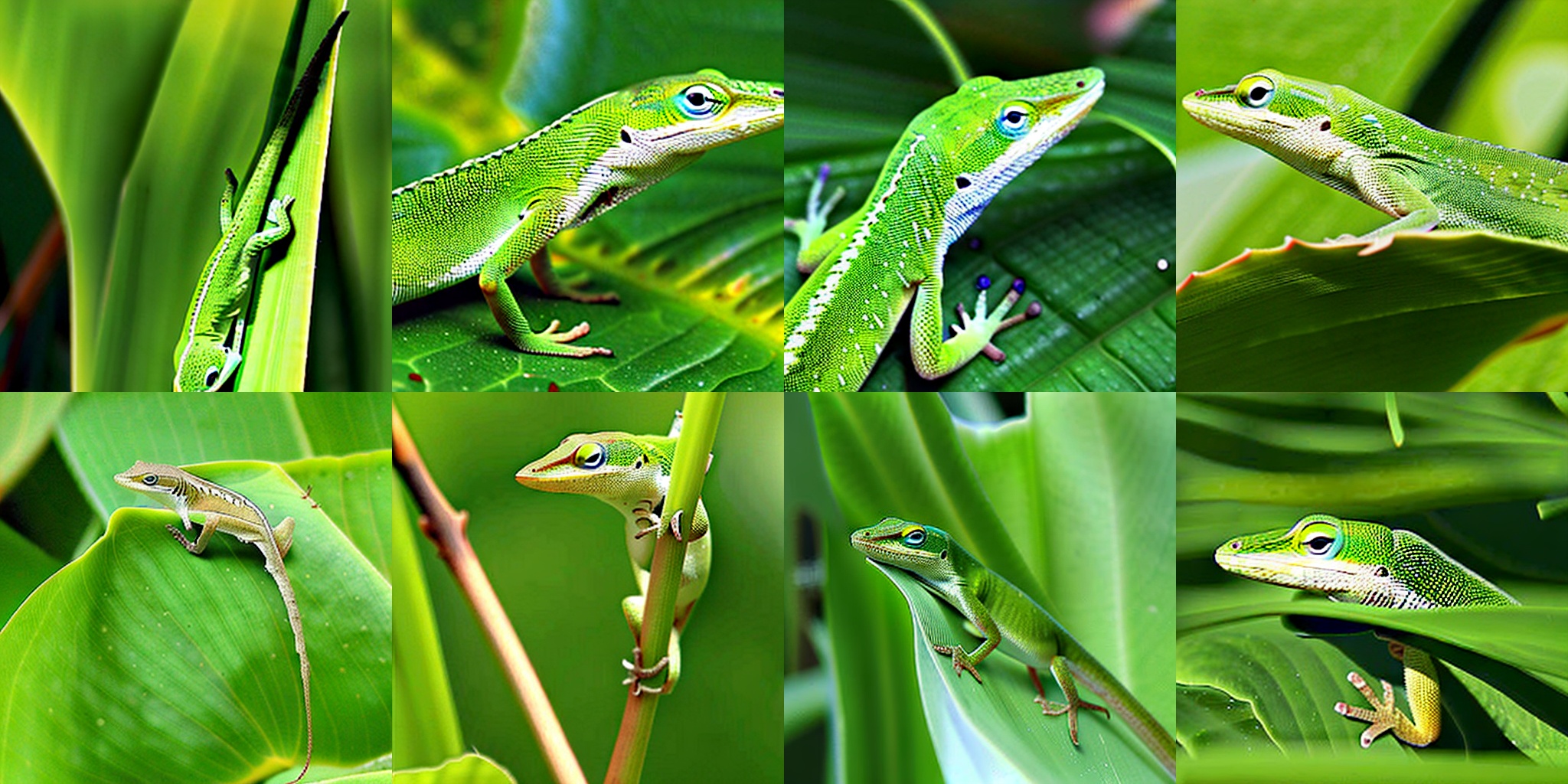}
                    };
                    \node[rotate=0, align=center, font=\normalsize, anchor=base, yshift=4pt] at (image.north) {CFG \strut};
                \end{tikzpicture}
            \end{subfigure}
            \hfill
            \begin{subfigure}{0.49\textwidth}
                \begin{tikzpicture}
                    \node[anchor=south west, inner sep=0] (image) at (0,0) {
                        \includegraphics[width=\textwidth]{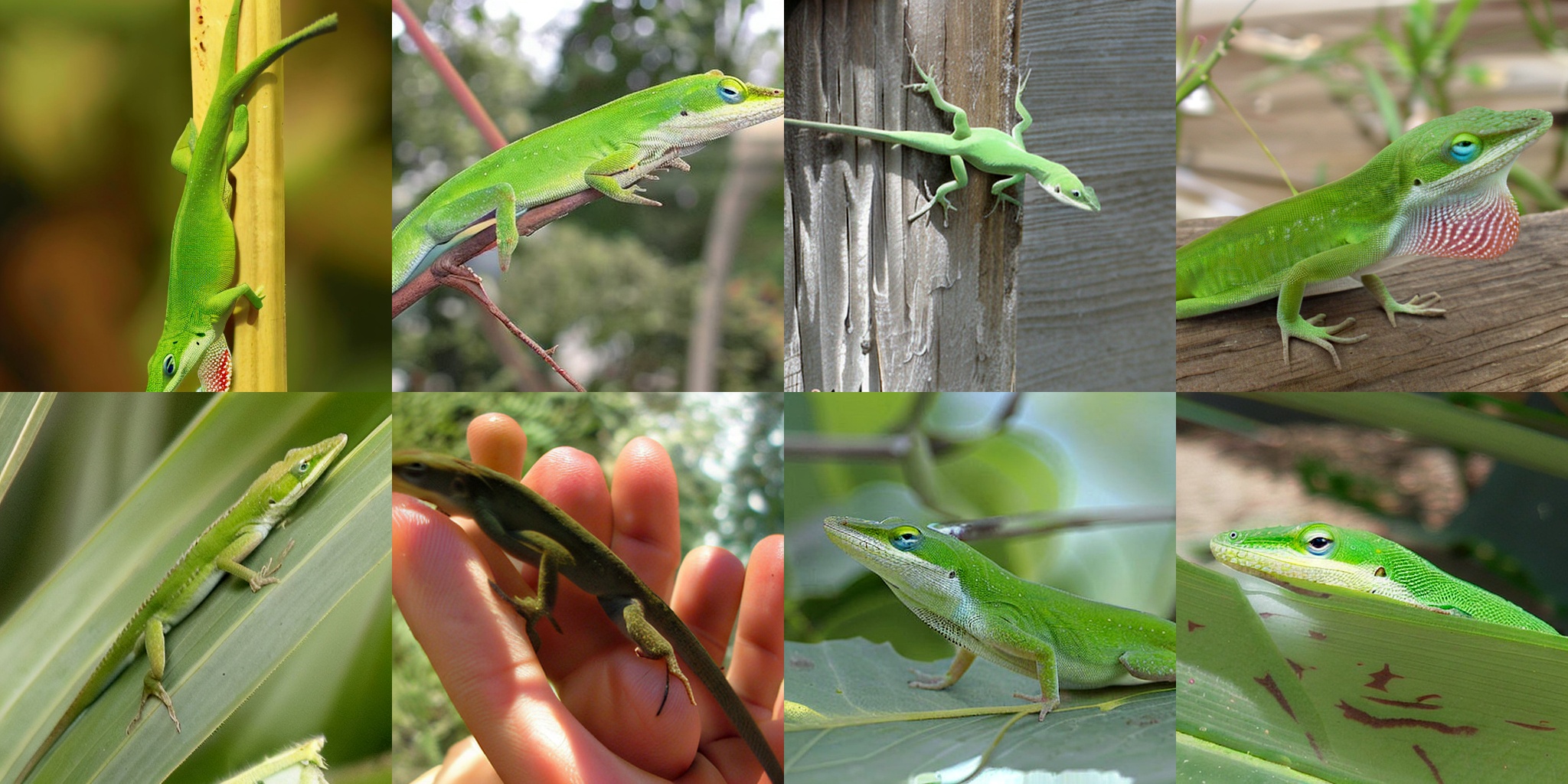}
                    };
                    \node[rotate=0, align=center, font=\normalsize, anchor=base, yshift=4pt] at (image.north) {\gls{ncg} (Ours) \strut};
                \end{tikzpicture}
            \end{subfigure}
            \hfill
            \begin{subfigure}{0.49\textwidth}
                \begin{tikzpicture}
                    \node[anchor=south west, inner sep=0] (image) at (0,0) {
                        \includegraphics[width=\textwidth]{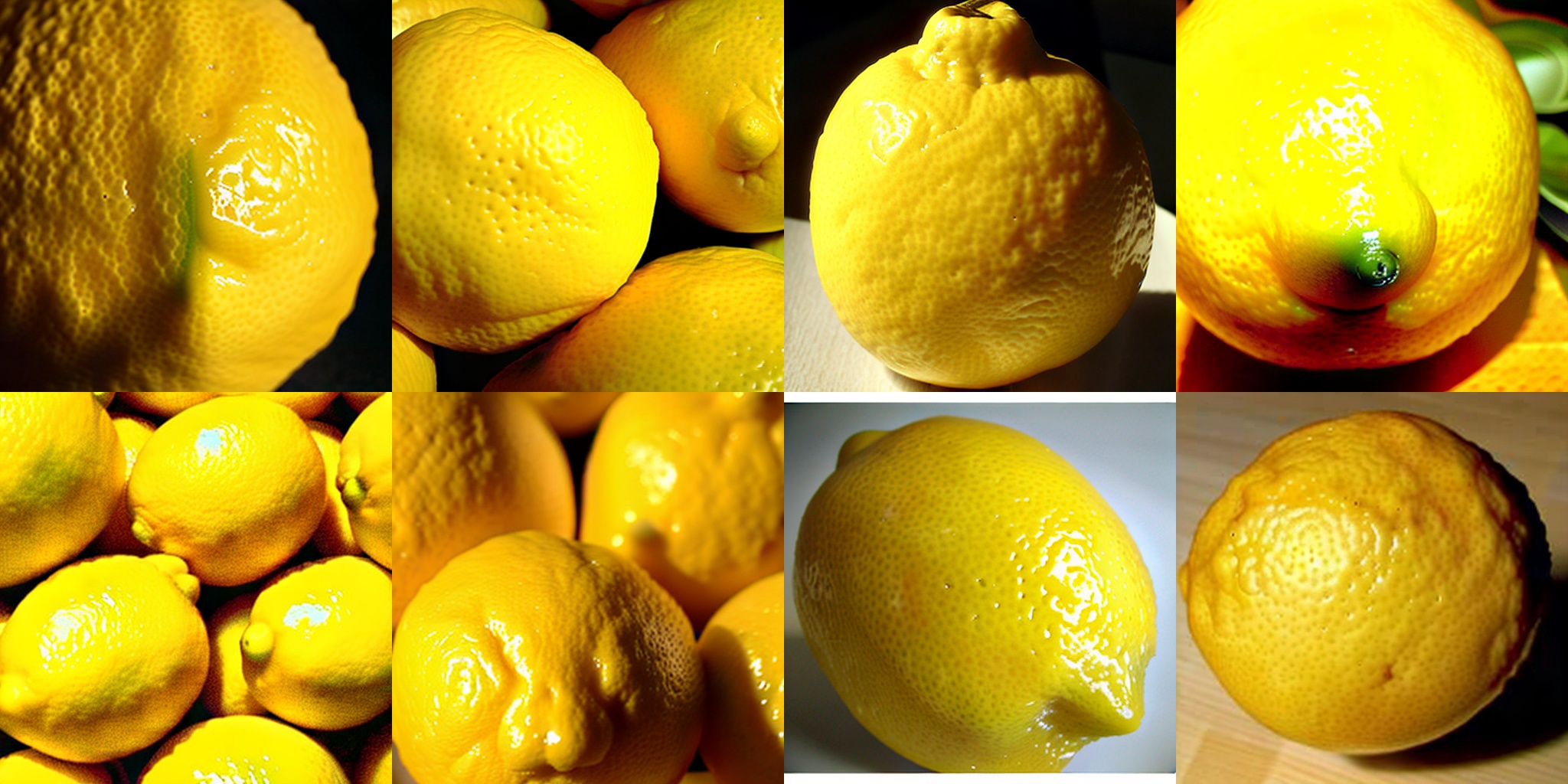}
                    };
                    \node[rotate=0, align=center, font=\normalsize, anchor=base, yshift=4pt] at (image.north) {CFG \strut};
                \end{tikzpicture}
            \end{subfigure}
            \begin{subfigure}{0.49\textwidth}
                \begin{tikzpicture}
                    \node[anchor=south west, inner sep=0] (image) at (0,0) {
                        \includegraphics[width=\textwidth]{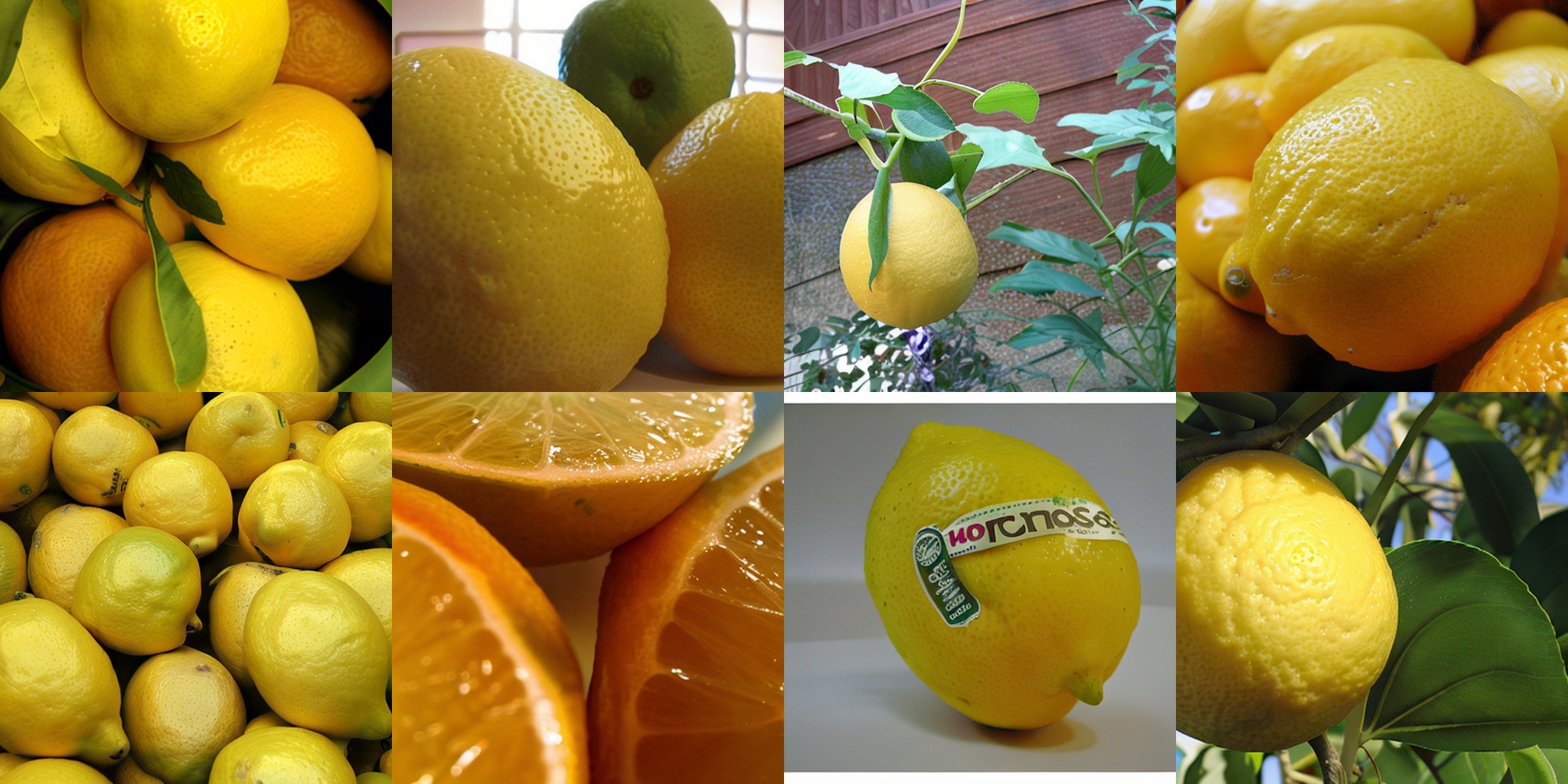}
                    };
                    \node[rotate=0, align=center, font=\normalsize, anchor=base, yshift=4pt] at (image.north) {\gls{ncg} (Ours) \strut};
                \end{tikzpicture}
            \end{subfigure}
            \hfill
            \begin{subfigure}{0.49\textwidth}
                \begin{tikzpicture}
                    \node[anchor=south west, inner sep=0] (image) at (0,0) {
                        \includegraphics[width=\textwidth]{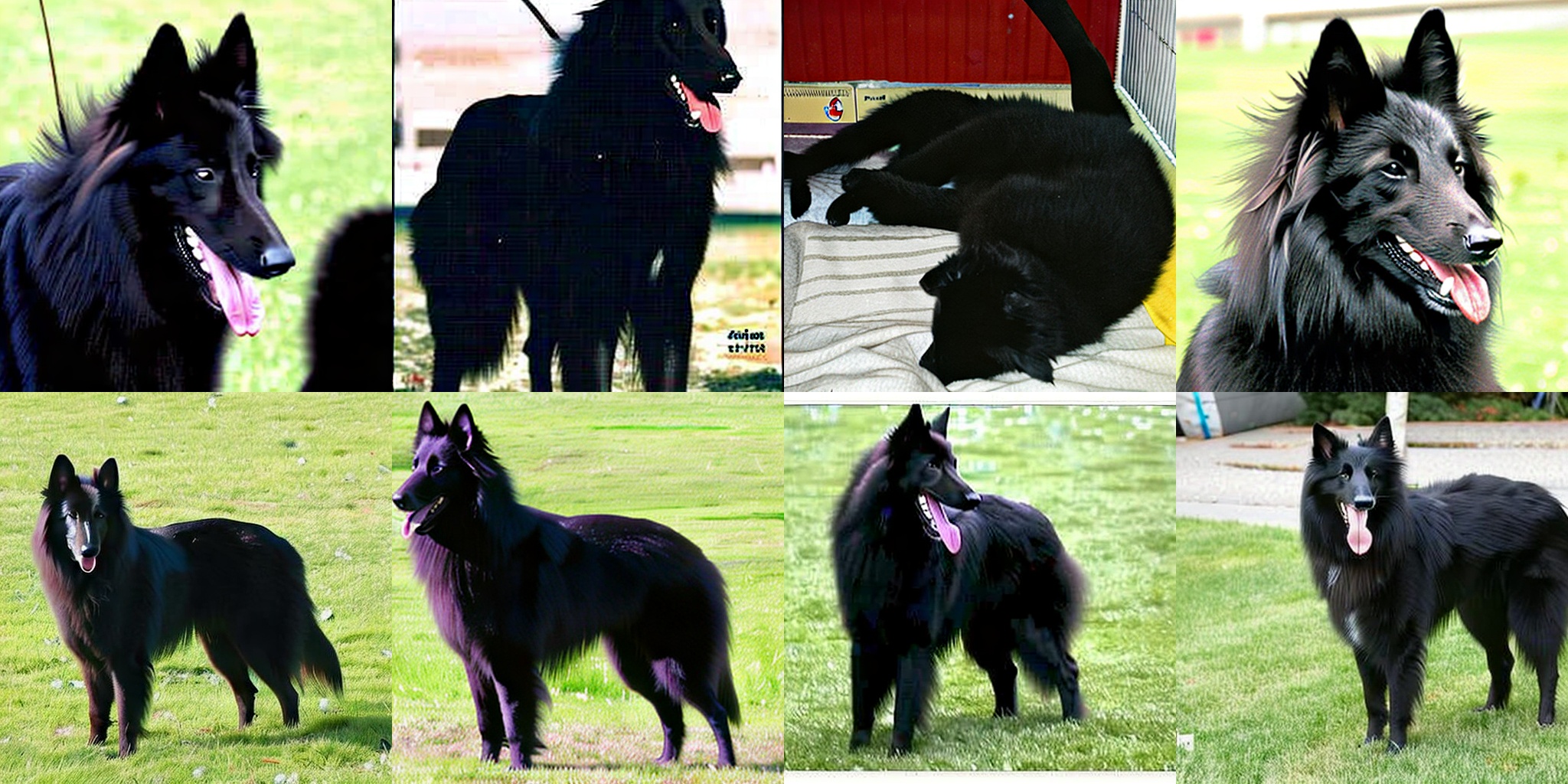}
                    };
                    \node[rotate=0, align=center, font=\normalsize, anchor=base, yshift=4pt] at (image.north) {CFG \strut};
                \end{tikzpicture}
            \end{subfigure}
            \hfill
            \begin{subfigure}{0.49\textwidth}
                \begin{tikzpicture}
                    \node[anchor=south west, inner sep=0] (image) at (0,0) {
                        \includegraphics[width=\textwidth]{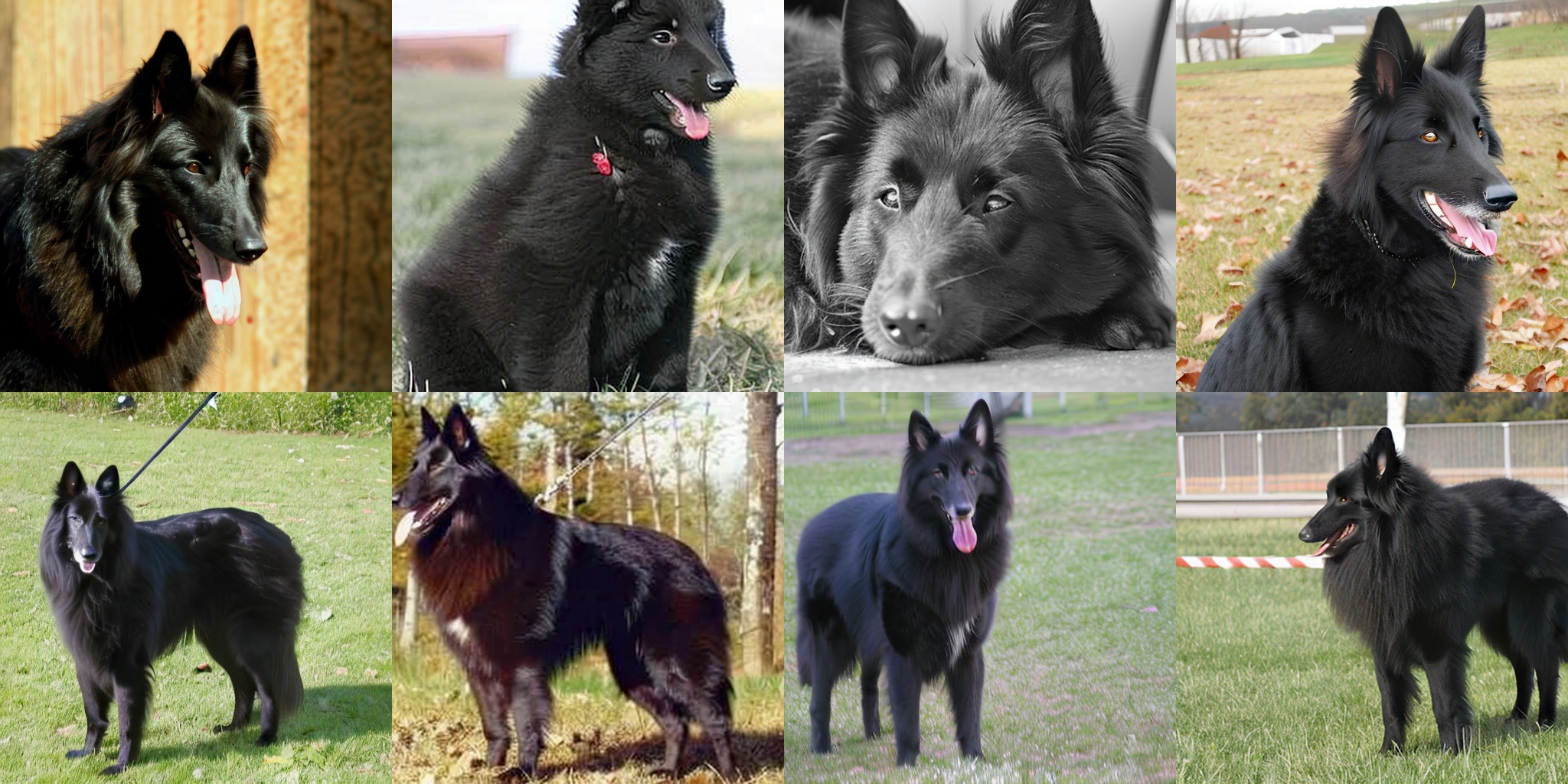}
                    };
                    \node[rotate=0, align=center, font=\normalsize, anchor=base, yshift=4pt] at (image.north) {\gls{ncg} (Ours) \strut};
                \end{tikzpicture}
            \end{subfigure}
            \begin{subfigure}{0.49\textwidth}
                \begin{tikzpicture}
                    \node[anchor=south west, inner sep=0] (image) at (0,0) {
                        \includegraphics[width=\textwidth]{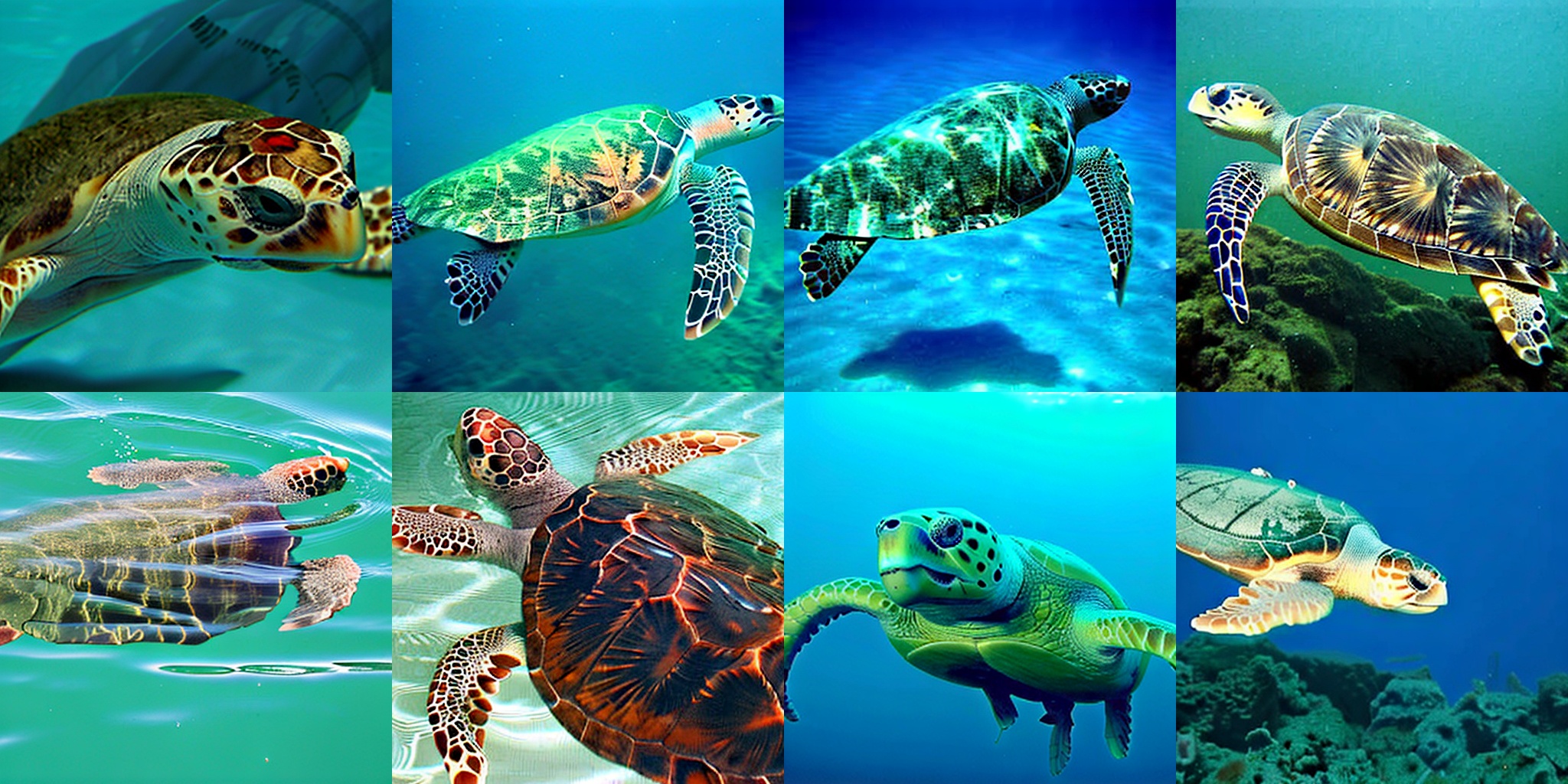}
                    };
                    \node[rotate=0, align=center, font=\normalsize, anchor=base, yshift=4pt] at (image.north) {CFG \strut};
                \end{tikzpicture}
            \end{subfigure}
            \hfill
            \begin{subfigure}{0.49\textwidth}
                \begin{tikzpicture}
                    \node[anchor=south west, inner sep=0] (image) at (0,0) {
                        \includegraphics[width=\textwidth]{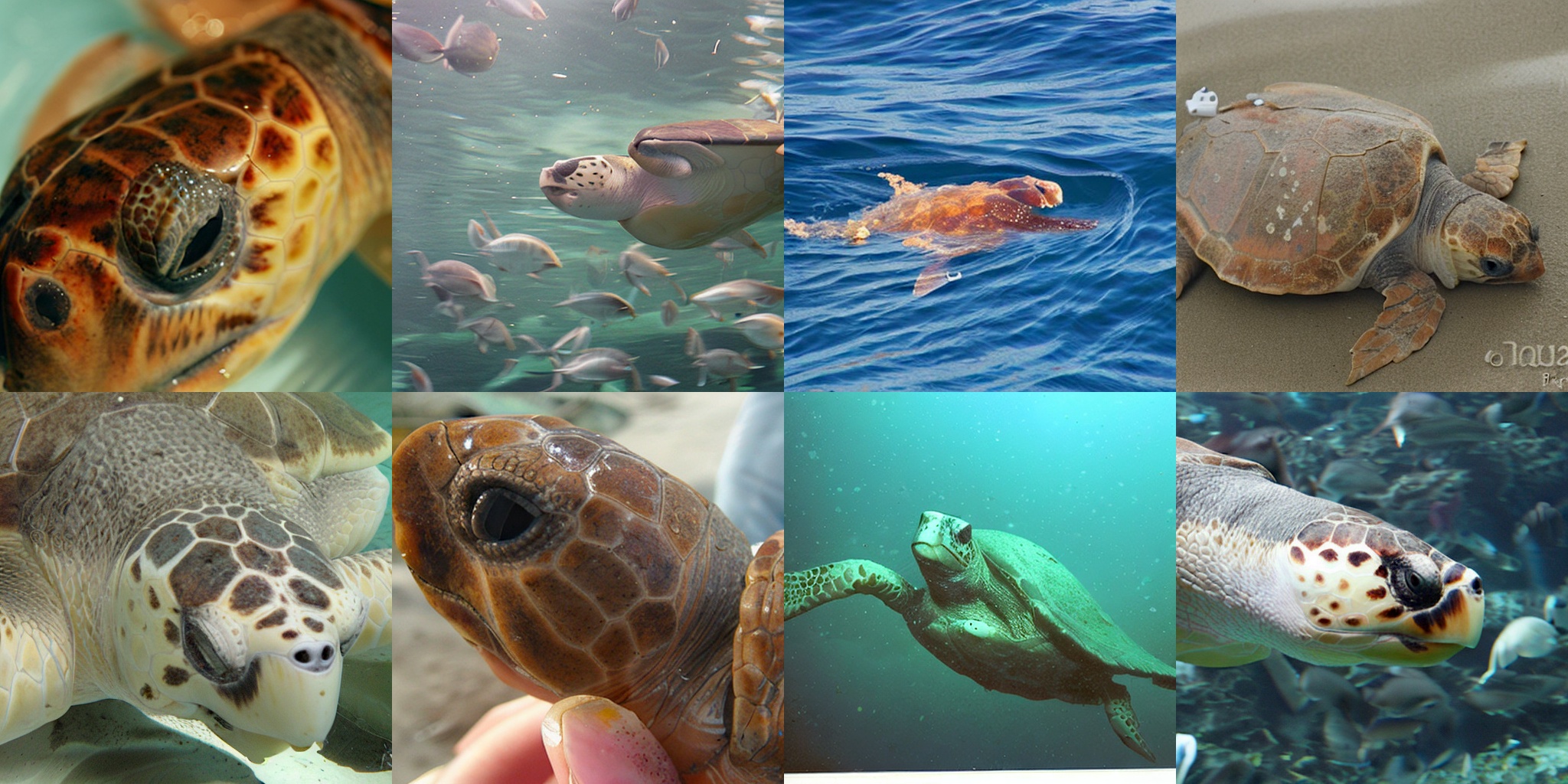}
                    };
                    \node[rotate=0, align=center, font=\normalsize, anchor=base, yshift=4pt] at (image.north) {\gls{ncg} (Ours) \strut};
                \end{tikzpicture}
            \end{subfigure}
        \end{minipage}
        \caption{More visual results comparing \gls{ncg} and CFG using EDM2.}
        \label{fig:edm-appendix}
    \end{figure}
}

\newcommand{\tabResultMain}{
    \begin{table}[t]
        \centering
        \caption{Quantitative comparison between CFG and \gls{ncg}. \gls{ncg} consistently improves FID, recall and color metrics while maintaining similar or better precision compared to CFG.}
        \label{table:main-results}
        \resizebox{\textwidth}{!}{
            \begin{tabular}{llccccccc}
                \toprule
                Model & Guidance & FID $\downarrow$ & Precision $\uparrow$ & Recall $\uparrow$ & Saturation $\downarrow$  & Contrast $\downarrow$\\
                \midrule
                
                \multirow{2}{*}{EDM2-S ($\wcfg=4$)} & CFG & 10.42  & \textbf{0.85} & 0.48 & 0.46 & 0.27 \\
                & \mycc \gls{ncg} (Ours) & \mycc \textbf{6.49} & \mycc \textbf{0.85} & \mycc \textbf{0.62} & \mycc \textbf{0.33} & \mycc \textbf{0.21}\\
                \midrule

                \multirow{2}{*}{EDM2-XXL ($\wcfg=2$)} & CFG & 8.65  & \textbf{0.84} & 0.57 & 0.37 & 0.23 \\
                & \mycc \gls{ncg} (Ours) & \mycc \textbf{4.94} & \mycc {0.83} & \mycc \textbf{0.67} & \mycc \textbf{0.31} & \mycc \textbf{0.21}\\
                \midrule

                \multirow{2}{*}{DiT-XL/2 ($\wcfg=4$)} & CFG & 19.14 & \textbf{0.92} & 0.35 & 0.37 & 0.25 \\
                & \mycc \gls{ncg} (Ours) & \mycc \textbf{9.34} & \mycc 0.89  & \mycc \textbf{0.56} & \mycc \textbf{0.30} & \mycc \textbf{0.20}\\
                \midrule

                \multirow{2}{*}{Stable Diffusion 2.1 ($\wcfg=10$)} & CFG & 27.53 & {0.65} & 0.41 & 0.36  & 0.27  \\
                & \mycc \gls{ncg} (Ours) &  \mycc \textbf{22.21} & \mycc \textbf{0.67} & \mycc \textbf{0.49} & \mycc \textbf{0.27} & \mycc \textbf{0.22} \\
                \midrule
                \multirow{2}{*}{Stable Diffusion XL ($\wcfg=15$)} & CFG & 26.29 & {0.62} & 0.49 & 0.28 & 0.24 \\
                & \mycc \gls{ncg} (Ours) &  \mycc \textbf{25.35} & \mycc \textbf{0.64} & \mycc \textbf{0.50} & \mycc \textbf{0.18} & \mycc \textbf{0.17} \\
                \bottomrule
            \end{tabular}
        }
        \vspace*{-0.3cm}

    \end{table}
}

\newcommand{\tabIGCads}{
    \begin{table}[t]
        \centering
        \caption{Compatibility of \gls{ncg} with CADS \citep{sadat2024cads} and IG \citep{intervalGuidance}. Combining \gls{ncg} with other methods that improve diversity results in better FID than each method in isolation.}
        \label{table:ig-cads}
            \begin{subtable}[t]{0.495\textwidth}
                \caption{CADS}
                \resizebox{\linewidth}{!}{
                \begin{tabular}{p{1.1cm}ccccccc}
                    \toprule
                    Guidance & FID $\downarrow$ & Precision $\uparrow$ & Recall $\uparrow$ & Saturation $\downarrow$  & Contrast $\downarrow$\\
                    \midrule    
                    CFG & 10.42  & \textbf{0.85} & 0.48 & 0.46 & 0.27 \\
                    +CADS & 8.65  & \textbf{0.85} & 0.56 & 0.43 & 0.26 \\
                    +\gls{ncg} & \mycc {6.49} & \mycc \textbf{0.85} & \mycc {0.62} & \mycc {0.33} & \mycc \textbf{0.21}\\
                    +both & \mycc \textbf{5.56} & \mycc {0.84} & \mycc \textbf{0.64} & \mycc \textbf{0.32} & \mycc \textbf{0.21}\\
                    \bottomrule
                \end{tabular}
        }
            \end{subtable}
            \begin{subtable}[t]{0.495\textwidth}
                \caption{Interval guidance (IG)}
                \resizebox{\linewidth}{!}{
                \begin{tabular}{p{1.1cm}ccccccc}
                    \toprule
                    Guidance & FID $\downarrow$ & Precision $\uparrow$ & Recall $\uparrow$ & Saturation $\downarrow$  & Contrast $\downarrow$\\
                    \midrule
                    CFG & 10.42  & \textbf{0.85} & 0.48 & 0.46 & 0.27 \\
                    +IG & 7.49  & {0.84} & 0.60 & 0.39 & 0.25 \\
                    +\gls{ncg} & \mycc {6.49} & \mycc \textbf{0.85} & \mycc {0.62} & \mycc \textbf{0.33} & \mycc \textbf{0.21}\\
                    +both & \mycc \textbf{5.29} & \mycc {0.84} & \mycc \textbf{0.65} & \mycc \textbf{0.33} & \mycc {0.22}\\
                    \bottomrule
                \end{tabular}
        }
            \end{subtable}
    \end{table}
}

\newcommand{\tabAblationHPs}{
    \begin{table}[t!]
        \vspace*{0.1cm}
        \centering
        \caption{Ablation study examining various design elements in \gls{ncg}.}
        \begin{subtable}[t]{0.3\textwidth}
            \centering
            \caption{Influence of \(\eta\)}
            \scalebox{0.8}{
                \begin{booktabs}{colspec = {Q[c, 0.7cm]Q[c, 1.0cm]Q[c, 1.65cm]}, row{1-Z}={font=\footnotesize}}
                    \toprule
                    \(\eta\) & FID $\downarrow$ & Saturation $\downarrow$ \\
                    \midrule
                    0.0 & \textbf{6.49} & \textbf{0.33} \\
                    0.25 & \textbf{6.49} & 0.34 \\
                    0.5 & \textbf{6.49} & 0.36 \\
                    1.0 &  6.63 & 0.37\\
                    \bottomrule
                \end{booktabs}
            }
            \label{table:ablation-eta}
        \end{subtable}
        \begin{subtable}[t]{0.34\textwidth}
            \centering
            \caption{Impact of rescaling \(r\)}
            \scalebox{0.8}{
                \begin{booktabs}{colspec = {Q[c, 0.7cm]Q[c, 1.0cm]Q[c, 1.65cm]}, row{1-Z}={font=\footnotesize}}
                    \toprule
                    \(r\) & FID $\downarrow$ & Recall $\uparrow$\\
                    \midrule
                    0.25 & {7.45} & \textbf{0.72} \\
                    2.5 & \textbf{{6.49}} & 0.62 \\
                    10 & 7.97 & 0.57  \\
                    $\infty$ & 7.93 & 0.56 \\
                    \bottomrule
                \end{booktabs}
            }
            \label{table:ablation-r}
        \end{subtable}
        \begin{subtable}[t]{0.34\textwidth}
            \centering
            \caption{Effect of momentum \(\beta\)}
            \scalebox{0.8}{
                \begin{booktabs}{colspec = {Q[c, 0.7cm]Q[c, 1.0cm]Q[c, 1.65cm]}, row{1-Z}={font=\footnotesize}}
                    \toprule
                    $\beta$ & FID $\downarrow$ & Recall $\uparrow$ \\
                    \midrule
                    $\num{-1.5}$  & 13.38 & \textbf{0.73} \\
                     $\num{-0.75}$ & \textbf{6.49} & 0.62 \\
                     $\num{0.0}$ & 6.84 & 0.60 \\
                     $\num{0.5}$ & 7.10 & 0.59 \\
                    \bottomrule
                \end{booktabs}
            }

            \label{table:ablation-beta}
        \end{subtable}
    \end{table}
}

\newcommand{\tabSamplers}{
    \begin{table}[t]
        \centering
        \caption{Impact of using \gls{ncg} with popular diffusion samplers using the class-conditional ImageNet model (DiT-XL/2). Compared to CFG, \gls{ncg} showes improved metrics across all samplers.}
        \label{table:samplers}
        \resizebox{\textwidth}{!}
        {
            \small
            \begin{tabular}{lcccccccc}
                \toprule
                & \multicolumn{3}{c}{{\gls{ncg} (Ours)}} & \multicolumn{3}{c}{{CFG}} \\
                \cmidrule(lr){2-4} \cmidrule(lr){5-7}
                Sampler & FID $\downarrow$ & Recall $\downarrow$ & Saturation $\downarrow$ & FID $\downarrow$ & Recall $\downarrow$ & Saturation $\downarrow$  \\
                \midrule
                DDIM \citep{songDenoisingDiffusionImplicit2022} & \mycc \textbf{6.69} & \mycc \textbf{0.62} & \mycc \textbf{0.30} & 17.45 & 0.38 & 0.42 \\
                DPM++ \citep{lu2022dpm} & \mycc \textbf{6.87} & \mycc \textbf{0.62} & \mycc \textbf{0.32} & 17.65 & 0.38 & 0.43  \\
                SDE-DPM++ \citep{lu2022dpm} & \mycc \textbf{8.53} & \mycc \textbf{0.57} & \mycc \textbf{0.32} & 19.01 & 0.36 & 0.43 \\
                PNDM \citep{liu2022pseudo} & \mycc \textbf{5.37} & \mycc \textbf{0.68} & \mycc \textbf{0.32} & 16.50 & 0.40 & 0.43 \\
                UniPC \citep{zhao2023unipc} & \mycc \textbf{6.91} & \mycc \textbf{0.62}  & \mycc \textbf{0.32} & 17.65 & 0.38 & 0.43 \\
                \bottomrule
            \end{tabular}
            \vspace*{-0.4cm}
        }
    \end{table}
}

\newcommand{\tabPredicitonFormulas}{
\begin{table}[t!]
    \centering
    \caption{Summary of calculating denoised predictions $\pred$ for different diffusion models.}
    \label{tab:x0-formula}
        \maxsizebox{\textwidth}{!}{
        \begin{booktabs}{
            colspec = {Q[l, m]Q[l, m]Q[l, m]Q[l, m]}
        }
        \toprule
        Config & Forward process $\vz_t$ & Model prediciton & Denoised prediction $\pred$\\
        \midrule
        DDPM & \(\alpha_t \vx + \sigma_t \mepsilon\) & $\prednoise$ & $\prn{\vz_t - \sigma_t \prednoise}/{\alpha_t}$\\
        \midrule
        DDPM & \(\alpha_t \vx + \sigma_t \mepsilon\) & $\predv$ & $\alpha_t \vz_t - \sigma_t \predv$\\
        \midrule
        EDM & \(\vx + \sigma(t) \mepsilon\) & $F_{\mtheta} (c_{\mathrm{in}}(t) \vz_t, c_{\mathrm{noise}}(t), \vy)$ & $c_{\mathrm{skip}}(t) \vz_t + c_{\mathrm{out}}(t) F_{\mtheta} (c_{\mathrm{in}}(t) \vz_t, c_{\mathrm{noise}}(t), \vy)$\\
        \midrule
        Rectified flow & \((1 - t) \vx + t \mepsilon\) & $\predv$ & $\vz_t - t \predv$\\
        \bottomrule
        \end{booktabs}

    }
    \end{table}
}

\newcommand{\tabICG}{
    \begin{table}[t!]
        \centering
        \caption{Compatibility of \gls{ncg} and ICG. Combining \gls{ncg} with ICG significantly improves FID, recall, and saturation scores while maintaining similar precision.}
        \label{table:icg-results}
        {
            \small
            \begin{tabular}{llccccccc}
                \toprule
                Guidance & FID $\downarrow$ & Precision $\uparrow$ & Recall $\uparrow$ & Saturation $\downarrow$  & Contrast $\downarrow$\\
                \midrule
                
                ICG & 17.63  & \textbf{0.85} & 0.32 & 0.49 & 0.28 \\
                +\mycc \gls{ncg} (Ours) & \mycc \textbf{5.73} & \mycc \textbf{0.85} & \mycc \textbf{0.63} & \mycc \textbf{0.33} & \mycc \textbf{0.22}\\
                \bottomrule
            \end{tabular}
        }
    \end{table}
}

\newcommand{\tabTSG}{
    \begin{table}[t!]
        \centering
        \caption{Compatibility of \gls{ncg} and TSG with. Combining \gls{ncg} with TSG improves FID, recall, and saturation scores while maintaining similar precision.}
        \label{table:tsg-results}
        {
            \small
            \begin{tabular}{lccccccc}
                \toprule
                Guidance & FID $\downarrow$ & Precision $\uparrow$ & Recall $\uparrow$ & Saturation $\downarrow$  & Contrast $\downarrow$\\
                \midrule
                
                 TSG & 14.00  & \textbf{0.81} & 0.52 & 0.37 & 0.28 \\
                +\mycc \gls{ncg} (Ours) & \mycc \textbf{5.84} & \mycc \textbf{0.81} & \mycc \textbf{0.66} & \mycc \textbf{0.30} & \mycc \textbf{0.20}\\
                \bottomrule
            \end{tabular}
        }
    \end{table}
}
\newcommand{\tabParameters}{
    \begin{table}[t]
        \begin{minipage}{\textwidth}
        \centering
            \caption{Hyperparameters used in the main experiment (\Cref{table:main-results}).}
            \label{tab:ncg-hp}
        {
            \small
        \begin{booktabs}{lcccc}
            \toprule
            Model & $\wcfg$ & $\eta$ & \(r\) &  $\beta$ \\
            \midrule
            EDM2-S & 4 & 0 & 2.5 & $\num{-0.75}$\\
            EDM2-XL & 2 & 0 & 2.5 & $\num{-0.75}$\\
             DiT-XL/2 &  4 & 0 & 5 & $\num{-0.50}$\\
             Stable Diffusion 2.1 & 10 & 0 & 7.5 & $\num{-0.75}$\\
             Stable Diffusion XL & 15 & 0 & 15 & $\num{-0.50}$\\
            \bottomrule
          \end{booktabs}
        }
        \end{minipage}
    \end{table}
}

\newcommand{\ncgGuidnacePlot}{
    \begin{figure}[t!]
        \centering
        \begin{minipage}{\textwidth}
            \resizebox{\textwidth}{!}{
                \begin{tikzpicture}
                    \begin{groupplot}[
                            group style={
                                    group size=4 by 1,
                                    horizontal sep=1.25cm,
                                },
                            xlabel={$w$},
                            ymajorgrids=true,
                            xmajorgrids=true,
                            grid style=dashed,
                            major grid style = {lightgray},
                            tick label style={font=\normalsize},
                            label style={font=\Large},
                            title style={font=\Large},
                            legend pos={north west}, legend cell align={left},
                            legend style={font=\large},
                            scale only axis,
                        ]
                        \nextgroupplot[title={FID}]
                        \addplot [C0, mark=*, very thick, mark options={solid}, dashed] table [x index=0, y index=1, col sep=comma] {data/edm_cfg.csv};
                        \addplot [C1, mark=*, very thick, mark options={solid}] table [x index=0, y index=1, col sep=comma] {data/edm_ncg.csv};
                        \nextgroupplot[title={Precision}]
                        \addplot [C0, mark=*, very thick, mark options={solid}, dashed] table [x index=0, y index=3, col sep=comma] {data/edm_cfg.csv};
                        \addplot [C1, mark=*, very thick, mark options={solid}] table [x index=0, y index=3, col sep=comma] {data/edm_ncg.csv};
                        \nextgroupplot[title={Recall}]
                        \addplot [C0, mark=*, very thick, mark options={solid}, dashed] table [x index=0, y index=4, col sep=comma] {data/edm_CFG.csv};
                        \addplot [C1, mark=*, very thick] table [x index=0, y index=4, col sep=comma] {data/edm_ncg.csv};
                        \nextgroupplot[title={Saturation}]
                        \addplot [C0, mark=*, very thick, mark options={solid}, dashed] table [x index=0, y index=5, col sep=comma] {data/edm_CFG.csv};
                        \addplot [C1, mark=*, very thick] table [x index=0, y index=5, col sep=comma] {data/edm_ncg.csv};
                        \legend{CFG, \gls{ncg} (Ours)}

                    \end{groupplot}
                \end{tikzpicture}
            }
        \end{minipage}
            
        \caption{Comparison between CFG and \gls{ncg} as the guidance scale increases. \gls{ncg} offers better FID and recall while maintaining similar or better precision to CFG at higher guidance scales.}
        \label{fig:ncg-guidance}
    \end{figure}
}

\newcommand{\ncgNFEPlot}{
    \begin{figure}[t!]
        \centering
        \begin{minipage}{\textwidth}
            \resizebox{\textwidth}{!}{
                \begin{tikzpicture}
                    \begin{groupplot}[
                            group style={
                                    group size=4 by 1,
                                    horizontal sep=1.25cm,
                                },
                            xlabel={\# Steps},
                            ymajorgrids=true,
                            xmajorgrids=true,
                            grid style=dashed,
                            major grid style = {lightgray},
                            tick label style={font=\normalsize},
                            label style={font=\huge},
                            title style={font=\huge},
                            legend pos={north east}, legend cell align={left},
                            legend style={font=\large},
                            scale only axis,
                        ]
                        \nextgroupplot[title={FID}]
                        \addplot [C0, mark=*, very thick, mark options={solid}, dashed] table [x index=0, y index=1, col sep=comma] {data/nfe_cfg.csv};
                        \addplot [C1, mark=*, very thick, mark options={solid}] table [x index=0, y index=1, col sep=comma] {data/nfe_ncg.csv};
                        \nextgroupplot[title={Precision}]
                        \addplot [C0, mark=*, very thick, mark options={solid}, dashed] table [x index=0, y index=3, col sep=comma] {data/nfe_cfg.csv};
                        \addplot [C1, mark=*, very thick, mark options={solid}] table [x index=0, y index=3, col sep=comma] {data/nfe_ncg.csv};
                        \nextgroupplot[title={Recall}]
                        \addplot [C0, mark=*, very thick, mark options={solid}, dashed] table [x index=0, y index=4, col sep=comma] {data/nfe_cfg.csv};
                        \addplot [C1, mark=*, very thick, mark options={solid}] table [x index=0, y index=4, col sep=comma] {data/nfe_ncg.csv};

                        \nextgroupplot[title={Saturation}]
                        \addplot [C0, mark=*, very thick, mark options={solid}, dashed] table [x index=0, y index=5, col sep=comma] {data/nfe_cfg.csv};
                        \addplot [C1, mark=*, very thick] table [x index=0, y index=5, col sep=comma] {data/nfe_ncg.csv};
                        \legend{CFG, \gls{ncg} (Ours)}

                    \end{groupplot}
                \end{tikzpicture}
            }
        \end{minipage}
            
        \caption{Comparison of CFG and \gls{ncg} across different numbers of sampling steps. \gls{ncg} consistently achieves better FID and recall while maintaining comparable or superior precision to CFG.}
        \label{fig:ncg-nfe}
    \end{figure}
}

\newcommand{\pixelDistPlot}{
    \begin{figure}[t!]
        \centering
        \begin{minipage}{\textwidth}
            \resizebox{\textwidth}{!}{
                \begin{tikzpicture}

                    \begin{groupplot}[
                        group style={
                                group size=4 by 1,
                                horizontal sep=1.25cm,
                            },
                            ylabel={Density},
                            ymajorgrids=true,
                            xmajorgrids=true,
                            grid style=dashed,
                            major grid style = {lightgray},
                            tick label style={font=\normalsize},
                            label style={font=\large},
                            title style={font=\large},
                            legend pos={north east}, 
                            legend cell align={left},
                            legend style={font=\normalsize, at={(0.5,0.975)},anchor=north},
                            scale only axis,
                            width=0.8\linewidth,
                            height=0.4\linewidth,
                    ]
                        \nextgroupplot[title={KDE of pixel values}]
                        \addplot[C0, mark=., very thick, mark options={solid}, dashed] table [col sep=space, x index=0, y index=1] {data/kde_cfg.txt};
                        \addplot[C1, mark=., very thick, mark options={solid}]  table [col sep=space, x index=0, y index=1] {data/kde_ncg.txt};
                        \addplot[C2, mark=., very thick, mark options={solid}, densely dashdotted]  table [col sep=space, x index=0, y index=1] {data/kde_real.txt};

                        \nextgroupplot[title={KDE of saturation values}]
                        \addplot[C0, mark=., very thick, mark options={solid}, dashed] table [col sep=space, x index=0, y index=1] {data/kde_saturation_cfg.txt};
                        \addlegendentry{CFG};
                        \addplot[C1, mark=., very thick, mark options={solid}]  table [col sep=space, x index=0, y index=1] {data/kde_saturation_ncg.txt};
                        \addlegendentry{\gls{ncg} (Ours)};
                        \addplot[C2, mark=., very thick, mark options={solid}, densely dashdotted]  table [col sep=space, x index=0, y index=1] {data/kde_saturation_real.txt};
                        \addlegendentry{Real data};

                    \end{groupplot}
                \end{tikzpicture}
            }
        \end{minipage}
            
        \caption{Kernel density estimates of pixel and saturation values for two sets of samples generated with CFG and \gls{ncg}. Compared to CFG, images generated with \gls{ncg} show less concentration around saturated pixels, indicated by the spikes at the extreme values in both plots.}
        \label{fig:pixel-dist}
        \vspace*{-0.25cm}
    \end{figure}
    
}

\newcommand{\figAppendixToy}{
    \begin{figure}[t]
        \centering
        \begin{subfigure}[b]{0.325\textwidth}
            \includegraphics[width=\textwidth]{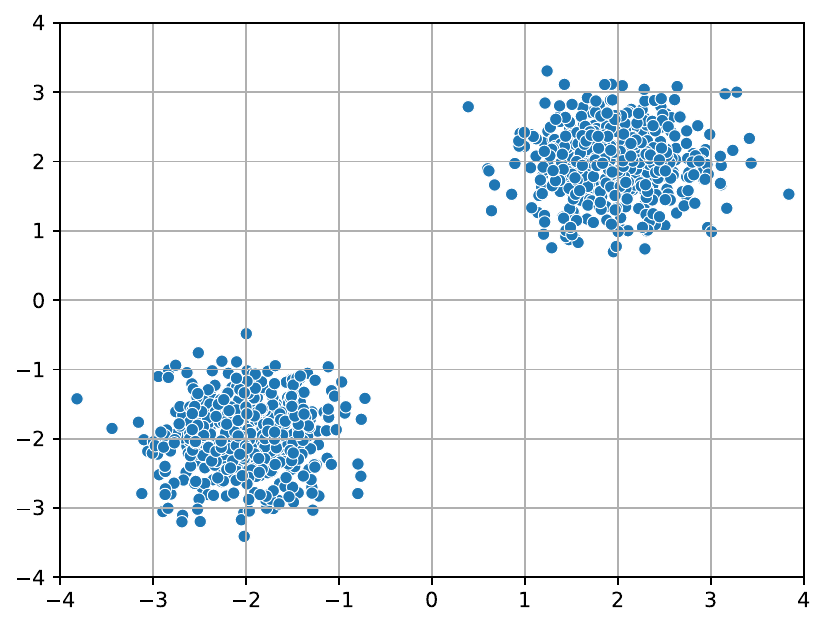}
            \caption{Real data}
            \label{fig:real-data}
        \end{subfigure}
        \hfill
        \begin{subfigure}[b]{0.325\textwidth}
            \includegraphics[width=\textwidth]{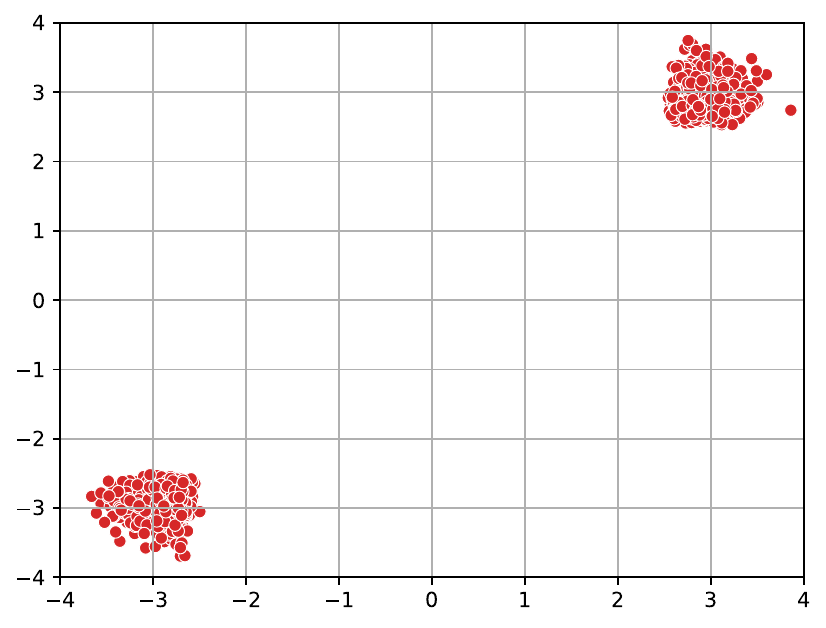}
            \caption{CFG}
            \label{fig:toy-ddpm}
        \end{subfigure}
        \hfill
        \begin{subfigure}[b]{0.325\textwidth}
            \includegraphics[width=\textwidth]{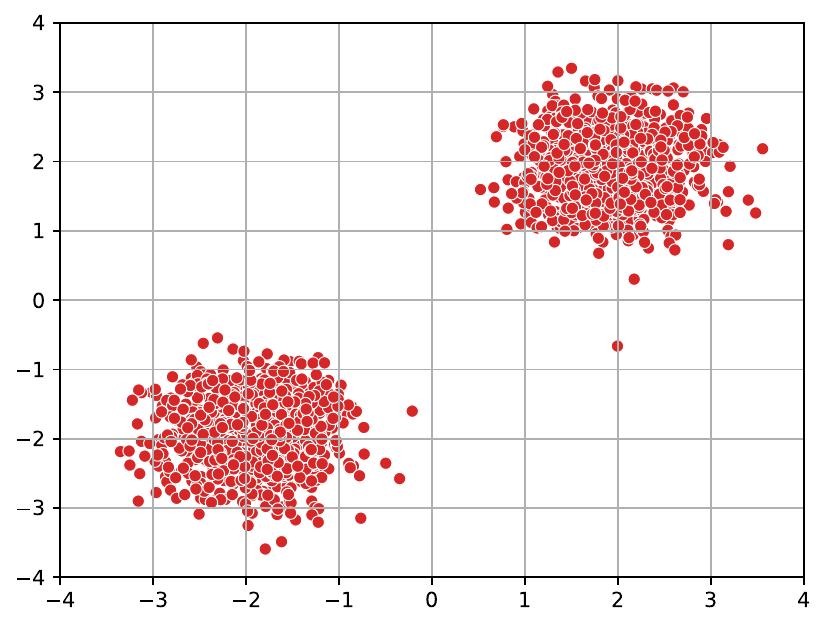}
            \caption{\gls{ncg} (Ours)}
            \label{fig:toy-ours}
        \end{subfigure}
        \caption{Visualizing the effect of \gls{ncg} on the sampling process using a toy problem. The real samples from the data distribution are shown in (a). When sampling with high guidance, CFG leads to a drift away from the true mean of the data distribution and results in reduced mode coverage in the generated samples (b). In contrast, sampling with \gls{ncg} eliminates the drift and increases the coverage of the distribution (c). We used the EDM sampler \citep{karras2022elucidating} for this experiment.}
        \label{fig:toy-example}

    \end{figure}
}
\newacronym{ldm}{LDM}{latent diffusion model}
\newacronym{vae}{VAE}{variational autoencoder}
\newacronym{sdvae}{SD-VAE}{Stable Diffusion VAE}
\newacronym{ncg}{APG}{adaptive projected guidance}

\begin{document}

\maketitle

\vspace{-0.45cm}
\begin{abstract}
Classifier-free guidance (CFG) is crucial for improving both generation quality and alignment between the input condition and final output in diffusion models. While a high guidance scale is generally required to enhance these aspects, it also causes oversaturation and unrealistic artifacts. In this paper, we revisit the CFG update rule and introduce modifications to address this issue. We first decompose the update term in CFG into parallel and orthogonal components with respect to the conditional model prediction and observe that the parallel component primarily causes oversaturation, while the orthogonal component enhances image quality. Accordingly, we propose down-weighting the parallel component to achieve high-quality generations without oversaturation. Additionally, we draw a connection between CFG and gradient ascent and introduce a new rescaling and momentum method for the CFG update rule based on this insight. Our approach, termed \gls{ncg}, retains the quality-boosting advantages of CFG while enabling the use of higher guidance scales without oversaturation. \gls{ncg} is easy to implement and introduces practically no additional computational overhead to the sampling process. Through extensive experiments, we demonstrate that \gls{ncg} is compatible with various conditional diffusion models and samplers, leading to improved FID, recall, and saturation scores while maintaining precision comparable to CFG, making our method a superior plug-and-play alternative to standard classifier-free guidance.\footnote{All visual results in the paper are best viewed in color and when zoomed in.}
\end{abstract}

\figTeaser
\section{Introduction}
Diffusion models \citep{sohl2015deep,hoDenoisingDiffusionProbabilistic2020,score-sde} are a class of generative models that learn the data distribution by reversing a forward process that adds noise to the data until the samples are indistinguishable from pure noise. Although the theory suggests that simulating the backward process in diffusion models should result in correct sampling from the data distribution, unguided sampling from diffusion models often results in low-quality images that do not align well with the input condition. Accordingly, classifier-free guidance \citep{hoClassifierFreeDiffusionGuidance2022} has been established as an essential tool in modern diffusion models for increasing the quality of generations and the alignment between the condition and the generated image, albeit at the cost of reduced diversity \citep{hoClassifierFreeDiffusionGuidance2022,sadat2024cads}.

Modern text-to-image models, such as Stable Diffusion \citep{rombachHighResolutionImageSynthesis2022}, generally require high guidance scales in order for the generations to have better quality and align well with the input prompt. However, high guidance scales often result in oversaturated colors and simplified image compositions \citep{saharia2022photorealistic, intervalGuidance}. \edit{Despite these disadvantages, high CFG scales are still used in practice due to their superior image quality compared to alternatives.}

In this paper, we analyze the update rule of CFG and show that with a few modifications to how the CFG update is applied at inference, we can vastly mitigate the oversaturation and artifacts of high guidance scales. First, we show that the CFG update rule can be decomposed into two components, one that is parallel to the conditional model prediction, and one that is orthogonal to this prediction. We show that the orthogonal element is mainly responsible for improving image quality, while the parallel part primarily adds contrast and saturation to the output. To the best of our knowledge, this is the first study that disentangles these two effects in CFG. 

Additionally, we establish a connection between the CFG update rule and stochastic gradient ascent. This insight leads us to explore a rescaled version of the CFG update direction and incorporate a momentum term, similar to adaptive optimization methods. The rescaling is motivated by the need to control large update norms, which can cause significant drifts in the sampling process. To prevent this, we constrain the updates to lie within a sphere. For the momentum term, unlike with traditional optimization, we apply a \emph{negative} value to introduce a repulsive effect between consecutive updates, effectively down-weighting components already present in previous steps. We refer to this as \emph{reverse momentum}. \edit{By combining rescaling, reverse momentum, and projection, we introduce a new method, called \acrfull{ncg}, which allows the use of higher guidance scales without oversaturation or degradation in image quality.}

Through extensive experiments with several diffusion models, such as EDM2 \citep{karras2023analyzing} and Stable Diffusion \citep{rombachHighResolutionImageSynthesis2022}, we demonstrate that \gls{ncg} can utilize high guidance scales without encountering oversaturation. As a result, we conclude that \gls{ncg} significantly expands the usable guidance range in practice and mitigates the harmful effects of CFG at high guidance scales. Our quantitative analysis shows that replacing CFG with \gls{ncg} improves FID, recall, and saturation scores while maintaining precision similar to CFG. Furthermore, when combined with Stable Diffusion 3 \citep{esser2024scaling}, \gls{ncg} enhances the consistency of text rendering in generated images. We also demonstrate that \gls{ncg} is compatible with distilled models that use fewer sampling steps, such as SDXL-Lightning \citep{lin2024sdxl}. A representative visual comparison between CFG and \gls{ncg} is shown in \Cref{fig:teaser}.
\section{Related work}

Score-based diffusion models \citep{DBLP:conf/nips/SongE19,score-sde,sohl2015deep,hoDenoisingDiffusionProbabilistic2020} learn data distributions by reversing a forward diffusion process that gradually corrupts data into Gaussian noise. These models have rapidly outperformed previous generative modeling methods in terms of fidelity and diversity \citep{nichol2021improved,dhariwalDiffusionModelsBeat2021}, setting new benchmarks across various domains. They have achieved state-of-the-art results in unconditional image generation \citep{dhariwalDiffusionModelsBeat2021,karras2022elucidating}, text-to-image generation \citep{dalle2,saharia2022photorealistic,balaji2022ediffi,rombachHighResolutionImageSynthesis2022,sdxl,yu2022scaling}, video generation \citep{blattmann2023align,stableVideoDiffusion,gupta2023photorealistic}, image-to-image translation \citep{saharia2022palette,liu20232i2sb}, and audio generation \citep{WaveGrad,DiffWave,huang2023noise2music}.

Since the introduction of the DDPM model \citep{hoDenoisingDiffusionProbabilistic2020}, numerous advancements have been made, such as improved network architectures \citep{hoogeboom2023simple,karras2023analyzing,peeblesScalableDiffusionModels2022,dhariwalDiffusionModelsBeat2021}, enhanced sampling algorithms \citep{songDenoisingDiffusionImplicit2022,karras2022elucidating,plms,dpm_solver,salimansProgressiveDistillationFast2022}, and new training techniques \citep{nichol2021improved,karras2022elucidating,score-sde,salimansProgressiveDistillationFast2022,rombachHighResolutionImageSynthesis2022}. Despite these advancements, diffusion guidance, including both classifier and classifier-free guidance \citep{dhariwalDiffusionModelsBeat2021,hoClassifierFreeDiffusionGuidance2022}, remains crucial in enhancing generation quality and improving alignment between the condition and the output image \citep{glide}, albeit at the cost of reduced diversity and oversaturated outputs.

A recent line of work, such as CADS \citep{sadat2024cads} and interval guidance (IG) \citep{intervalGuidance}, has focused on enhancing the diversity of generations at higher guidance scales. In contrast, our proposed method, \gls{ncg}, specifically addresses the oversaturation issue in CFG, as these diversity-boosting methods still struggle with oversaturation at higher guidance scales. \edit{In \Cref{sec:cads-ig}, we demonstrate that \gls{ncg} can be combined with CADS and IG to achieve diverse generations without encountering oversaturation problems.}

Dynamic thresholding \citep{saharia2022photorealistic} was introduced to mitigate the saturation effect in CFG, but it is not directly applicable to latent diffusion models (since it assumes pixel values are between $[-1, 1]$) and tends to produce images lacking in detail. Another approach, CFG Rescale \citep{zerosnrdiff}, aims to reduce overexposure in generated images by rescaling the standard deviation of the predictions after applying CFG. However, we demonstrate that our method is noticeably more effective at reducing oversaturation compared to CFG Rescale.

Orthogonal projection has been explored in the context of text-to-3D generation \citep{armandpour2023re} and non-linear guidance \citep{zheng2023characteristic}, but none of these methods tackle the saturation issue at higher guidance scales. We also demonstrate that naive projection has minimal impact on CFG behavior, as it must be applied to the denoised predictions to be effective. Additionally, we incorporate rescaling and reverse momentum to further mitigate the adverse effects of CFG at higher guidance scales. We show that \gls{ncg} can be applied to various conditional diffusion models while adding practically no overhead to the sampling process.

\section{Background}\label{sec:background}

We provide a brief overview of diffusion models in this section. Let \(\vx \sim \pdata(\vx)\) represent a data point, and let \(\vz_t = \vx + \sigma(t) \mepsilon\) describe a forward process of the diffusion model that introduces noise to the data, where \(t \in [0,1]\) is the time step. Here, \(\sigma(t)\) is the noise schedule, which determines the amount of information destroyed at each time step \(t\), with \(\sigma(0) = 0\) and \(\sigma(1) = \sigma_{\textnormal{max}}\). \citet{karras2022elucidating} demonstrated that this forward process is equivalent to the following ordinary differential equation (ODE):
\begin{equation}\label{eq:diffusion-ode}
    \dd\vz_t = - \dot{\sigma}(t) \sigma(t) \grad_{\vz_t} \log p_t(\vz_t) \dd t,
\end{equation}
where \(p_t(\vz_t)\) denotes the time-dependent distribution of noisy samples, with \(p_0 = \pdata\) and \(p_1 = \mathcal{N}(\mathbf{0}, \sigma_{\textnormal{max}}^2 \mathbf{I})\). With access to the time-dependent score function \(\nabla_{\vz_t} \log p_t(\vz_t)\), one can sample from the data distribution \(\pdata\) by solving the ODE backward in time (from \(t = 1\) to \(t = 0\)). The unknown score function \(\nabla_{\vz_t} \log p_t(\vz_t)\) is estimated using a neural denoiser \(D_{\mtheta}(\vz_t, t)\), which is trained to predict the clean samples \(\vx\) from the corresponding noisy samples \(\vz_t\). This framework also allows for conditional generation by training a denoiser \(D_{\mtheta}(\vz_t, t, \vy)\) that incorporates additional input signals \(\vy\), such as class labels or text prompts.

\paragraph{\textbf{Classifier-free guidance} (CFG)} CFG is an inference method designed to enhance the quality of generated outputs by combining the predictions of a conditional model and an unconditional model \citep{hoClassifierFreeDiffusionGuidance2022}. Given a null condition \(\vy_{\textnormal{null}} = \varnothing\) for the unconditional case, CFG modifies the denoiser's output at each sampling step as follows:
\begin{equation}\label{eq:cfg}
    \predcfg = \prednull + \wcfg \prn{\pred - \prednull},
\end{equation}
where \(\wcfg = 1\) represents the non-guided case. The unconditional model \(\prednull\) is trained by randomly applying the null condition \(\vy_{\textnormal{null}} = \varnothing\) to the denoiser's input for a portion of training. Alternatively, a separate denoiser can be trained to estimate the unconditional score in \Cref{eq:cfg} \citep{karras2023analyzing}. Similar to the truncation method used in GANs \citep{brockLargeScaleGAN2019}, CFG improves the quality of images but reduces diversity \citep{pml2Book}.

\section{\Acrlong{ncg}}\label{sec:methodology}
We now present our method for addressing oversaturation and artifacts in CFG at high guidance scales. 
Let $\dpred = \pred - \prednull$ be the CFG update direction at time step $t$. Note that \Cref{eq:cfg} can now be written as 
\begin{align}
    \predcfg
    &= \pred + (\wcfg - 1) \dpred. \label{eq:cfgnew}
\end{align}
(See \Cref{sec:cfg-detail} for the derivation.) We use \Cref{eq:cfgnew} for the rest of this paper to motivate our changes. \gls{ncg} has three elements: (1) projection, (2) rescaling, and (3) reverse momentum. We discuss each component below.

\paragraph{Orthogonal projection} First, note that we can decompose \(\dpred\) into two different components: \(\dpredpar\), which is parallel to \(\pred\), and \(\dpredorth\), which is orthogonal to \(\pred\), i.e., \(\dpred = \dpredorth + \dpredpar\). We can compute \(\dpredpar\) via orthogonal projection, with 
\begin{equation}
    \dpredpar = \frac{\inner{\dpred}{\pred}}{\inner{\pred}{\pred}} \pred.
\end{equation}
We then have \(\dpredorth = \dpred -  \dpredpar\). We observe that the orthogonal component is chiefly responsible for improvements in image quality, while the parallel component increases saturation in the generations as shown in \Cref{fig:method-visual-projection}. 

Accordingly, we modify the update direction to form \(\dpred(\eta) = \dpredorth + \eta \dpredpar\), where \(\eta \leq 1\) is a hyperparameter. Note that $\dpred(1)$ is identical to the unmodified CFG update direction described above. We show that reducing the strength of the parallel component (i.e.\ setting $\eta$ close to zero) significantly reduces saturation and results in more realistic generations at higher guidance scales.

The intuition behind the saturating effect of the parallel component is helped by thinking of the output $\pred$ as an image with a typical range of values.\footnote{This intuition also holds for the image-like representations in latent diffusion models.} When an update parallel to this image is added, it serves to create a ``gain,'' pushing the values toward the extremes of their range. This gain effect can be seen by direct calculation (assuming $\pred$ and $\dpredpar$ have the same direction): 
\begin{equation}
    \pred + (\wcfg - 1) \dpredpar = \brk{1 + (\wcfg-1) \frac{\norm{\dpredpar}}{\norm{\pred}}} \pred,
\end{equation}
where we note that the term in brackets on the right-hand side is greater than one for $\wcfg > 1$.
Thus, this term only adds saturation to the predictions \(\pred\) during each inference step, much like multiplying pixel values by a number greater than one. We show in \Cref{sec:ablation} that reducing \(\eta\) and leaning more heavily on the orthogonal component significantly attenuates this saturation side effect in generations while maintaining the quality-boosting benefits of CFG.

\paragraph{Adding rescaling} Next, we argue that the CFG update rule in \Cref{eq:cfgnew} can be interpreted as one step of gradient ascent on the $\ell_2$ distance between the conditional and unconditional prediction, i.e., one step of gradient ascent on $\frac{1}{2} \norm{\pred - \prednull}^2$ with a learning rate of $\wcfg - 1$. (See \Cref{sec:cfg-detail} for proof.) Inspired by this interpretation and normalized gradient ascent, we rescale the CFG update rule at each step to regulate the impact of each update. Specifically, we constrain $\dpred$ to be inside a sphere with radius $r$ via
\begin{equation}
    \dpred \leftarrow \dpred \cdot \min\prn{1, \frac{r}{\norm{\dpred}}},
\end{equation}
where $r$ is a hyperparameter. This rescaling ensures that the CFG update \(\dpred\) stays closer to \(\pred\), limiting drift at each sampling step if $\norm{\dpred}$ is large. As demonstrated in \Cref{sec:ablation}, this adjustment improves both FID and recall.

\paragraph{Adding reverse momentum}  Finally, leveraging the connection to gradient ascent, we introduce a reverse momentum term to the CFG update rule. We define the momentum for the CFG update direction as \(\overline{\dpred} \leftarrow \dpred + \beta \overline{\dpred}\), where \(\overline{\dpred} = 0\) initially. The momentum term accounts for the average values of past updates; however, unlike standard optimization methods, we use a \emph{negative} momentum strength \(\beta < 0\). Intuitively, this pushes the model away from previous CFG update directions and encourages the model to focus more on the current update direction. As shown in \Cref{sec:ablation}, incorporating reverse momentum further enhances image quality (i.e., lower FID scores).

\gls{ncg} is easy to implement, and we provide the source code in \Cref{fig:imp-ncg} (appendix). As shown in \Cref{sec:ablation}, it is crucial to convert the diffusion model's outputs (e.g., predicted noise) into the denoised prediction $\pred$ in order to perform the projection. Further details on obtaining $\pred$ for common prediction types are discussed in \Cref{sec:denoised-pred}. \Cref{fig:method-visual} demonstrates that using \gls{ncg} instead of CFG produces high-quality generations without oversaturation or the undesirable artifacts associated with high guidance scales.
\figMethod
\section{Experiments and results}\label{sec:results}

\paragraph{Setup} We mainly experiment with text-to-image generation with Stable Diffusion \citep{rombachHighResolutionImageSynthesis2022} and class-conditional ImageNet \citep{imagenet} generation using EDM2 \citep{karras2023analyzing} and DiT-XL/2 \citep{peeblesScalableDiffusionModels2022}. For all experiments, we use the default diffusion sampler from each model (e.g., Euler scheduler for Stable Diffusion XL) along with pretrained checkpoints and corresponding codebases to ensure consistency in weights and the sampling process with the original frameworks.

\paragraph{Distribution metrics} We use Fréchet Inception Distance (FID) \citep{fid} as our primary metric for evaluating both the quality and diversity of generated images due to its alignment with human judgment. Since FID is sensitive to small implementation details, we ensure that all models are evaluated under the same setup. For completeness, we also report precision \citep{improvedPR} as an additional quality metric and recall \citep{improvedPR} as a diversity metric.

\paragraph{Color metrics} While FID measures the overall quality of generated images, we introduce specific metrics to directly assess saturation and contrast. To measure saturation, we convert each image from RGB to HSV and compute the mean of the saturation channel. We define contrast (also known as RMS contrast) as the standard deviation of pixel values after converting the image to grayscale. The final metrics are derived by averaging the saturation and contrast values across all generated images.

\subsection{Main results}
\figEDMMain
\paragraph{Qualitative results}
\Cref{fig:edm-main,fig:sd-main} present our qualitative results comparing \gls{ncg} with CFG for EDM2 and Stable Diffusion XL. We observe that, compared to CFG, \gls{ncg} generates more realistic images with noticeably lower saturation. Furthermore, \gls{ncg} appears to produce fewer artifacts in the final outputs, as illustrated in \Cref{fig:artifacts}. Additional visual results can be found in \Cref{sec:more-visual-results}.

\figSDArtifacts

\paragraph{Quantitative results} We next present a quantitative comparison between \gls{ncg} and CFG in \Cref{table:main-results}. The table shows that \gls{ncg} outperforms CFG across multiple models, consistently achieving better FID and recall scores, as well as lower saturation and contrast. Moreover, \gls{ncg} demonstrates similar precision to CFG, indicating that the reduction in saturation does not compromise the quality of individual samples.
\tabResultMain

\pixelDistPlot

\paragraph{Distribution of pixel values} {\Cref{fig:pixel-dist} presents the kernel density estimate (KDE) plot of RGB and saturation values for 100 images generated using CFG and \gls{ncg}, along with KDE plots for 100 real samples drawn from the evaluation subset of ImageNet. Compared to CFG, \gls{ncg} plots are more broadly distributed across the spectrum with less concentration at the extremes. This indicates that images generated with \gls{ncg} are closer to real data in terms of saturation and color composition.}

\paragraph{\acrshort{ncg} vs guidance scale}
{In \Cref{fig:ncg-guidance}, we demonstrate that as the guidance scale increases, \gls{ncg} consistently achieves lower FID and higher recall while maintaining similar or better precision compared to CFG. Additionally, CFG exhibits increasing saturation at higher guidance scales, whereas \gls{ncg} maintains a relatively constant saturation level. Therefore, \gls{ncg} allows the usage of higher guidance scales, achieving better FID and diversity without oversaturation.}
\ncgGuidnacePlot
\figDistilledMain

\paragraph{Improving diversity} While \gls{ncg} is designed to address oversaturation at high guidance scales, we also observed that it can enhance the diversity of generations. As shown in \Cref{table:main-results} and \Cref{fig:ncg-guidance}, \gls{ncg} improves distribution coverage (i.e., higher recall) while maintaining precision comparable to CFG. Additional qualitative results illustrating the enhanced diversity are provided in \Cref{fig:diversity} (appendix).

\paragraph{Using \acrshort{ncg} with distilled models}
{A common issue with CFG is that it degrades the quality of final outputs when applied to distilled models with fewer sampling steps (e.g., 8-step SDXL-Lightning \citep{sdlightning}). In this section, we show that \gls{ncg} does not encounter this problem and can be effectively applied to distilled models. \Cref{fig:distilled} demonstrates that replacing CFG with \gls{ncg} significantly improves generation quality. Extended results with additional models are provided in \Cref{sec:distilled-appendix}, along with more visual examples in \Cref{sec:more-visual-results}.}

\paragraph{Text spelling with Stable Diffusion 3}
Next, we demonstrate that integrating \gls{ncg} with Stable Diffusion 3 \citep{esser2024scaling} enhances the consistency of text rendering in generated images. As shown in \Cref{fig:sd3-text}, \gls{ncg} produces more accurate spelling in the generated images compared to standard CFG. \edit{More visual results are given in \Cref{fig:sd3-appendix}} (appendix).

\figSDthreeMain
\figRescale

\paragraph{Comparison with CFG Rescale}
CFG Rescale was introduced in \citep{zerosnrdiff} as a method to reduce saturation at high guidance scales. In this section, we demonstrate that \gls{ncg} is more effective than CFG Rescale. The comparison in \Cref{fig:cfg-rescale} shows that \gls{ncg} outputs have significantly less saturation and are more realistic than those produced with CFG Rescale.

\paragraph{Computational cost of \gls{ncg}} The computational cost of \gls{ncg} is practically identical to that of CFG, as the rescaling and projection steps incur negligible overhead compared to querying the denoiser. Specifically, in the case of Stable Diffusion XL, the forward pass through the diffusion network takes approximately 130 milliseconds on an RTX 3090 GPU for a single image, while the guidance step requires only about 0.45 milliseconds.

\subsection{Ablation studies}\label{sec:ablation}
\begin{wraptable}[7]{r}{0.375\textwidth}
    \vspace{-0.475cm}
    \begin{adjustbox}{valign=t}
        \begin{minipage}{\linewidth}
            \captionsetup{skip=5pt}
            \caption{Importance of different components in \gls{ncg}.}
              \centering
                \resizebox{\linewidth}{!}{
                  \large
                    \begin{booktabs}{
                        colspec = {Q[l, m]Q[c, m]Q[c, m]Q[c, m]}
                    }
                    \toprule
                    Config & FID $\downarrow$ & Recall $\uparrow$ & Saturation $\downarrow$\\
                    \midrule
                    \gls{ncg} ($\wcfg=4$) & \textbf{6.49} &  \textbf{0.62} & \textbf{0.33}\\
                    w/o projection & 6.63 & 0.60 & 0.37   \\
                    w/o rescaling & 7.93 & 0.56 & 0.34  \\
                    w/o momentum & {6.85} & 0.61 & \textbf{0.33} \\
                    \bottomrule
                    \end{booktabs}
                }
              \label{tab:ablation-main}
            \end{minipage}%
    \end{adjustbox}
\end{wraptable} 
We now present our ablation studies in this section. The experiments are based on class-conditional generation using the EDM2 model \citep{karras2023analyzing}, with FID as the primary metric to justify our design choices. First, \Cref{tab:ablation-main} highlights the importance of each component in \gls{ncg}. We observe that removing projection, rescaling, or reverse momentum results in higher FID scores. Additionally, note that the projection component is primarily responsible for reducing saturation while rescaling and reverse momentum mainly improve FID and recall. \Cref{sec:extended-ablation} gives extended ablation results on the effect of each component in \gls{ncg}.

\paragraph{Importance of the model prediction type} 
While CFG works the same across all model prediction types, we observed that our method performs best when applied to the denoised predictions $\pred$, rather than, for example, the noise prediction $\prednoise$. This is illustrated in \Cref{fig:projection-ablation}, where projecting onto $\prednoise$ produces results nearly identical to CFG, while projecting onto $\pred$ significantly reduces saturation. Note that as discussed in \Cref{sec:denoised-pred}, this is not a bottleneck for \gls{ncg} as various prediction types can be readily converted to $\pred$ at each step.

\figAblationProjection
\section{Conclusion and discussion}
In this work, we investigated the oversaturation effect of high CFG scales and introduced a new method, \acrfull{ncg}, that achieves the same quality-boosting benefits as CFG without causing oversaturation. The key idea behind \gls{ncg} is to project the CFG update onto the denoised prediction of the diffusion model \(\pred\) and remove or down-weight the component parallel to that prediction. Additionally, by linking CFG to gradient ascent, we demonstrated that its performance can be further enhanced by incorporating rescaling and reverse momentum. Through extensive experiments, we showed that \gls{ncg} improves FID, recall, and saturation metrics compared to CFG, while maintaining similar or better precision. Thus, \gls{ncg} offers a plug-and-play alternative to standard CFG capable of delivering superior results with practically no additional computational overhead. Like CFG, challenges remain in accelerating \gls{ncg} so that the sampling cost approaches that of the unguided sampling (i.e., removing the need to query the diffusion network twice at each sampling step). We consider this a promising direction for future research. 

\clearpage

\subsubsection*{Ethics Statement}\label{sec:impact-statement}
As generative modeling continues to evolve, the potential to create fake or erroneous data increases. While advancements in AI-generated content can enhance efficiency and foster creativity, it is crucial to address the associated ethical concerns. For a more detailed discussion on ethics and creativity in computer vision, we recommend \cite{rostamzadeh2021ethics}.

\subsubsection*{Reproducibility Statement}\label{sec:impact-statement}
This work builds on the official implementations of the pretrained models referenced in the main text. The source code for implementing \gls{ncg} is provided in \Cref{fig:imp-ncg}, and \Cref{sec:imp-detail} outlines additional implementation details, including the hyperparameters used in the main experiments.

\bibliography{ref}
\bibliographystyle{iclr2025_conference}

\clearpage
\appendix

\section{Details on CFG as gradient ascent} \label{sec:cfg-detail}
In this section, we discuss how CFG can be interpreted as one step of gradient ascent. To begin, note that the CFG update rule can be expressed as:
\begin{align}
    \predcfg &= \prednull + \wcfg (\pred - \prednull) \\
    &= \wcfg \pred + (1 - \wcfg) \prednull \\
    &= \pred + (\wcfg - 1)\pred + (1 - \wcfg) \prednull\\
    &= \pred + (\wcfg - 1)\prn{\pred - \prednull}\\
    &= \pred + \gcfg{\dpred},
\end{align}
where \(\gcfg = \wcfg - 1\), and $\dpred = \pred - \prednull$. Next, observe that we can write:
\begin{equation}
    \pred - \prednull = \grad_{\pred} \brk{\frac{1}{2} \norm{\pred - \prednull}^2}.
\end{equation}
Thus, if we define the CFG objective function as 
\begin{equation}
    f_{\mathrm{CFG}}(\pred, \prednull) = \frac{1}{2} \norm{\pred - \prednull}^2,
\end{equation}
the CFG update rule becomes equivalent to:
\begin{equation}
    \predcfg = \pred + \gcfg \grad_{\pred} f_{\mathrm{CFG}}(\pred, \prednull).
\end{equation}
Hence, we have shown that the CFG update rule corresponds to a single step of gradient \emph{ascent} with respect to the objective function \(f_{\mathrm{CFG}}(\pred, \prednull)\). 

This interpretation motivated us to incorporate rescaling into standard CFG. Since the objective function \(f_{\mathrm{CFG}}(\pred, \prednull)\) does not have a maximum, the CFG update step may result in arbitrary drift from \(\pred\). By applying rescaling, we constrain the CFG update to remain within a ball of limited radius around \(\pred\). The reverse momentum method is similarly inspired by this interpretation, where each update is pushed away from previous predictions.

\section{Denoised prediction for different diffusion models}\label{sec:denoised-pred}
We next briefly outline the process of computing the denoised prediction \(\pred\) for various diffusion models. For further details, we refer readers to \citet{kingma2023vdm}. In the following, let \(\vx\) represent the clean data, \(\vy\) a condition or class, and \(\mepsilon \sim \normal{\zero}{\mI}\) the noise. Given a noisy sample \(\vz_t\) at time step \(t\), the objective is to recover the clean data \(\vx\) that produced \(\vz_t\). The denoised version of \(\vz_t\), which approximates \(\vx\), is estimated by a neural network, denoted as \(\pred\). Before applying \gls{ncg}, we always convert all model predictions to \(\pred\). This conversion is compatible with most samplers based on the denoising framework, such as EDM \citep{karras2022elucidating} and DPM++ \citep{lu2022dpm}. The conversions for various models are derived below, and a summary is provided in \Cref{tab:x0-formula}.

\paragraph{DDPM} For models using the DDPM framework \citep{hoDenoisingDiffusionProbabilistic2020}, the forward diffusion process is defined as \(\vz_t = \alpha_t \vx + \sigma_t \mepsilon\), where $\sigma_t^2 + \alpha_t^2 = 1$. These models typically predict the total added noise $\mepsilon$ via a neural  network \(\prednoise\). Given the prediction of the model, the denoised prediction can be estimated via
\begin{equation}
    \pred = \frac{\vz_t - \sigma_t \prednoise}{\alpha_t}.
\end{equation}
If the model predicts the velocity $\vv = \alpha_t \mepsilon - \sigma_t \vx$, we have
 \begin{equation}
    \vv = \alpha_t \frac{\vz_t - \alpha_t \vx}{\sigma_t} - \sigma_t \vx = \frac{\alpha_t \vz_t - \alpha_t^2 \vx - \sigma_t^2 \vx}{\sigma_t} =\frac{\alpha_t \vz_t - \vx}{\sigma_t}.
 \end{equation}
 This leads to the following formulation for the denoised prediction:
 \begin{equation}
    \pred = \alpha_t \vz_t - \sigma_t \predv.
 \end{equation}

\paragraph{EDM framework} For the EDM framework \citep{karras2022elucidating}, the forward process is described by $\vz_t = \vx + \sigma(t) \mepsilon$, and the denoised prediction $\pred$ is formulated via
\begin{equation}
    \pred = c_{\mathrm{skip}}(t) \vz_t + c_{\mathrm{out}}(t) F_{\mtheta} (c_{\mathrm{in}}(t) \vz_t, c_{\mathrm{noise}}(t), \vy),
\end{equation}
where $F_{\mtheta} (c_{\mathrm{in}}(t) \vz_t, c_{\mathrm{noise}}(t), \vy)$ is the output of the neural network. The EDM framework uses $\sigma(t) \propto t$; thus, $\sigma$ and $t$ can be used interchangeably in this framework.

\paragraph{Rectified flow models} For rectified flow models \citep{liu2022flow}, such as Stable Diffusion 3 \citep{esser2024scaling}, the forward process is given by $\vz_t = (1 - t) \vx + t \mepsilon$. The model predicts the velocity field given by $\vv = \mepsilon - \vx$. Accordingly, we have
\begin{align}
    \vv &= \mepsilon - \vx 
    = \frac{\vz_t - (1 - t)\vx}{t} - \vx
    = \frac{\vz_t - (1 - t)\vx - t\vx}{t}
    = \frac{\vz_t - \vx}{t}.
\end{align}
Thus, the denoised prediction can be determined by:
\begin{equation}
    \pred = \vz_t - t \predv = \vz_t - \sigma_t \predv,
\end{equation}
where we define $\sigma_t = t$.

\tabPredicitonFormulas

This section demonstrates the effect of applying \gls{ncg} on a toy example to illustrate the differences between \gls{ncg} and CFG. We use a mixture of two high-dimensional Gaussians as the data distribution, which allows us to analytically compute the score functions during the diffusion process, eliminating potential errors introduced by a denoiser network. Specifically, the data distribution \(\pdata\) is defined as:
\begin{equation}
    \pdata(\vx) = \frac{1}{2} \normal{\bmmu_1}{\sigma^2\mI}(\vx) + \frac{1}{2} \normal{\bmmu_2}{\sigma^2\mI}(\vx),
\end{equation}
where \(\bmmu_1=[-2, -2, \dots, -2]\), \(\bmmu_2=[2, 2, \dots, 2]\), and $\sigma=0.25$. We use a dimensionality of 500 for each component. Accordingly, the conditional distributions are equal to
\begin{equation}
    \pdata(\vx \cond y = 1) = \normal{\bmmu_1}{\sigma^2\mI}(\vx) \qquad \mathrm{and} \qquad \pdata(\vx \cond y = 2) = \normal{\bmmu_2}{\sigma^2\mI}(\vx).
\end{equation}
The sampling results are shown in \Cref{fig:toy-example} (visualizing the first two dimensions of each Gaussian). When CFG is applied with a high guidance scale, it results in a drift toward regions less likely according to the data distribution. In contrast, applying \gls{ncg} corrects this drift and improves mode coverage. While this is a simplified example, we argue that a similar phenomenon occurs when applying CFG to images, leading to artifacts and oversaturation in the final outputs.
\figAppendixToy

\section{Additional experiments}
Additional experiments and ablation studies are included in this section. Unless stated otherwise, the experiments are conducted using class-conditional ImageNet \citep{imagenet} generation.
\subsection{Compatibility with CADS and IG}\label{sec:cads-ig}
We first demonstrate that \gls{ncg} is compatible with CADS \citep{sadat2024cads} and interval guidance (IG) \citep{intervalGuidance}, both of which are designed to enhance the diversity of generations at high guidance scales. The results, shown in \Cref{table:ig-cads}, indicate that replacing CFG with \gls{ncg} leads to improved FID, recall, and saturation scores for both methods.
\tabIGCads

\subsection{\gls{ncg} vs number of sampling steps}
We now present the performance comparison between \gls{ncg} and CFG across different numbers of sampling steps using the EDM2 model. \Cref{fig:ncg-nfe} indicates that \gls{ncg} consistently provides better FID, recall, and saturation while maintaining the same level of precision.
\ncgNFEPlot

\subsection{Using \gls{ncg} with distilled models}\label{sec:distilled-appendix}
In this section, we show the compatibility of \gls{ncg} with distilled models using  \textsc{PixArt}-$\delta$ \citep{chen2024pixart}, SDXL-Lightning \citep{sdlightning}, and SDXL-Flash \citep{sdxl-flash}. Consistent with the main text, \Cref{fig:distilled-extended} demonstrates that replacing CFG with \gls{ncg} significantly improves generation quality and saturation level across all models. This is also consistent with \Cref{fig:ncg-nfe}, where \gls{ncg} outperforms CFG at fewer sampling steps (e.g., 8-16).
\figDistilled

\subsection{Compatibility with different samplers}
While the main experiment in \Cref{table:main-results} used the default sampler of each model, we next separately show that \gls{ncg} is compatible with different sampling algorithms widely used with diffusion models. As shown in \Cref{table:samplers}, using \gls{ncg} with different samplers results in improved FID, recall, and saturation scores, consistent with the main findings in \Cref{table:main-results}. We used class-conditional generation using DiT-XL/2 for this experiment.
\tabSamplers

\subsection{Compatibility with ICG}
Independent condition guidance (ICG) \citep{sadat2024no} is a method to apply CFG without the need to query an unconditional model. In \Cref{table:icg-results}, we show that \gls{ncg} is compatible with ICG, and similar to CFG, using \gls{ncg} with ICG results in improved FID, recall, and saturation scores while maintaining similar precision. We use class-conditional ImageNet generation with EDM2-S for this experiment.
\tabICG

\subsection{Compatibility with TSG}
Time-step guidance (TSG) \citep{sadat2024no} is an extension of CFG that leverages the time-step information learned by the diffusion model to enhance the quality of generations. We next demonstrate that applying the update rule in \gls{ncg} further improves the performance of TSG. \Cref{table:tsg-results} shows that \gls{ncg} improves FID, recall, and saturation metrics, while maintaining similar precision to TSG. This experiment is based on class-conditional ImageNet generation using DiT-XL/2.
\tabTSG

\subsection{Alignment with the condition}
\begin{wraptable}[7]{r}{0.35\textwidth}
    \vspace{-0.35cm}
    \begin{adjustbox}{valign=t}
        \begin{minipage}{\linewidth}
            \captionsetup{skip=5pt}
            \caption{Condition alignment comparison between CFG and \gls{ncg}.}
              \centering
                \resizebox{\linewidth}{!}{
                \begin{booktabs}{colspec = {Q[l, 3.4cm]Q[c, 1cm]Q[c, 1cm]}}
                \toprule
                Alignment metric & CFG & \gls{ncg} \\ 
                \midrule
                Class Accuracy $\uparrow$ & {0.97} & {0.96} \\
                CLIP-Score $\uparrow$ & 0.31 & {0.31}\\
                \bottomrule
              \end{booktabs}
                }
              \label{table:alignment}
            \end{minipage}%
    \end{adjustbox}
\end{wraptable} 
We next demonstrate that replacing CFG with \gls{ncg} does not compromise the alignment between the input condition and the output. To validate this, we measure the classification accuracy of the generated results for the ImageNet task and the CLIP score for Stable Diffusion. The results in \Cref{table:alignment} show that both CFG and \gls{ncg} achieve comparable alignment metrics. Thus, \gls{ncg} reduces saturation and improves FID without compromising condition alignment.

\subsection{Extended ablation studies}\label{sec:extended-ablation}
\paragraph{Effect of the parallel component}
We next demonstrate the effect of $\eta$ on the generated images in \Cref{table:ablation-eta}. As hypothesized in \Cref{sec:methodology}, increasing the strength of the parallel component leads to higher saturation levels and increased FID. We recommend setting $\eta=0$ by default and only increasing it if more saturation is desired in the generated images.

\paragraph{Effect of the rescaling threshold}
The effect of the rescaling radius $r$ on the generated images is shown in \Cref{table:ablation-r}. Excessive rescaling degrades image quality, while high values of $r$ result in no noticeable change, as the rescaling function approaches the identity function. Therefore, midrange values for $r$ yield better FID scores. We suggest observing the norm of $\dpred$ during the inference process and choosing $r$ in a way that is comparable (on average) to the norm of $\dpred$.

\paragraph{Effect of the momentum strength}
\Cref{table:ablation-beta} shows the effect of momentum strength \(\beta\) on generation quality. Note
 Negative values for \(\beta\) result in better FID compared to positive momentum, and excessive momentum degrades image quality. This aligns with our hypothesis that moving away from the previous directions helps limit the drift that can occur during sampling with higher guidance scales. Empirically, we found that \(\beta \in [-0.75, -0.25]\) works well in most setups.
 \tabAblationHPs

\section{Implementation details}\label{sec:imp-detail}
We provide the code for \gls{ncg} in \Cref{fig:imp-ncg}. Compared to CFG, \gls{ncg} only includes a few additional lines of code without noticeable computational overhead. As discussed in \Cref{sec:methodology}, we always convert the predictions of the diffusion model to $\pred$, compute the guided prediction, and convert it back to the initial output type at each sampling step. 

We mainly use the ADM evaluation suite \citep{dhariwalDiffusionModelsBeat2021} for computing FID, precision, and recall. The FID is computed using $\num{10000}$ generated images and the whole training set for class-conditional ImageNet models. For text-to-image models, the FID is evaluated using the evaluation subset of MS COCO 2017 \citep{lin2014microsoft}. The hyperparameters used for the main experiment are given in \Cref{tab:ncg-hp}.

\tabParameters

\begin{algorithm}[h]
\caption{PyTorch implementation of \acrshort{ncg}.}
    \label{fig:imp-ncg}
    \centering
    \begin{minted}
[
% frame=lines,
framesep=2mm,
baselinestretch=1.2,
bgcolor=LG,
fontsize=\footnotesize,
% linenos
]
{python}
import torch

class MomentumBuffer:
    def __init__(self, momentum: float):
        self.momentum = momentum
        self.running_average = 0

    def update(self, update_value: torch.Tensor):
        new_average = self.momentum * self.running_average
        self.running_average = update_value + new_average


def project(
    v0: torch.Tensor, # [B, C, H, W] 
    v1: torch.Tensor, # [B, C, H, W]
):
    dtype = v0.dtype
    v0, v1 = v0.double(), v1.double()
    v1 = torch.nn.functional.normalize(v1, dim=[-1, -2, -3])
    v0_parallel = (v0 * v1).sum(dim=[-1, -2, -3], keepdim=True) * v1
    v0_orthogonal = v0 - v0_parallel
    return v0_parallel.to(dtype), v0_orthogonal.to(dtype)


def adaptive_projected_guidance(
    pred_cond: torch.Tensor,   # [B, C, H, W]
    pred_uncond: torch.Tensor, # [B, C, H, W]
    guidance_scale: float,
    momentum_buffer: MomentumBuffer = None,
    eta: float = 1.0,
    norm_threshold: float = 0.0,
):
    diff = pred_cond - pred_uncond
    if momentum_buffer is not None:
        momentum_buffer.update(diff)
        diff = momentum_buffer.running_average
    if norm_threshold > 0:
        ones = torch.ones_like(diff)
        diff_norm = diff.norm(p=2, dim=[-1, -2, -3], keepdim=True)
        scale_factor = torch.minimum(ones, norm_threshold / diff_norm)
        diff = diff * scale_factor
    diff_parallel, diff_orthogonal = project(diff, pred_cond)
    normalized_update = diff_orthogonal + eta * diff_parallel
    pred_guided = pred_cond + (guidance_scale - 1) * normalized_update
    return pred_guided
\end{minted}
\end{algorithm}

\section{More visual results}\label{sec:more-visual-results} This section presents extended visual comparisons between \gls{ncg} and CFG. Additional results using EDM2 are provided in \Cref{fig:edm-appendix}, with an example of how \gls{ncg} enhances diversity shown in \Cref{fig:diversity}. Further images for Stable Diffusion 2.1 and Stable Diffusion XL are included in \Cref{fig:sd2-appendix,fig:sdxl-appendix}. Moreover, \Cref{fig:sd3-appendix} illustrates how \gls{ncg} improves text spelling in Stable Diffusion 3. Finally, more examples of \gls{ncg} applied to distilled models are shown in \Cref{fig:pixart-lcm-appendix,fig:sdxl-lightning-appendix,fig:sdxl-flash-appendix}.
\figEDMAppendix
\figDiversity
\figSDtwoAppendix
\figSDXLAppendix
\figSDthreeAppendix
\figPixartLCMAppendix
\figSDXLFlashAppendix
\figSDXLLightningAppendix

\end{document}